\documentclass{article}

\usepackage[preprint, nonatbib]{neurips_2025}

\usepackage[utf8]{inputenc} 
\usepackage{hyperref}       
\usepackage{url}            
\usepackage{booktabs}       
\usepackage{amsfonts}       
\usepackage{nicefrac}       
\usepackage{microtype}      
\usepackage{xcolor}         
\usepackage{eccvabbrv}
\usepackage[most]{tcolorbox}
\usepackage{float}
\usepackage{graphicx}
\usepackage{capt-of}
\usepackage{cuted}
\usepackage{multirow}
\usepackage{cleveref}
\usepackage{multicol}
\usepackage[dvipsnames]{xcolor}

\usepackage{subcaption}
\usepackage{colortbl}

\newcommand{\vggt}{\textcolor{BlueViolet}{\textbf{VGGT}}}
\newcommand{\da}{\textcolor{Mahogany}{\textbf{DA3}}}
\newcommand{\duster}{\textcolor{OliveGreen}{\textbf{DUSt3R}}}

\newtcolorbox{geobox}{
  colback=fBlue30,  
  colframe=fBlue, 
  rounded corners,          
  boxrule=1.5pt,            
}

\newtcolorbox{databox}{
  colback=fOrange30,  
  colframe=fOrange, 
  rounded corners,          
  boxrule=1.5pt,            
}

\definecolor{cvprblue}{rgb}{0.21,0.49,0.74}

\definecolor{fBlue}{HTML}{004A99}   
\definecolor{fOrange}{HTML}{E98300} 

\colorlet{fBlue80}{fBlue!80!white}
\colorlet{fBlue60}{fBlue!60!white}
\colorlet{fBlue40}{fBlue!40!white}
\colorlet{fBlue30}{fBlue!30!white}
\colorlet{fBlue20}{fBlue!20!white}

\colorlet{fOrange80}{fOrange!80!white}
\colorlet{fOrange60}{fOrange!60!white}
\colorlet{fOrange40}{fOrange!40!white}
\colorlet{fOrange30}{fOrange!30!white}
\colorlet{fOrange20}{fOrange!20!white}

\title{On Geometric Understanding and Learned Priors in Feed-forward 3D Reconstruction Models}

\author{Jelena Bratuli\'c\textsuperscript{1}\thanks{The work was done during a PhD visit to the University of Oxford within the ELLIS PhD program. Correspondence to \texttt{bratulic@cs.uni-freiburg.de}} \hspace{0.2cm} Sudhanshu Mittal\textsuperscript{1} \hspace{0.2cm} Thomas Brox\textsuperscript{1} \hspace{0.2cm} Christian Rupprecht\textsuperscript{2}\\[.5em]
 \textsuperscript{1}University of Freiburg \hspace{0.2cm} \textsuperscript{2}University of Oxford
}

\begin{document}

\maketitle

\begin{abstract}

Feed-forward 3D reconstruction models such as DUSt3R, VGGT, and Depth Anything 3 (DA3) are transformer--based foundation models that infer camera geometry and dense scene structure in a single forward pass. Trained at scale in a supervised fashion, they raise a central question: do these models build upon geometric principles akin to traditional multi--view pipelines, or do they primarily rely on learned priors arising from the large--scale training setup? We find that epipolar geometry emerges within the intermediate layers of all three models and is causally linked to correspondence patterns in attention heads. To study this, we perform a systematic analysis of their internal representations across three real--world datasets and a controlled synthetic dataset. We quantify geometric understanding by probing intermediate features, analyzing attention patterns to identify correspondence matching patterns, and performing targeted interventions at the attention level. Further, we assess the role of learned priors by applying challenging input--level perturbations, such as occlusions, scene ambiguities, and varying camera configurations, and compare them against classical multi--stage reconstruction pipelines. 
\end{abstract}
\section{Introduction} \label{sec:introduction}

Estimating 3D geometry from a set of 2D images has long been a fundamental challenge in computer vision. Traditional approaches rely on geometry--constrained pipelines that require a multi--step, sequential process: feature extraction, matching, and refinement, followed by camera pose estimation and 3D point cloud reconstruction.  
These approaches, although mathematically precise, struggle with ambiguities and lack robustness to scene dynamics and partial visibility, and can be computationally expensive due to runtime optimization.

Recent deep learning approaches such as DUSt3R~\cite{dust3r_cvpr24}, MASt3R~\cite{mast3r_eccv24}, VGGT ~\cite{wang2025vggt}, and DA3~\cite{depthanything3}, have shown that 3D reconstruction can be achieved in a single unified feed--forward step, replacing the traditional multi--stage pipeline. 
These methods predict camera poses and dense 3D point maps from image sequences using a functional mapping learned from large--scale labeled datasets. 
As all optimization is shifted to training, they run much faster than classical frameworks, and the priors learned from the data make them significantly more robust to ambiguities and input variations. 
However, such learned functional mappings are prone to fail in out--of--distribution cases. 
It is worth noting that the laws of 3D geometry are universal and should not depend on the distribution of the training data. 
Therefore, the question arises whether these models, while learning from data,
internally learn any geometric reasoning mechanisms similar to classical multi--view pipelines, or do their capabilities primarily arise from learned priors? 
If a geometric mechanism is indeed learned, it would guarantee a similar level of robustness to out--of--distribution cases as with traditional pipelines, while leveraging the complex learned priors in in--distribution scenarios. 

In this work, we examine the underlying mechanisms of three representative feed--forward 3D reconstruction models: DUSt3R, VGGT, and Depth Anything 3 (DA3). 
We aim to mechanistically understand how these models compute their predictions and to determine whether their performance arises from internalized geometric reasoning or predominantly from learned priors.
In particular, we investigate whether their representations encode fundamental principles of multi--view geometry, such as correspondences and epipolar geometry. 

To this end, we design a controlled study using different object-centric and scenery datasets, and a controlled synthetic dataset built on ShapeNet~\cite{shapenet2015} objects under various conditions, with complete scene information and all 3D attributes being available. 
Our systematic analysis adopts two perspectives: (1) a \textbf{\textcolor{fBlue}{geometrical}} perspective, where we probe the internal representations and perform interventions to assess the causal effect of these representations on the fidelity of the network representation in terms of epipolar geometry, (2) a \textbf{\textcolor{fOrange}{data}} perspective, where we test the model's robustness to interventions on input images to study the (typically positive) influence of learned priors from data and training.

The three studied models are jointly trained to predict dense scene structure and camera geometry directly from a set of images in a single forward pass. Although their architectural designs differ, none is explicitly trained to estimate classical geometric entities such as the fundamental matrix. To evaluate their use of geometric principles, we focus on a core property: epipolar geometry. The fundamental matrix $\mathbf{F} \in \mathbb{R}^{3 \times 3}$ provides a natural lens for this analysis, as it formalizes the epipolar geometry between different camera views and corresponding points observed in them. We therefore use $\mathbf{F}$ prediction via probing as a proxy to assess how much geometric information is encoded in the model's internal representations.

\textbf{Contributions.}
Our extensive study across three real-world and a synthetic dataset reveals several key insights. 
(1) We show that the intermediate representations in DUSt3R, VGGT, and DA3 encode epipolar geometry.
(2) We identify a geometric phase transition in the middle layers, where fundamental matrix recovery coincides with the emergence of point correspondences in the attention space.
(3) Through targeted attention-knockout interventions, we establish a causal link between correspondence formation and the encoding of epipolar geometry.
(4) From a data perspective, we show, using VGGT, that such models remain robust to moderate disturbances, often outperforming classical pipelines. Under occlusions, they infer correspondences despite missing evidence, revealing reliance on learned priors and a trade-off between robustness and geometric consistency.

\section{Related work} \label{sec:related_work}

\textbf{Classical 3D reconstruction and multi-view geometry.}
Classical structure-from-motion (SfM) pipelines~\cite{Hartley2004, snavely2006photo, schoenberger2016revisiting, agarwal2011building} comprise several stages: feature detection, correspondence matching, camera pose estimation, and bundle adjustment, and rely on hand-crafted features such as SIFT~\cite{lowe2004distinctive}. 
A central component of these pipelines is estimating the fundamental matrix $\mathbf{F}$, which encodes the epipolar geometry between two views~\cite{Hartley2004}. Minimal solvers such as the five-point~\cite{nister2004fivepoint} and eight-point~\cite{hartley1997eightpoints} algorithms recover $\mathbf{F}$ from sparse correspondences, usually combined with robust estimators like RANSAC~\cite{fischler1981random} to handle noisy correspondences and outliers. 
Recent learned matching methods such as SuperGlue~\cite{sarlin20superglue}, LightGlue~\cite{lindenberger2023lightglue}, and LoFTR~\cite{sun2021loftr} significantly improve correspondence matching. However, geometric relations such as the fundamental matrix are still estimated using classical algorithms. 
In this work, we use epipolar geometry as a principled lens to study whether modern feed-forward reconstruction models internally encode such geometric structure.

\textbf{Feed-forward reconstruction models.}
Recent deep learning approaches challenge the traditional multi-stage paradigm for 3D reconstruction. 
DUSt3R~\cite{dust3r_cvpr24} introduced a feed-forward formulation for dense 3D reconstruction from image pairs without requiring known camera poses. 
MASt3R~\cite{mast3r_eccv24} improves robustness to large viewpoint changes by augmenting the architecture with a dense local feature head, while preserving DUSt3R's robustness.
VGGSfM~\cite{wang2024vggsfm} further demonstrated that fully differentiable end-to-end reconstruction pipelines can outperform classical SfM methods.
More recent foundation models, such as VGGT~\cite{wang2025vggt} and Depth Anything 3 (DA3)~\cite{depthanything3}, directly infer depth and scene geometry from image collections in a single feed-forward pass, achieving state-of-the-art performance while significantly reducing inference time. 
Follow-up work further accelerates inference~\cite{Yang_2025_Fast3R} or extends these architectures to downstream tasks such as SLAM and real-time reconstruction~\cite{streamVGGT, maggio2025vggtslamdensergbslam}. 
Despite these advances, the internal mechanisms by which these models represent and reason about geometry remain poorly understood, a gap our work aims to address.
 
\textbf{Interpretability and causal analysis.} Understanding neural network mechanisms has become increasingly important, with much of the work centered on LLMs~\cite{elhage2021mathematical, olsson2022context}. 
A common approach is the use of probing classifiers~\cite{belinkov2017neural, alain2018understandingintermediatelayersusing, laina2022measuring} to test whether information is decodable from intermediate representations.
For computer vision tasks, Network Dissection and related work~\cite{netdissect2017, Zhou2018ECCV, bau2020units} quantify the CNNs' interpretability at individual units, whereas Chefer et al.~\cite{Chefer2021CVPR} showed that naive attention visualization can be misleading for transformers.
Different interventions, such as activation patching or causal mediation analysis~\cite{heimersheim2024useinterpretactivationpatching, zhang2024best_practices}, provide more substantial evidence by measuring the effects of modified activations within a model.
However, little work has been conducted on interpretability in geometric understanding for 3D vision models. 
Recently, Stary et al.~\cite{stary2025understandingmultiviewtransformers} analyzed a variant of DUSt3R and studied the role of individual layers and correspondences in multi-view transformers. Maggio et al.~\cite{maggio2026vggtslam20realtimedense} analyze the VGGT's layers for image retrieval verification, while Han et al.~\cite{han2025emergent} study the emergent outlier detection ability in VGGT.

\section{Background} \label{sec:background}
\subsection{Epipolar Geometry}\label{sec:background_epipolar}

Here we provide a brief overview of epipolar geometry, but an extensive analysis can be found in~\cite{Hartley2004}. Epipolar geometry describes the projective relationship between two cameras viewing the same 3D point. If a 3D point $\mathbf{X} \in \mathbb{R}^3$ is observed by two pinhole cameras, its image projections are
\begin{equation}
    \mathbf{x}_1 = \mathbf{K}_1 [\, \mathbf{I} \mid \mathbf{0} \,] \mathbf{X}, 
    \quad 
    \mathbf{x}_2 = \mathbf{K}_2 [\, \mathbf{R} \mid \mathbf{t} \,] \mathbf{X},
\end{equation}
where $\mathbf{K}_i$ are the camera intrinsics and $(\mathbf{R}, \mathbf{t})$ is the relative pose. 
A ray from the first camera center through $\mathbf{x}_1$ defines an \emph{epipolar plane} with the two camera centers. 
Its intersection with the second image plane gives the \emph{epipolar line} $\ell_2$, on which the corresponding point $\mathbf{x}_2$ must lie. All epipolar lines intersect at the epipole.
This geometric relationship is expressed by the \emph{epipolar constraint}:
\begin{equation}
    \mathbf{x}_2^\top \mathbf{E}\, \mathbf{x}_1 = 0, 
    \quad 
    \mathbf{E} = [\mathbf{t}]_\times \mathbf{R},
\end{equation}
where $\mathbf{E}$ is the \textit{essential matrix} and $[\mathbf{t}]_\times$ is the skew-symmetric matrix representing the cross product with $\mathbf{t}$. For uncalibrated cameras, the relation becomes
\begin{equation}
    \mathbf{x}_2^\top \mathbf{F}\, \mathbf{x}_1 = 0, 
    \quad 
    \mathbf{F} = \mathbf{K}_2^{-\top} \mathbf{E} \mathbf{K}_1^{-1},
\end{equation}
where $\mathbf{F}$ is the \textit{fundamental matrix}. 

Both the essential matrix $\mathbf{E}$ and fundamental matrix $\mathbf{F}$ have rank~2, as $\mathbf{E} = [\mathbf{t}]_\times \mathbf{R}$ inherits the rank deficiency from the skew-symmetric matrix $[\mathbf{t}]_\times$. This constraint reflects the fact that all epipolar lines intersect at the epipole. When estimating $\mathbf{F}$ from point correspondences, the rank-2 constraint is typically enforced by projecting the unconstrained solution onto the manifold of rank-2 matrices. Given an initial estimate $\hat{\mathbf{F}}$, we compute its singular value decomposition  $\hat{\mathbf{F}} = \mathbf{U} \text{diag}(\sigma_1, \sigma_2, \sigma_3) \mathbf{V}^\top$ where $\sigma_1 \geq \sigma_2 \geq \sigma_3$, and set the smallest singular value to zero:
\begin{equation}
    \mathbf{F} = \mathbf{U}\, \text{diag}(\sigma_1, \sigma_2, 0)\, \mathbf{V}^\top.
\end{equation}
This produces the closest rank-2 approximation in the Frobenius norm.

\textbf{Evaluating Epipolar Geometry.} We can assess the quality of an estimated fundamental matrix using two metrics that quantify deviations from epipolar consistency. 

The algebraic error
\begin{equation}
    e_{\text{alg}} = \mathbf{x}_2^\top \mathbf{F}\, \mathbf{x}_1
\end{equation}
measures the constraint satisfaction (which should be 0), but has no geometric meaning directly and is scale-dependent. More geometrically meaningful measures are the Sampson error and the smallest singular value of $\mathbf{F}$. 
The \emph{Sampson error}~\cite{sampson1982fitting} represents a first-order approximation of the true geometric error:
\begin{equation}
d_{\text{Sampson}} = \frac{(\mathbf{x}_2^\top \mathbf{F}\, \mathbf{x}_1)^2}
{\|(\mathbf{F} \mathbf{x}_1)_{1:2}\|^2 + \|(\mathbf{F}^\top \mathbf{x}_2)_{1:2}\|^2}
\end{equation}
Sampson error approximates the reprojection error, avoiding iterative optimization. 
In practice, well-calibrated systems typically yield Sampson errors below $1$--$2$ pixels, though acceptable values depend on image resolution, focal length, and the application's precision requirements.  
The \textit{smallest singular value} of $\mathbf{F}$ must be zero to satisfy the rank-2 constraint of the fundamental matrix. In practice, the smallest singular value that is less than $10^{-3}$ of the largest singular value is generally considered acceptable.

\subsection{Feed-forward 3D Reconstruction Models}

We analyze three representative feed-forward 3D reconstruction models covering different architectural paradigms.

\textbf{DUSt3R}~\cite{dust3r_cvpr24} introduced the paradigm of feed-forward dense 3D reconstruction from arbitrary image pairs without requiring prior camera calibration or pose information. The model predicts pointmaps expressed in the reference frame of one view using an asymmetric transformer encoder--decoder architecture. For multi-view inputs, pairwise predictions are aligned into a common coordinate frame through a global alignment step performed at test time. This formulation unifies depth, pose, and dense reconstruction within a single model and laid the foundation for later feed-forward geometry models such as VGGT and DA3.

\textbf{Visual Geometry Grounded Transformer (VGGT)}~\cite{wang2025vggt} is a transformer-based model that jointly predicts camera intrinsics, extrinsics, depth maps, pointmaps, and 3D point trajectories from multiple views.
It builds on a DINOv2~\cite{oquab2023dinov2} backbone with alternating frame-wise and global self-attention layers to aggregate information across views. VGGT is trained on a large mixture of publicly available 3D datasets using direct supervision for camera pose estimation, depth prediction, 3D point cloud reconstruction, and 3D point tracking. 

\textbf{Depth Anything 3}~\cite{depthanything3} predicts scene geometry from arbitrary visual inputs using a transformer backbone trained in a teacher–student framework. Unlike the multi-task supervision used in VGGT, DA3 formulates the problem as the joint prediction of depth maps and ray maps using a unified prediction head. The model can optionally condition on input camera poses at inference and achieves state-of-the-art performance on camera pose estimation and geometric accuracy benchmarks, surpassing prior models, including VGGT. 

\subsection{Dataset Design}\label{sec:background_dataset}
We conduct our study on synthetic and real-world datasets. Since all three models are trained on large, diverse datasets, ensuring meaningful out-of-distribution evaluation and analysis of certain scenarios requires careful dataset selection. Thus, we construct a custom ShapeNet-based synthetic dataset that provides complete geometric ground truth and allows systematic scene manipulation. Here we briefly describe the three real-world and custom synthetic datasets, while more details and visual examples are provided in the supplementary.

\textbf{Real-world datasets.} \textit{DTU MVS}~\cite{jensen2014large} is a controlled dataset of 124 
object-centric scenes photographed from 49 fixed positions under structured-light ground truth.  \textit{ETH3D}~\cite{schoeps2017cvpr} is a precise, high-resolution, multi-view dataset covering diverse indoor and outdoor scenes. \textit{MipNeRF360}~\cite{barron2022mipnerf360} comprises nine unbounded outdoor and indoor scenes captured with a consumer camera and reconstructed via COLMAP. Camera poses and correspondences are derived from COLMAP sparse reconstructions, while for DTU, we use depth maps from MVSNet~\cite{yao2018mvsnet}. To ensure diversity in viewpoint separation while improving control over the data distribution, we bin image pairs by the true 3D angular distance between the cameras and sample from the bins for MipNeRF360 and DTU. For ETH3D, we include all pairs since its scenes, although diverse, contain only a relatively small number of images. We provide additional analysis using the semantic correspondences from the SPair71k dataset~\cite{min2019spair} in the supplementary.

\textbf{Custom ShapeNet dataset.} To model diverse scenarios, we generated a controlled synthetic dataset from ShapeNet~\cite{shapenet2015} assets rendered in Blender, which enables precise control over scene properties and complete geometric ground truth (camera parameters, depth, correspondences, fundamental and essential matrices) for precise analysis. We selected two types of objects: 10 different categories of \emph{unique} assets with distinctive features and 5 different categories of \emph{symmetric} assets (e.g., bottles, vases) that introduce correspondence ambiguity. We used focal lengths ranging from 24 mm to 100 mm with a standard 36 mm camera sensor, and distinguish between 4 camera configurations: a stereo setup with parallel image planes and a horizontal baseline, and three different non-parallel image planes with different angular baselines, categorized as small (10--25 degrees), medium (45--75), and large (90--120). For robustness experiments, we created ambiguous scenes with repetitive objects.  
\section{Geometric interpretability} \label{sec:geometry_point}

\duster,~\vggt, and Depth Anything 3~(\da) are trained to predict camera geometry and dense 3D structure through supervised objectives specific to their respective formulations. To evaluate whether these models develop geometric reasoning beyond their explicit training targets, we examine their ability to represent the fundamental matrix via a small MLP probing. Since the fundamental matrix is never directly supervised, this independent probing avoids confounding effects from task-specific heuristics and strong supervision on camera pose and intrinsics. Further, layer-wise probing also allows us to localize where this geometric information is encoded in the network layers, providing deeper insight into how geometric structure develops across the network.

\subsection{Do these models encode geometry? If so, where?}
We probe intermediate layers to determine where, and to what extent, the fundamental matrix can be recovered from these representations.

\begin{figure}[b!]
\centering
\newcommand{\rowlabelwidth}{0.55cm}
\newcommand{\imwidth}{0.305\textwidth}
\setlength{\tabcolsep}{2pt} 
\renewcommand{\arraystretch}{0} 

\newcommand{\rowlab}[1]{%
  \makebox[\rowlabelwidth][c]{\raisebox{2.5ex}{\rotatebox{90}{\textbf{#1}}}}%
}

\begin{tabular}{@{}c c c c@{}}
  & \textbf{VGGT} & \textbf{Depth Anything 3} & \textbf{DUSt3R} \\

  \rowlab{Synthetic} &
  \includegraphics[width=\imwidth]{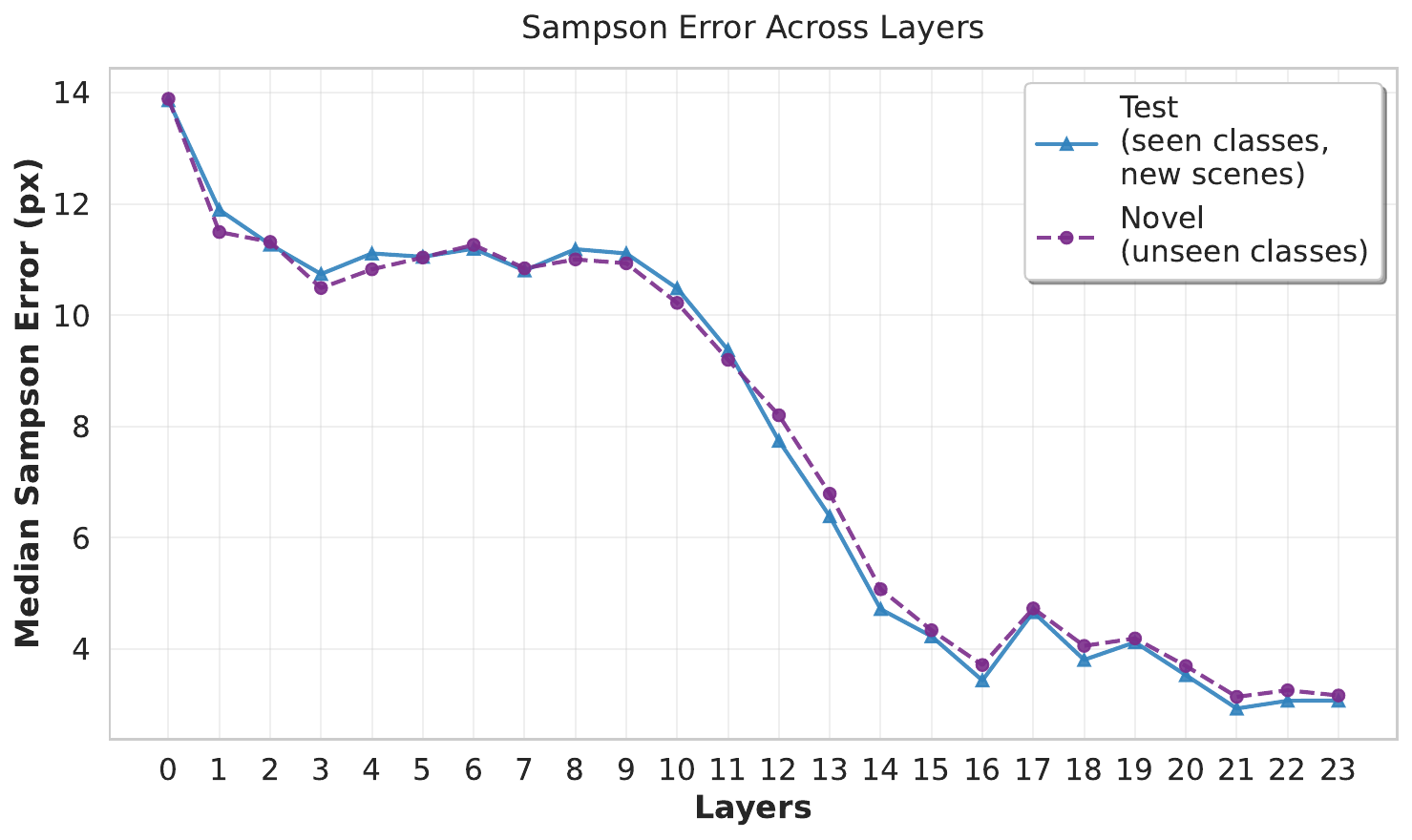} &
  \includegraphics[width=\imwidth]{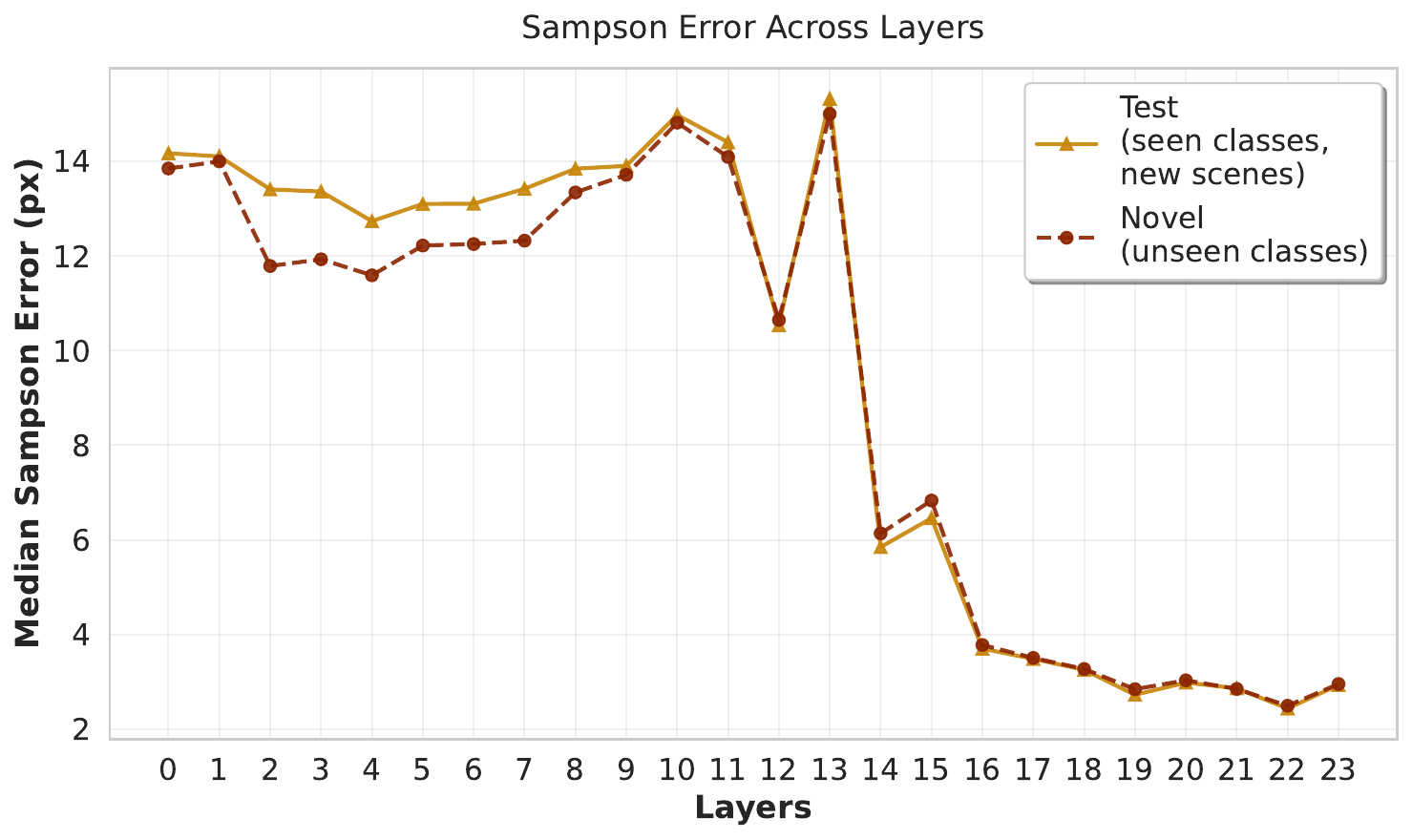} &
  \includegraphics[width=\imwidth]{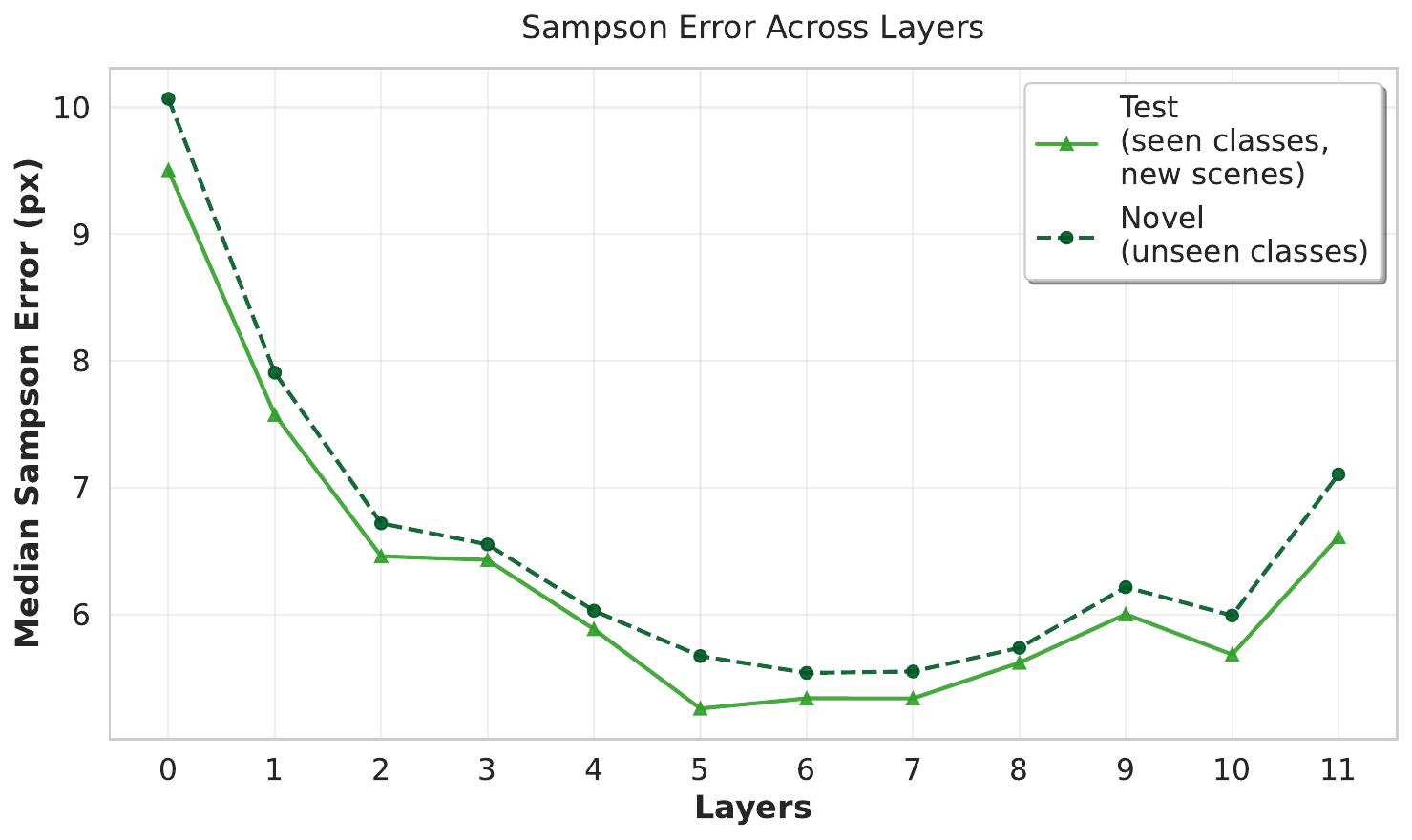} \\

  \rowlab{Real data} &
  \includegraphics[width=\imwidth]{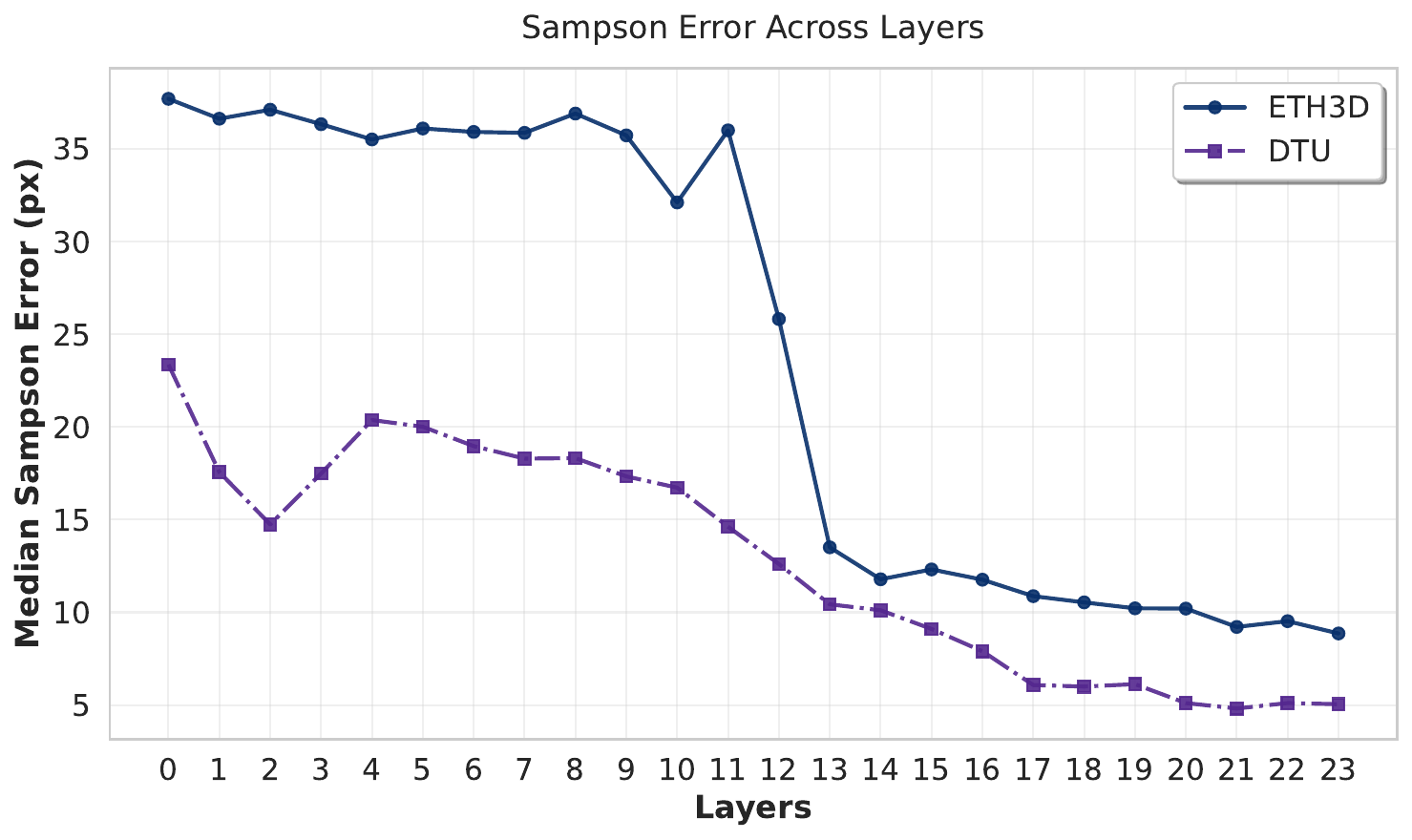} &
  \includegraphics[width=\imwidth]{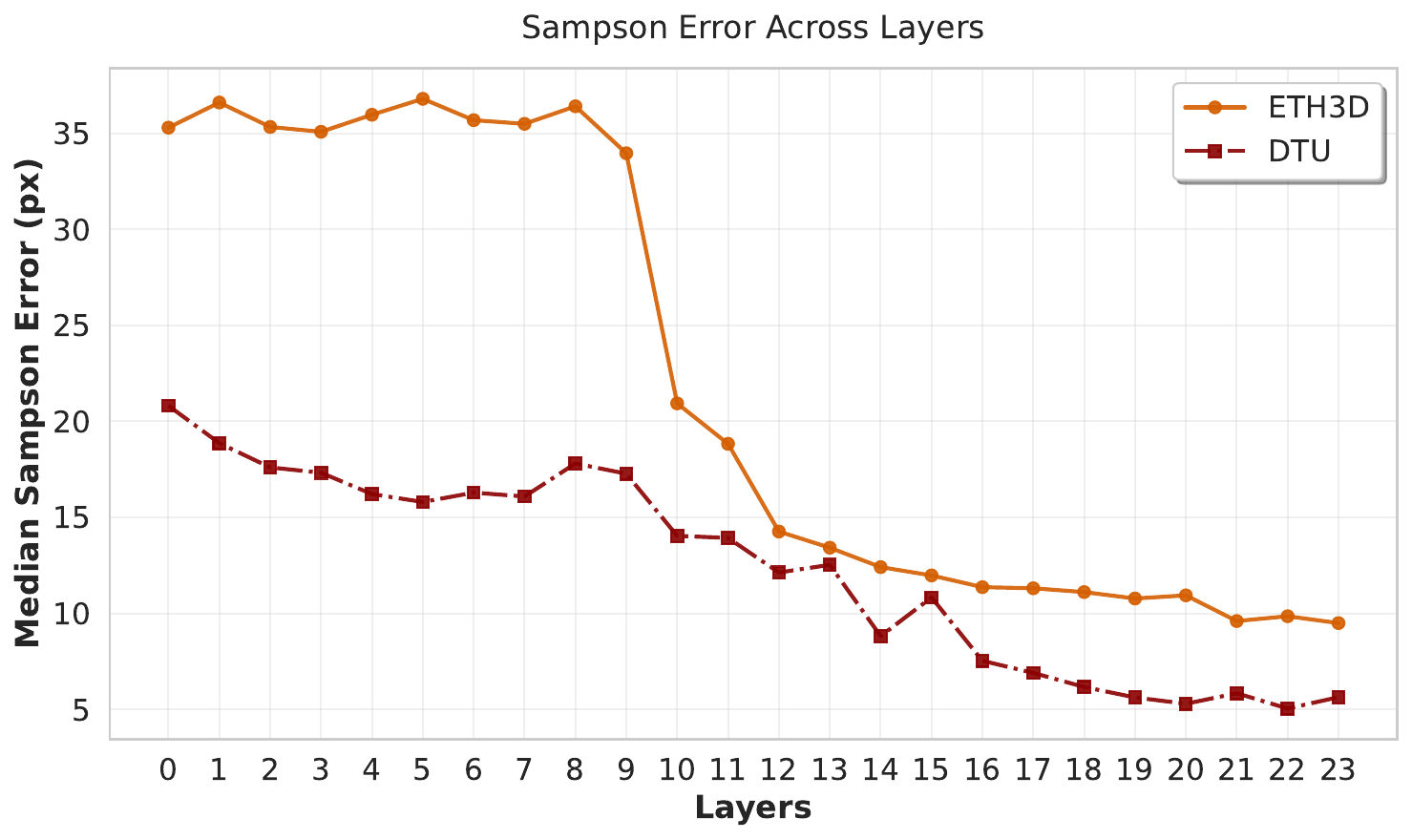} &
  \includegraphics[width=\imwidth]{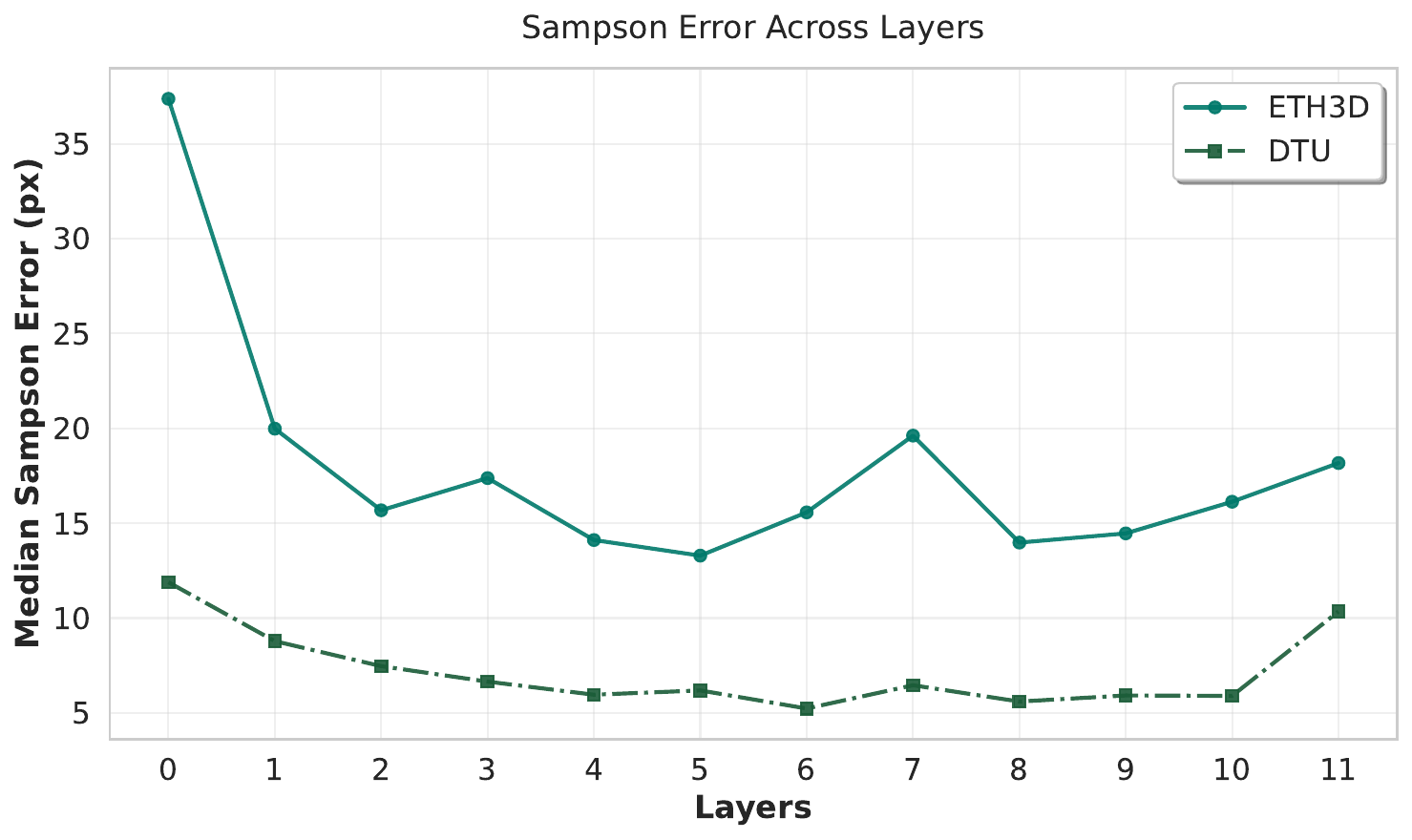} \\
\end{tabular}

\caption{\textbf{Probing internal representations for fundamental matrix approximation.} We successfully recover the fundamental matrix from all three models on both synthetic and real-world datasets. However, the emergence trends differ across models: models with camera-tokens (\vggt~and \da) exhibit a similar pattern, with clear change in the middle layers and strong performance in the final layers. \duster~shows strong performance already in the early layers, with slight degradation towards the final layers. }
    \label{fig:probing_results}
\end{figure}

\textbf{Experiment setup.} We train simple two-layer MLP probes on intermediate representations at each layer to predict the fundamental matrix. Given the differences between the architectures, we use camera tokens from each layer for \vggt~and \da, and averaged features after each decoder block for~\duster. Each probe utilizes a hidden dimension of 512 and is optimized for 30 epochs using the squared Sampson distance as the training loss on ground-truth correspondences, without directly enforcing the rank-2 constraint. We train probes on each dataset separately, and report the root Sampson error in pixel space across unseen data splits.
We further perform a similar analysis using image triplets and probing for all available fundamental matrices. Additional details and results are provided in the supplementary.

\textbf{Results.}
~\Cref{fig:probing_results} shows that the fundamental matrix can be reliably recovered from all three models (\vggt,~\da,~\duster) across different datasets (ShapeNet, ETH3D, DTU). The location and the trends of the emergence, however, differ. 
In \vggt~and \da, the ability to recover the fundamental matrix emerges roughly in the middle layers (layers 12–14).
Performance continues to be refined through the final layers. The sudden improvement in error within a few layers suggests that the necessary information for establishing epipolar geometry is formed in these layers. The layers in \da~correspond to those immediately after global attention is introduced. As a second criterion, we also check the smallest singular value of the predicted $\mathbf{F}$. We observe that it attains its minimum within the same set of layers and is approximately $2.5\mathrm{e}{-4}$ smaller than the maximum value, further supporting the emergence of epipolar geometry within these layers (see supplementary).

\duster, on the other hand, exhibits a different trend, in which the fundamental matrix emerges in the early decoder blocks and refines throughout the layers with slight divergence in the last layers.
We speculate that this divergence arises due to independent target heads in the model.

Since all models encode the fundamental matrix in their representations, with the emergence followed by a sudden improvement over a few layers, this suggests that these layers serve a specific purpose. An obvious candidate is that the network’s attention maps capture a correspondence structure across views.

\begin{geobox}
\textit{Epipolar geometry emerges in the internal representations of feed-forward multi-view transformers. In camera-token based architectures, it appears in the middle layers and is progressively refined towards the final layers.}
\end{geobox}

\subsection{How do attention maps encode correspondences?}

Probing reveals the emergence of learned epipolar geometry in the intermediate layers. However, we do not yet know how the model recovers this information. To this end, we aim to discover the underlying correspondence-matching patterns in the attention maps of those layers across two views. We hypothesize that these correspondences are computed by global (cross) attention layers.

\textbf{Experiment setup.} We analyze the query-key (QK) attention space to identify correspondence matching. For each patch token in the source image, we locate the patch in the target image that receives the highest attention and calculate accuracy based on whether this target patch corresponds to the true matching patch. When a source patch maps to multiple target patches due to patch discretizations, we count a hit if any of the correct positions receive the maximum attention. We evaluate all unique patch correspondence pairs and report the average correspondence accuracy over our controlled ShapeNet's test set of 5 scenes from different object categories, taken at a 50 mm focal length, averaged across sampling modes that cover different camera-pair configurations.  We analyze all available cross- and global-attention layers in the model: 24 for~\vggt, 8 for~\da~(starting at layer 9), and 12 for~\duster. We compare the forward direction (view 1$\xrightarrow{}$view 2) and reverse (view 2$\xrightarrow{}$view 1). 
Further, we analyze feature-space similarity to assess whether correspondence-matching information is retained in the representation outside of attention space. Details, real-world results, and feature-similarity results are provided in the supplementary.

\begin{figure}[b]
\centering
\newcommand{\rowlabelwidth}{0.55cm}
\newcommand{\imwidth}{0.305\textwidth}
\setlength{\tabcolsep}{2pt} 
\renewcommand{\arraystretch}{0} 

\newcommand{\rowlab}[1]{%
  \makebox[\rowlabelwidth][c]{\raisebox{2.5ex}{\rotatebox{90}{\textbf{#1}}}}%
}

\begin{tabular}{@{}c c c c@{}}
  & \textbf{VGGT} & \textbf{Depth Anything 3} & \textbf{DUSt3R} \\

  \rowlab{F (1$\to$2)} &
  \includegraphics[width=\imwidth]{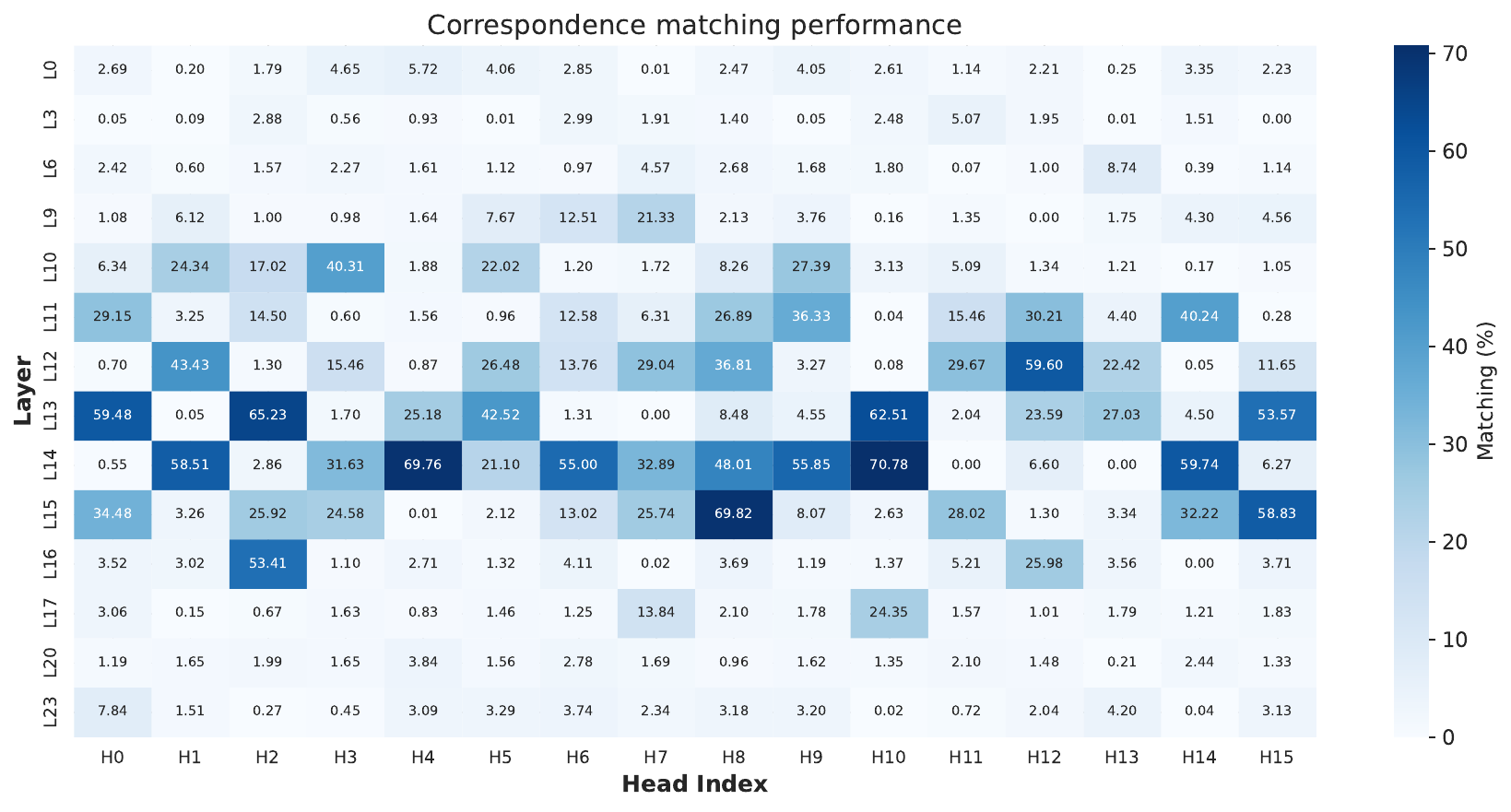} &
  \includegraphics[width=\imwidth]{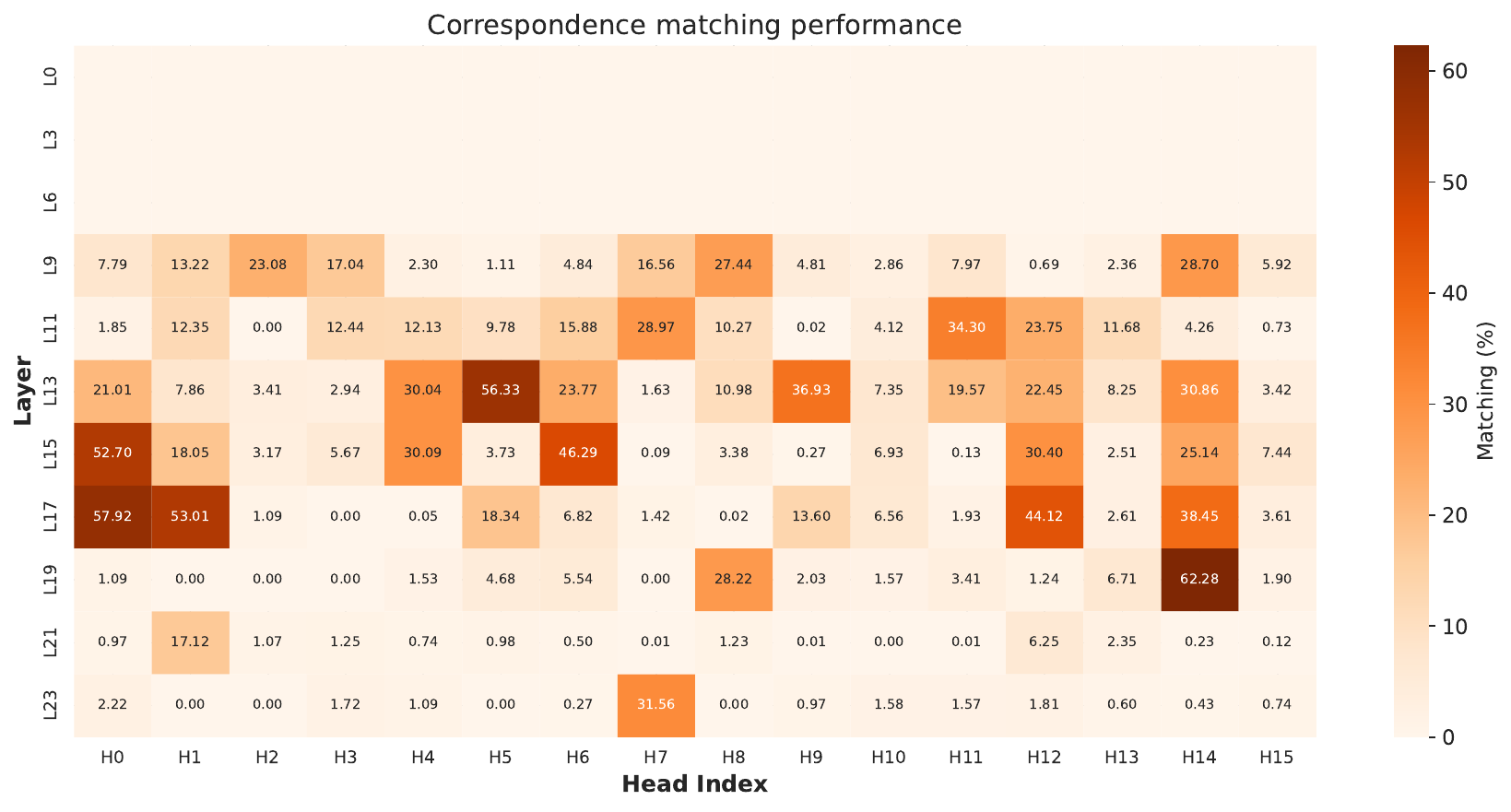} &
  \includegraphics[width=\imwidth]{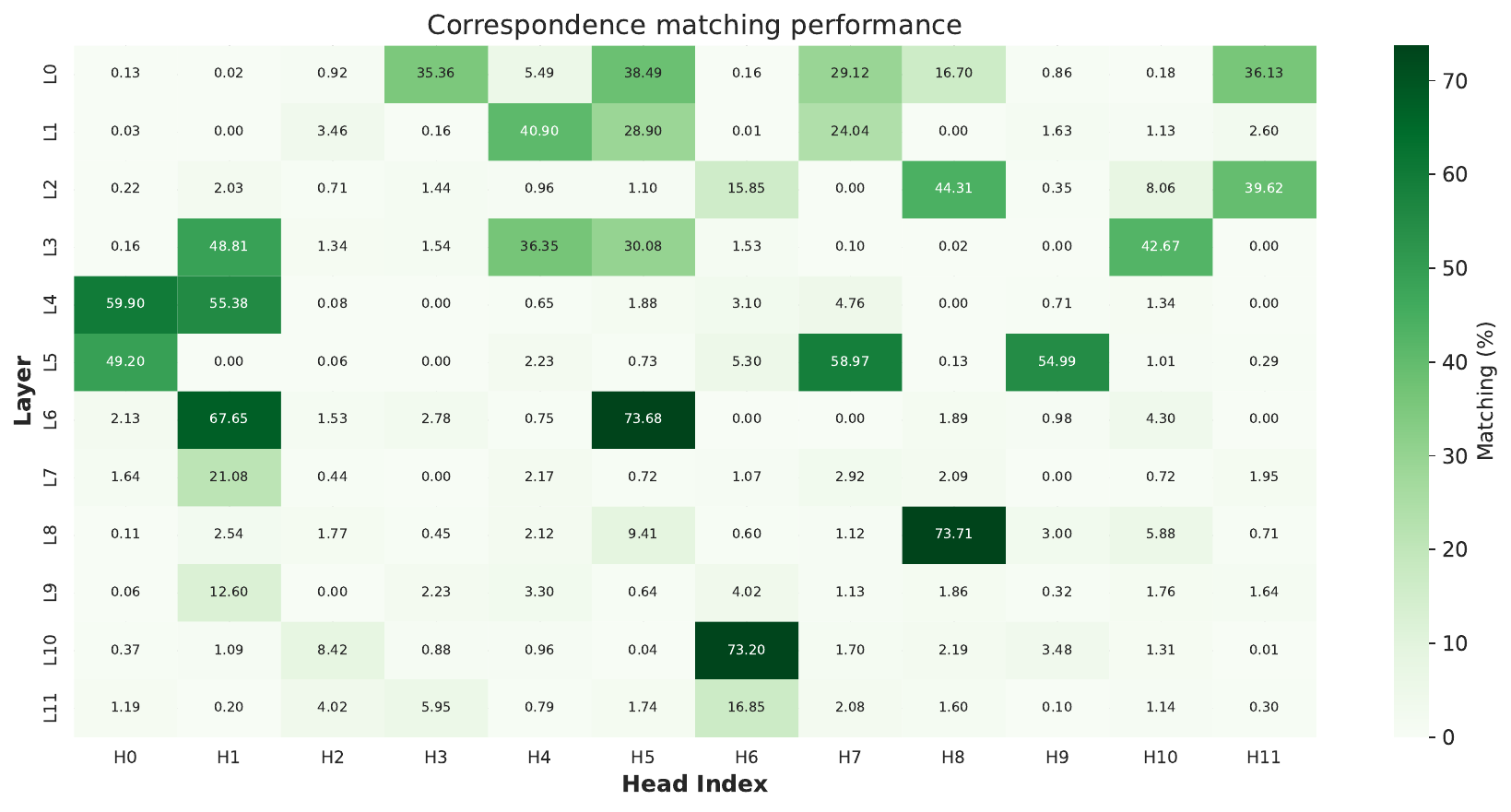} \\

  \rowlab{R (2$\to$1)} &
  \includegraphics[width=\imwidth]{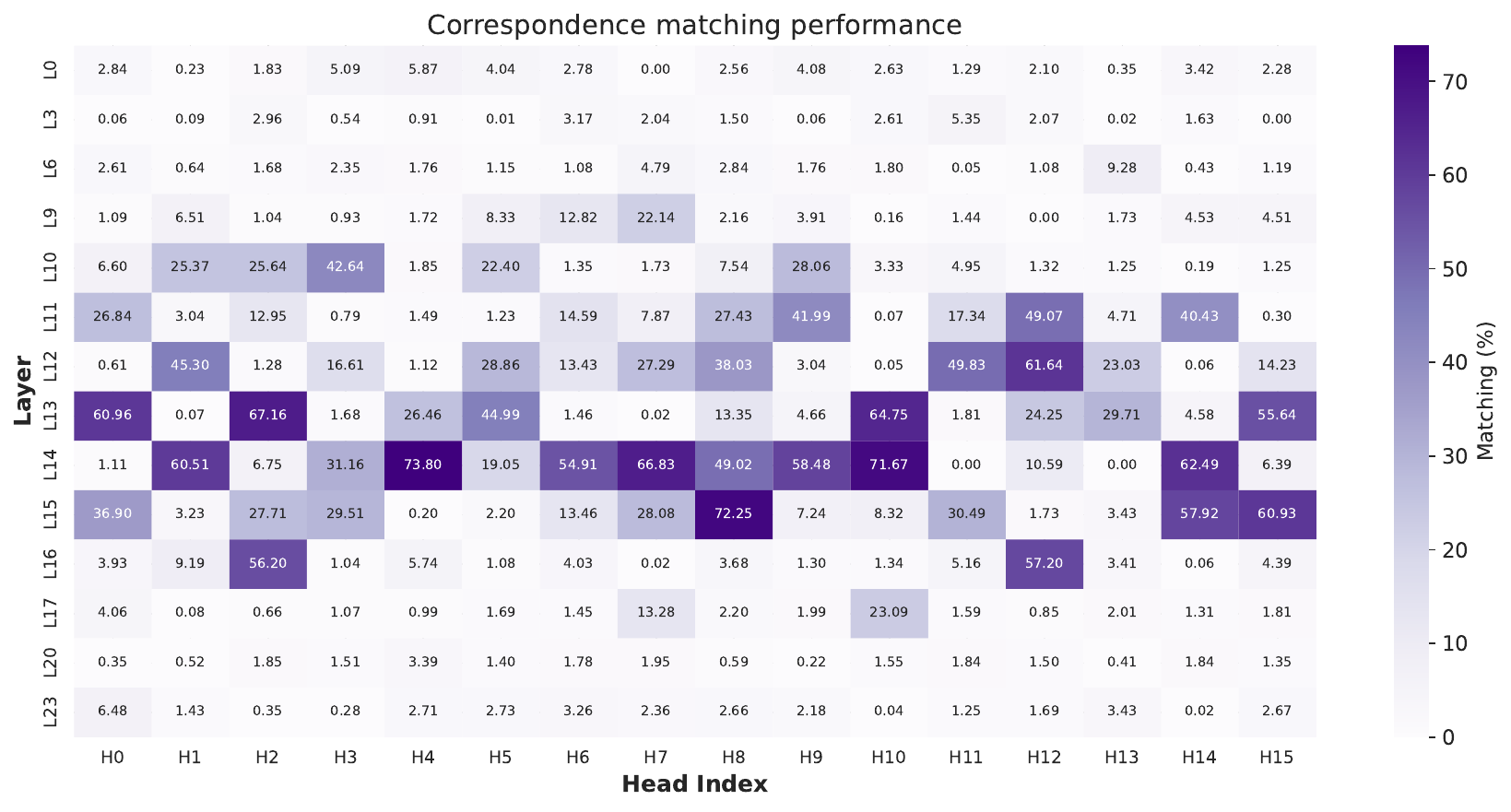} &
  \includegraphics[width=\imwidth]{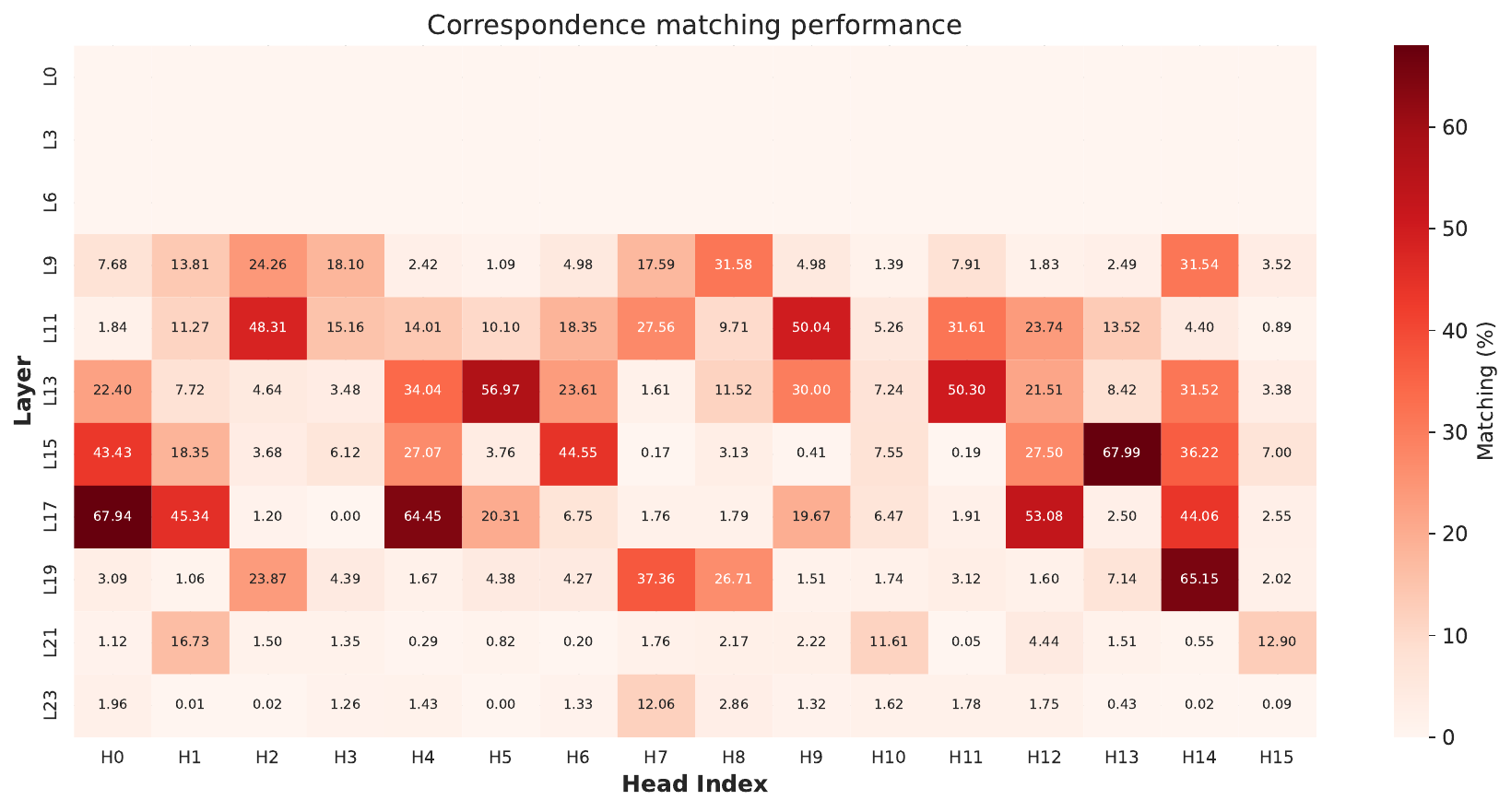} &
  \includegraphics[width=\imwidth]{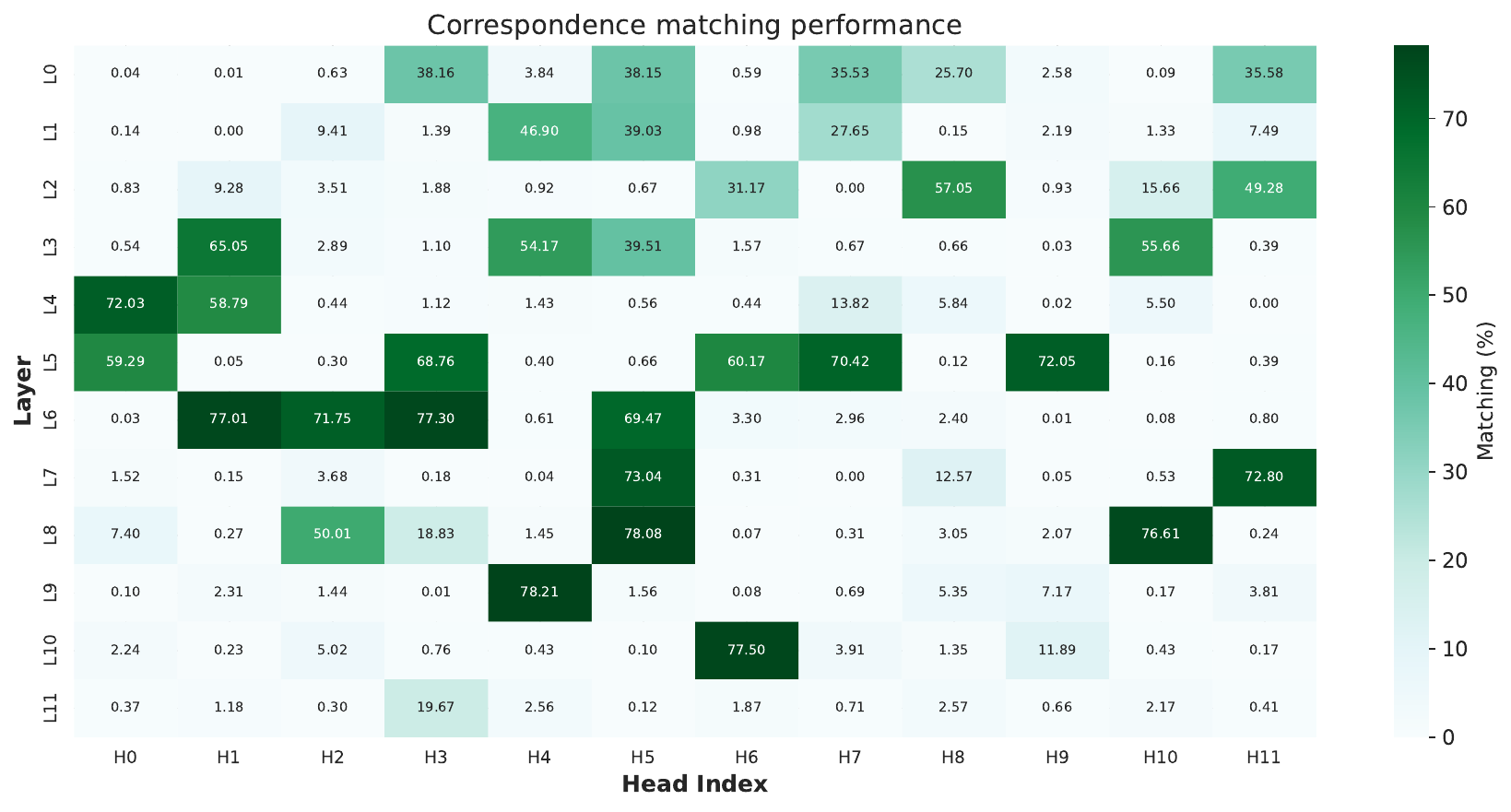} \\
\end{tabular}

\caption{\textbf{Correspondence matching per attention head.} We compare correspondence matching performance using the QK attention space for forward (view 1$\xrightarrow{}$view 2) and reverse (view 2$\xrightarrow{}$view 1) directions. 
Strong matching performance is established in the layers, which immediately precede the transition observed in~\cref{fig:probing_results}. Middle layers are active for both \vggt~and \da, whereas \duster~exhibits correspondence matching already in the first block and asymmetrically active heads in different directions, stemming from its asymmetric decoders.}
\label{fig:corr_matching}
\end{figure}

\textbf{Results.}~\Cref{fig:corr_matching} shows strong correspondence matching activity in multiple layers and heads across all three models.
In \vggt, strong correspondence matching across multiple attention heads is observed in the middle layers, roughly between layers 10 and 16. This behavior begins around layer 10 and precedes the sharp improvement in fundamental matrix recovery at layers 13 and 14 (see~\cref{fig:probing_results}).
This suggests a connection between correspondence estimation and the emergence of epipolar geometry within the network. We observe a similar trend for \da, which, despite having layer 9 as the first global-attention layer, has the strongest correspondence matching across the middle layers of the complete architecture.  \duster, however, exhibits strong correspondence-matching performance from the early decoder layers, while performance continues to improve in the middle layers. This is again well aligned with the probing results in~\cref{fig:probing_results}, where \duster~exhibited some decent fundamental matrix recovery already in the early layers. 

Further, both \vggt~and~\da~have symmetrical patterns, meaning that the same heads and layers perform correspondence matching equally well in both forward (F 1$\xrightarrow{}$2) and reverse (R 2$\xrightarrow{}$1) directions. On the other hand, \duster's activation patterns are asymmetric, with the reverse direction exhibiting stronger correspondence-matching performance and a greater number of active heads. We believe the difference stems from \vggt~and \da~using global attention layers, whereas \duster~has asymmetric cross-attention decoders.

We also analyzed QK-space patterns across other layers to identify interpretable behavior. 
We observed that early layers in~\vggt~exhibit semantic alignment, where visually similar parts are matched (e.g.~a chair leg in one view matches another chair leg in the second view), but not necessarily to the correct instance. More details are included in the supplementary.

\begin{geobox}
\textit{Point correspondence matching appears in the mid-layer attention heads as a key mechanism, enabling geometric alignment across views.} 
\end{geobox}

\subsection{Do these attention patterns play a causal role?}

The previous observations establish a correlation between the emergence of epipolar geometry via fundamental matrix recovery and the correspondence-matching patterns in the same set of intermediate layers. To determine whether the attention patterns are truly a cause of the geometric encoding observed in~\cref{fig:probing_results}, we perform targeted interventions on the attention heads that we hypothesize are responsible for the epipolar representation.

\begin{figure}[t!]
\centering
\newcommand{\rowlabelwidth}{0.55cm}
\newcommand{\imwidth}{0.305\textwidth}
\setlength{\tabcolsep}{2pt} 
\renewcommand{\arraystretch}{0} 

\newcommand{\rowlab}[1]{%
  \makebox[\rowlabelwidth][c]{\raisebox{9.5ex}{\rotatebox{90}{\textbf{#1}}}}%
}

\newcommand{\rowlabf}[1]{%
  \makebox[\rowlabelwidth][c]{\raisebox{4.5ex}{\rotatebox{90}{\textbf{#1}}}}%
}

\begin{tabular}{@{}c c c c@{}}
  & \textbf{VGGT} & \textbf{Depth Anything 3} & \textbf{DUSt3R*} \\

  \rowlabf{Whole heads} &
  \includegraphics[width=\imwidth]{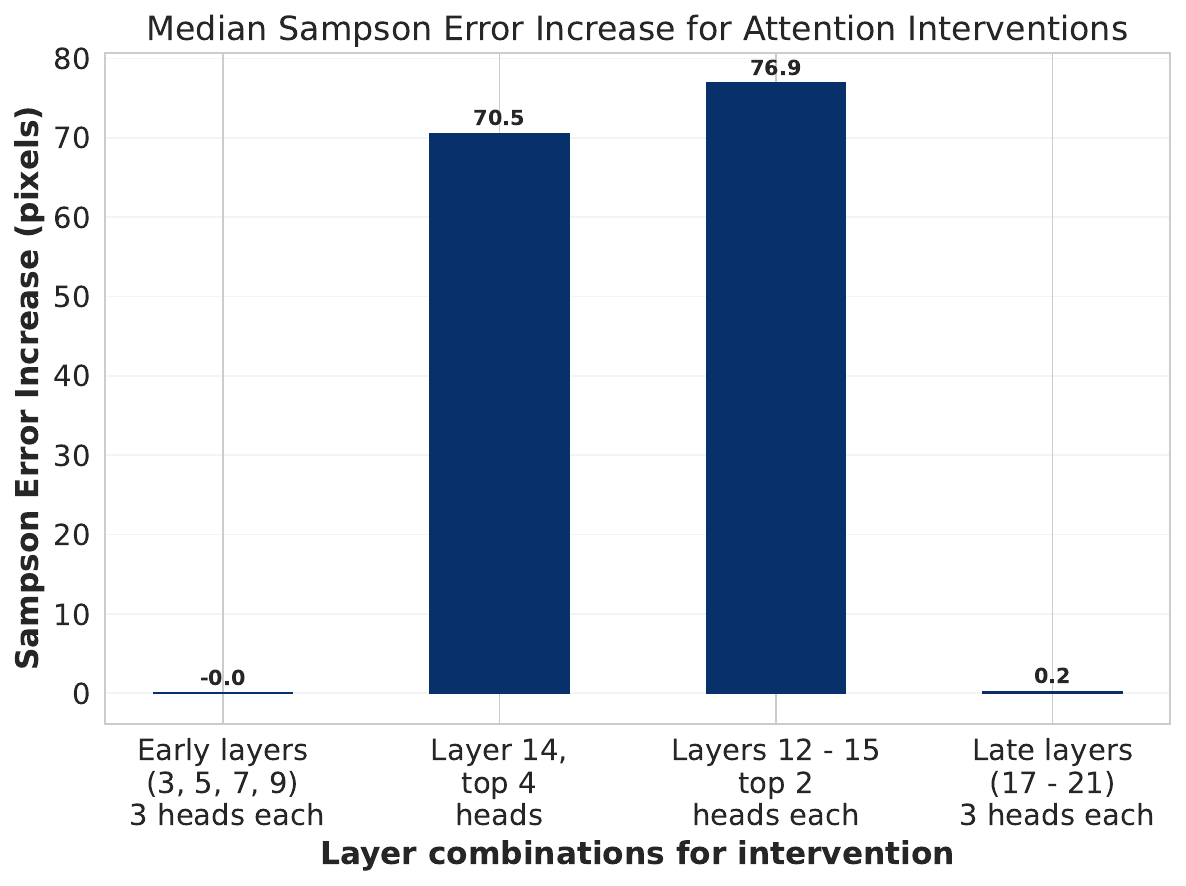} &
  \includegraphics[width=\imwidth]{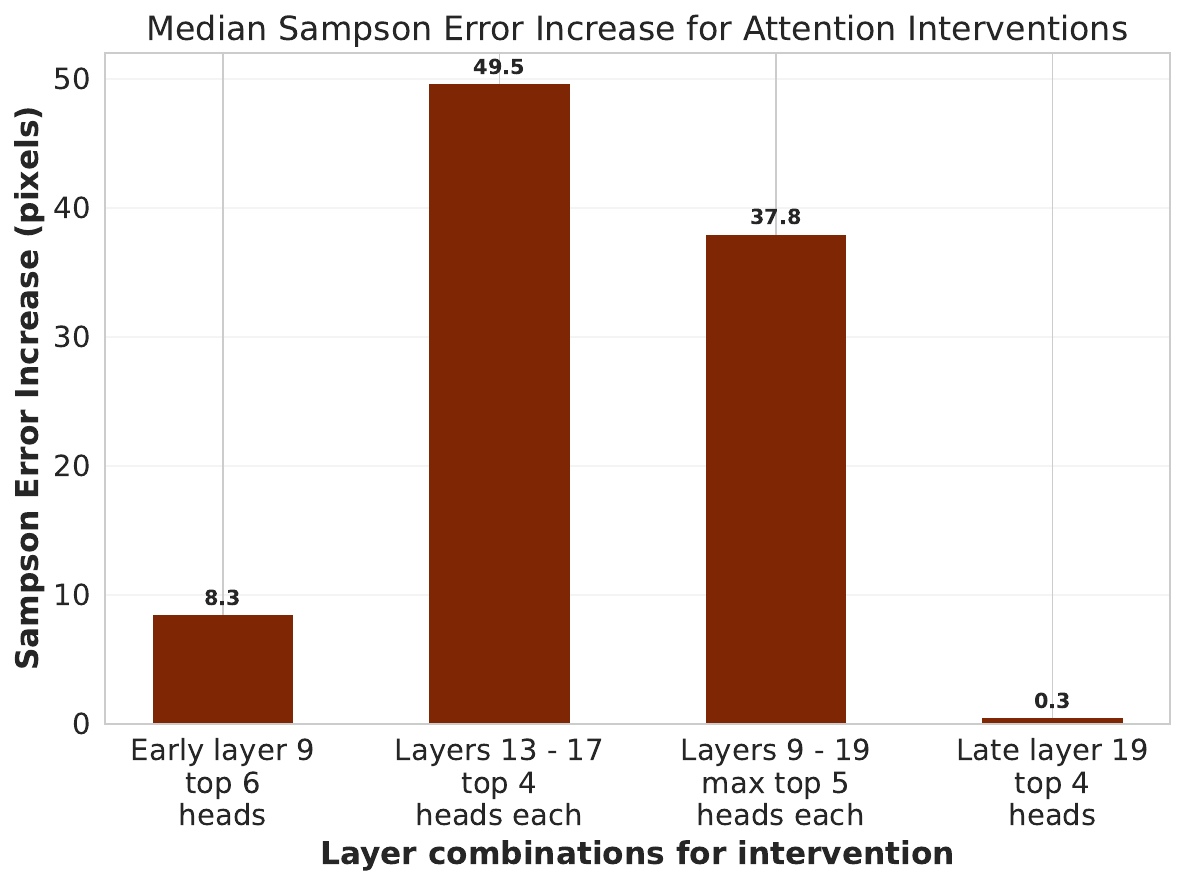} &
  \includegraphics[width=\imwidth]{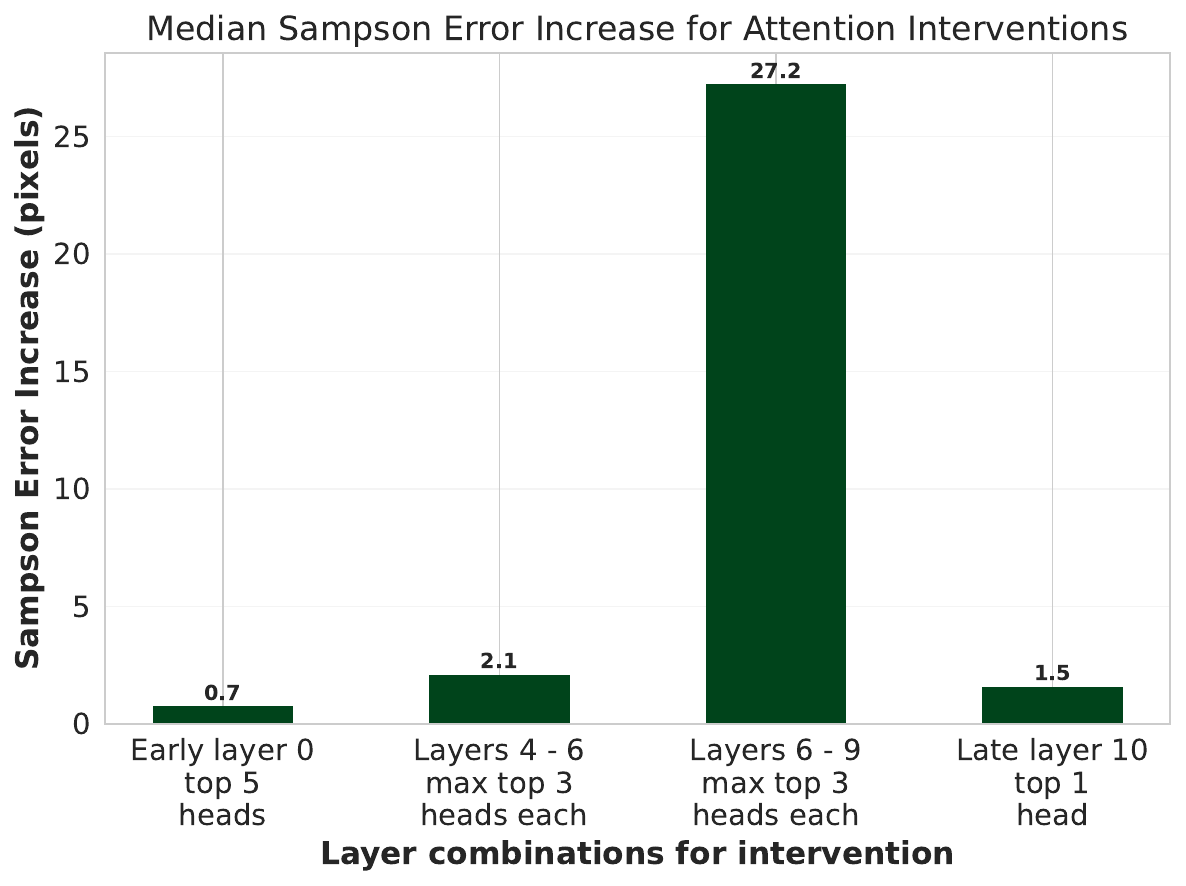} \\

  \rowlabf{Image map} &
  \includegraphics[width=\imwidth]{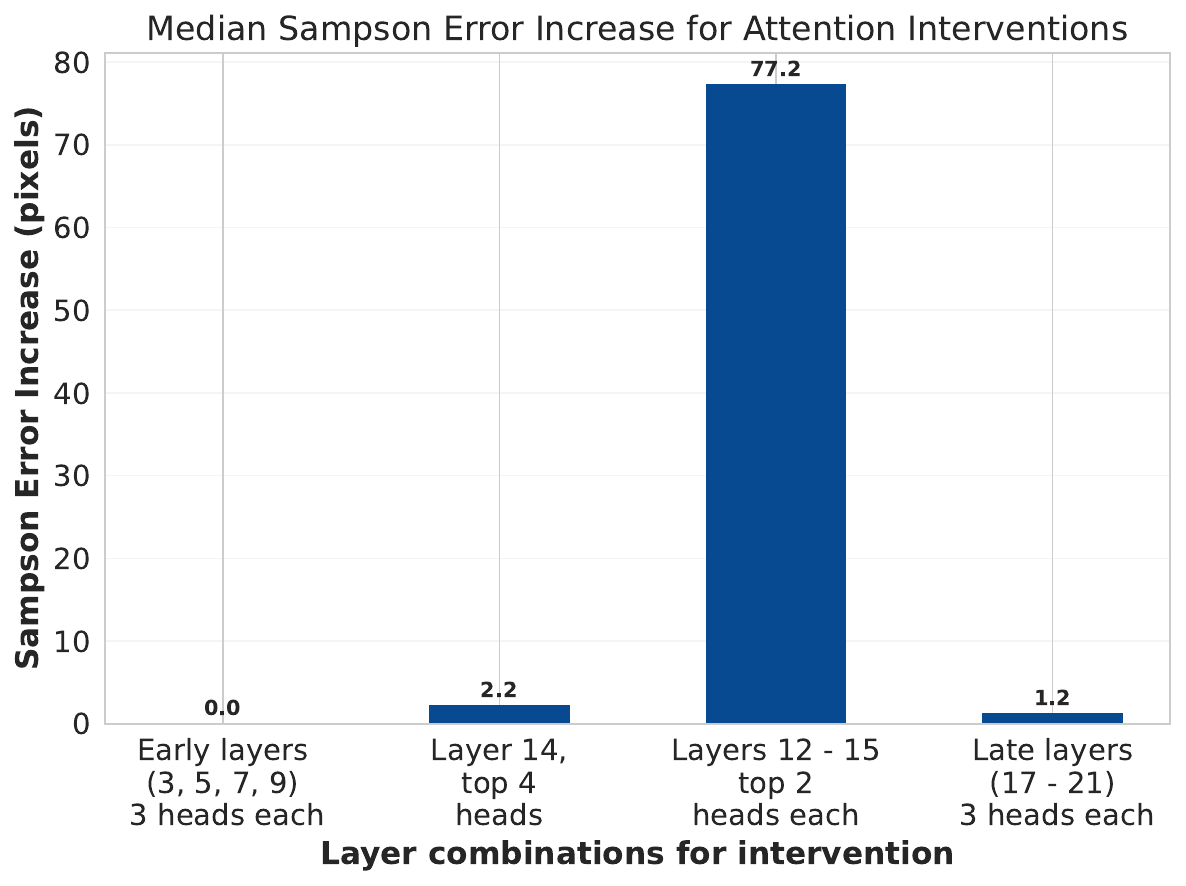} &
  \includegraphics[width=\imwidth]{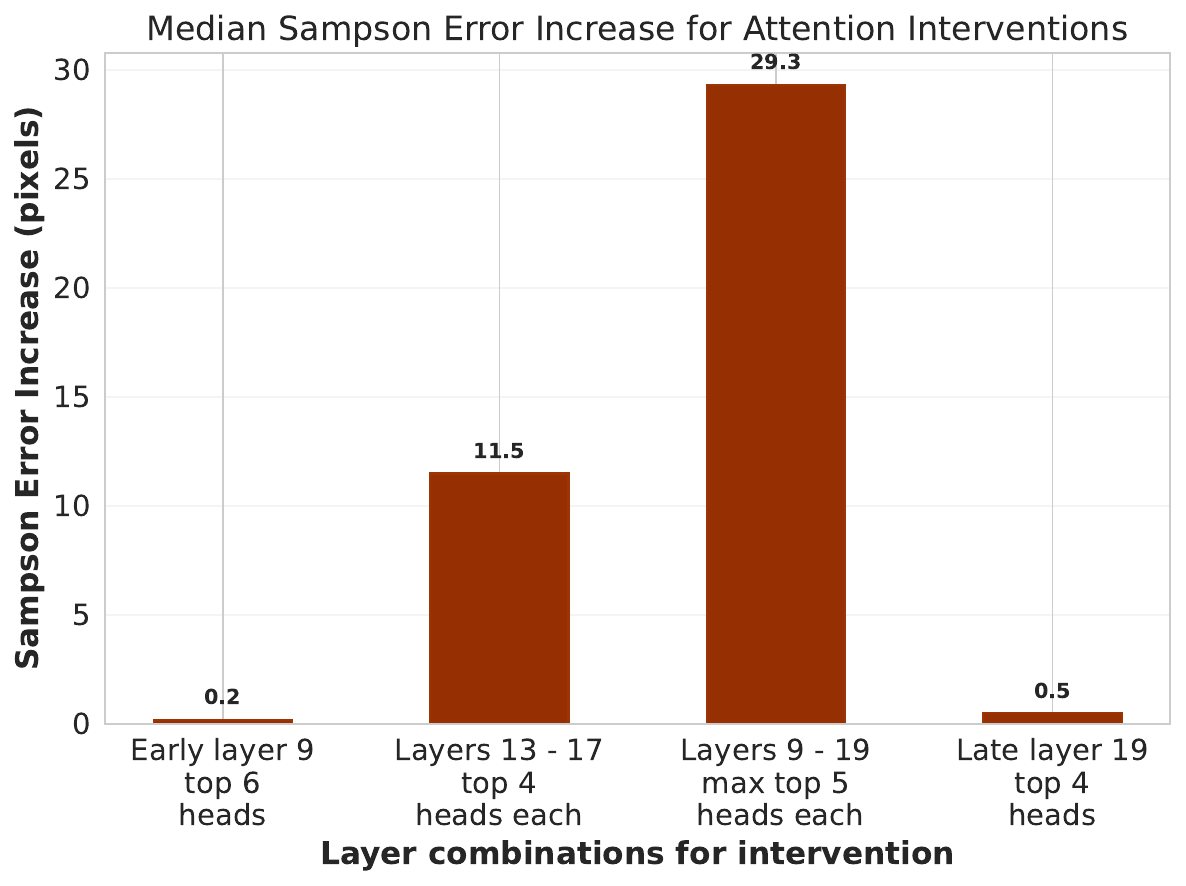} & 
  \includegraphics[width=\imwidth]{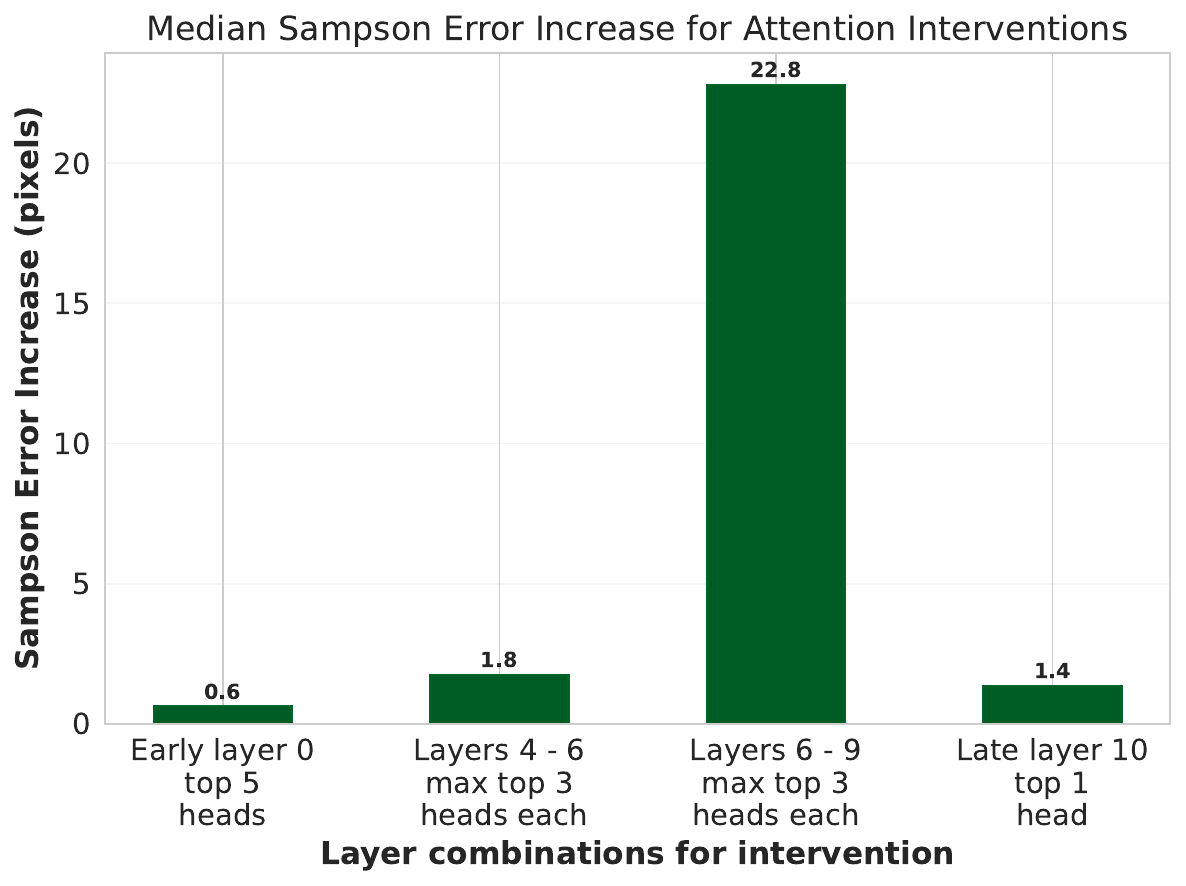}\\

  \rowlab{Block} &
  \includegraphics[width=\imwidth]{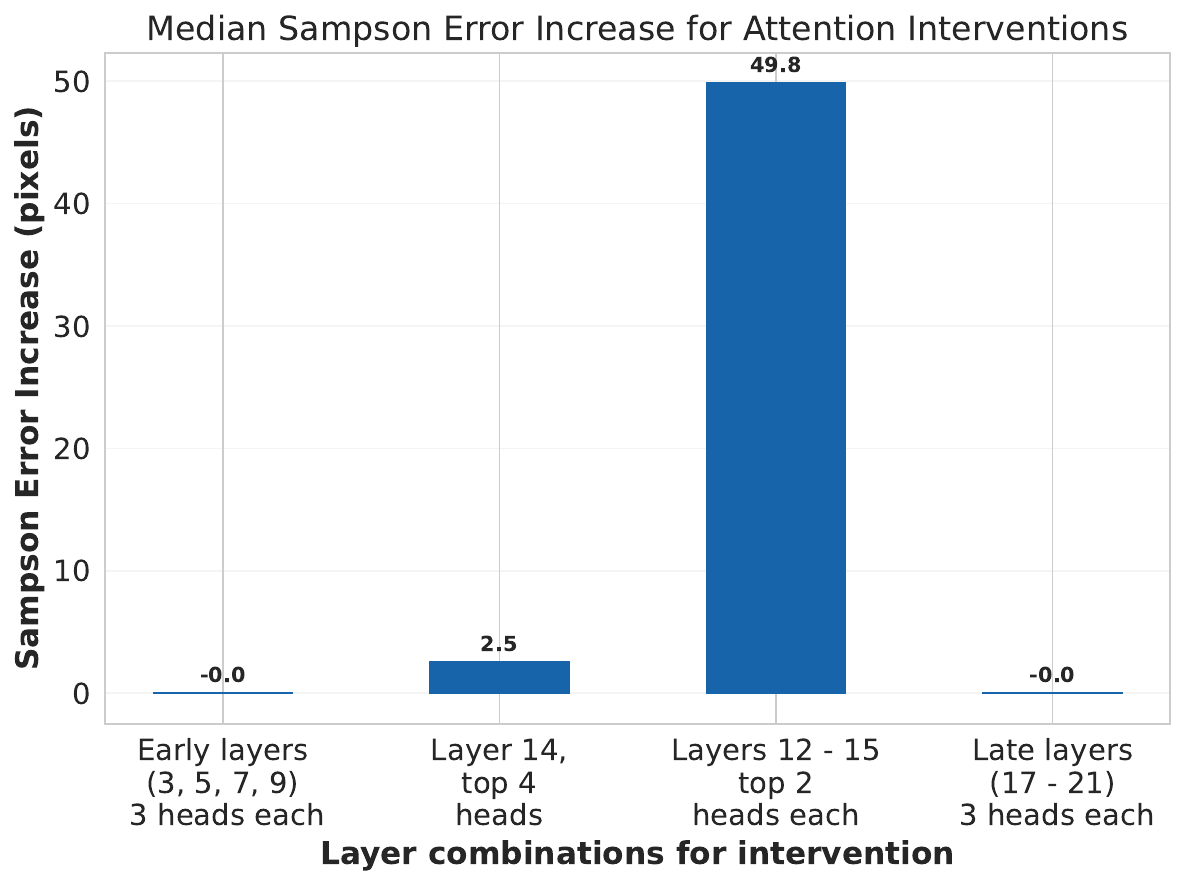} &
  \includegraphics[width=\imwidth]{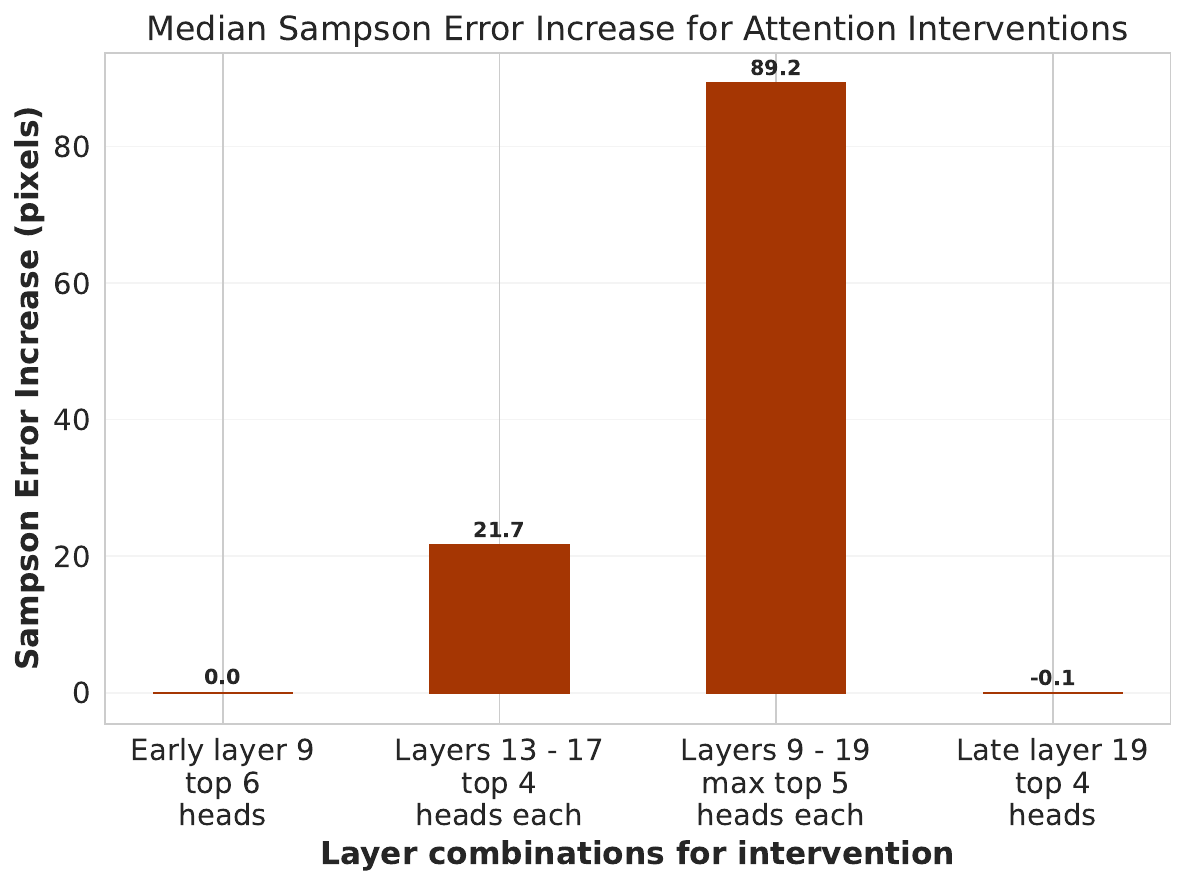} &
  \includegraphics[width=\imwidth]{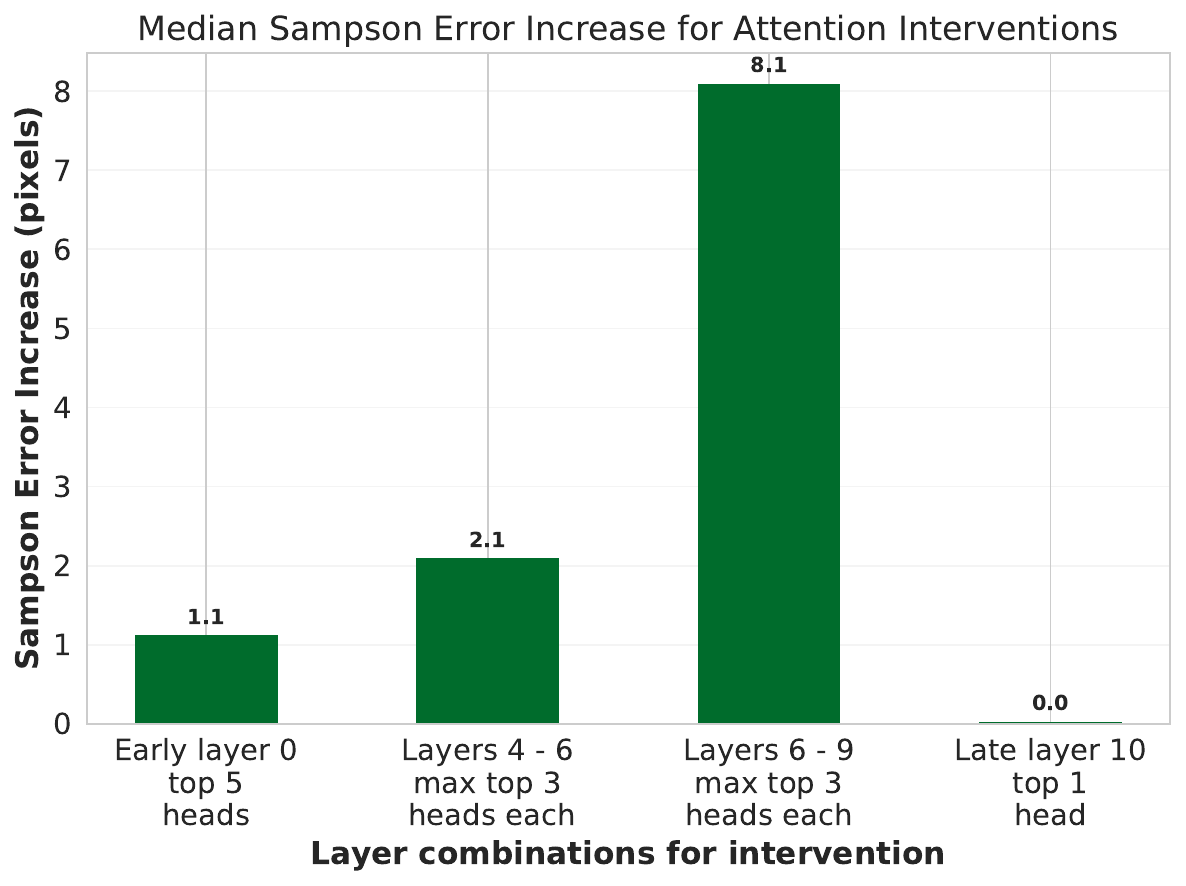} \\

  \rowlab{Patch} &
  \includegraphics[width=\imwidth]{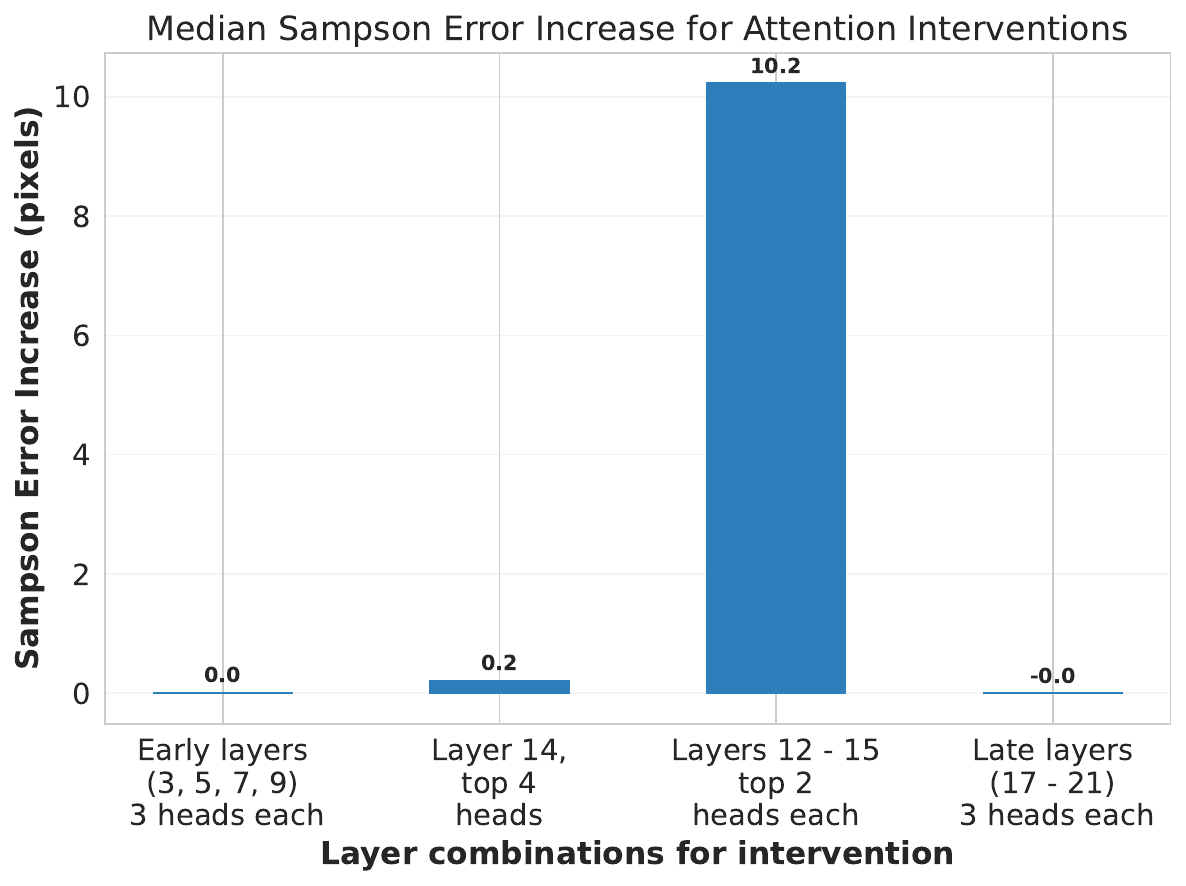} &
  \includegraphics[width=\imwidth]{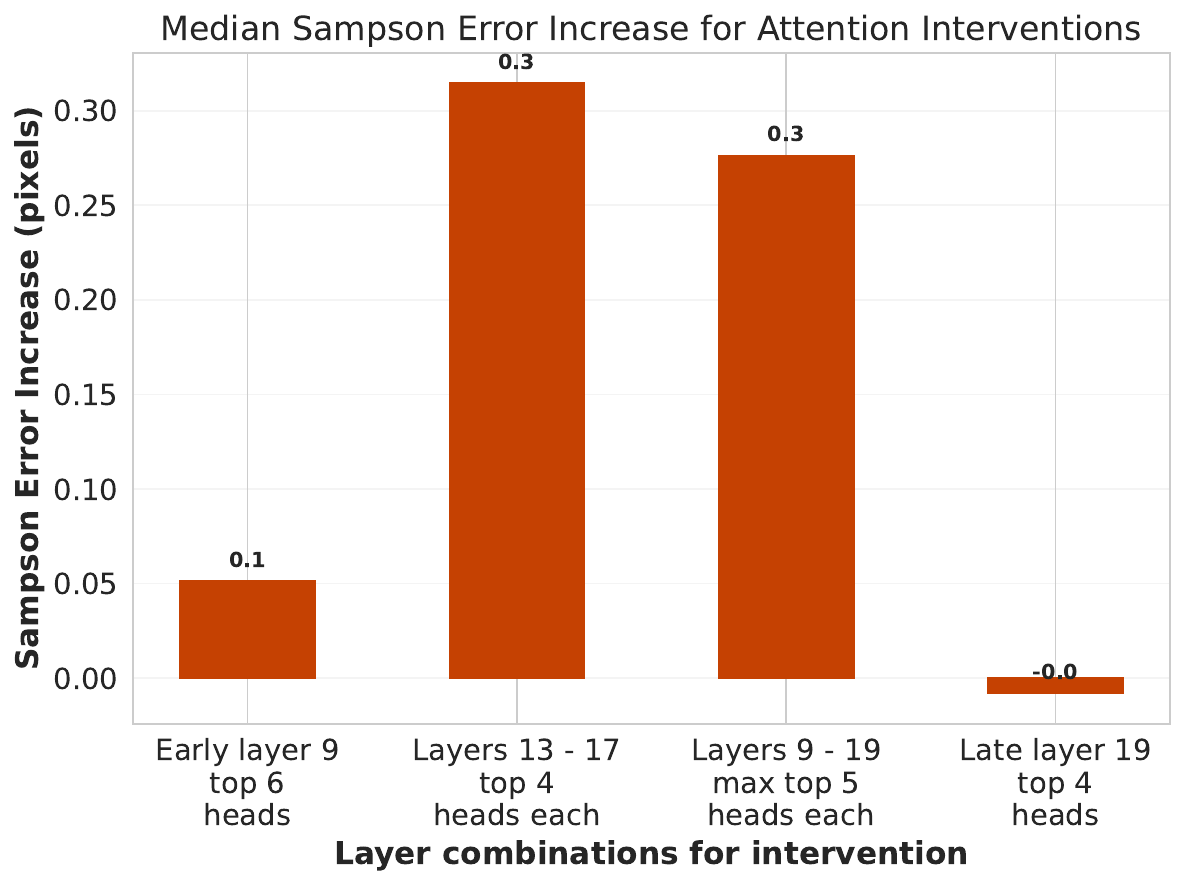} &
  \includegraphics[width=\imwidth]{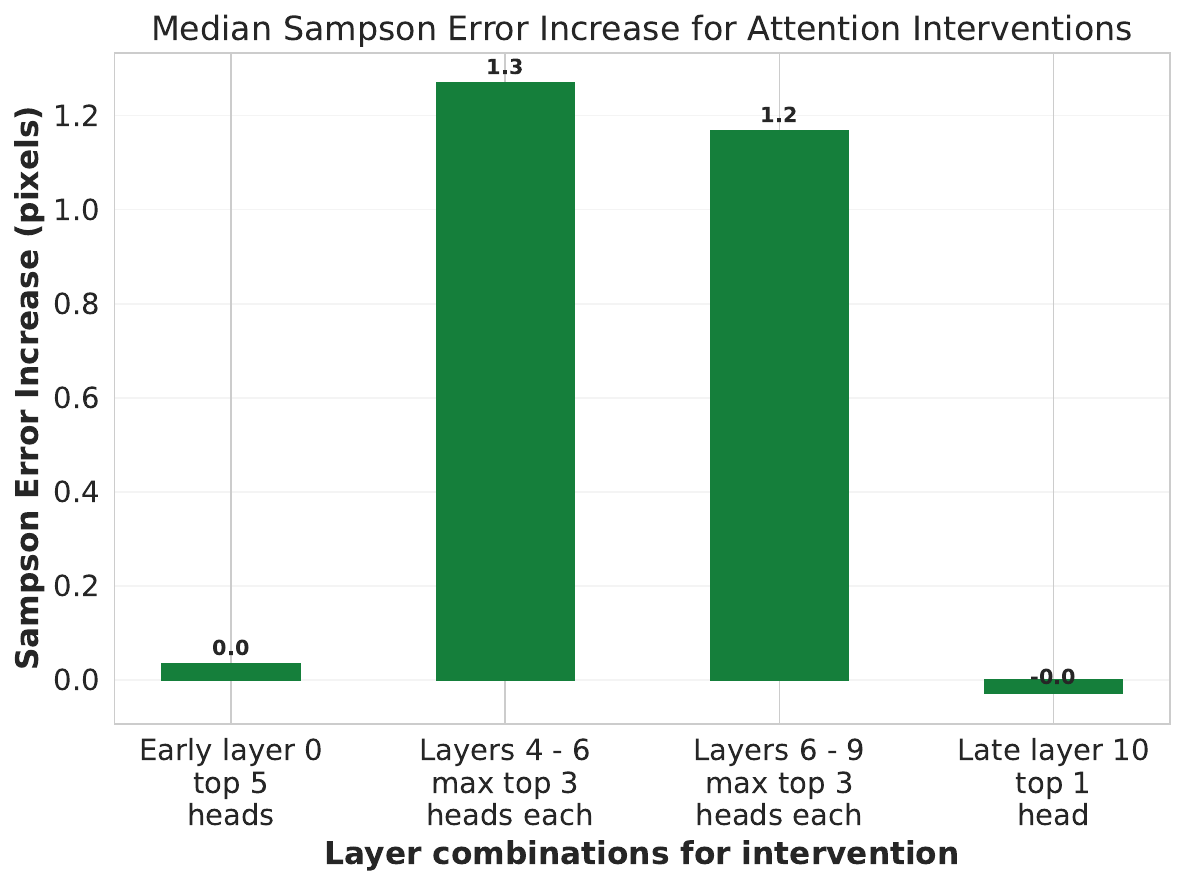} \\
\end{tabular}

\caption{\textbf{Performance degradation under attention interventions on 4 localities.} Disrupting random early or late layers has minimal impact on performance, whereas intervention on high correspondence-matching-layers cause significant degradation. The degradation is consistent across all models and at all levels of locality.  *For \duster, we report results only for valid cases for Image map and Whole heads, since 72\% of cases result in invalid predictions.} 
\label{fig:attn_intervention_results}
\end{figure}

\textbf{Experiment setup.} We intervene on the QK attention space using attention knockout, a method similar to Geva et al.~\cite{geva-etal-2023-dissecting}, where we zero out activations for specific layer-head combinations of the attention map for different levels of locality, ranging from zeroing out the whole attention head (\textit{Whole head}), to the most local with just the corresponding patches (\textit{Patch} in~\cref{fig:attn_intervention_results}).  We select interventions based on the high correspondence-matching performance of the layer-head combinations and two baselines of random early and late layers.

To perform the analysis, we require dense correspondence ground truth. There, we conduct the experiment on our controlled ShapeNet data and report median results across all sampling modes at a 50 mm focal length over the test set of 5 categories. Additional combinations, details on the intervention locality, and results are provided in the supplementary material. 
We report the difference in the Sampson distance between the baseline and the intervened model, where we approximate the fundamental matrix using the model's camera pose predictions. For \duster, this requires post-processing steps that can result in invalid camera poses, which we consider a failure case.

\textbf{Results.} ~\Cref{fig:attn_intervention_results} demonstrates clear causal effects consistently for all three models and different levels of locality. 
In \vggt~and \da, disrupting the high-performing attention heads in the middle layers degrades the Sampson distance error to such a large degree that it can be regarded as a complete failure. Intervening on random early or late layers, on the other hand, has no significant impact. 
There are two types of effective interventions:
(1) multiple heads within a single layer, and (2) fewer heads distributed across multiple layers. These correspond to the middle bars in \cref{fig:attn_intervention_results}. For instance, in \vggt, disabling only four heads in a single layer already leads to substantial degradation.

However, \duster~exhibits a slightly different behavior from \vggt~and \da. 
Intervening in the early layers significantly affects performance, consistent with the strong correspondence-matching observed in those layers. 
Thus, while the layer location differs, the overall trend remains aligned with the other models.
Moreover, for the first two intervention types (\textit{Whole heads} and \textit{Image map}), where we zero out the entire head or the full attention map (keys set to 0), \duster\ frequently produces invalid camera pose predictions. 
This suggests that its cross-attention mechanism is more sensitive to disruption than the global attention used in \vggt\ and \da\ in this setting.

These results support our hypothesis: disabling attention heads that exhibit strong correspondence matching leads to significant degradation in the encoded epipolar geometry. However, this does not conclusively prove that correspondence matching alone is responsible for the emergence of epipolar geometry. 
These heads may also encode additional mechanisms that contribute to the effect. This becomes evident in more localized interventions: while the same degradation trend persists, but the error scale change. Even when exact patch-to-patch correspondences in the QK space are removed, nearby patches still provide sufficient signals for the model to partially recover geometric information.

\begin{geobox}
    \textit{Intervening on the attention patterns for point correspondences shows a causal relationship between those attention heads and the epipolar geometry representation by the model.}
\end{geobox}

\section{The role of learned priors in VGGT} \label{sec:data_point}

While our geometric analysis includes three models, we focus on VGGT here to study how much robustness arises from learned priors rather than geometry. We consider VGGT a strong representative model for this study because it provides a cleaner experimental setup and operates without pose-conditioning at inference, unlike DA3. 
Additionally, there is no clear indication of a fundamental difference between VGGT and DA3; therefore, we focus on VGGT to get deeper insights. 

\begin{figure*}[bp!]
    \centering
    \includegraphics[width=\textwidth]{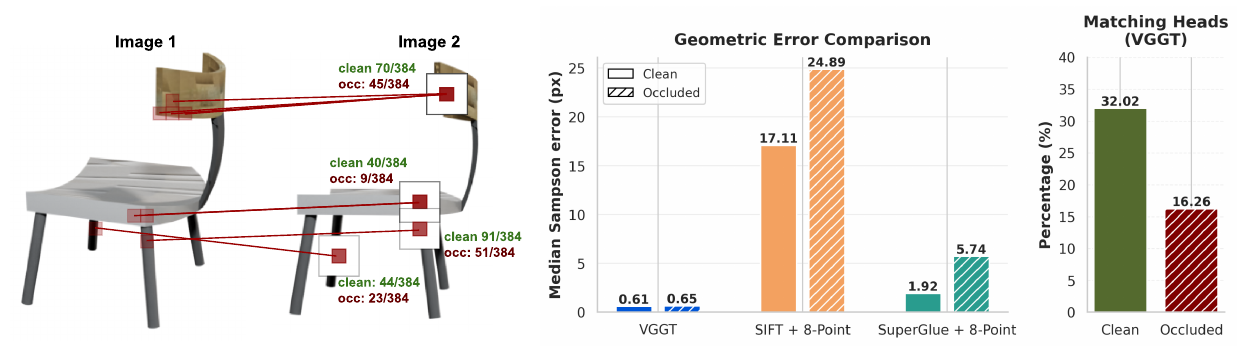}
    \caption{\textbf{Occlusion experiment.} \textbf{(Left)} We occlude image two from a pair with white patches and report the correspondence matching ability in all layers only for the masked patches. The occluded model shows lower matching performance yet remains above zero. \textbf{(Center)} We measure the changes in the Sampson error for clean and occluded image pairs where VGGT has a minimal degradation in performance, but others are significantly affected. \textbf{(Right)} We show the average number of heads with correspondence matching on occluded patches for the clean and occluded models.
    }
    \label{fig:inputimage}
\end{figure*}

Recent models such as VGGT and DA3 are trained on large datasets and leverage pretrained self-supervised models such as DINOv2. 
This raises an important question: to what extent does the model’s advantage come from data-based appearance priors and pre-training priors, and how well do these capabilities generalize beyond the training distribution?
To explore this, we evaluate the robustness of VGGT’s geometric predictions under controlled input perturbations.  We consider (i) appearance-based 2D perturbations, such as occlusions, and (ii) rendered 3D perturbations with structural ambiguities, light-source, and camera configuration changes. To assess the model's robustness, we compare VGGT against classical geometry-based methods and learned correspondence methods.

\subsection{How well does VGGT handle occlusions?}

To evaluate how well VGGT performs under missing visual information, we examine the impact of occlusions on correspondence matching and resulting geometric predictions. 
We simulate occlusions by masking key regions of the image. In principle, the model could address this challenge in multiple ways: by utilizing global geometric reasoning, inferring missing information from learned priors, or simply ignoring the occluded areas.

\textbf{Experiment setup.}
We intervene on input images by randomly selecting patches with known correspondences and masking them along with their surrounding regions (3$\times$3 patch neighborhoods); see \cref{fig:inputimage}. We perform inference twice: first on the clean image pair and then on a pair in which one image is partially occluded.  
We report the change in Sampson distance error (in pixels) and the number of attention heads in which the correspondence is matched, averaged over all intervened correspondences in a scene. 
Results are averaged over 40 scenes at a focal length of 50 mm, with small viewpoint differences (10-25 degrees) in our controlled ShapeNet dataset. We also compare VGGT's performance with the 8-point algorithm on all matched correspondences from SIFT~\cite{lowe2004distinctive} and SuperGlue~\cite{sarlin20superglue}.

\textbf{Results.}
\Cref{fig:inputimage} (left) shows the results for one case in which we masked four object patches for occlusion analysis, together with their corresponding ground-truth patches. 
For each occluded patch, we show the number of heads in which the occluded correspondence was matched (green for the clean version and red for the occluded version). 
Correspondence matching, measured by the number of matched heads (a maximum of 24 layers $\times$ 16 heads = 384 heads), shows that matched heads decrease for corrupted patches but do not collapse entirely. 
\Cref{fig:inputimage}~(right) reports the average number of matching heads before and after occlusion for the occluded patches. 
While there is a clear drop in matching heads, approximately 50\% of heads are retained, indicating that the model does not ignore occluded regions but recovers correspondences through some inpainting behavior. This phenomenon may arise from the mask-based IBOT~\cite{zhou2021ibot} loss in DINOv2~\cite{oquab2023dinov2}, which is trained to recover the masked content.

Additionally, VGGT shows only a slight degradation in the Sampson error, indicating that the epipolar geometry is still restored, whereas the other multi-stage baselines show clear degradation, shown in \Cref{fig:inputimage}~(center).

\begin{databox}
\textit{VGGT shows robustness to occlusions, preserving epipolar geometry by partially recovering matching heads for the occluded correspondences.}
\end{databox}

\subsection{How robust is VGGT to scene perturbations?}

We investigate the role of learned priors under controlled 3D perturbations that introduce structural ambiguity (e.g., symmetric and repetitive objects) and variations in lighting, focal length, and camera configuration. 
As expected, we observe that VGGT remains stable under moderate appearance and viewpoint changes, including focal length variations, with only limited degradation in geometric accuracy. 
Notably, when we generate challenging scenes with repetition and symmetry, for example, identical objects arranged in a circular pattern, classical pipelines struggle to establish reliable correspondences, whereas VGGT remains comparatively robust when asymmetric lighting introduces subtle cues that help disambiguate the scene. In contrast, under fully symmetric lighting and object configurations, where no disambiguating signals are present, all methods fail as expected, confirming that the task is geometrically underconstrained.

Overall, these results indicate that VGGT benefits from learned data-based appearance priors to resolve partial ambiguities, outperforming both classical and learned correspondence-based methods in such regimes, while still respecting the fundamental limits imposed by multi-view geometry when no disambiguating evidence exists. Detailed experimental setups, quantitative results, and additional analyses are provided in the supplementary material.

\begin{databox}
    \textit{VGGT leverages learned visual cues to resolve partially ambiguous scenes, but like classical methods, it fails when the scene is fully geometrically symmetric.}

\end{databox}

\section{Discussion} \label{sec:discussion}

Our geometric analysis provides evidence that modern feed-forward 3D reconstruction models such as DUSt3R, VGGT, and Depth Anything 3 encode epipolar structure in their intermediate layers. 
A closer examination of these layers shows that several attention heads also capture point correspondences across views. The strong alignment between these emergent correspondences and geometric constraints suggests that these models organize information (at least partially) geometrically. Intervening on the most involved attention heads further confirms their functional role in encoding geometry.

While the overall behavior is consistent across the models, we observe small differences in where exactly this information emerges. In VGGT and DA3, which use global attention in the decoder and predict in a unified 3D space, correspondences and geometric information appear in the middle layers. DUSt3R, in contrast, encodes this information early in the decoder layers. These results show that models share similar geometric mechanisms, but their architecture changes where this encoding occurs.

Our analysis of learned data priors reveals that VGGT partially fills in missing correspondences in scenes with occlusions. While input perturbations degrade both correspondence matching and geometric accuracy, the degradation is not significant. Particularly, the correspondence matching performance remains above zero in the occluded patches.
This non-zero matching indicates that the model hallucinates missing correspondences rather than ignoring the occluded areas. 
While hallucinations could be beneficial if correct, it is fundamentally impossible to resolve single-view ambiguity when one of the two images is occluded, making this behavior potentially undesirable for reconstruction tasks. At the same time, VGGT remains robust in scenes with local feature ambiguity, leveraging global context and subtle cues such as shadows or lighting that classical pipelines fail to exploit.

\section{Conclusion} \label{sec:conclusion}

We presented a systematic study of modern feed-forward 3D reconstruction models, focusing on DUSt3R, VGGT and Depth Anything 3, to determine whether they employ interpretable geometric concepts or rely primarily on learned priors from data and model. 
Our results show that these models encode geometric structure and strongly indicate that they use point correspondences to do so. 
Further, our input perturbation experiments reveal that the models rely strongly on learned priors, which provide robustness to input variations but can also lead to undesirable hallucinations.

\section*{Acknowledgments}
This research was funded by the Deutsche Forschungsgemeinschaft (DFG, German Research Foundation) under grant number 499552394--SFB 1597 Small Data, and grant number 539134284, through EFRE (FEIH\_2698644) and the state of Baden-Württemberg. This research was also funded by the German Ministry for Economy and Climate Protection via a decision by the German parliament (19A23014R). The compute used in this project was also funded by the German Research Foundation (DFG) under grants 417962828 and 539134284. 
Jelena Bratuli\'{c} is part of the European Lab for Learning and Intelligent Systems (ELLIS) PhD program, and the research visit within the ELLIS PhD program was supported through the ELSA mobility program.
We want to thank Jianyuan Wang, Paul Englester, Orest Kupyn, Simon Schrodi, Karim Farid, and Rajat Sahay for valuable discussions and feedback during the development of this work. 
\begin{center}
\includegraphics[width=0.3\textwidth]{figures/BaWue_Logo_Standard_rgb_pos.png}
\includegraphics[width=0.3\textwidth]{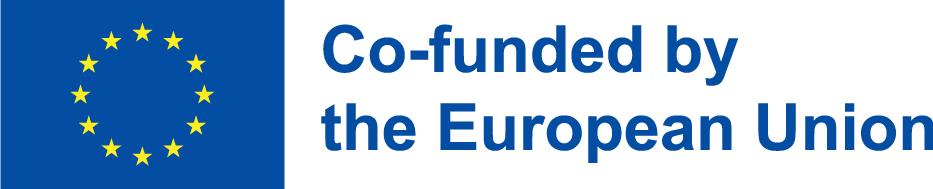}
\end{center}


\appendix

\section{Dataset details}\label{suppl:dataset}

\subsection{Real-world datasets}
\begin{figure}[t!]
    \centering
    \begin{subfigure}[b]{0.24\textwidth}
        \includegraphics[width=\textwidth]{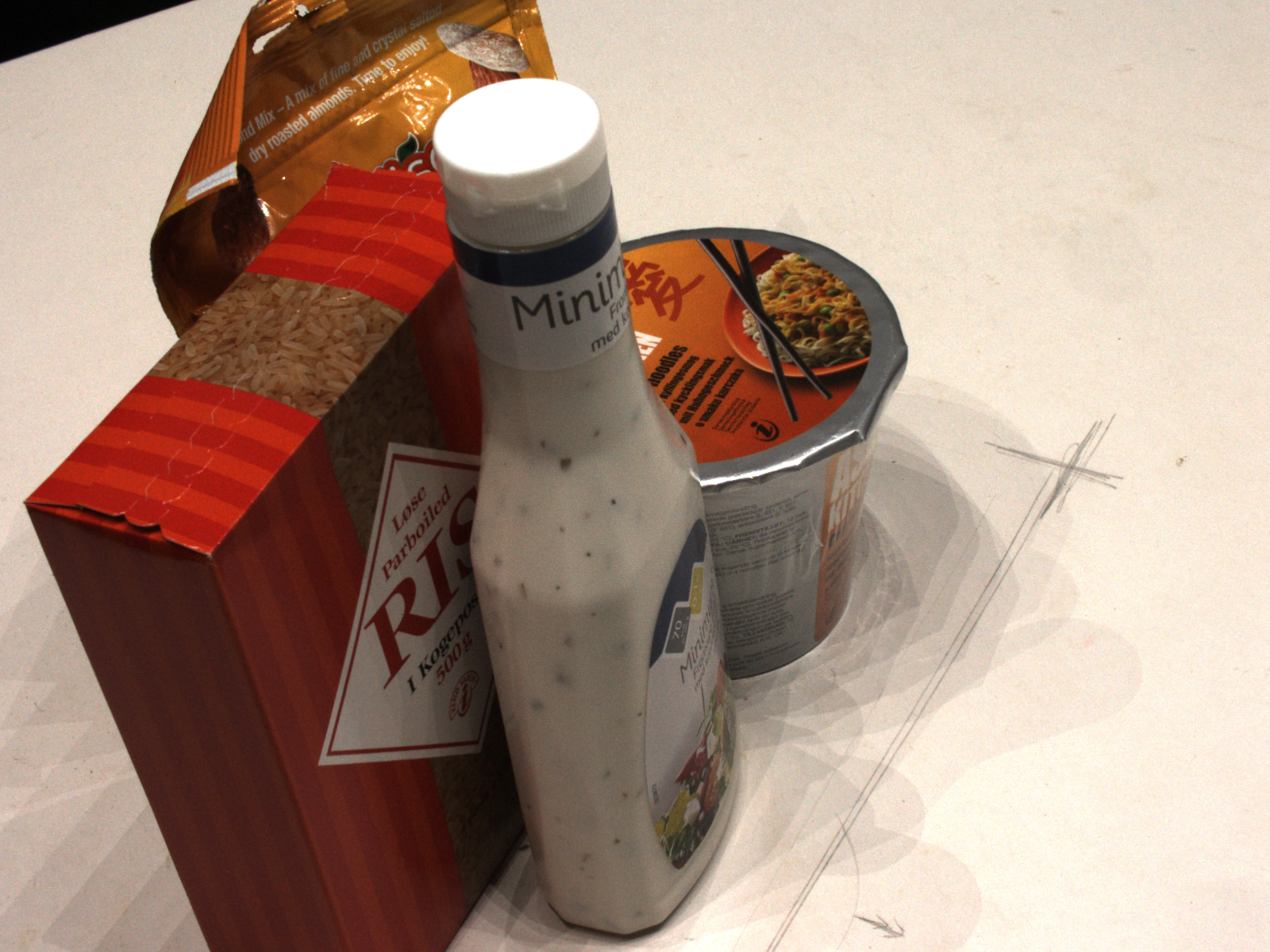}
    \end{subfigure}
    \begin{subfigure}[b]{0.24\textwidth}
        \includegraphics[width=\textwidth]{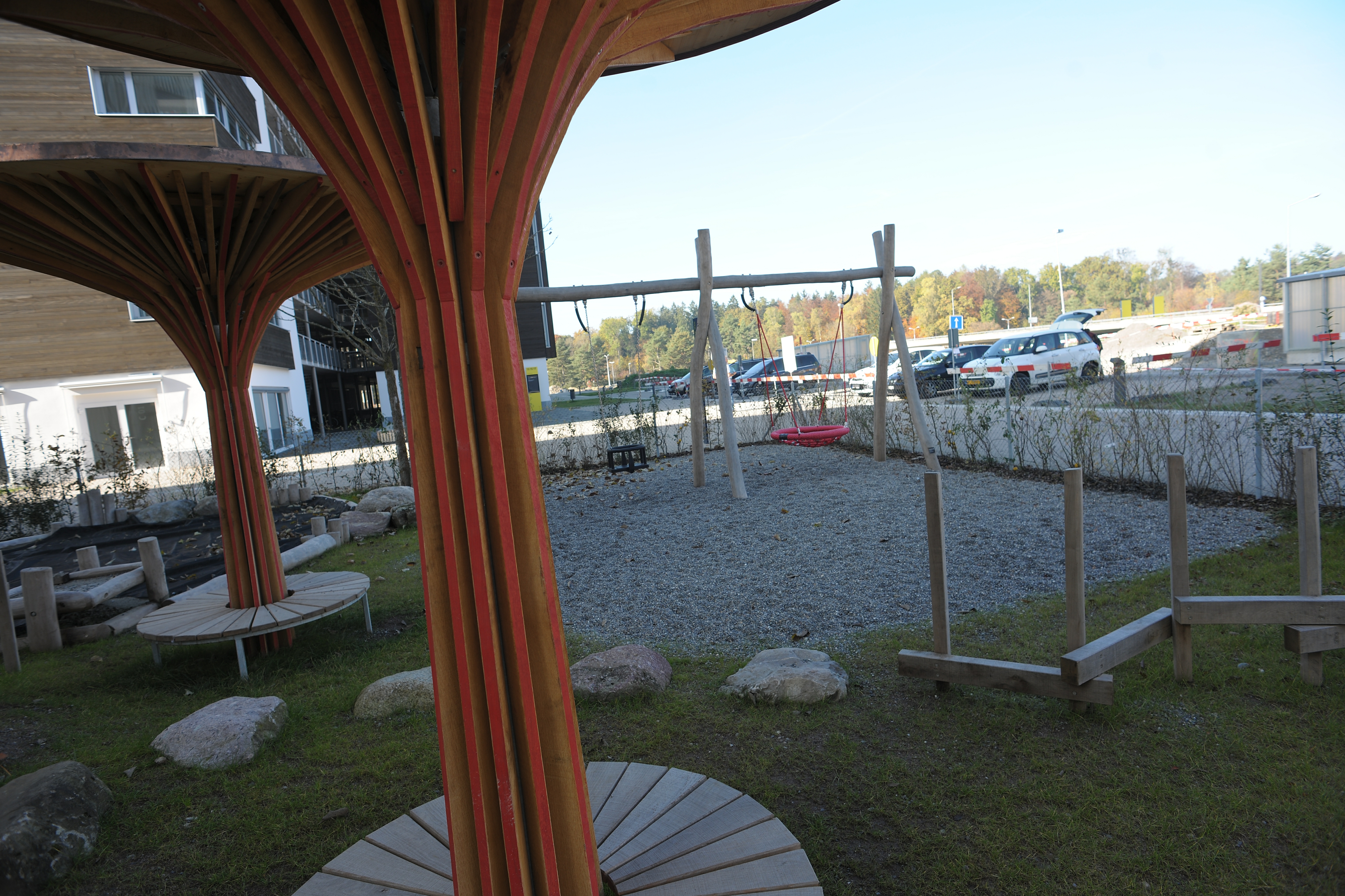}
    \end{subfigure}
    \begin{subfigure}[b]{0.24\textwidth}
        \includegraphics[width=\textwidth]{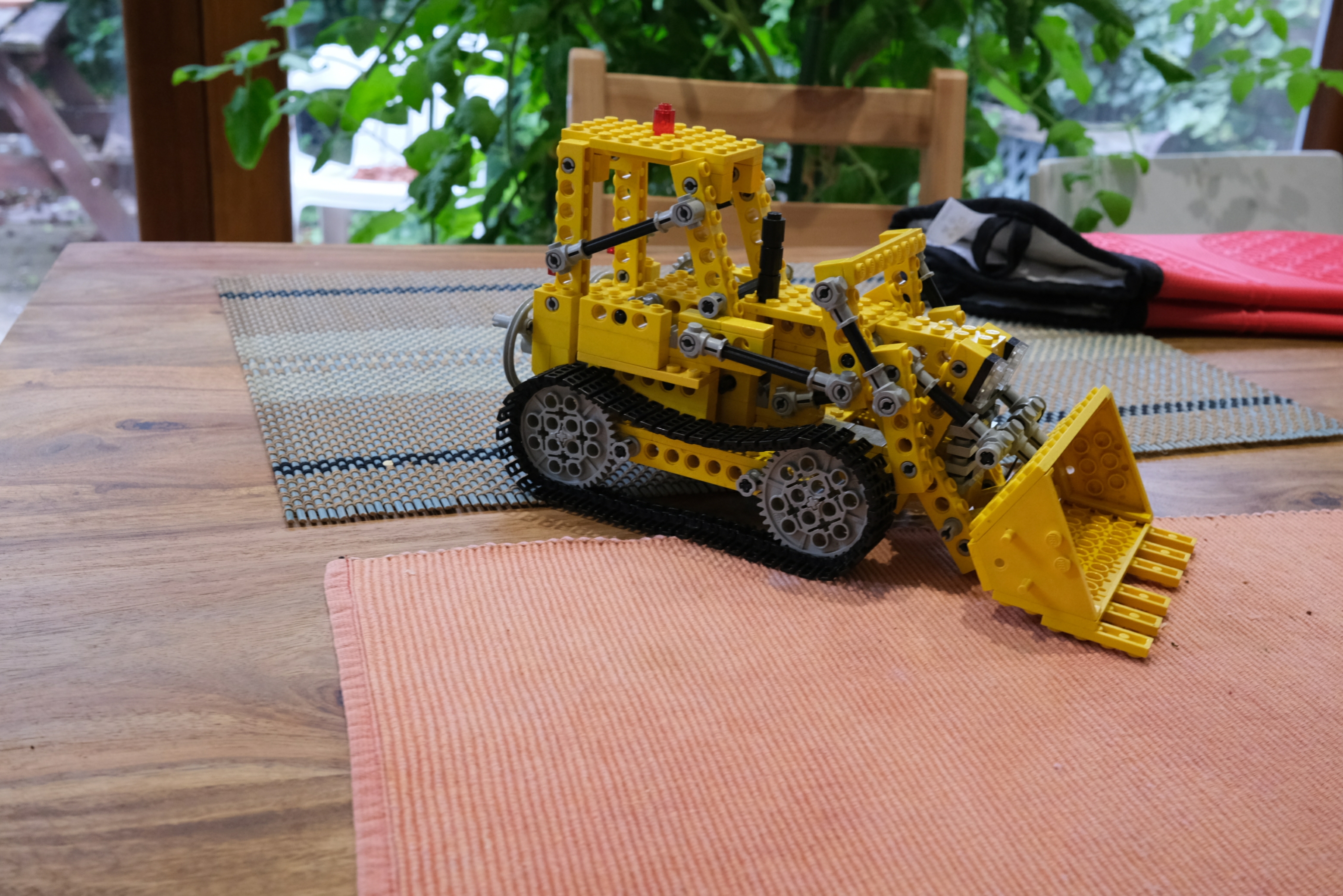}
    \end{subfigure}
    \begin{subfigure}[b]{0.24\textwidth}
        \includegraphics[width=\textwidth]{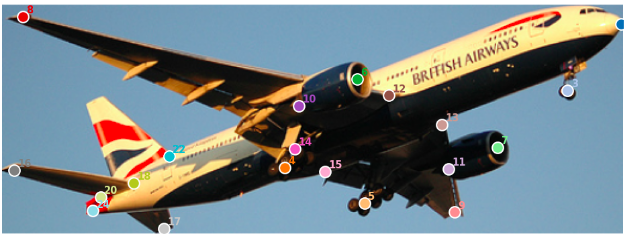}
    \end{subfigure}
    \begin{subfigure}[b]{0.24\textwidth}
        \includegraphics[width=\textwidth]{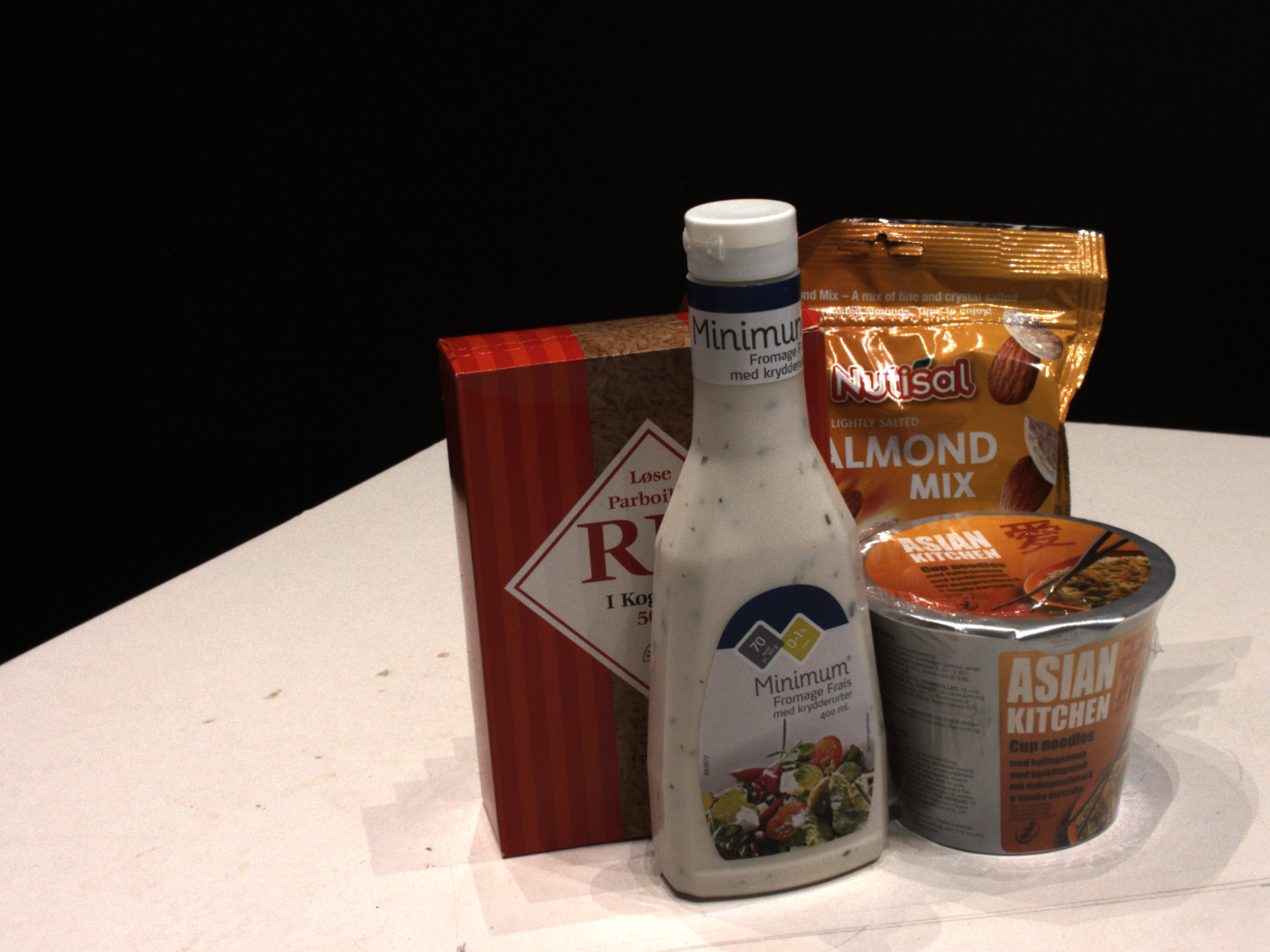}
    \end{subfigure}
    \begin{subfigure}[b]{0.24\textwidth}
        \includegraphics[width=\textwidth]{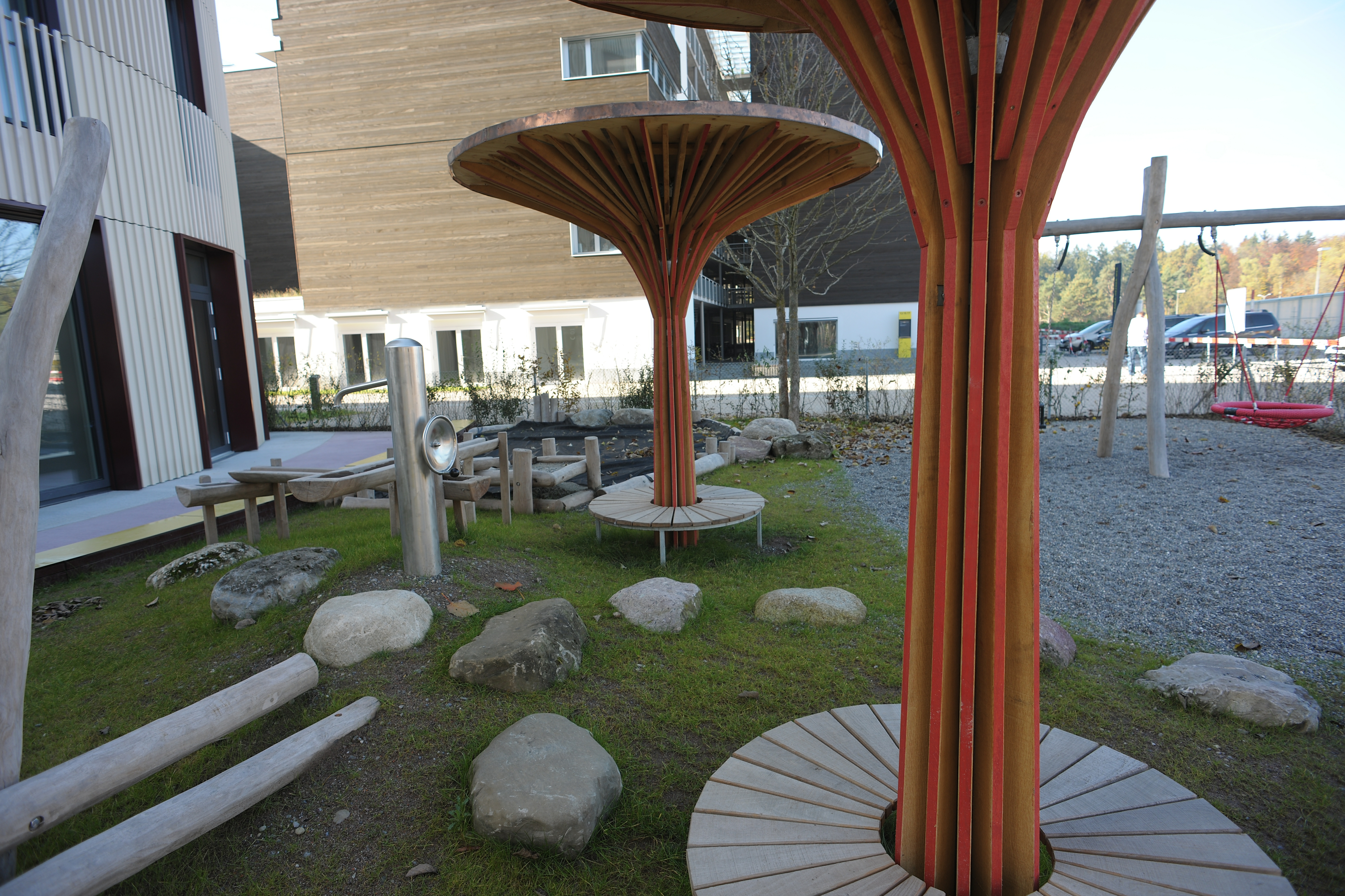}
    \end{subfigure}
    \begin{subfigure}[b]{0.24\textwidth}
        \includegraphics[width=\textwidth]{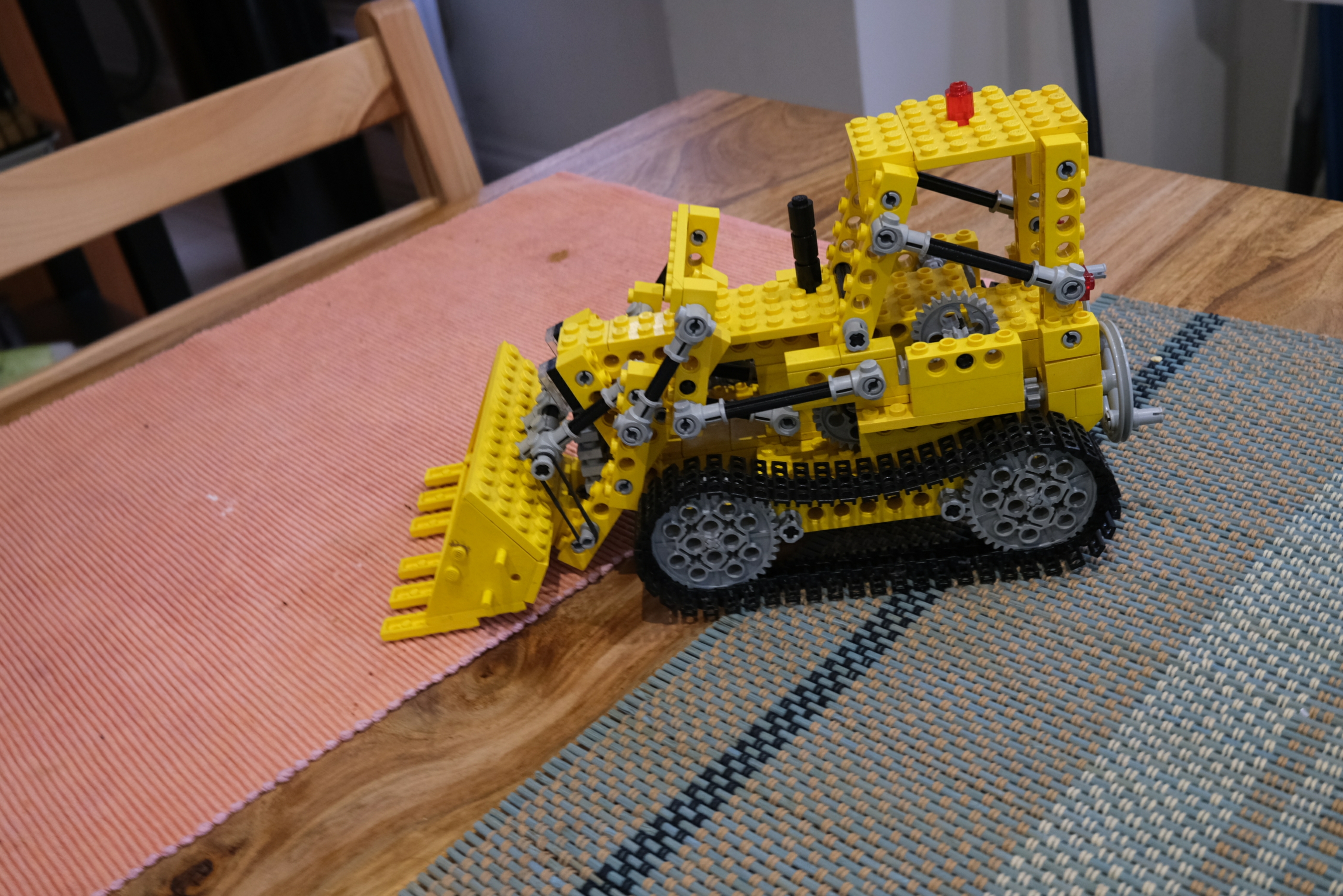}
    \end{subfigure}
    \begin{subfigure}[b]{0.24\textwidth}
        \includegraphics[width=\textwidth]{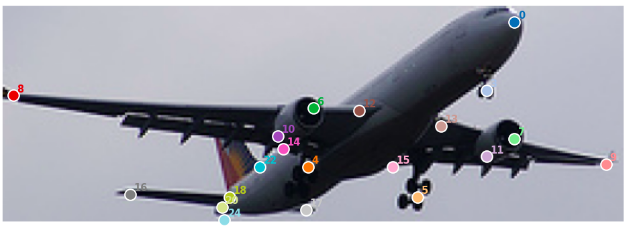}
    \end{subfigure}

    \caption{\textbf{Dataset examples used in analysis.} We leverage diverse real-world datasets:, DTU MVS objects~\cite{jensen2014large} (first column), ETH3D~\cite{schoeps2017cvpr} (second column), MipNeRF360~\cite{barron2022mipnerf360} (third column), and SPair71k~\cite{min2019spair} (last column).}
    \label{fig:real_data}
\end{figure}

We use 3 real-world datasets in our analysis, comprising object-centric and scenery scenarios with diverse camera poses. We show examples from each on~\cref{fig:real_data}.

\textbf{DTU MVS}~\cite{jensen2014large} is a dataset comprising 124 object-centric scenes photographed from 49 fixed positions under structured-light ground truth. We use depth maps from MVSNet~\cite{yao2018mvsnet} for dense correspondence calculation. We leverage official train, validation, and test sets. To ensure diversity in viewpoint separation while improving control over the data distribution, we bin image pairs by the true 3D angular distance between the cameras, using bins of 15°, 30°, 45°, 60°, 75°, 90°, 105°, and 120°. We sample 75 image pairs from each bin per scene.

\textbf{ETH3D}~\cite{schoeps2017cvpr} is a precise, high-resolution, multi-view dataset covering diverse indoor and outdoor scenes. We use only high-resolution scenes, using all the official training scenes and a subset of official validation scenes for training, and split the remaining validation scenes into a validation set (1 scene) and a test subset (3 scenes).  We include all pairs since the scenes, although diverse, contain only a relatively small number of images. Since we use part of the official validation scenes to train the probes, we use the official ground-truth data in COLMAP format and sparse correspondences.

\textbf{MipNeRF360}~\cite{barron2022mipnerf360} comprises nine outdoor and indoor scenes captured with a consumer camera. We use 7 scenes for training, 1 for validation, and 1 for testing. We use the official ground truth data for camera poses and sparse correspondences in COLMAP format. To ensure diversity in viewpoint separation while improving control over the data distribution, we bin image pairs by the true 3D angular distance between the cameras, using bins of 15°, 30°, 45°, 60°, 75°, 90°, 105°, and 120°. We sample 100 image pairs from each bin per scene. 

\textbf{SPair71k}~\cite{min2019spair} is a semantic correspondence dataset. For a given pair of images, where the images are from different scenes, we have annotations of semantic correspondences, such as the left wing of one bird to the left wing of another. We sample 350 pairs from the official test split based on the highest number of correspondences between the images and use them for our analysis to explain the role of layers in the network. We show an example with marked correspondences on~\cref{fig:real_data}.

\subsection{Synthetic ShapeNet dataset}
We generated a controlled synthetic dataset from ShapeNet~\cite{shapenet2015} assets rendered in Blender. We opted for a simpler synthetic dataset with complete control over the scenes since complex scenes with ambiguity could bias our analysis of the model's internal workings. Each scene was rendered with different camera configurations regarding focal length, angular distance between the cameras, and scene composition. 

We selected 13 different categories (bag, basket, bathtub, bed, bench, bottle, bowl, cap, car, chair, earphone, jar), where we define splits of \emph{unique} assets with distinctive features and \emph{symmetric} assets with object symmetry (basket, bottle, bowl, jar). 
For each category, we define splits for the probing experiment: 15 instances for training, 2 for validation, and 2 for testing, for a total of 10 categories (bottles, bowls, and symmetric jars are omitted). 

We further define smaller and larger testing splits, which we use for later experiments. Both testing splits are a combination of validation and testing splits used in the probing experiment. For the smaller version, we include only 5 instances from 5 categories (bed, cap, car, chair, and jar). For the larger version, we include all instances from the validation and test splits, yielding 40 instances across 10 categories. 
Finally, we also define another split with fully symmetric objects from the categories basket, bottle, bowl, and jar, which we use to create repetitive scenes.

\begin{figure}[b!]
    \centering
    \begin{subfigure}[b]{0.15\textwidth}
        \includegraphics[width=\textwidth]{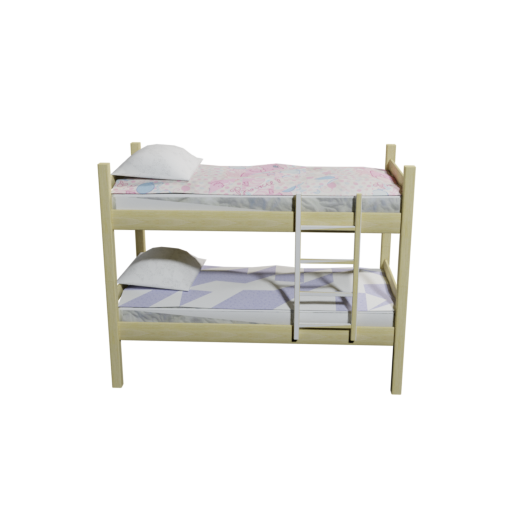}
    \end{subfigure}
    \begin{subfigure}[b]{0.15\textwidth}
        \includegraphics[width=\textwidth]{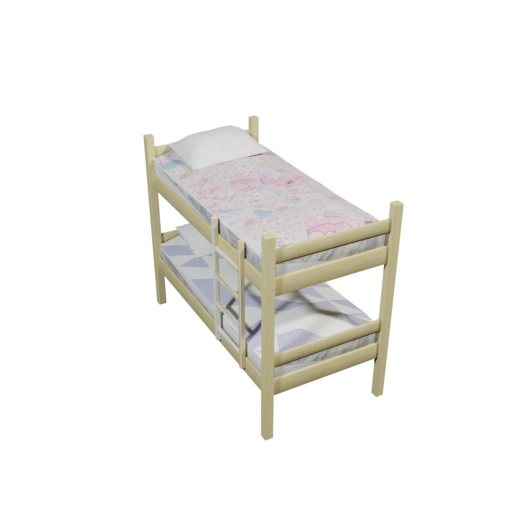}
    \end{subfigure}
    \begin{subfigure}[b]{0.15\textwidth}
        \includegraphics[width=\textwidth]{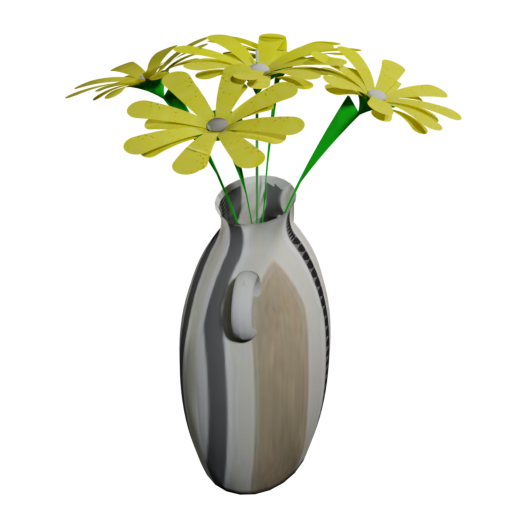}
    \end{subfigure}
    \begin{subfigure}[b]{0.15\textwidth}
        \includegraphics[width=\textwidth]{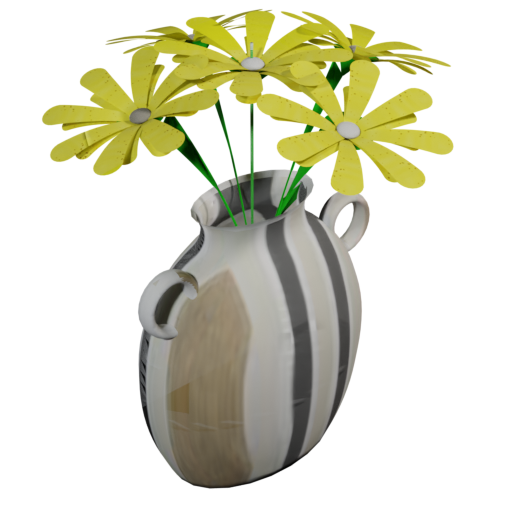}
    \end{subfigure}
    \begin{subfigure}[b]{0.15\textwidth}
        \includegraphics[width=\textwidth]{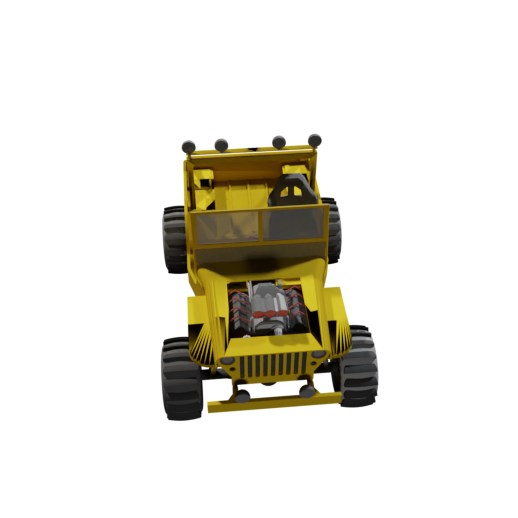}
    \end{subfigure}
    \begin{subfigure}[b]{0.15\textwidth}
        \includegraphics[width=\textwidth]{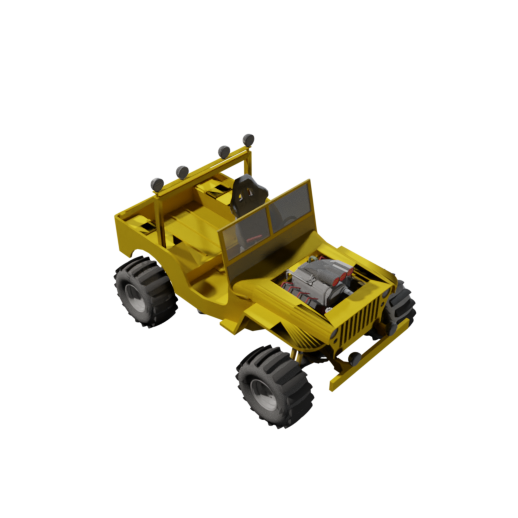}
    \end{subfigure}
    
     \begin{subfigure}[b]{0.15\textwidth}
        \includegraphics[width=\textwidth]{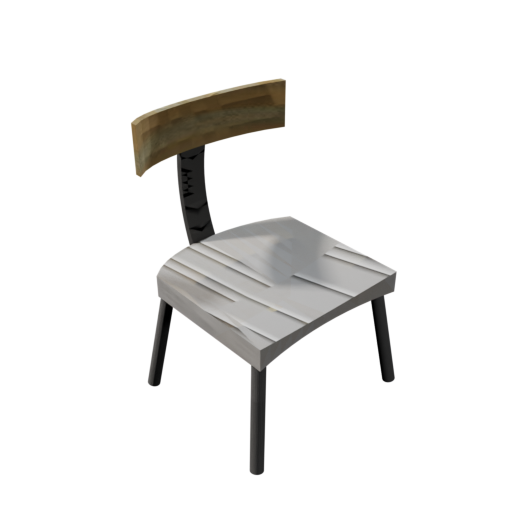}
    \end{subfigure}
    \begin{subfigure}[b]{0.15\textwidth}
        \includegraphics[width=\textwidth]{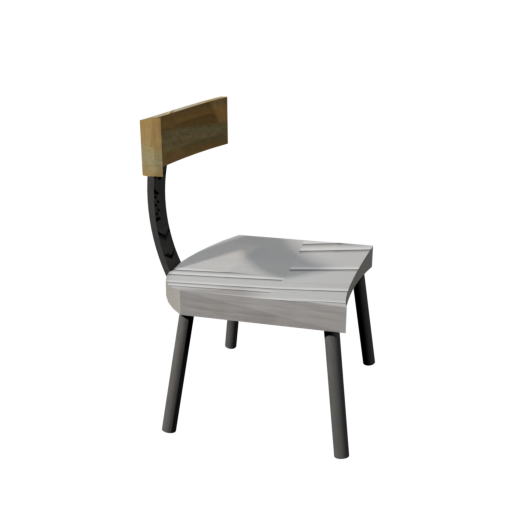}
    \end{subfigure}
    \begin{subfigure}[b]{0.15\textwidth}
        \includegraphics[width=\textwidth]{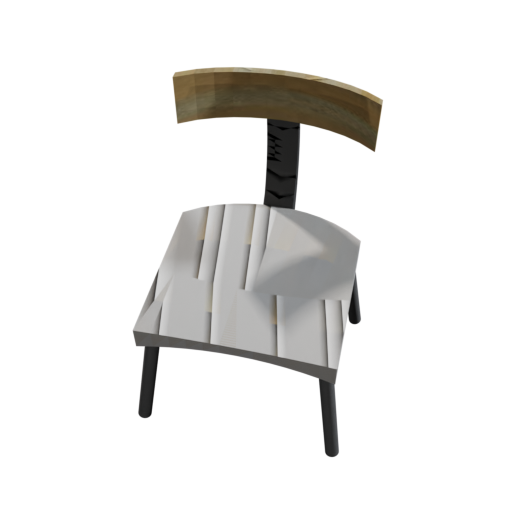}
    \end{subfigure}
    \begin{subfigure}[b]{0.15\textwidth}
        \includegraphics[width=\textwidth]{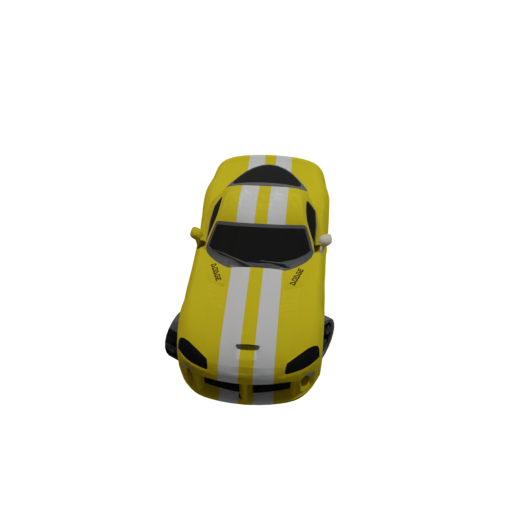}
    \end{subfigure}
    \begin{subfigure}[b]{0.15\textwidth}
        \includegraphics[width=\textwidth]{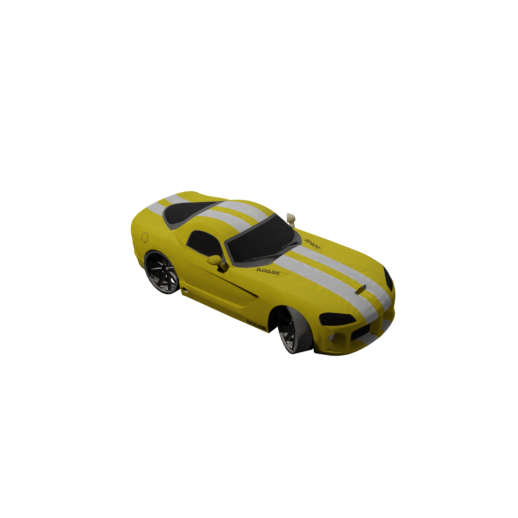}
    \end{subfigure}
    \begin{subfigure}[b]{0.15\textwidth}
        \includegraphics[width=\textwidth]{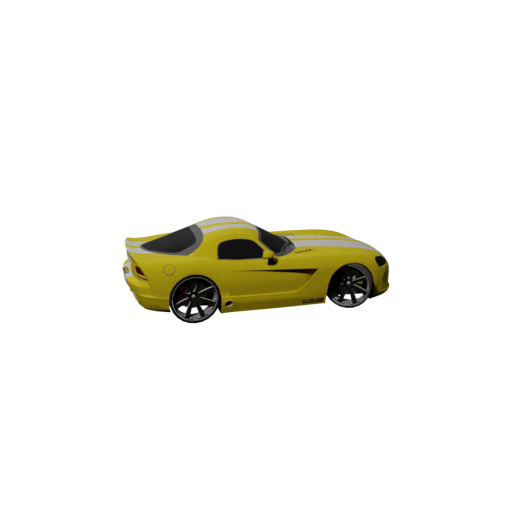}
    \end{subfigure}

    \caption{Few samples from our synthetic controlled dataset built from ShapeNet~\cite{shapenet2015} objects with 2-view pairs in the first row and 3-view triplets in the second row.}
    \label{fig:data}
\end{figure}

We generated the data in Blender by positioning the object at the origin and sampling five camera pairs in the scene. We first randomly chose a position of the first camera, and based on the selected camera configuration mode, we defined the position of the second camera in the scene. We rendered the scenes with different focal lengths (24 mm, 35 mm, 40 mm, 50 mm, 70 mm, 85 mm, 100 mm) and a sensor size of 36 mm, where larger focal lengths result in \emph{zoomed in} images. We calculated ground truth data for pair combinations in both directions (camera 1-camera 2 and camera 2-camera 1) with ground truth data including camera poses, correspondences, and fundamental matrix.
We included four camera configurations: a stereo camera pair with parallel image planes and a horizontal baseline, and non-parallel image planes with different angles between the cameras marked as small (10 - 25 degrees), medium (45 - 75), and large (90 - 120). Viewpoint differences present one aspect of task difficulty.

We further use another version of the datasets, in which we generated triplets of images with the reference camera in the middle, on the left, or on the right, and use 3 distinct camera viewpoints, small (10--25), medium (45--75), and large (90--140). We use the same splits as for the pair-wise version of the datasets.

\section{Two-view probing for fundamental matrix}\label{suppl:probing}
We probe model's intermediate representation to uncover where and how the geometry is represented. 

\subsection{Design and evaluation details}
We train simple two-layer MLP probes on \vggt's and \da's camera tokens at each layer to predict the fundamental matrix, whereas for \duster~we use features from each decoder block. Each probe uses a hidden dimension of 512 and is optimized for 30 epochs with the Adam optimizer at a learning rate of 1e-4 and a step scheduler with a step size of 10 and a gamma of 0.5. We use the Sampson distance (in square-pixel space) as the training loss over all available ground-truth correspondences, without directly enforcing the rank-2 constraint. We perform early stopping based on the validation subset and report the final performance of the best-performing probe on the test dataset.

We further report results for a much simpler linear probe, trained under the same conditions as the MLP probe, to confirm that the information about the fundamental matrix is present in the representation. With both probes, we observe the same trends, but we obtain smoother curves with the MLP probe, so we use the MLP probe as the default.

\subsection{Results on MipNeRF360} 
\begin{figure}[h!]
\centering
\newcommand{\rowlabelwidth}{0.55cm}
\newcommand{\imwidth}{0.305\textwidth}
\setlength{\tabcolsep}{2pt} 
\renewcommand{\arraystretch}{0} 

\newcommand{\rowlab}[1]{%
  \makebox[\rowlabelwidth][c]{\raisebox{2.5ex}{\rotatebox{90}{\textbf{#1}}}}%
}

\begin{tabular}{@{}c c c c@{}}
  & \textbf{VGGT} & \textbf{Depth Anything 3} & \textbf{DUSt3R} \\

  \rowlab{MipNeRF} &
  \includegraphics[width=\imwidth]{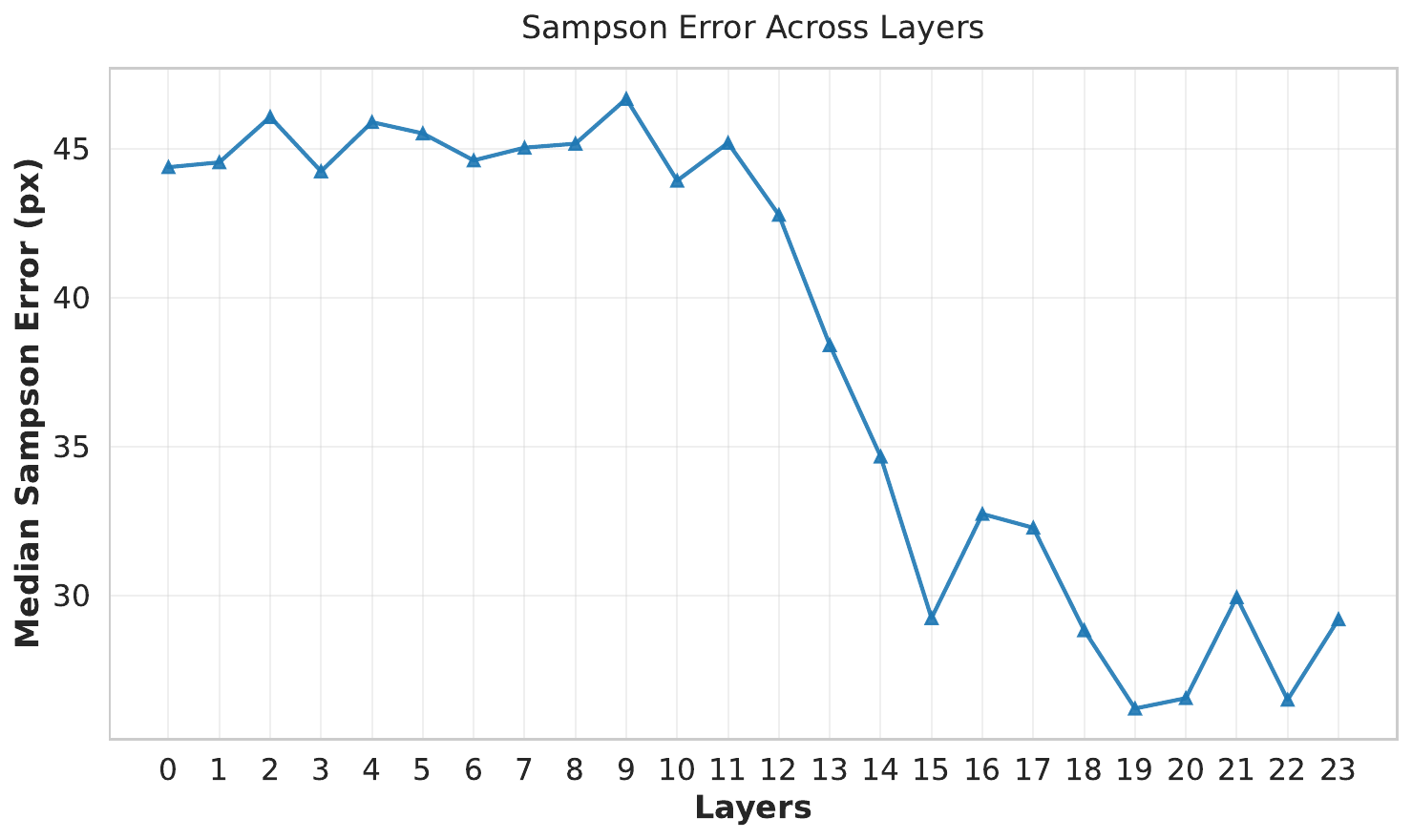} &
  \includegraphics[width=\imwidth]{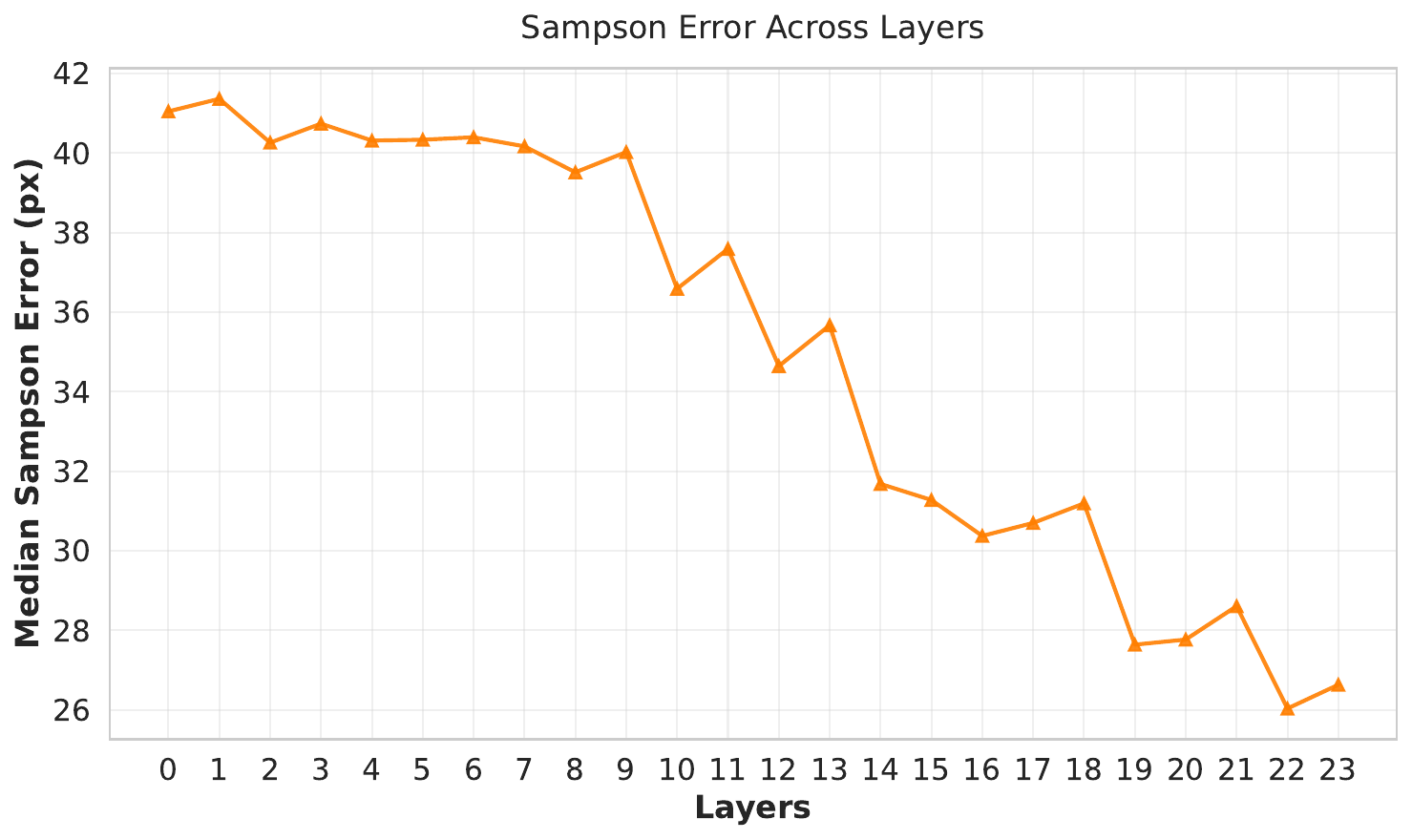} &
  \includegraphics[width=\imwidth]{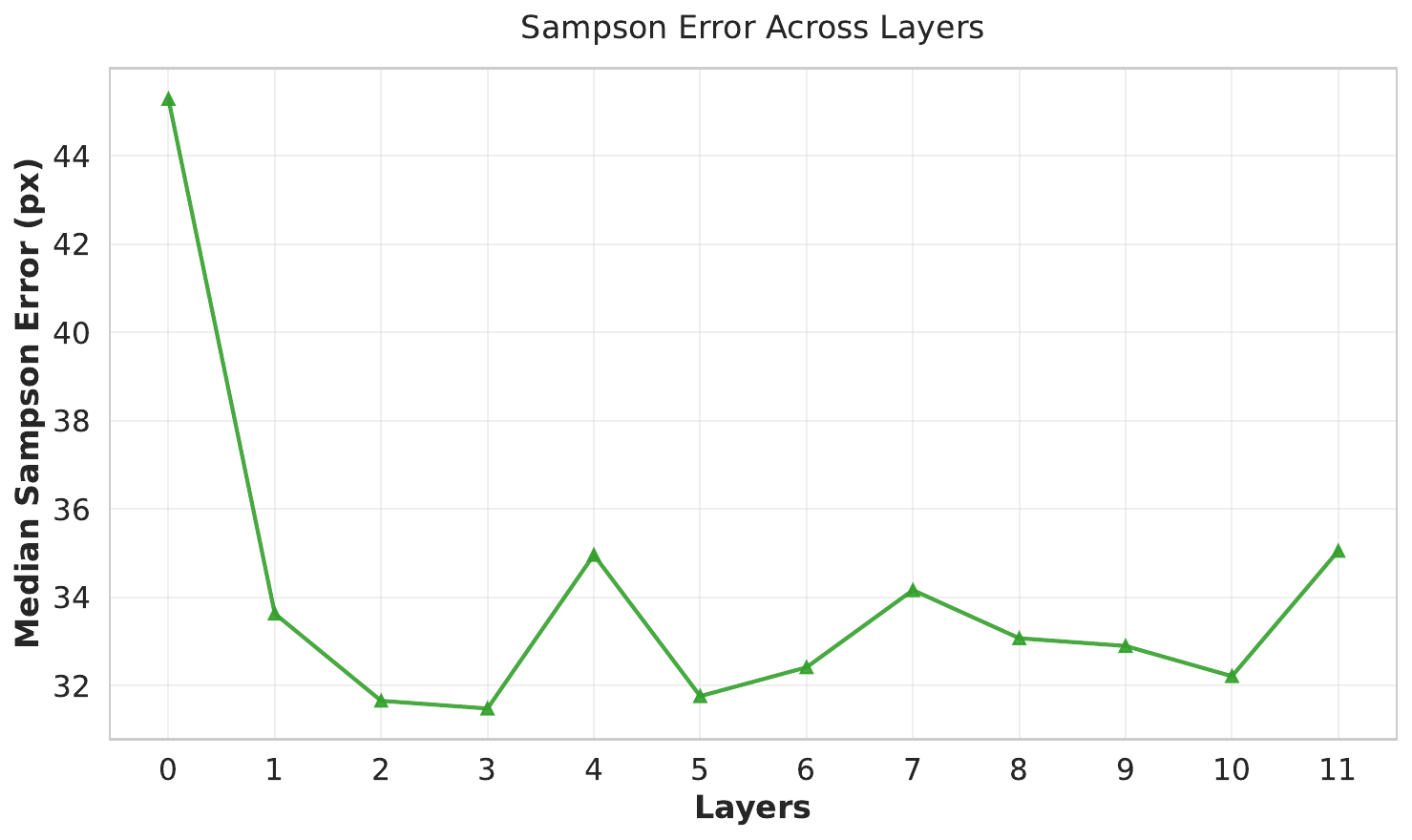} \\

\end{tabular}

\caption{\textbf{Probing internal representations for fundamental matrix approximation.} We successfully recover the fundamental matrix from all three models, also on the MipNeRF360 dataset, but with higher error values. However, the emergence trends remain the same as for other datasets.}
    \label{fig:probing_results_mip}
\end{figure}

We report the results on the MipNeRF360 datasets on~\cref{fig:probing_results_mip}. We did not include the results in the main paper because the error scale differs from other real-world datasets. We observe the same trends as for other datasets.

\subsection{Probe design ablations} 
\begin{figure}[t!]
\centering
\newcommand{\rowlabelwidth}{0.55cm}
\newcommand{\imwidth}{0.305\textwidth}
\setlength{\tabcolsep}{2pt} 
\renewcommand{\arraystretch}{0} 

\newcommand{\rowlab}[1]{%
  \makebox[\rowlabelwidth][c]{\raisebox{2.5ex}{\rotatebox{90}{\textbf{#1}}}}%
}

\begin{tabular}{@{}c c c c@{}}
  & \textbf{VGGT} & \textbf{Depth Anything 3} & \textbf{DUSt3R} \\

  \rowlab{Synthetic} &
  \includegraphics[width=\imwidth]{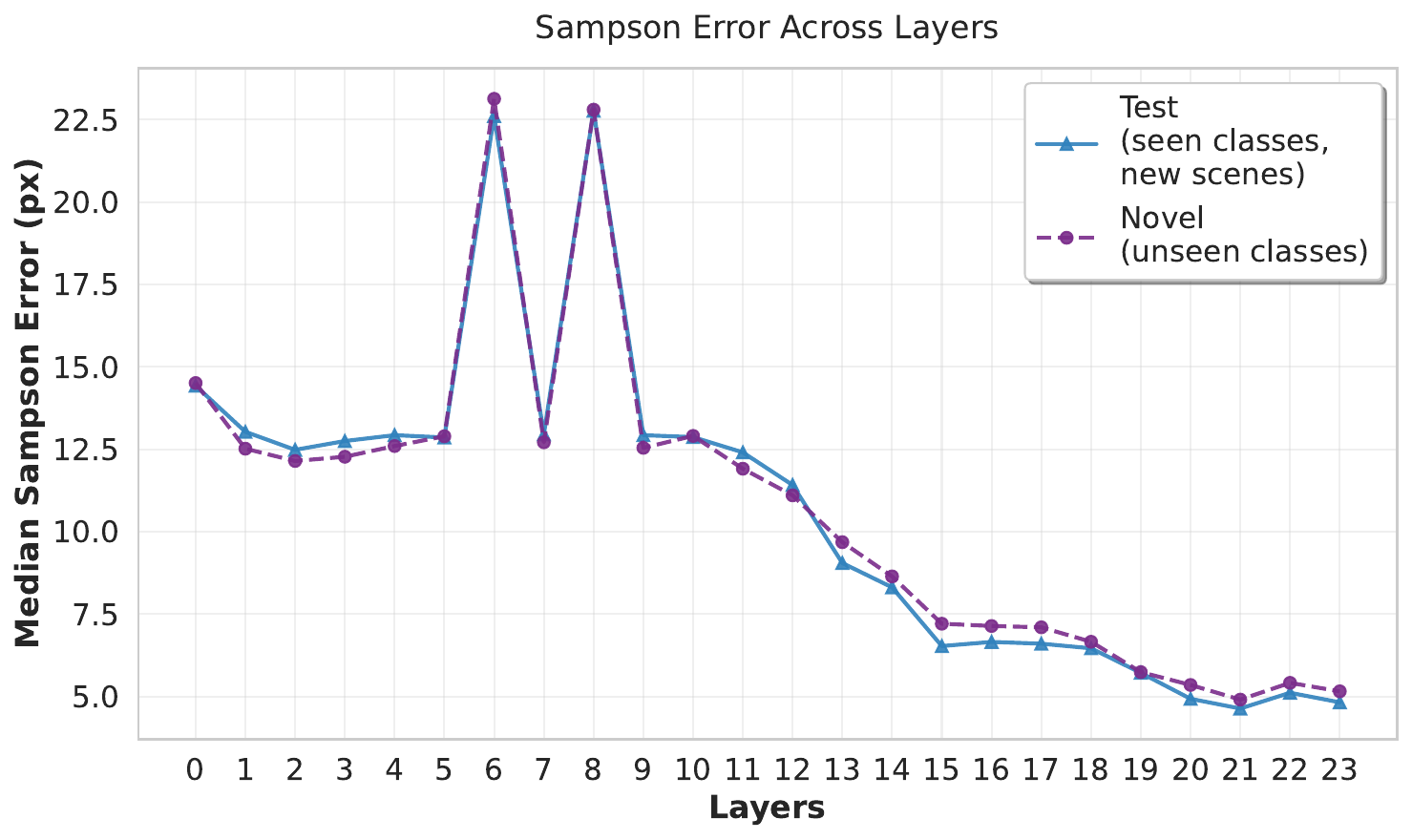} &
  \includegraphics[width=\imwidth]{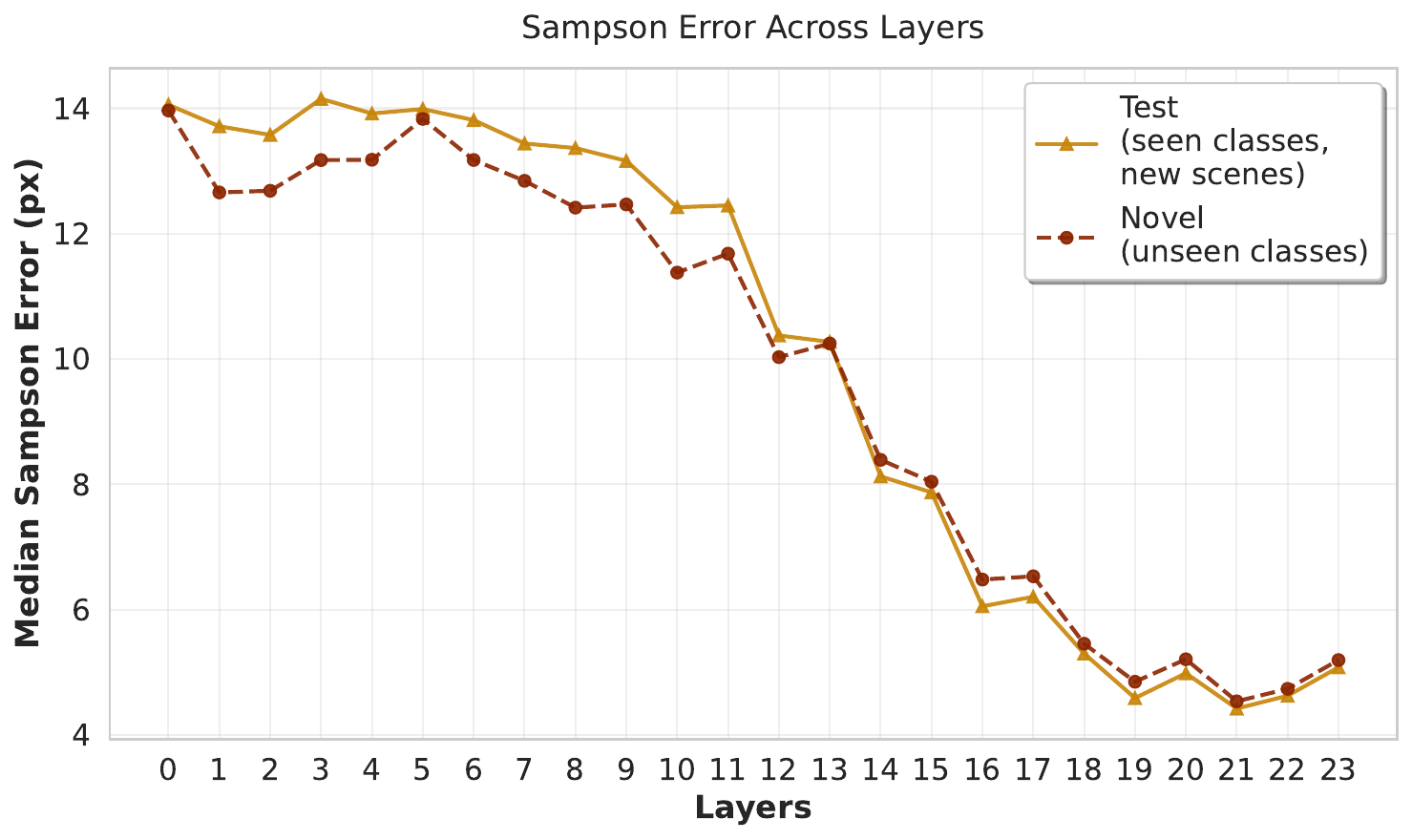} &
  \includegraphics[width=\imwidth]{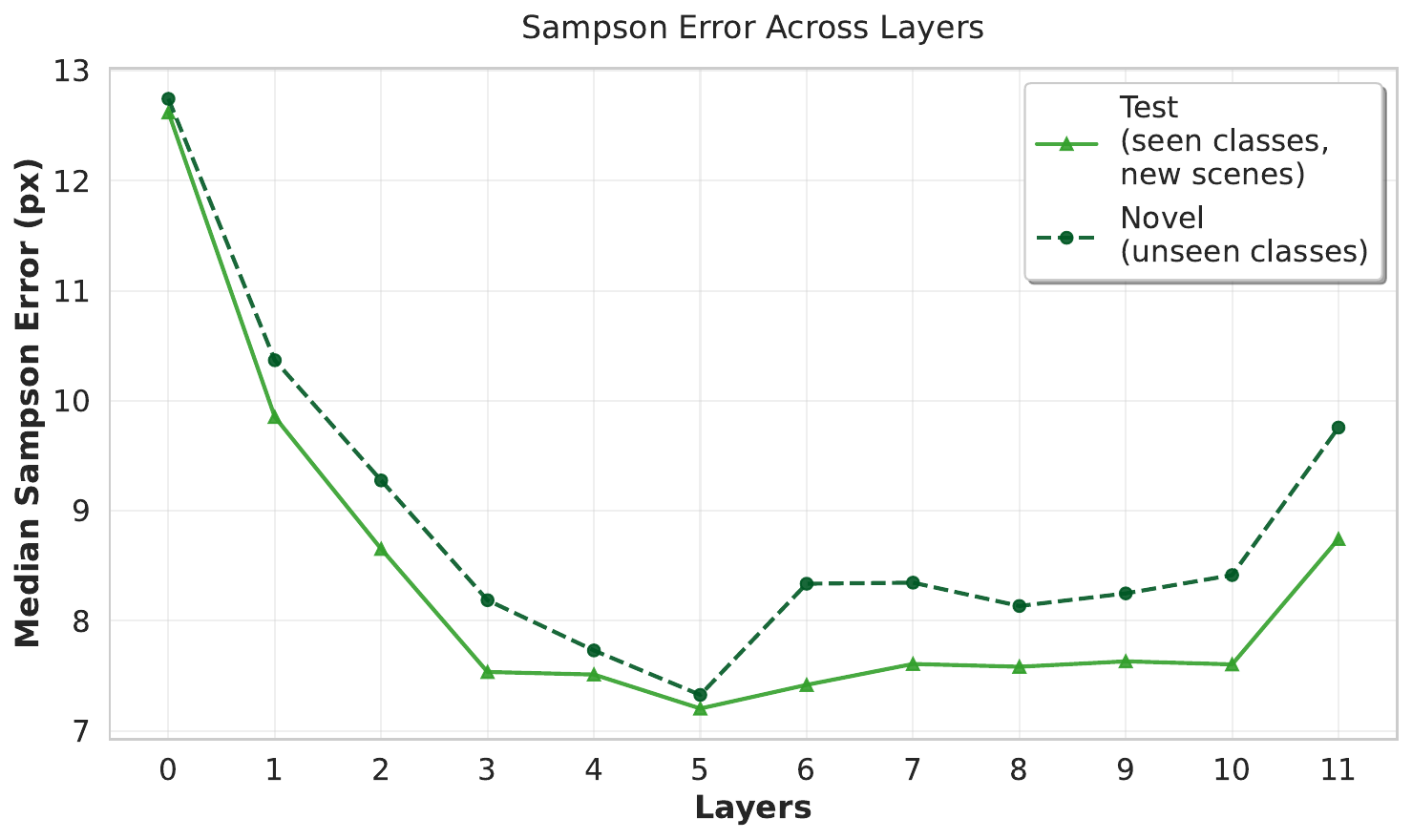} \\

  \rowlab{Real data} &
  \includegraphics[width=\imwidth]{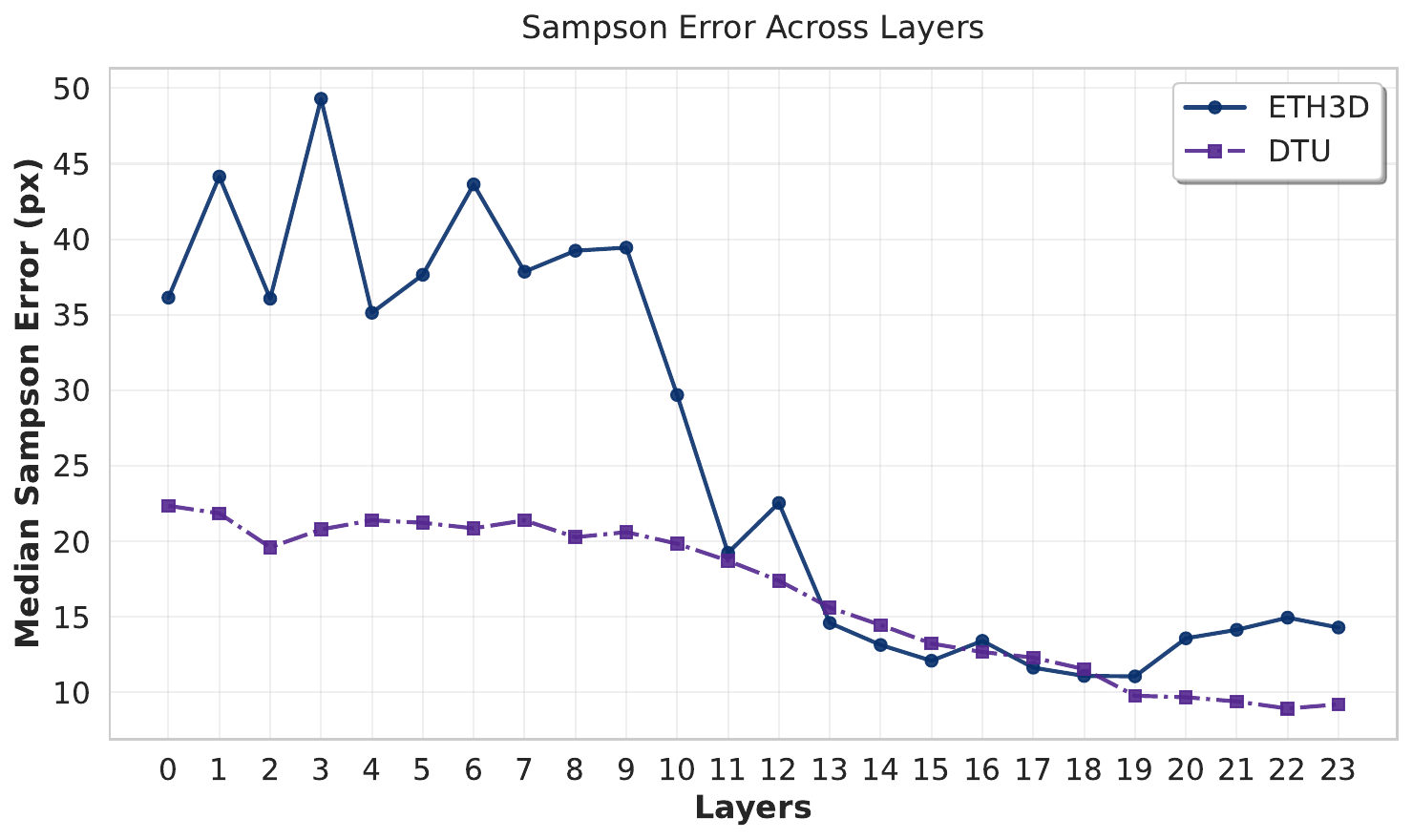} &
  \includegraphics[width=\imwidth]{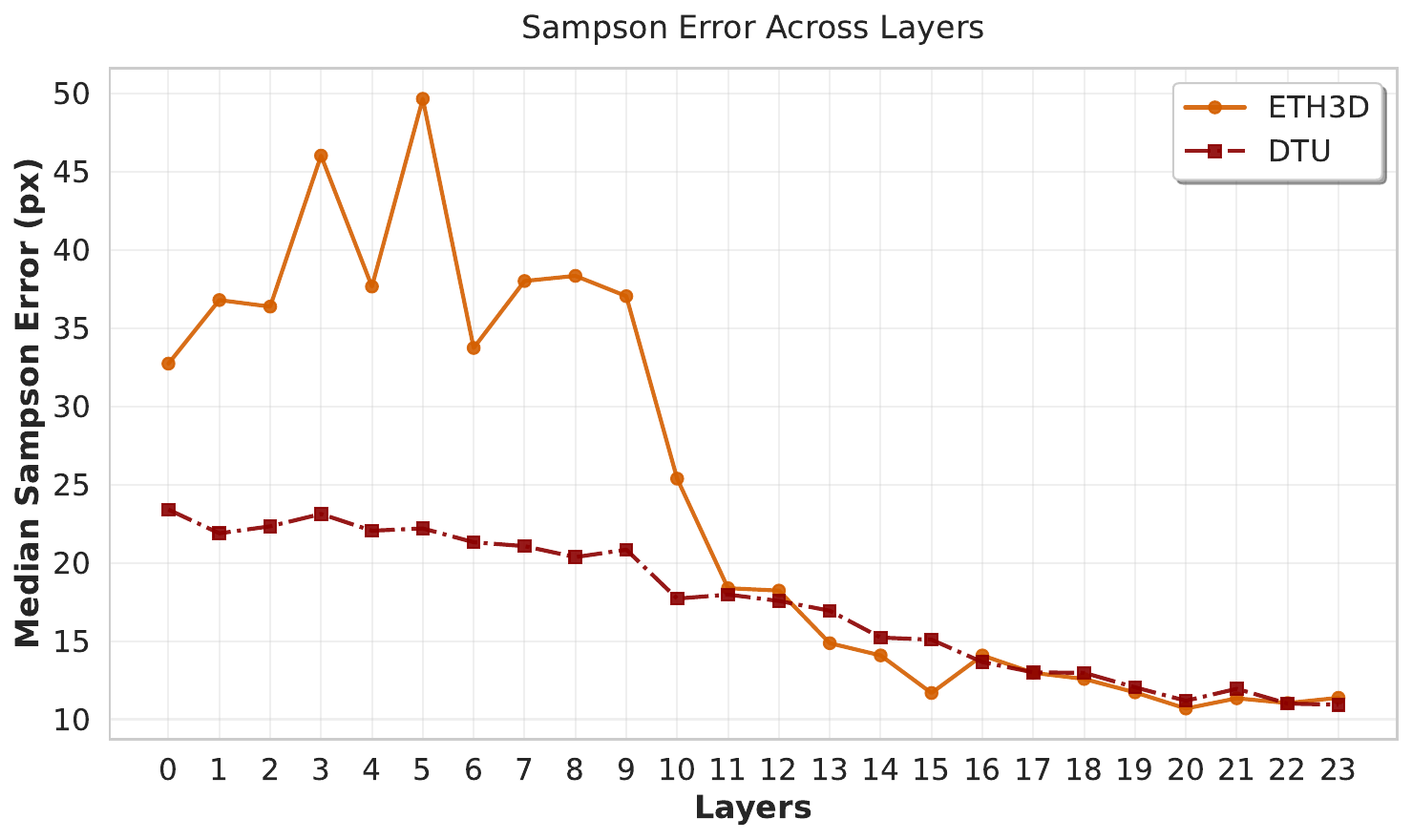} &
  \includegraphics[width=\imwidth]{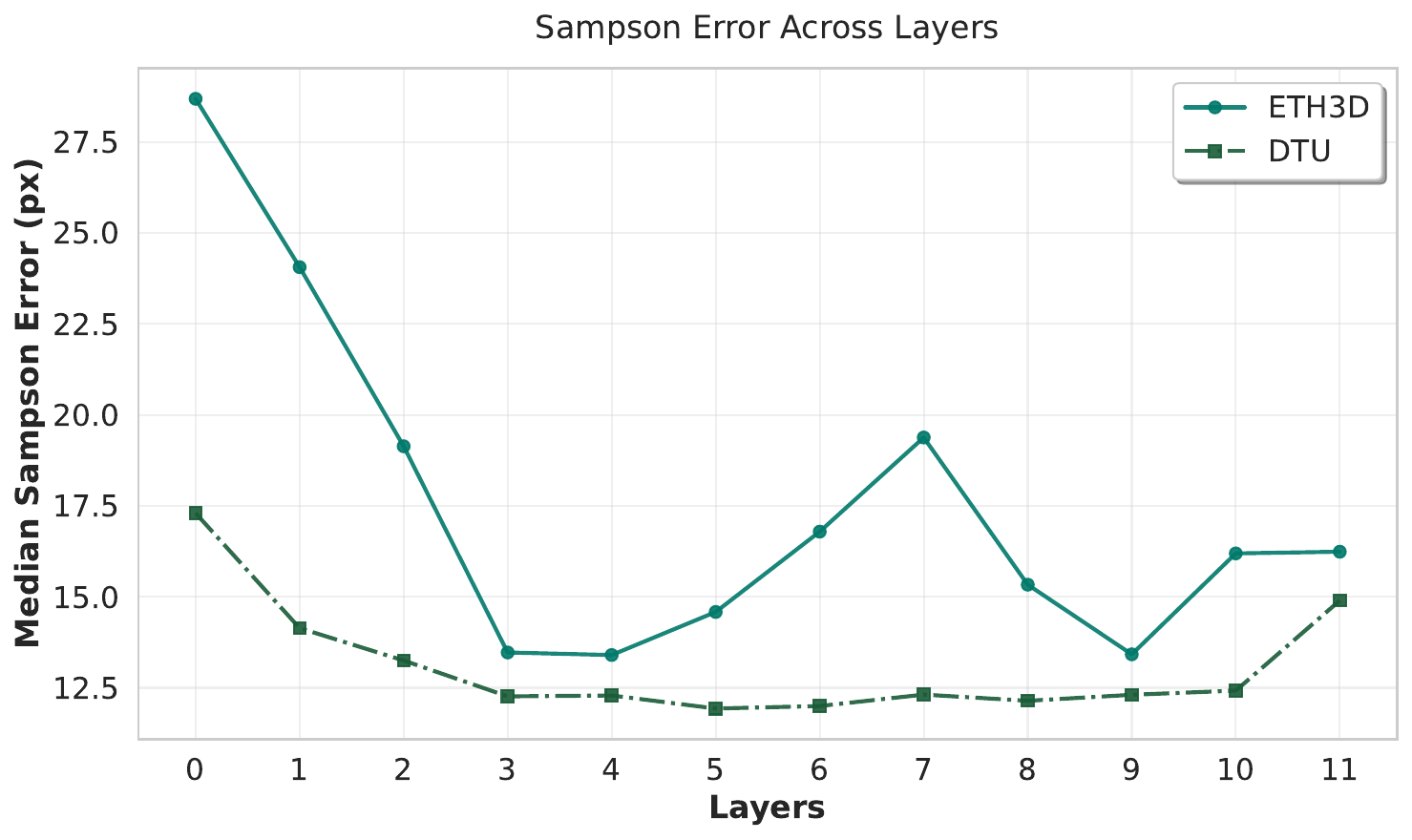} \\
\end{tabular}

\caption{\textbf{Probing internal representations for fundamental matrix approximation.} We successfully recover the fundamental matrix from all three models on both synthetic and real-world datasets when using a linear probe. We report the same trends as for the MLP probe, but with higher errors. }
    \label{fig:probing_results_linear}
\end{figure} 

We show the results of a linear probe on~\cref{fig:probing_results_linear}, where, compared to an MLP probe, we observe the same trends in the ability to retrieve the fundamental matrix, but the Sampson distance error is better for the MLP probe, and the general curve is less noisy.

\begin{figure}[t!]
\centering
\newcommand{\rowlabelwidth}{0.55cm}
\newcommand{\imwidth}{0.305\textwidth}
\setlength{\tabcolsep}{2pt} 
\renewcommand{\arraystretch}{0} 

\newcommand{\rowlab}[1]{%
  \makebox[\rowlabelwidth][c]{\raisebox{2.5ex}{\rotatebox{90}{\textbf{#1}}}}%
}

\begin{tabular}{@{}c c c c@{}}
  & \textbf{VGGT} & \textbf{Depth Anything 3} & \textbf{DUSt3R} \\

  \rowlab{Synthetic} &
  \includegraphics[width=\imwidth]{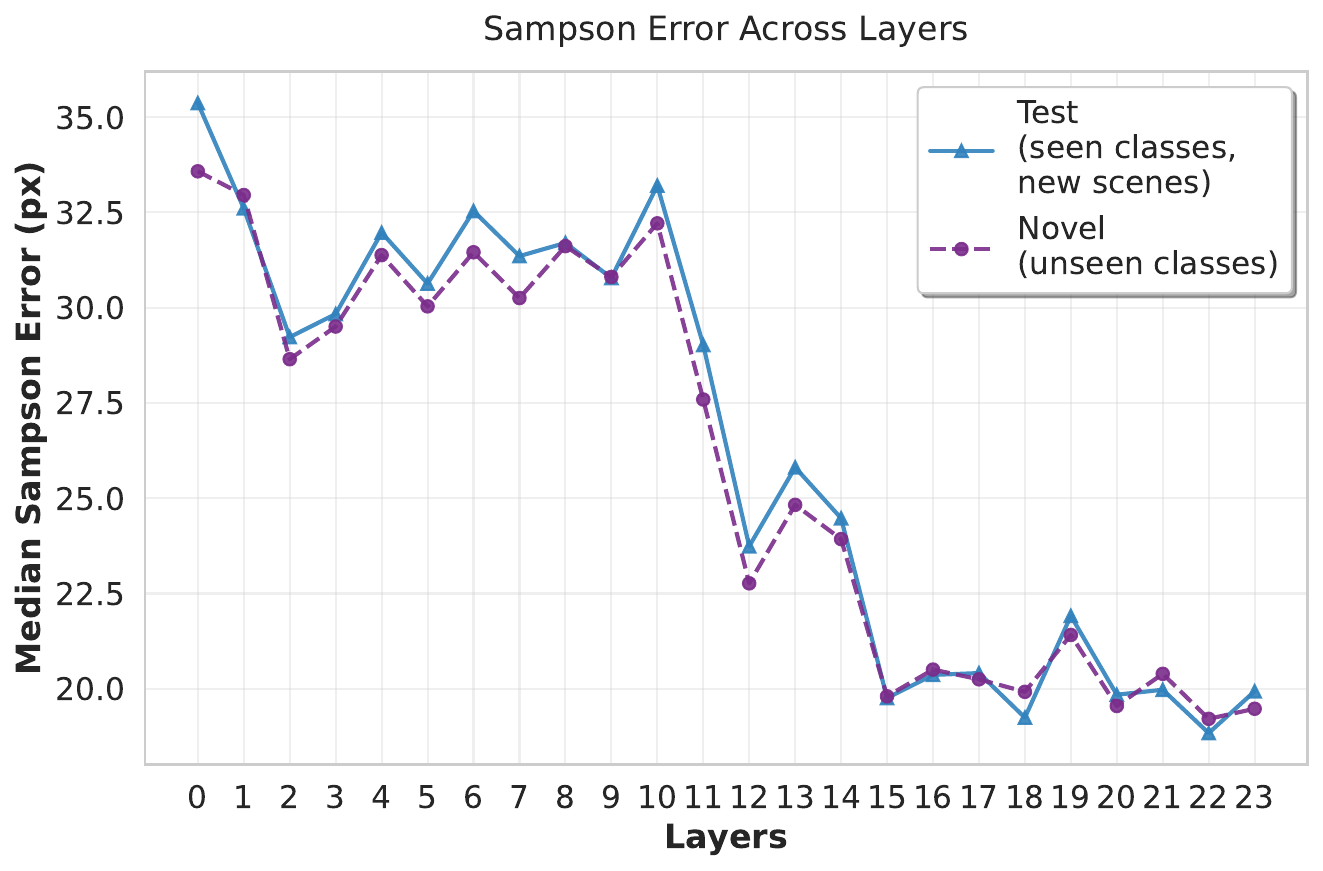} &
  \includegraphics[width=\imwidth]{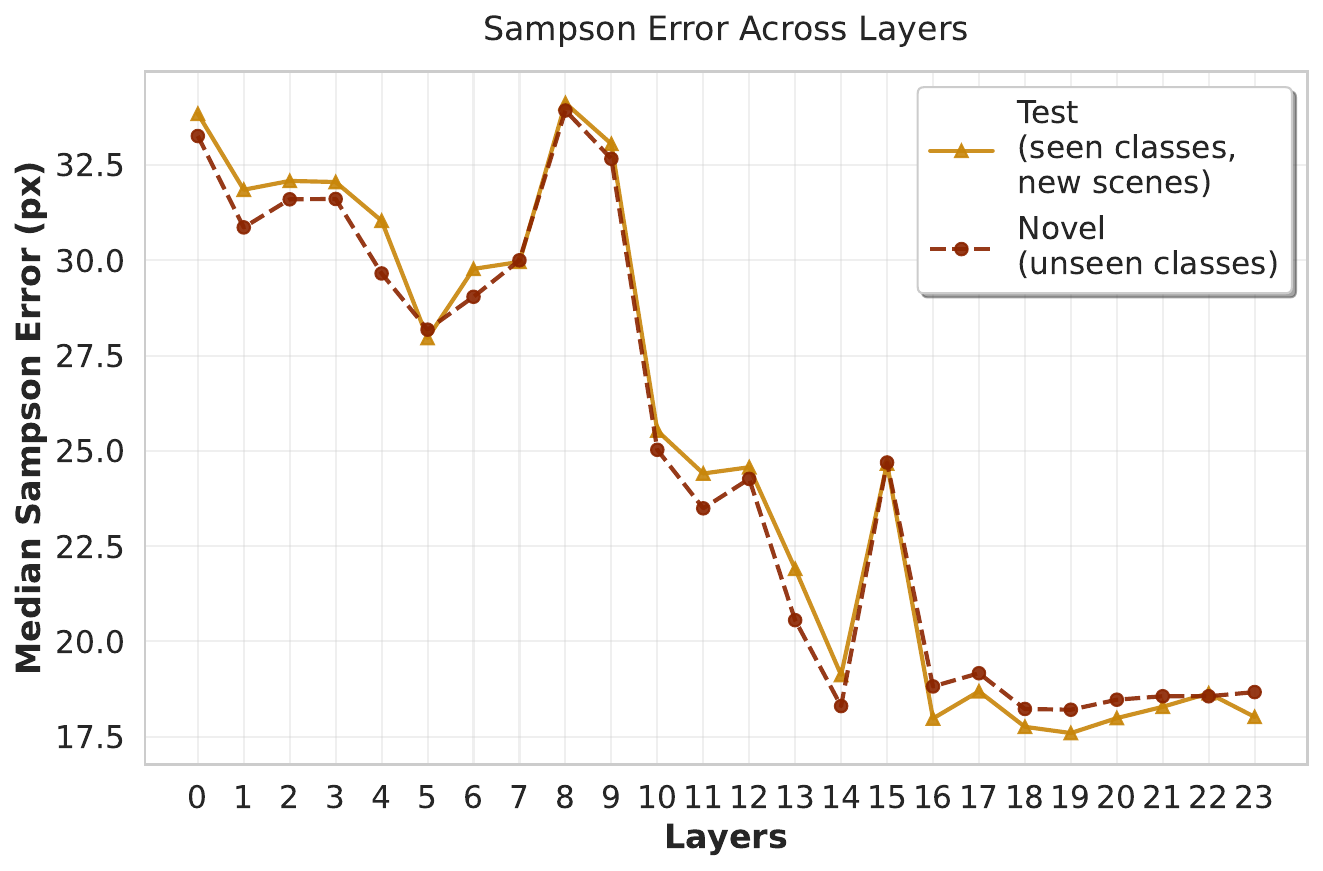} &
  \includegraphics[width=\imwidth]{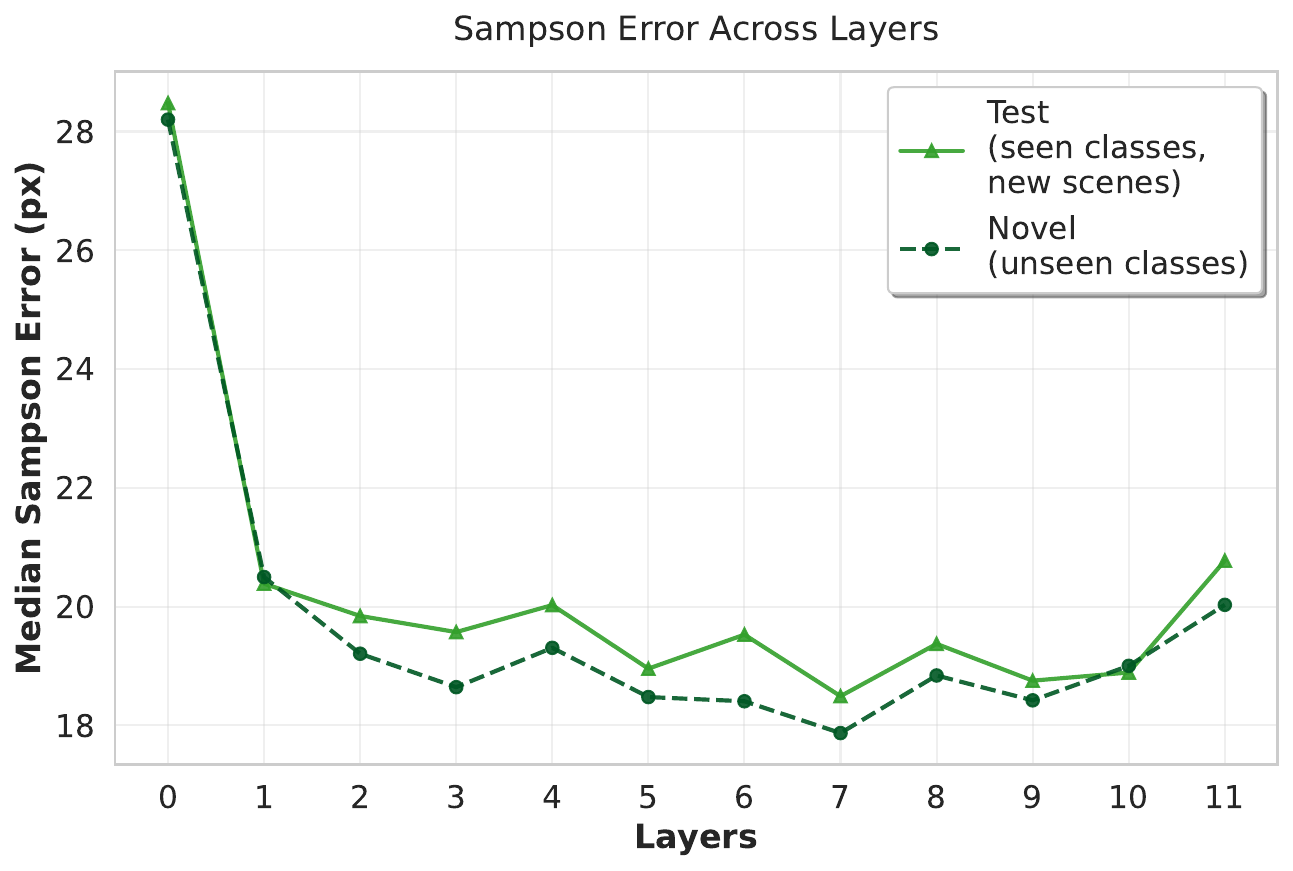} \\

  \rowlab{Real data} &
  \includegraphics[width=\imwidth]{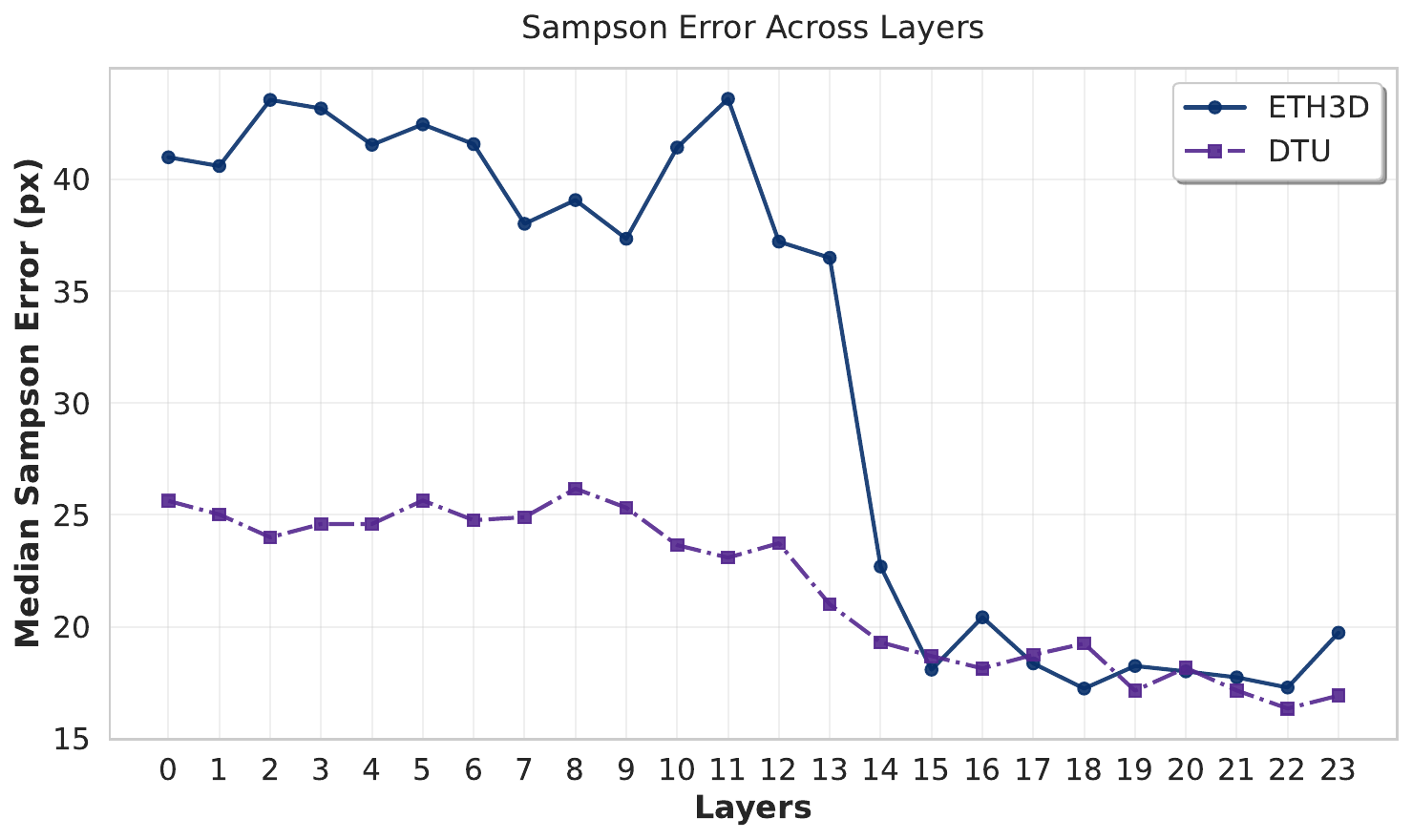} &
  \includegraphics[width=\imwidth]{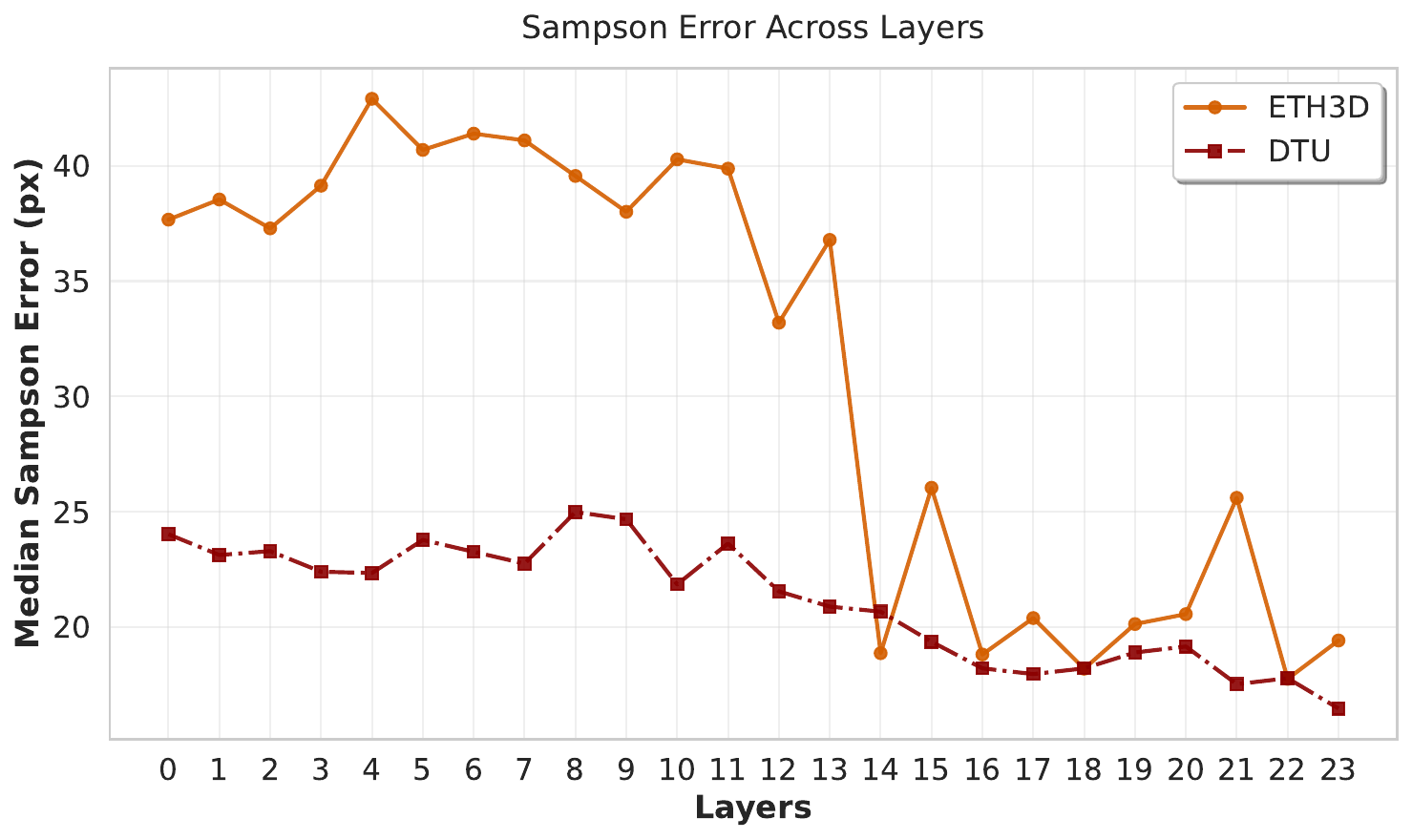} &
  \includegraphics[width=\imwidth]{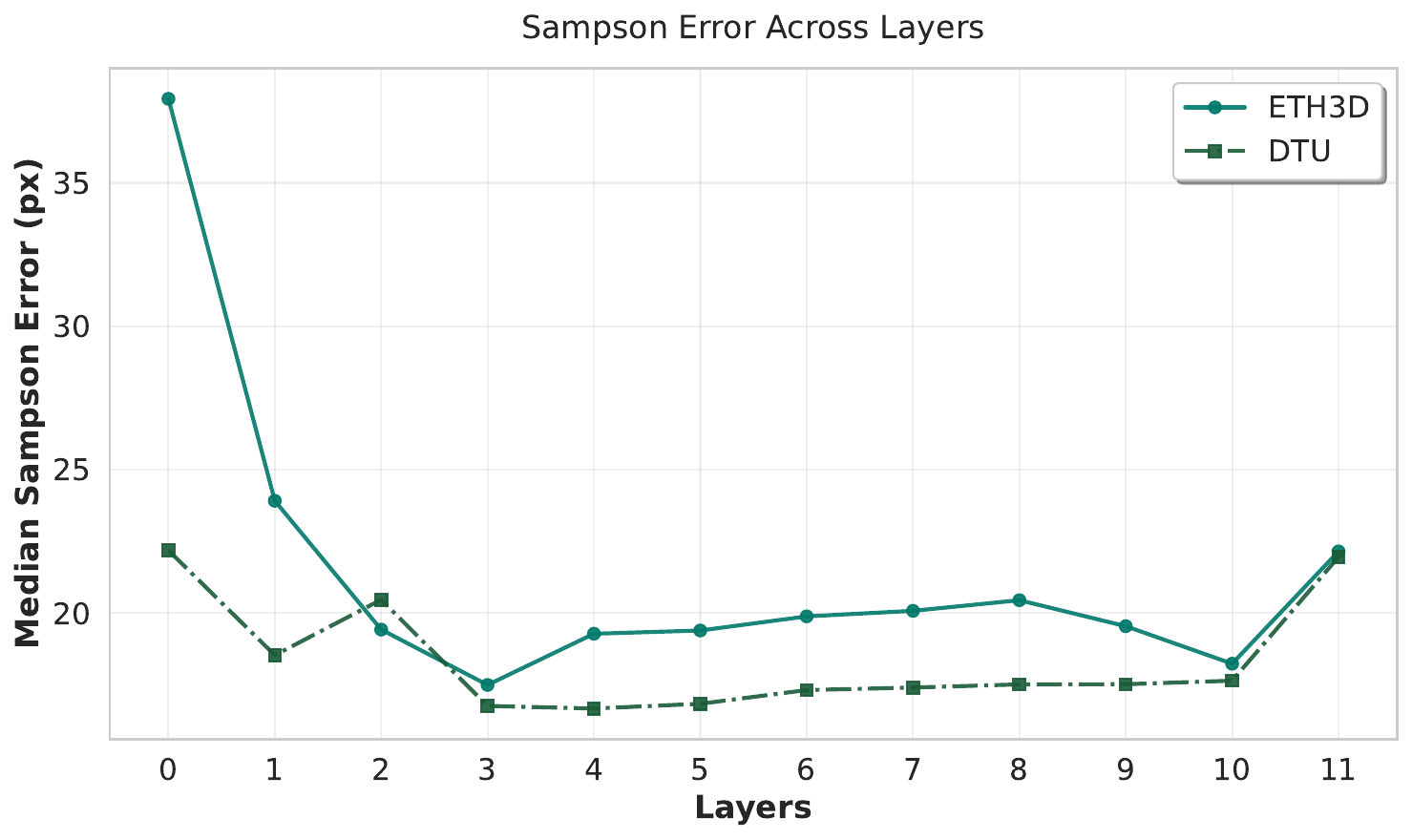} \\
\end{tabular}

\caption{\textbf{Probing internal representations for fundamental matrix approximation.} We successfully recover the fundamental matrix from all three models on both synthetic and real-world datasets with an MLP probe trained only for 1 epoch. We report the same trends, but the error values are significantly higher.}
    \label{fig:probing_results_ep1}
\end{figure}

Further, on~\cref{fig:probing_results_ep1} we report the results of an MLP probe after one epoch to show the same trends again, but with a larger total error. This confirms that the fundamental matrix information is present in the representation and that no overfitting occurred during probing. 

\subsection{Rank 2 and singular value analysis} 

\begin{figure}[b!]
\centering
\newcommand{\rowlabelwidth}{0.55cm}
\newcommand{\imwidth}{0.305\textwidth}
\setlength{\tabcolsep}{2pt} 
\renewcommand{\arraystretch}{0} 

\newcommand{\rowlab}[1]{%
  \makebox[\rowlabelwidth][c]{\raisebox{2.5ex}{\rotatebox{90}{\textbf{#1}}}}%
}

\begin{tabular}{@{}c c c c@{}}
  & \textbf{VGGT} & \textbf{Depth Anything 3} & \textbf{DUSt3R} \\

  \rowlab{Synthetic} &
  \includegraphics[width=\imwidth]{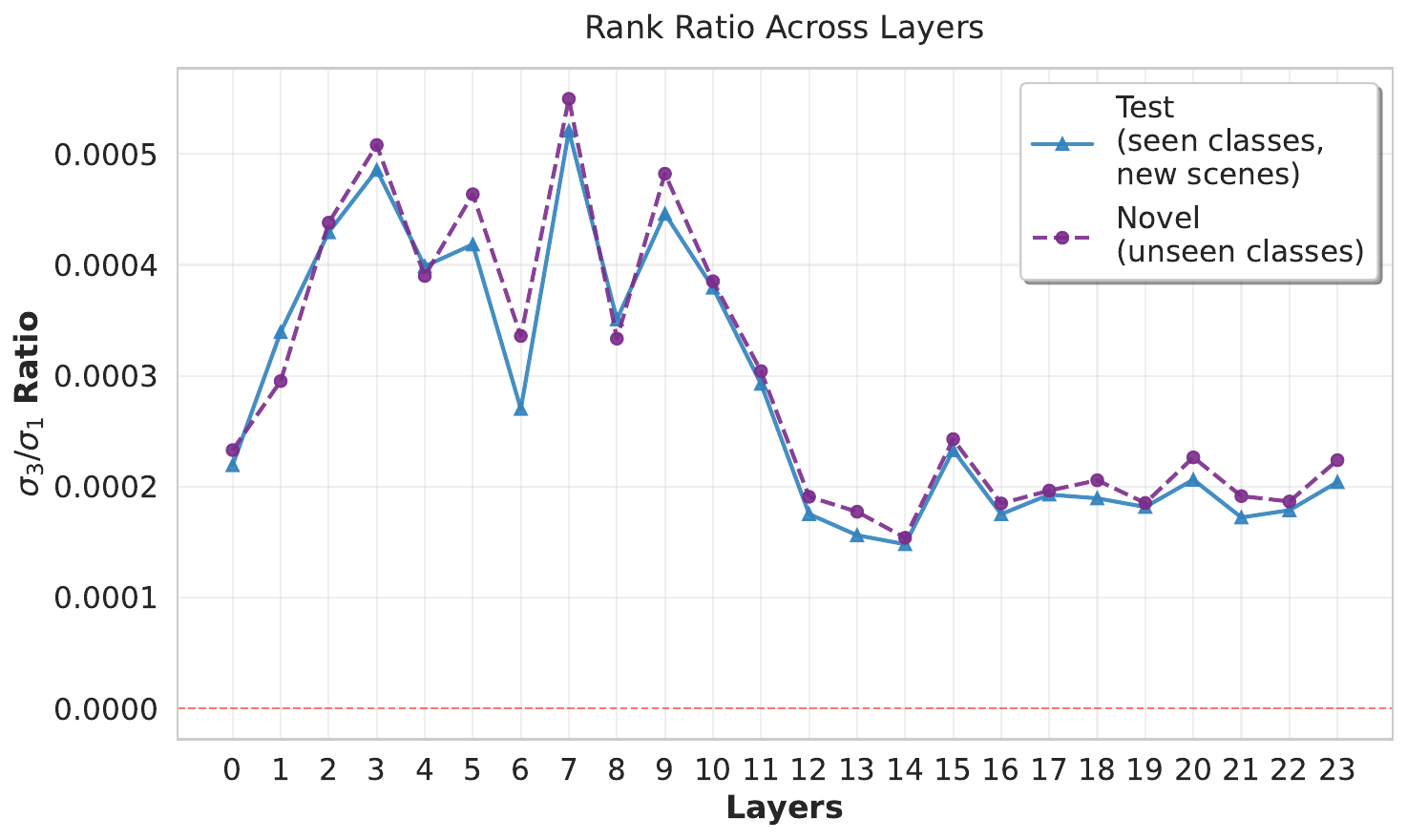} &
  \includegraphics[width=\imwidth]{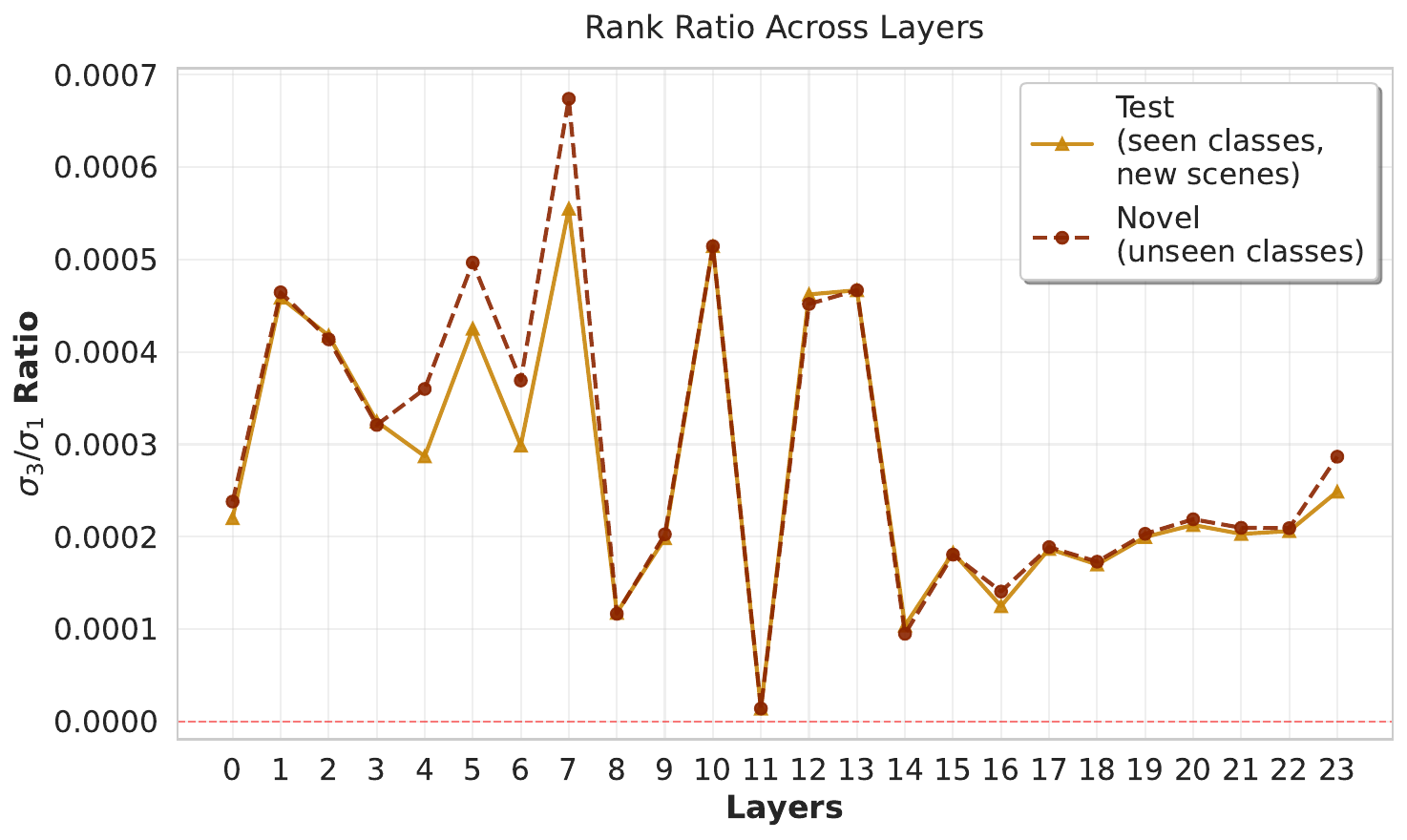} &
  \includegraphics[width=\imwidth]{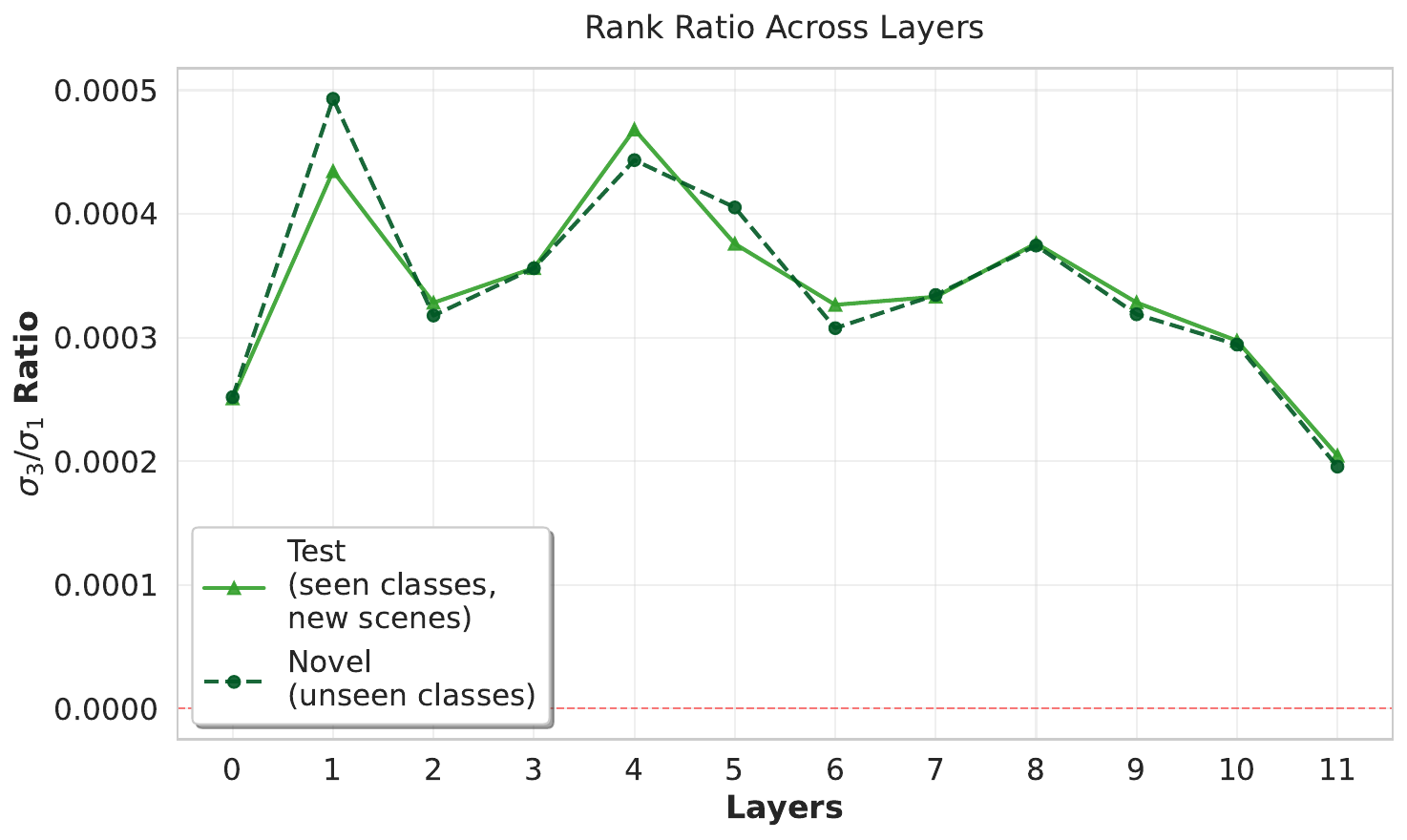} \\

\end{tabular}

\caption{\textbf{Rank ratio.} Ratio of smallest and largest singular values across layers on different data splits for the probing experiment from~\cref{fig:probing_results}. A good fundamental matrix should have a significant difference between the largest and smallest singular values.}
    \label{fig:suppl_rank2}
\end{figure}

We further analyze the rank-2 constraints. For \vggt~and \da, we show that the middle layers comprise a representation that enables recovery of the fundamental matrix $F$, as evidenced by a sudden drop in Sampson distance error, whereas for \duster~all layers exhibit reasonably good performance. Here, on~\cref{fig:suppl_rank2}, we show another criterion for evaluating the fundamental matrix -- the ratio between the smallest and largest singular values of the fundamental matrix.  Since the fundamental matrix is of rank 2~\cite{Hartley2004}, its smallest singular value should be around zero, and the largest singular value should be significantly larger for a well-estimated fundamental matrix. While the trend is not the cleanest between the models, we still observe the minimum values in the same set of layers where the sudden drop occurs. This further supports the emergence of epipolar geometry by satisfying the rank-2 constraint without directly enforcing it during probe training.

\subsection{Triplet 2-view consistency results} 
\begin{figure}[h!]
    \centering
        \begin{subfigure}{0.49\linewidth}
        \centering
        \includegraphics[width=\linewidth]{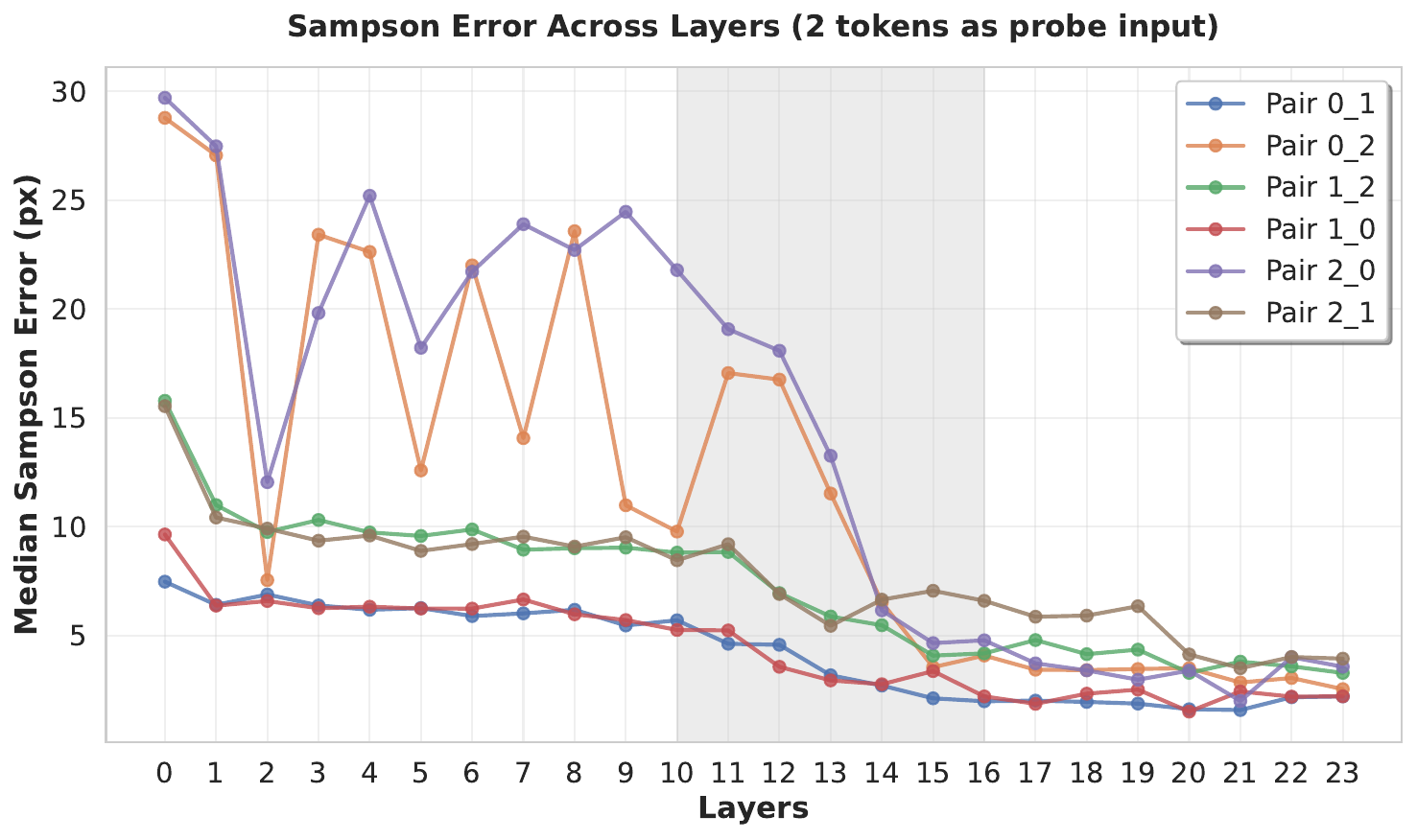}
      \end{subfigure}
      \hfill
      \begin{subfigure}{0.49\linewidth}
        \centering
        \includegraphics[width=\linewidth]{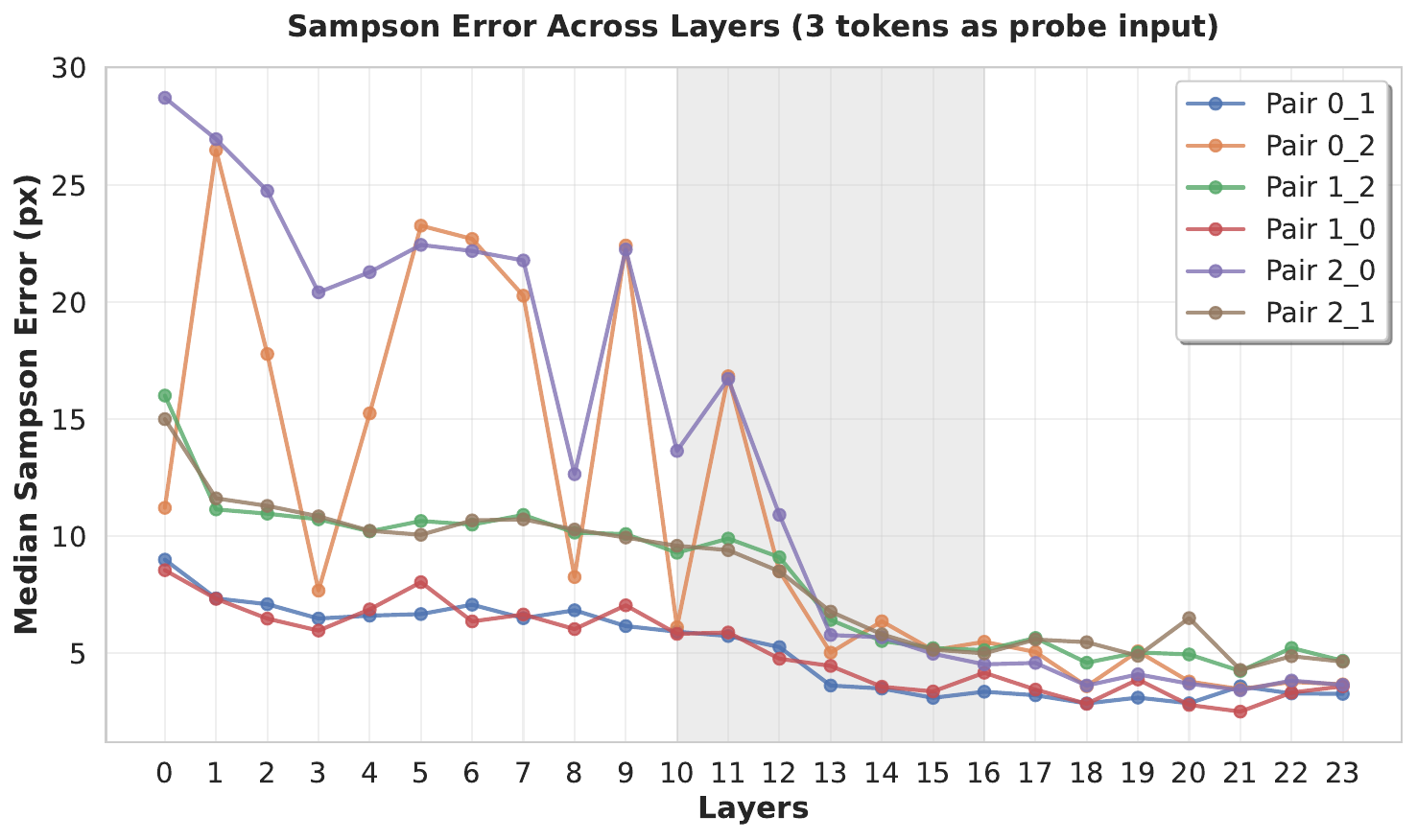}
      \end{subfigure}
    \caption{Multi-view probing on VGGT shows that the fundamental matrix can be extracted for all image pairs in a sequence.}
    \label{fig:multi_probing}
\end{figure}

Additional views add complementary constraints that strengthen correspondence estimation, so our results extend naturally to the multi-view setting. We confirm this by probing for fundamental matrices and computing correspondence matching across all pairs in a three-view sequence for VGGT and our ShapeNet dataset. Correspondence activation is consistent across the same layers and heads for all pairs, with comparable probing performance across views (Fig.~\ref{fig:multi_probing}), indicating that the identified mechanisms persist in multi-view configurations. For clarity, we focus on two views, though multi-view representations may also encode trifocal or multifocal tensors. We leave probing these higher-order structures for future work.

\section{Correspondence matching analysis of intermediate layers}

Probing reveals the emergence of learned epipolar geometry in the intermediate layers. However, we do not yet know how the model recovers this information. To this end, we aim to discover the underlying correspondence-matching patterns in the attention maps of those layers across two views. We hypothesize that these correspondences are computed by global (cross) attention layers, as evidenced by our analysis of the QK attention space in our ShapeNet dataset. 
Here, we provide the same analysis on real-life datasets and extend it to feature similarity analysis rather than the QK attention space. We explore the semantic correspondence-matching ability and, finally, compare different model sizes of \da.

\begin{figure}[t!]
\centering
\newcommand{\rowlabelwidth}{0.45cm}
\newcommand{\imwidth}{0.28\textwidth}
\setlength{\tabcolsep}{2pt} 
\renewcommand{\arraystretch}{0} 

\newcommand{\rowlab}[1]{%
  \makebox[\rowlabelwidth][c]{\raisebox{2.5ex}{\rotatebox{90}{\textbf{#1}}}}%
}

\newcommand{\datasetlab}[1]{%
  \makebox[\rowlabelwidth][r]{\raisebox{5.5ex}{\rotatebox{90}{\textbf{#1}}}}%
}

\begin{tabular}{@{}c c c c c@{}}
  && \textbf{VGGT} & \textbf{Depth Anything 3} & \textbf{DUSt3R} \\

  \multirow{2}{*}{\datasetlab{ETH3D}} &
  \rowlab{F (1$\to$2)} &
  \includegraphics[width=\imwidth]{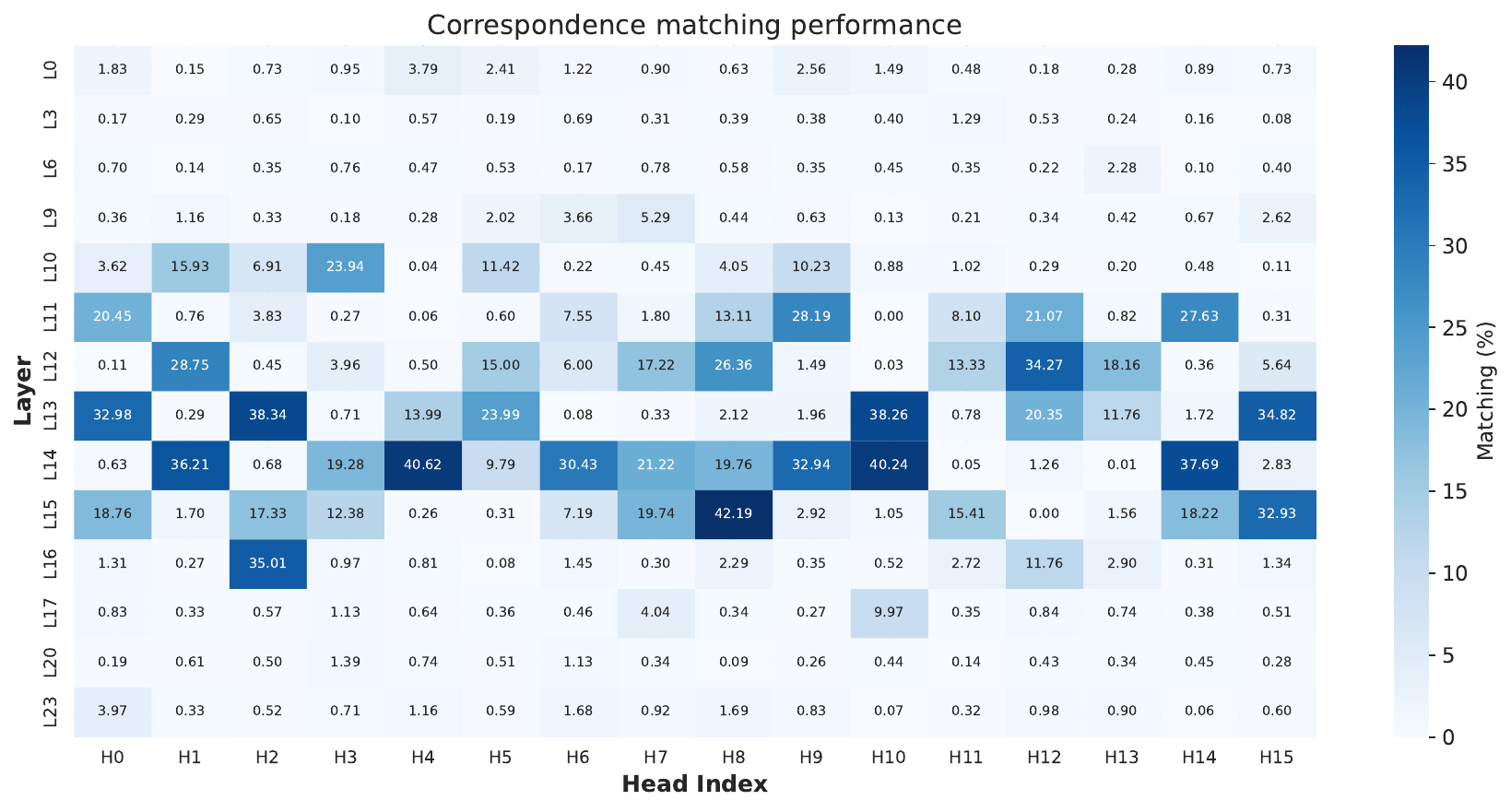} &
  \includegraphics[width=\imwidth]{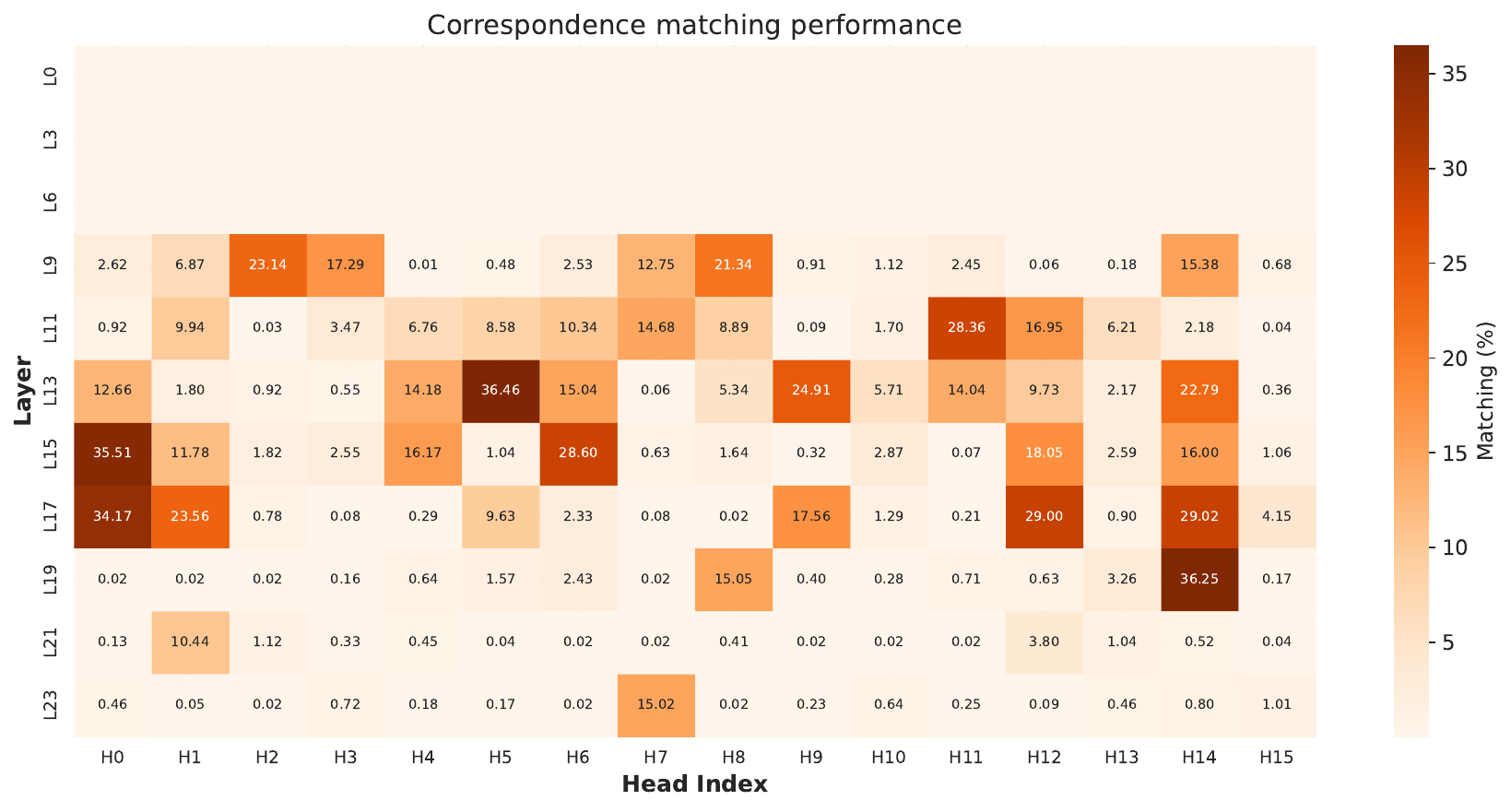} &
  \includegraphics[width=\imwidth]{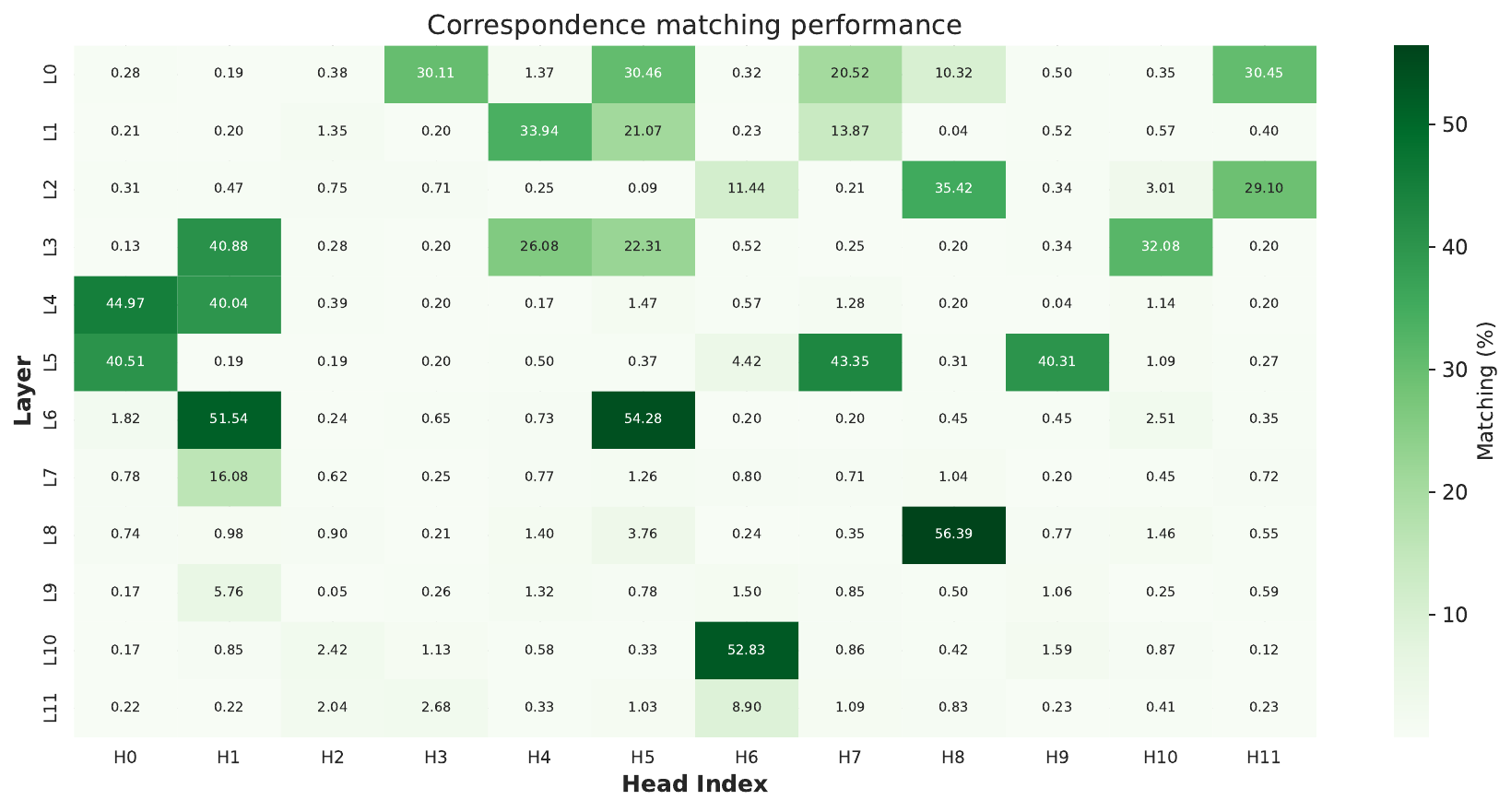} \\
  &
  \rowlab{R (2$\to$1)} &
  \includegraphics[width=\imwidth]{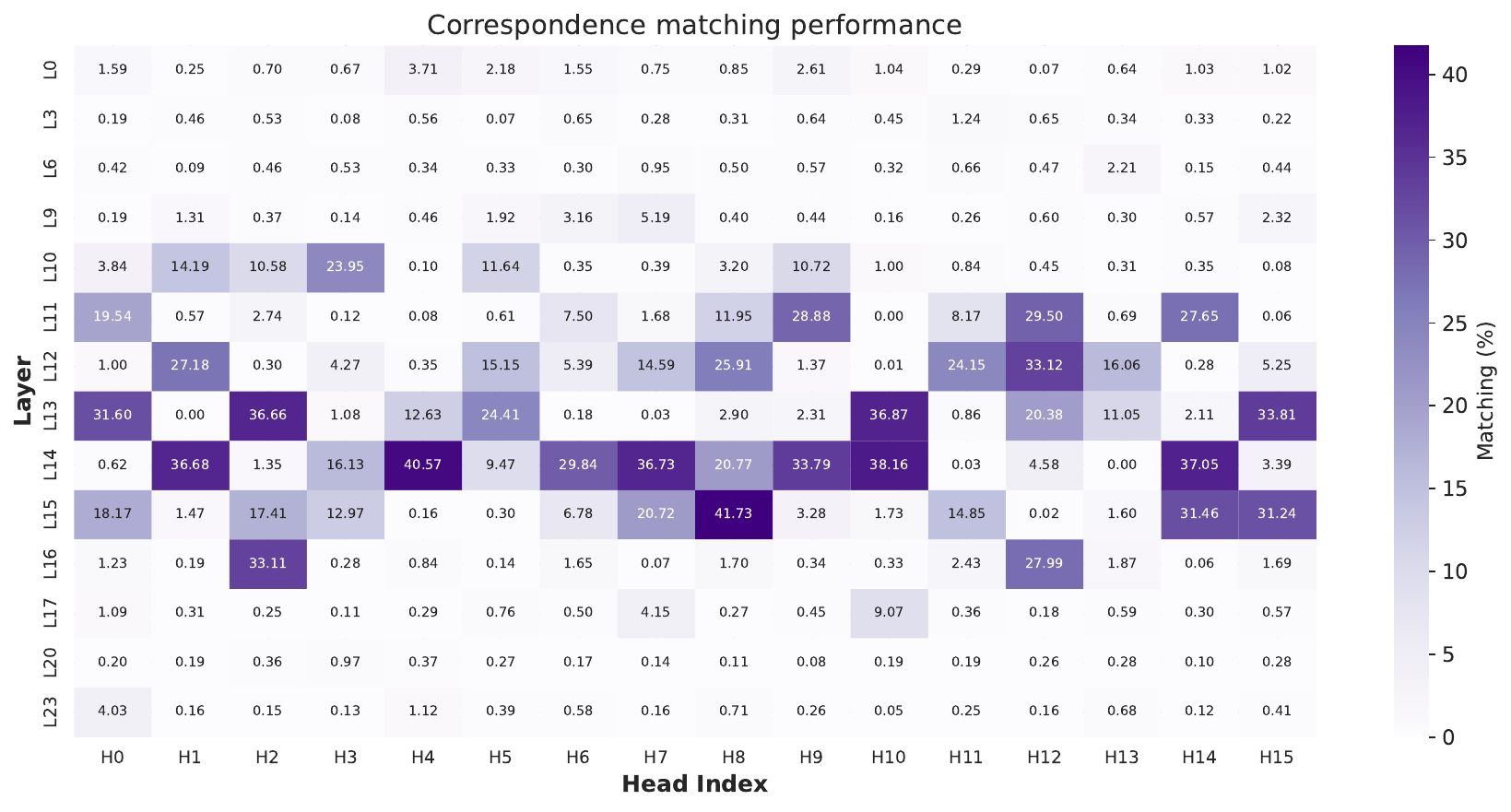} &
  \includegraphics[width=\imwidth]{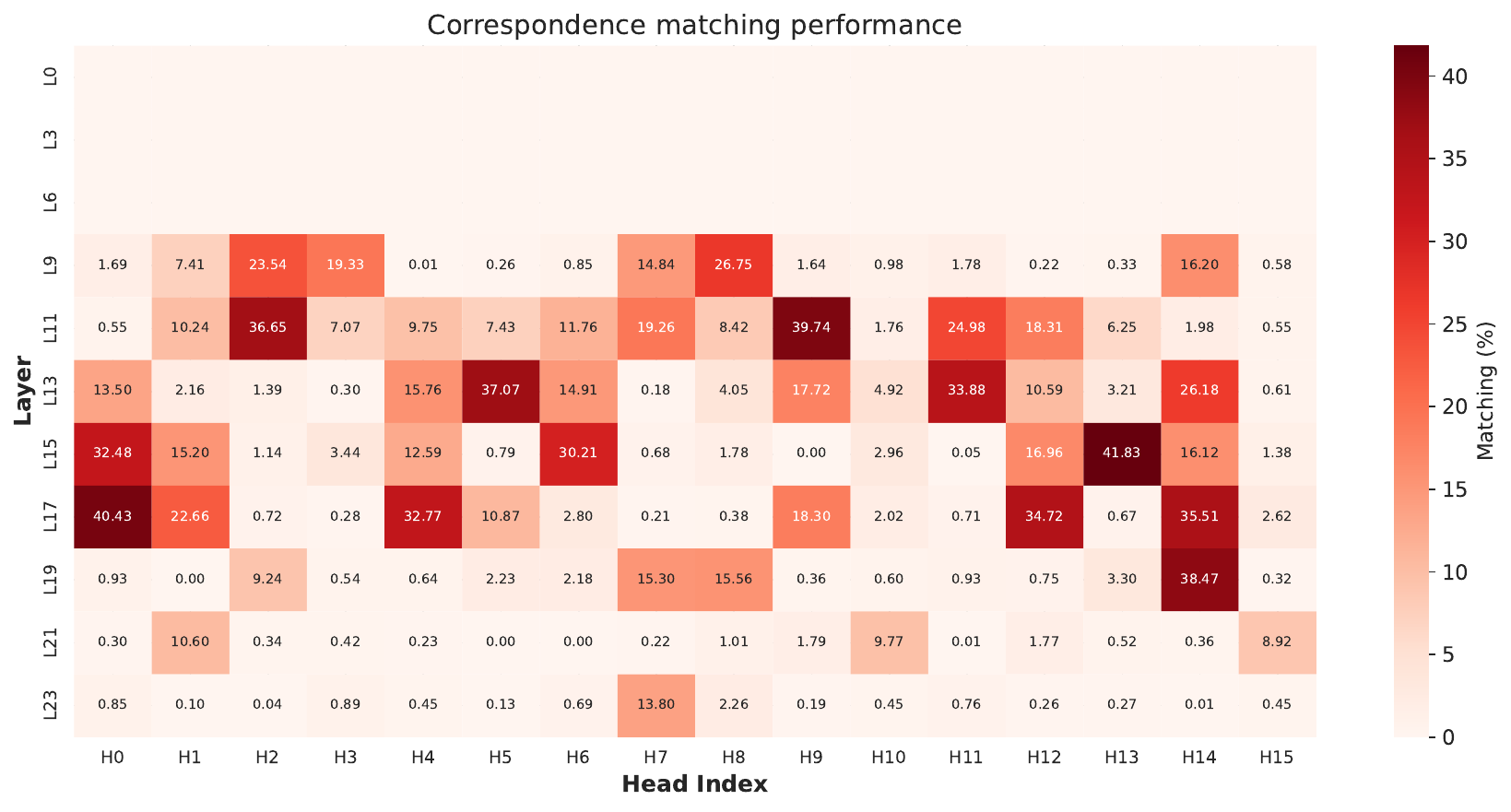} &
  \includegraphics[width=\imwidth]{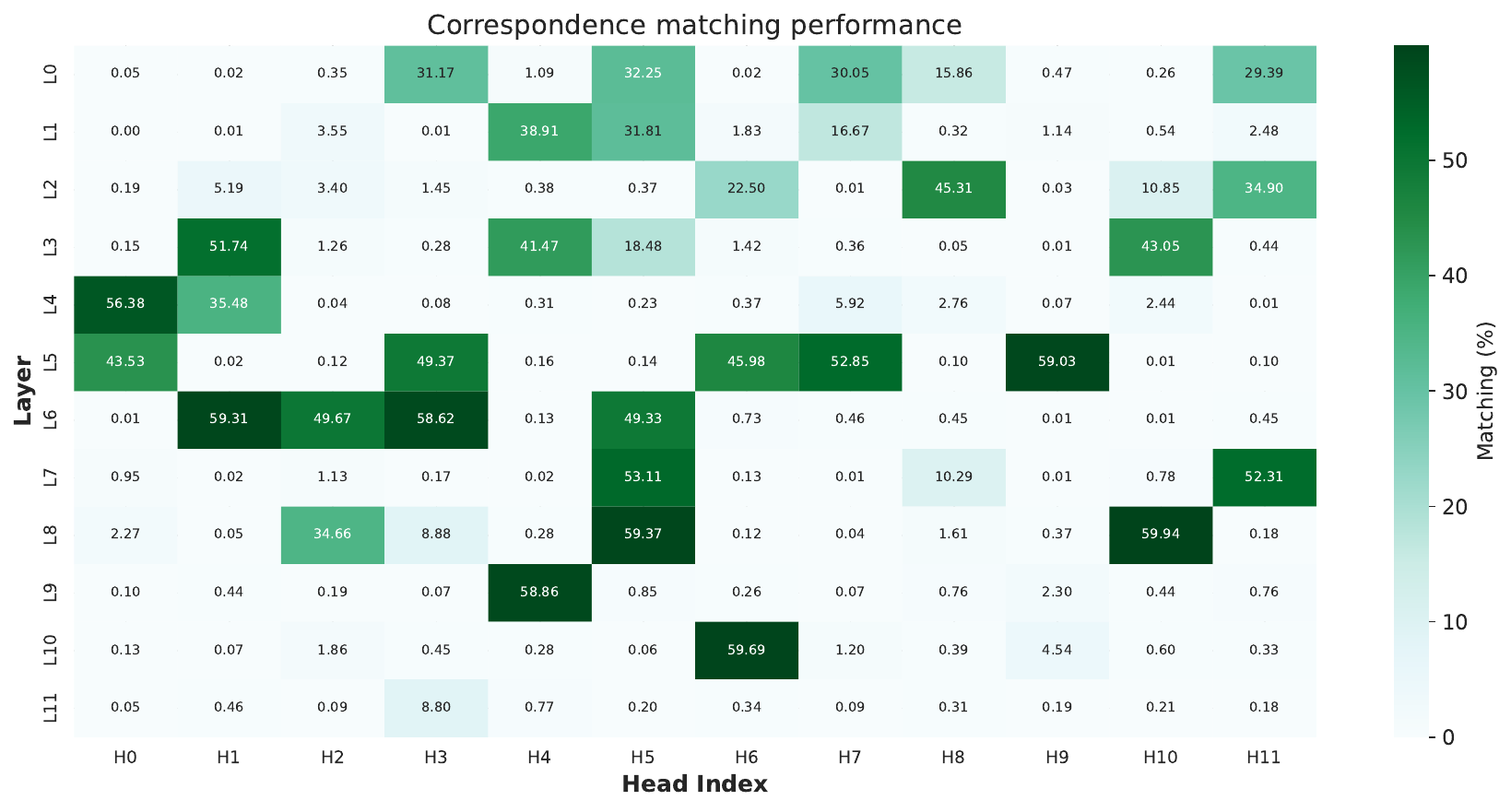} \\
\end{tabular}

\begin{tabular}{@{}c c c c c@{}}
  && \textbf{VGGT} & \textbf{Depth Anything 3} & \textbf{DUSt3R} \\
  \multirow{2}{*}{\datasetlab{DTU}} &
  \rowlab{F (1$\to$2)} &
  \includegraphics[width=\imwidth]{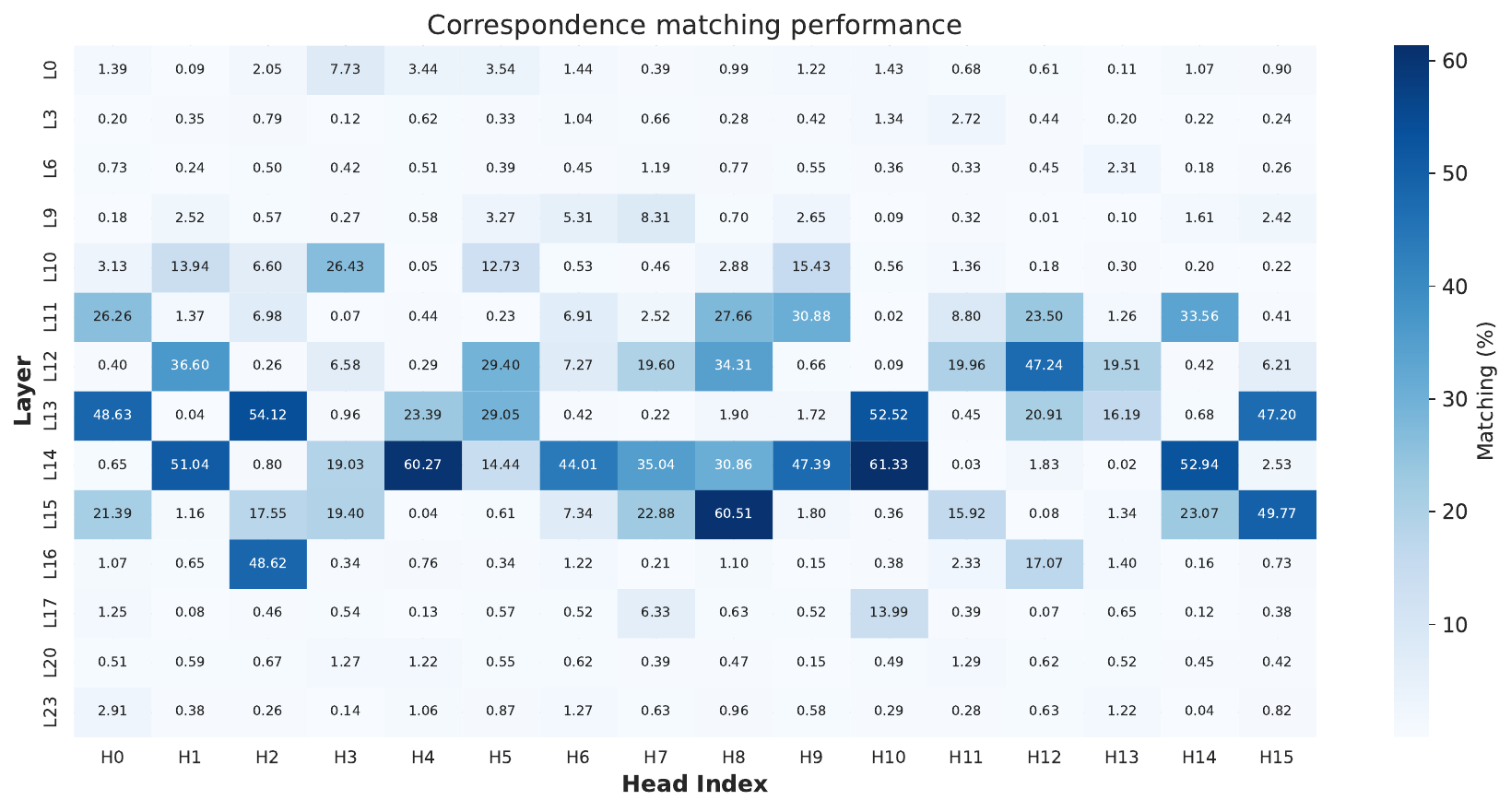} &
  \includegraphics[width=\imwidth]{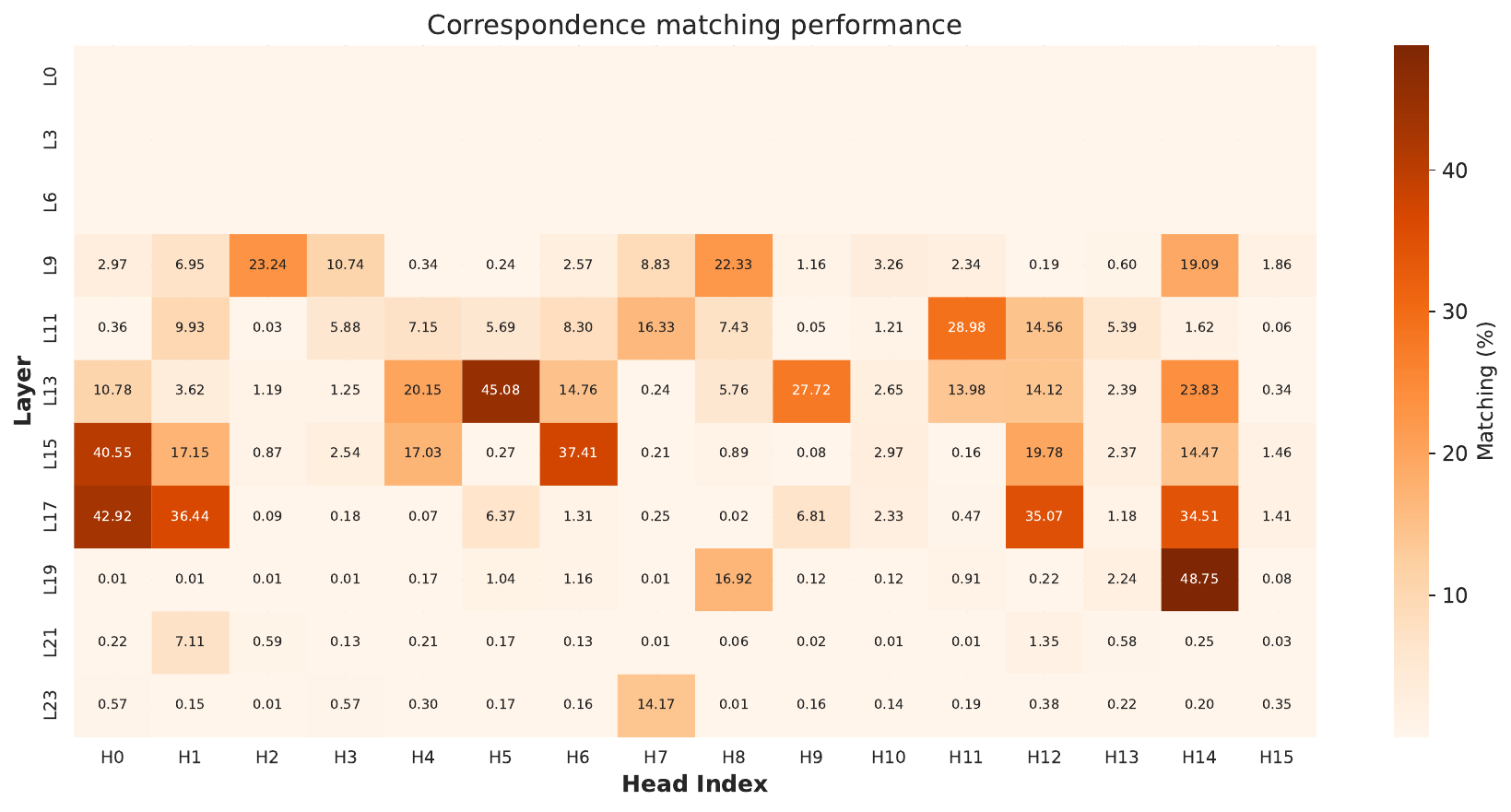} &
  \includegraphics[width=\imwidth]{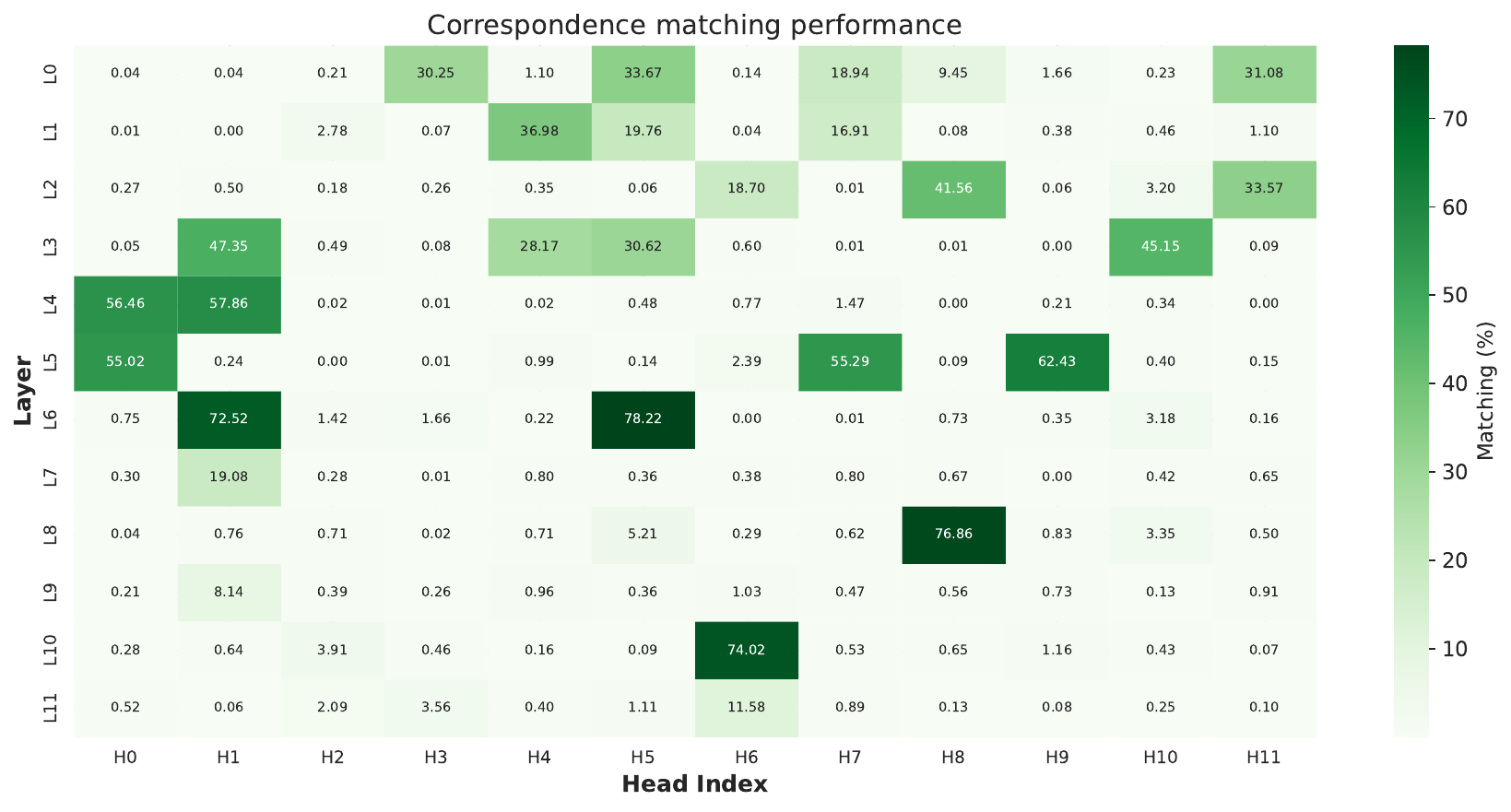} \\
  &
  \rowlab{R (2$\to$1)} &
  \includegraphics[width=\imwidth]{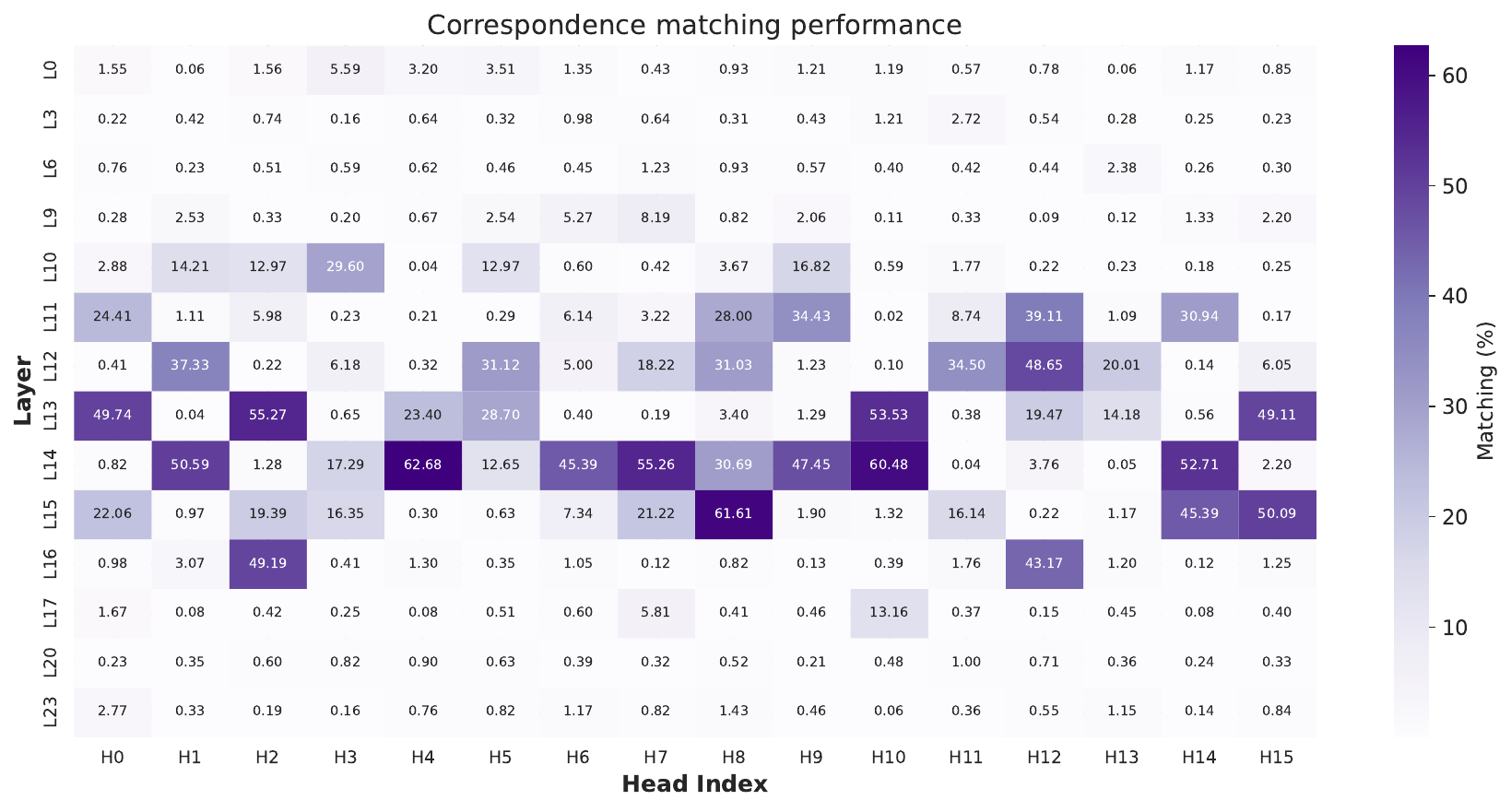} &
  \includegraphics[width=\imwidth]{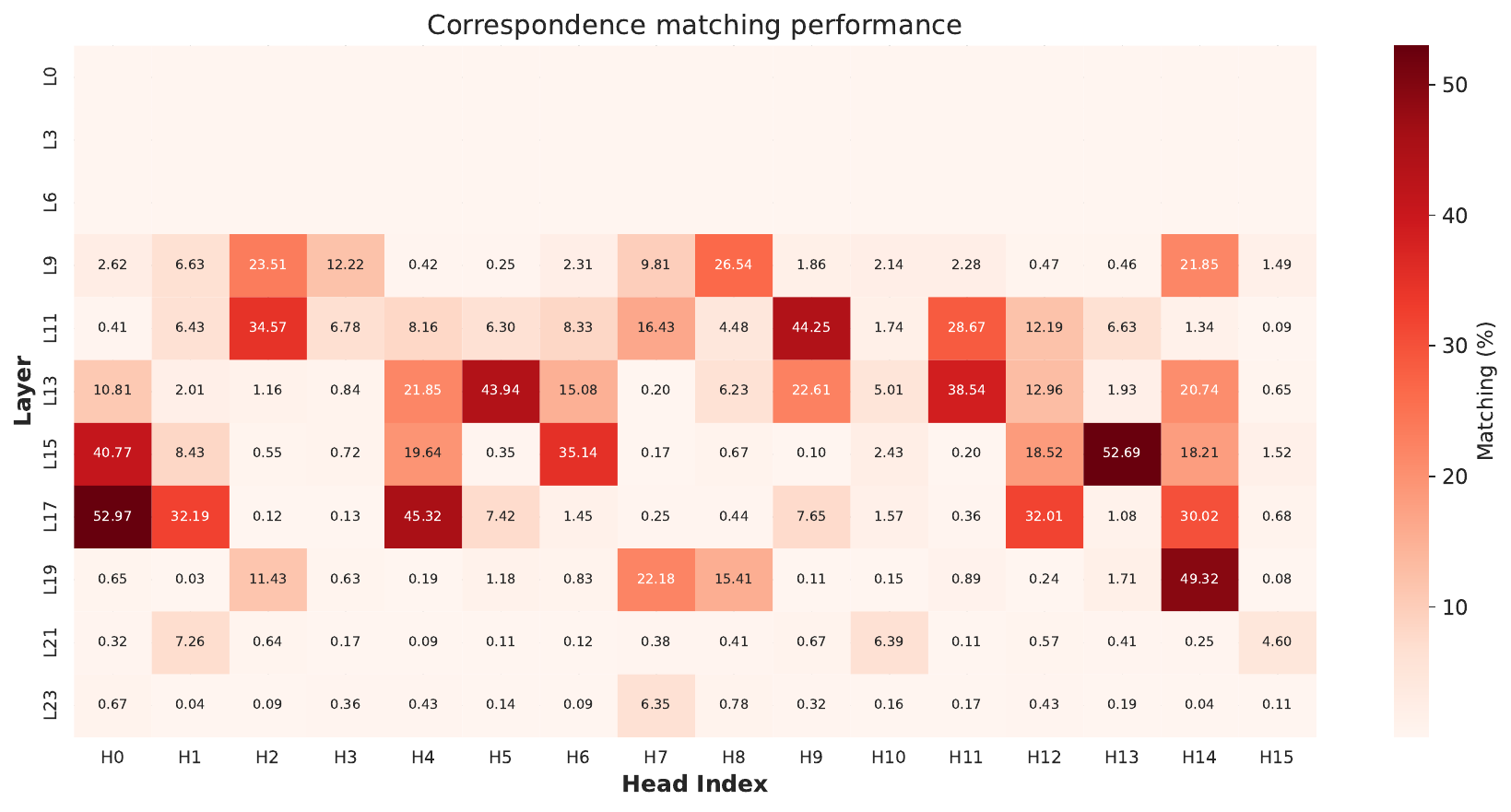} &
  \includegraphics[width=\imwidth]{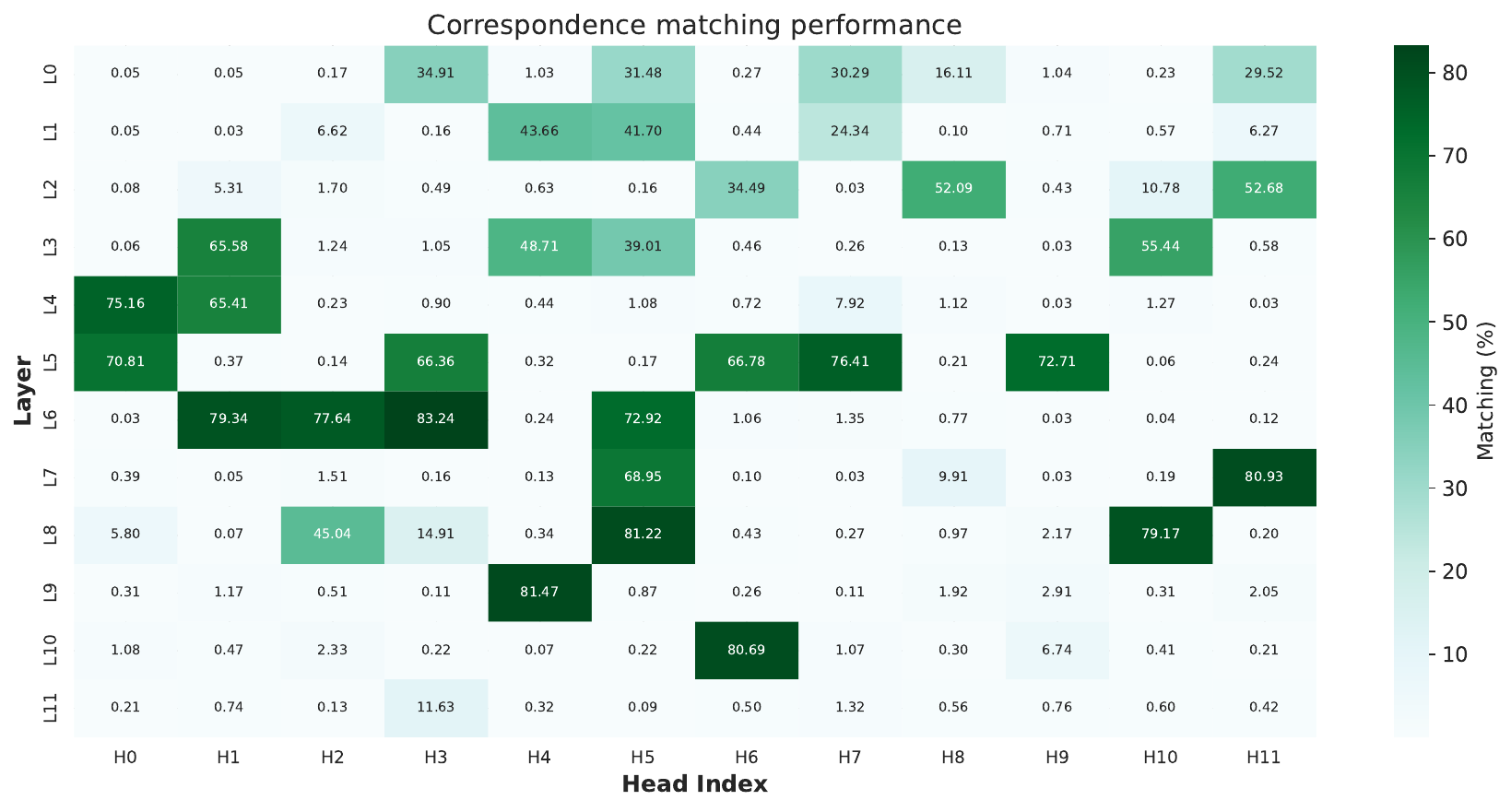} \\
\end{tabular}

\begin{tabular}{@{}c c c c c@{}}
  & &\textbf{VGGT} & \textbf{Depth Anything 3} & \textbf{DUSt3R} \\
  \multirow{2}{*}{\datasetlab{MipNeRF}} &
  \rowlab{F (1$\to$2)} &
  \includegraphics[width=\imwidth]{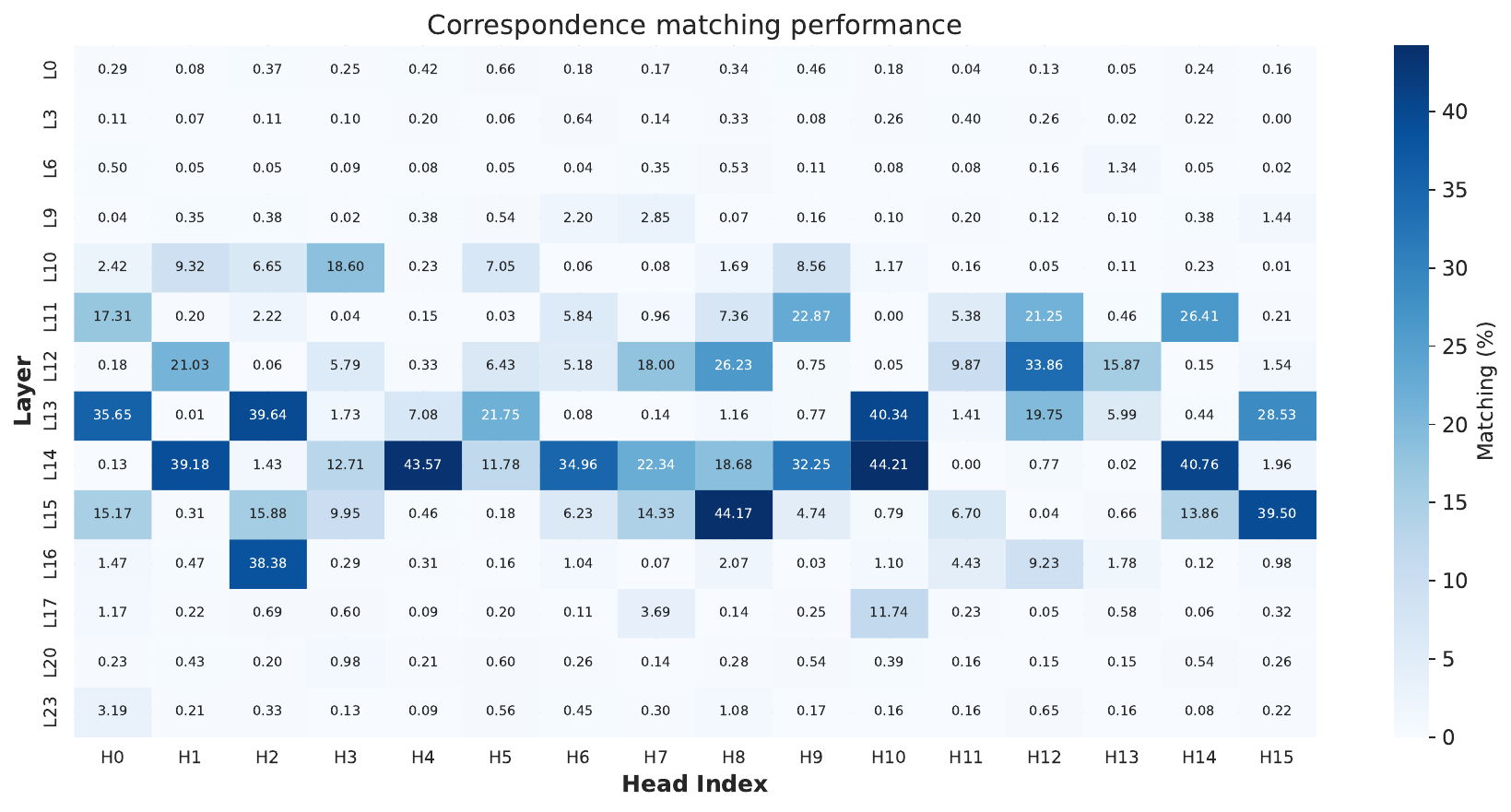} &
  \includegraphics[width=\imwidth]{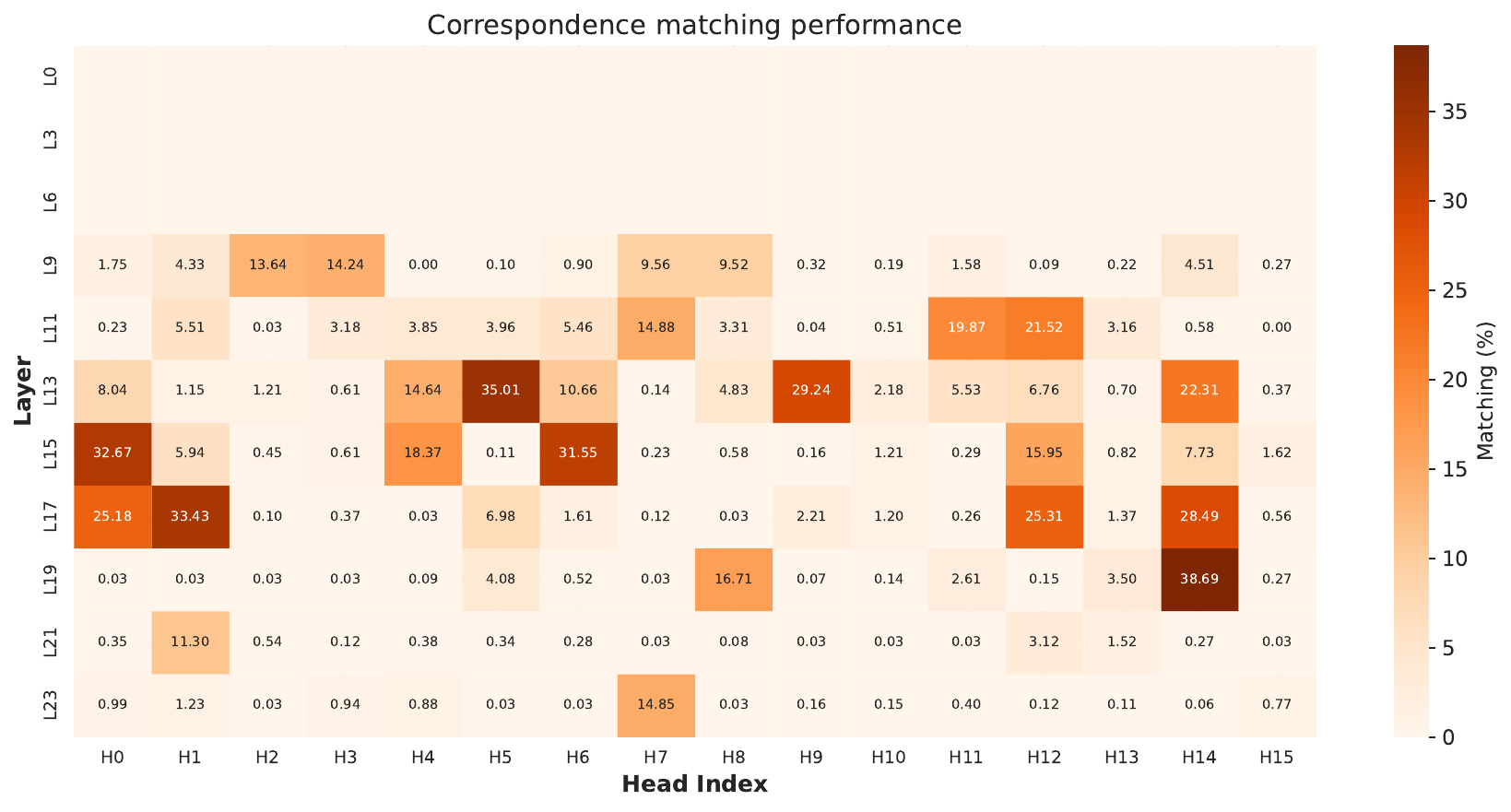} &
  \includegraphics[width=\imwidth]{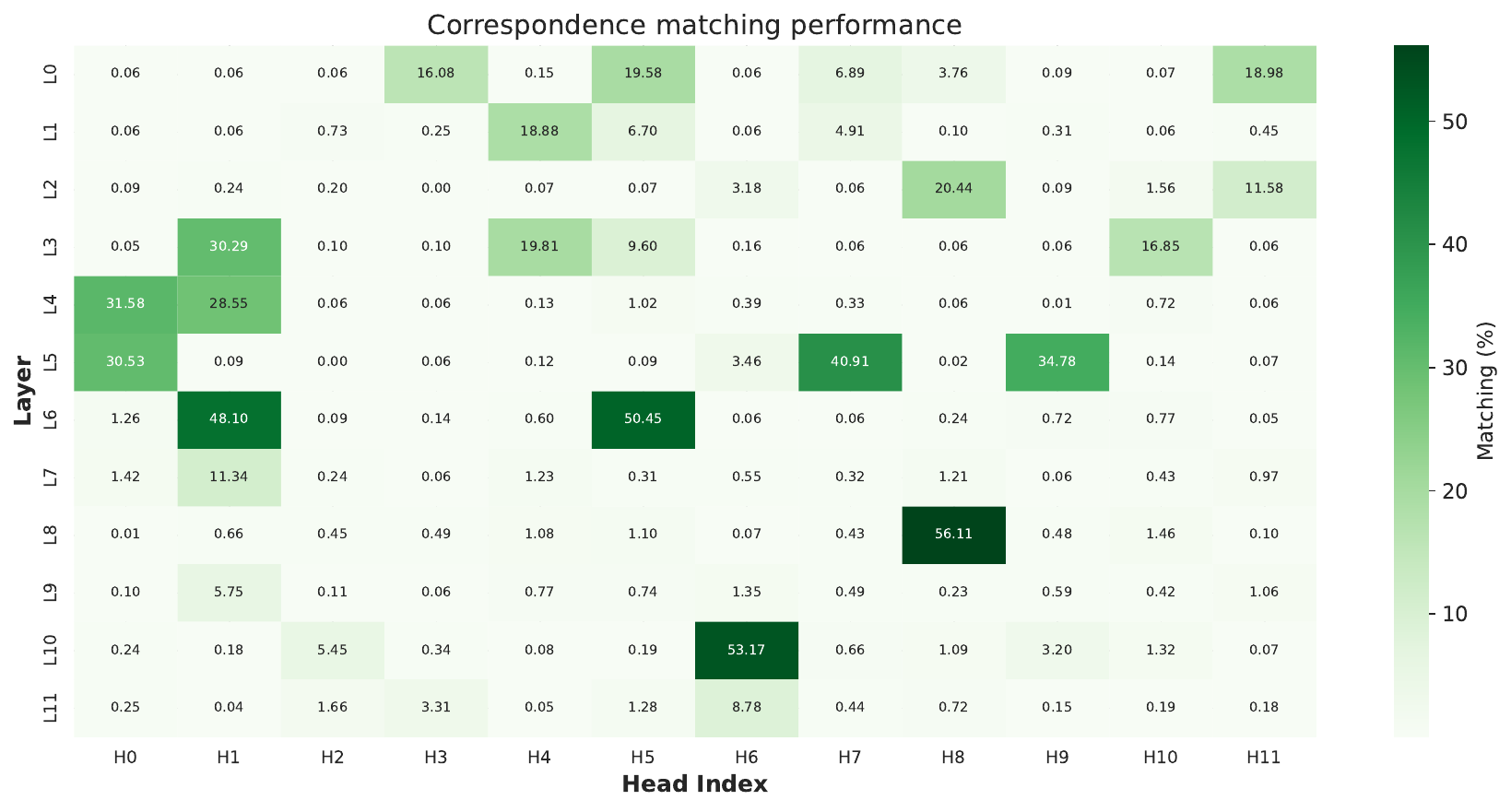} \\
  &
  \rowlab{R (2$\to$1)} &
  \includegraphics[width=\imwidth]{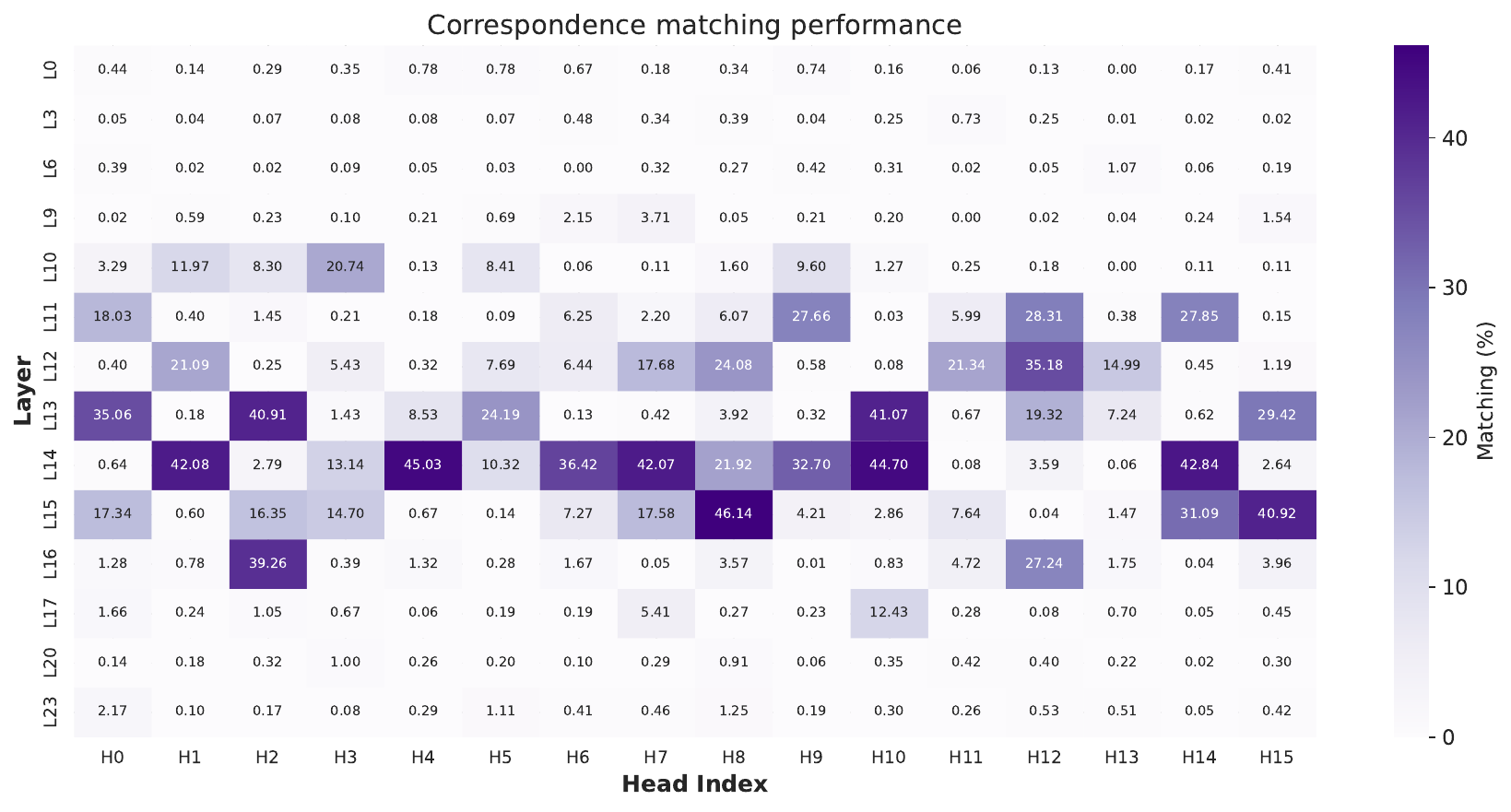} &
  \includegraphics[width=\imwidth]{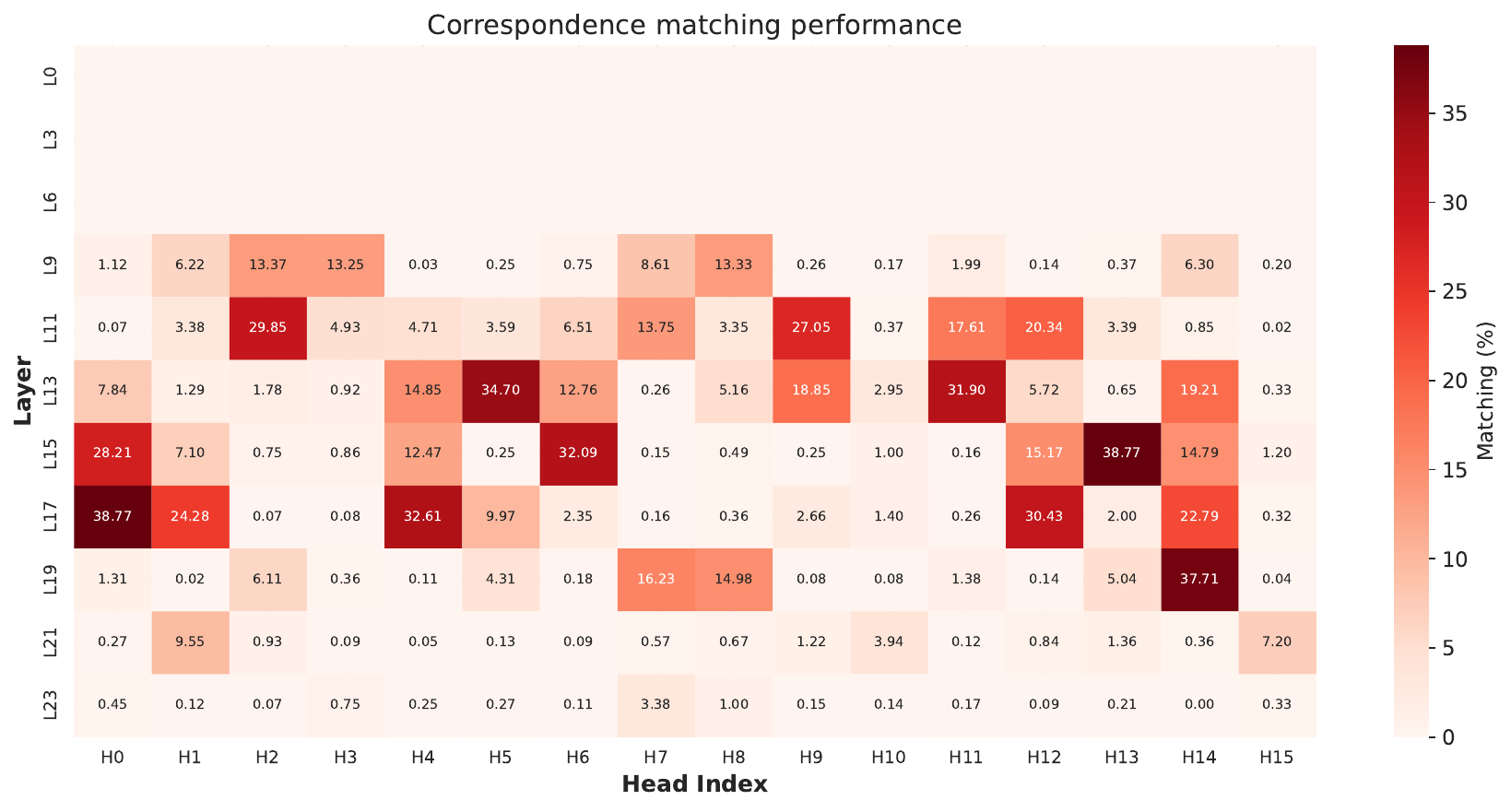} &
  \includegraphics[width=\imwidth]{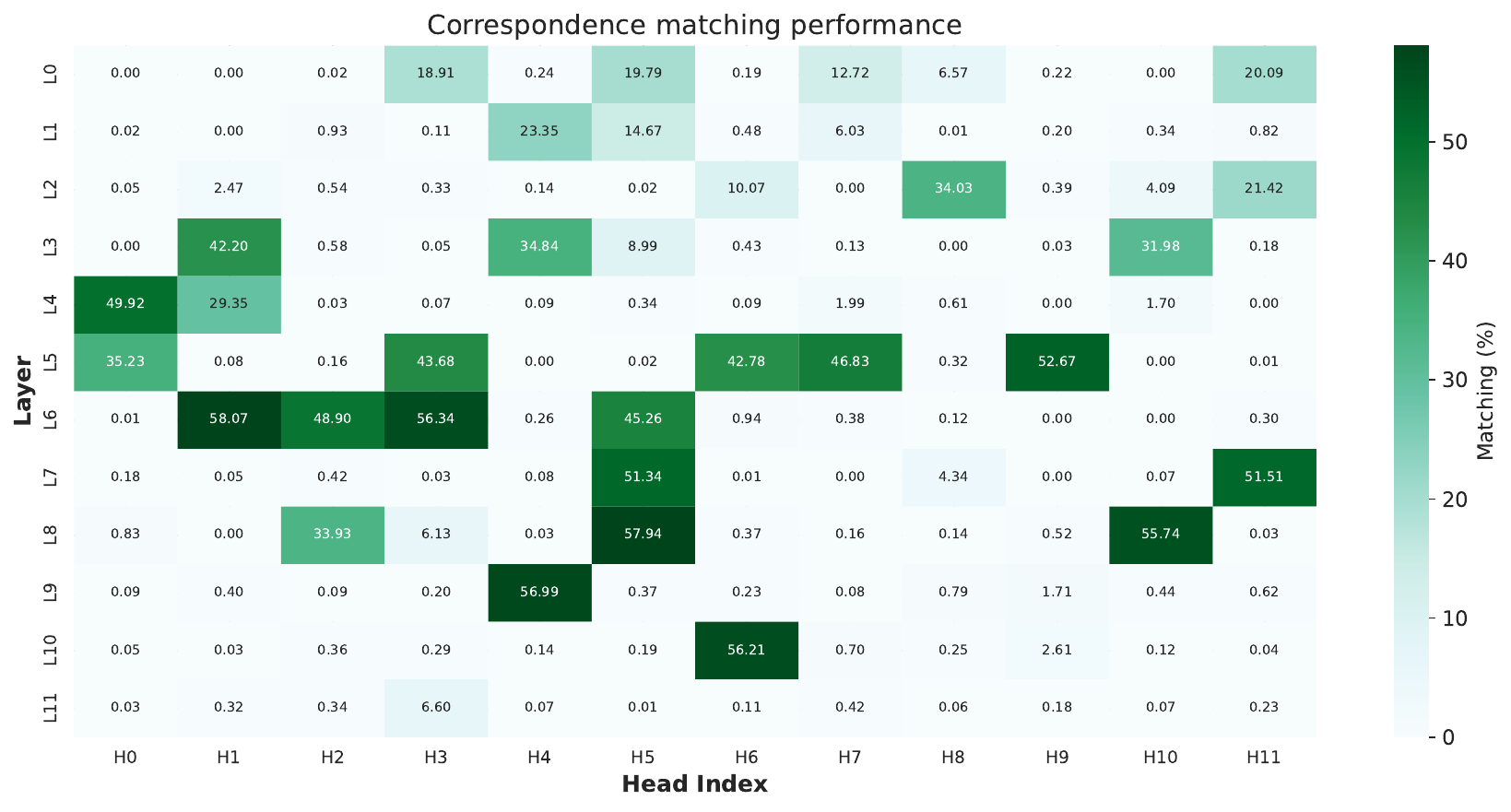} \\
\end{tabular}

\caption{\textbf{Correspondence matching per attention head for real-world datasets.} We compare correspondence matching performance using the QK attention space for forward (view 1$\xrightarrow{}$view 2) and reverse (view 2$\xrightarrow{}$view 1) directions. 
Strong matching performance is observed in the layers immediately preceding the transition in the probing experiment. We observe the same trends as for our ShapeNet dataset.}
\label{fig:corr_real_world}
\end{figure}

\subsection{Analysis in the attention space}

We analyze the attention maps of the \vggt's and \da's global attention layers for correspondence matching capabilities, while for \duster~we use its cross-attention layers from the asymmetrical decoder. Precisely, we analyze the query-key (QK) attention space after softmax, where for each pair of corresponding patches (patch from the first image and its corresponding patch in the second image and vice versa), we look for the location in the attention map with the highest value and calculate the accuracy whether the patch with highest accuracy corresponds to the true matching patches or not. Since there will be scenarios in which a patch from the first image has multiple matches in the second image, we count a hit if any of the correct positions receives the highest attention value.  We report the results on the test splits.

On~\cref{fig:corr_real_world}, we show the correspondence matching results for ETH3D (first 2 rows), DTU (middle 2 rows), and MipNeRF360 (last 2 rows). We observe the same trends as for our ShapeNet dataset, where correspondence-matching activity in the layers is connected with the well-performing layers in probing. For \vggt~and \da, the middle layers are the ones with high correspondence-matching activity, while for \duster, we observe correspondence-matching activity already from the early layers.

\begin{figure}[t!]
\centering
\newcommand{\rowlabelwidth}{0.55cm}
\newcommand{\dslabwidth}{0.45cm}
\newcommand{\imwidth}{0.305\textwidth}
\setlength{\tabcolsep}{2pt}
\renewcommand{\arraystretch}{0}
\newcommand{\rowlab}[1]{%
  \makebox[\rowlabelwidth][c]{\raisebox{2.5ex}{\rotatebox{90}{\textbf{#1}}}}%
}
\newcommand{\datasetlab}[1]{%
  \makebox[\dslabwidth][c]{\raisebox{5.5ex}{\rotatebox{90}{\textbf{#1}}}}%
}

\setlength{\tabcolsep}{3.5pt}
\begin{tabular}{@{}c l l c c r r@{}}
  \multicolumn{1}{c}{\hspace{\dslabwidth}} &
  \multicolumn{2}{c}{\textbf{VGGT}} &
  \multicolumn{2}{c}{\textbf{Depth Anything 3}} &
  \multicolumn{2}{c}{\textbf{DUSt3R}} \\[2pt]
  \datasetlab{Synthetic} &
  \includegraphics[width=0.12\textwidth]{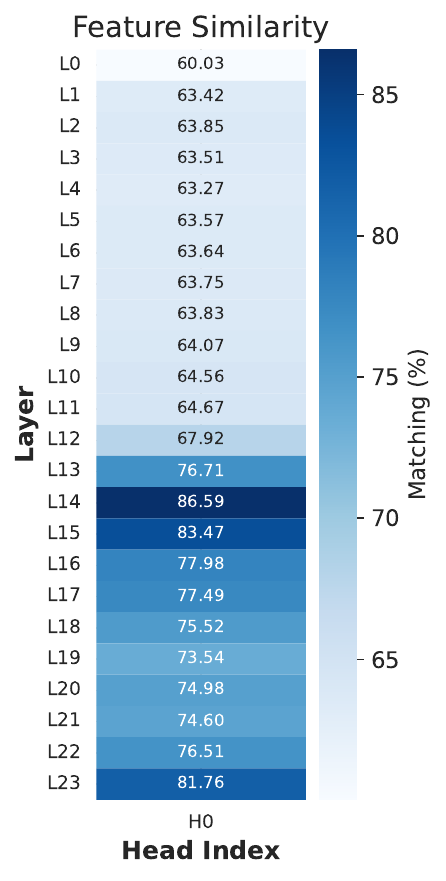} &
  \includegraphics[width=0.12\textwidth]{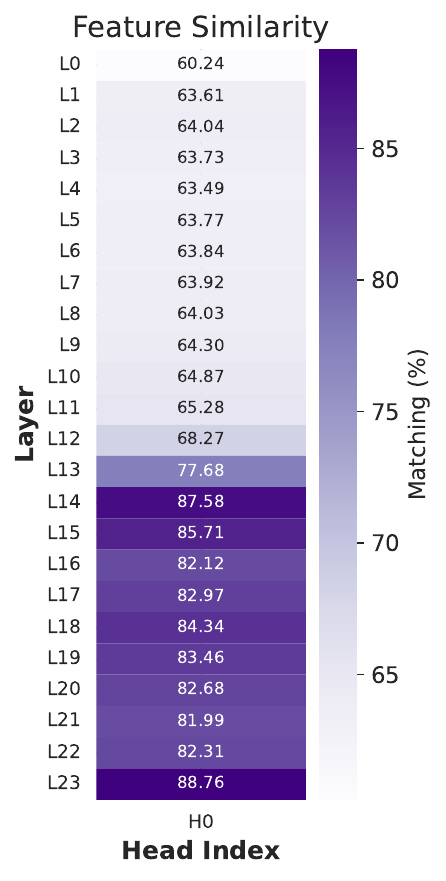} &
  \includegraphics[width=0.12\textwidth]{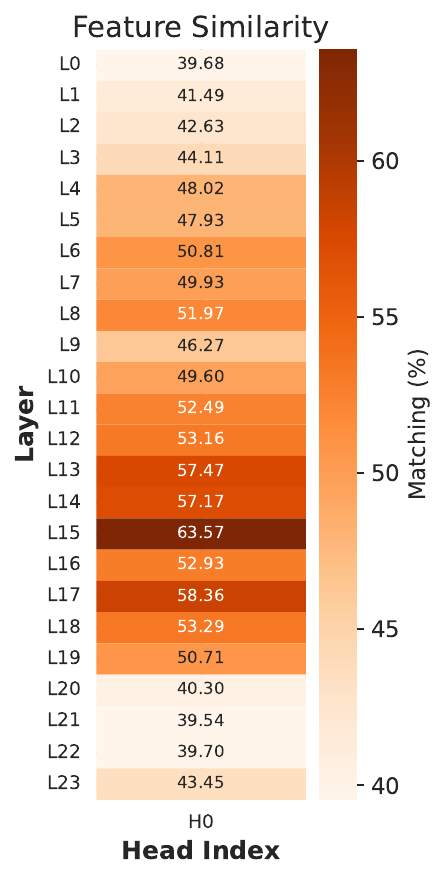} &
  \includegraphics[width=0.12\textwidth]{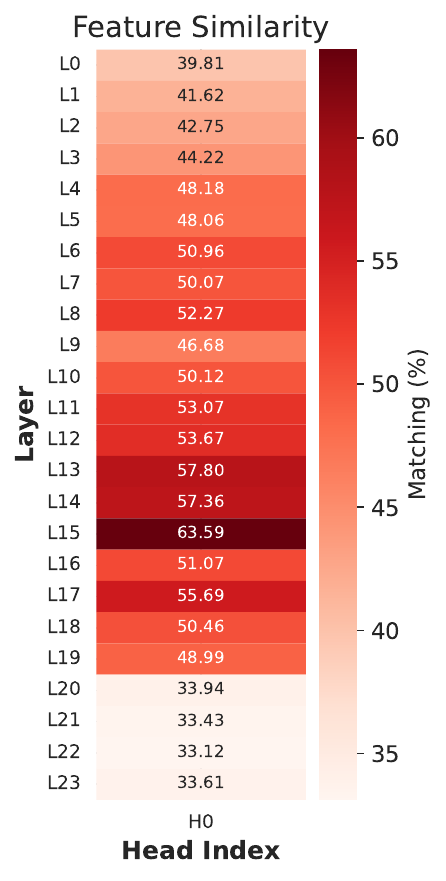} &
  \includegraphics[width=0.12\textwidth]{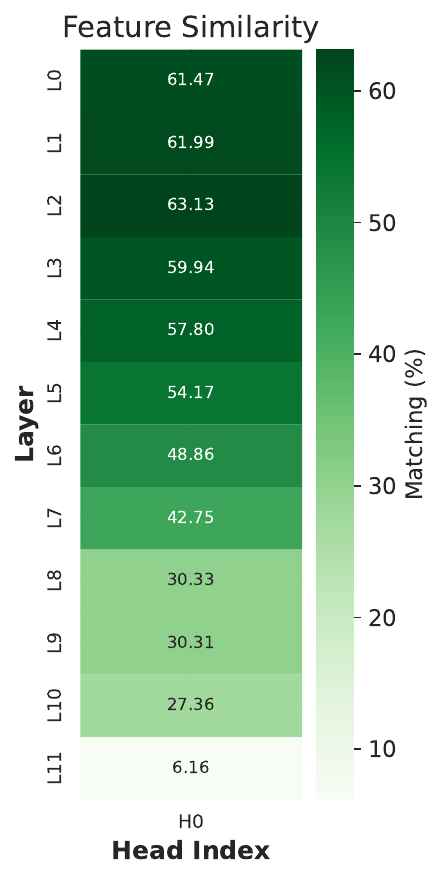} &
  \includegraphics[width=0.12\textwidth]{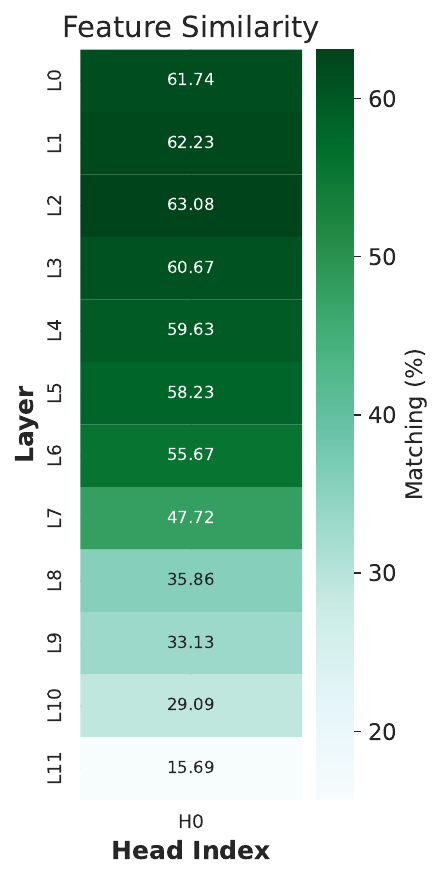} \\[2pt]
  \datasetlab{ETH3D} &
  \includegraphics[width=0.12\textwidth]{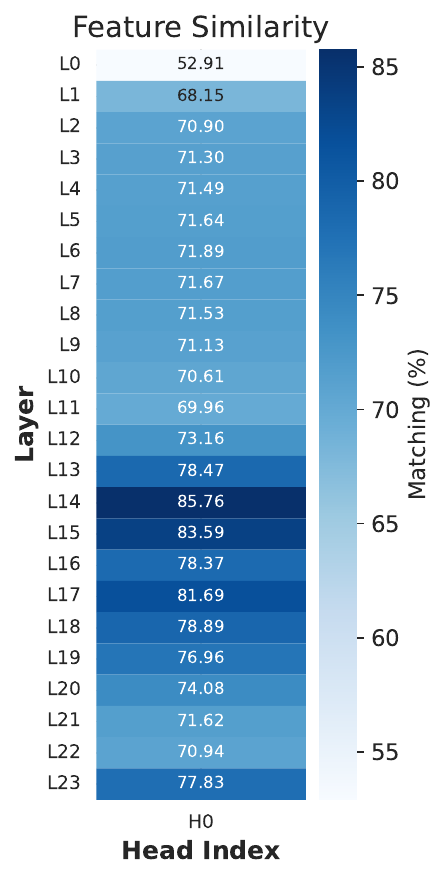} &
  \includegraphics[width=0.12\textwidth]{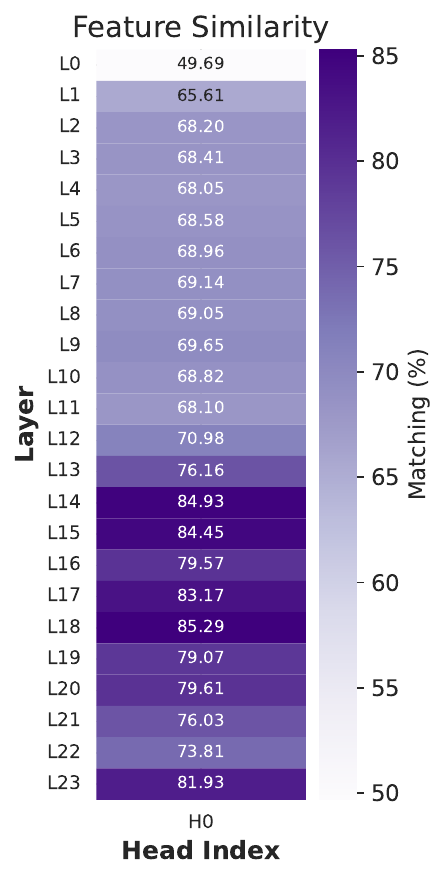} &
  \includegraphics[width=0.12\textwidth]{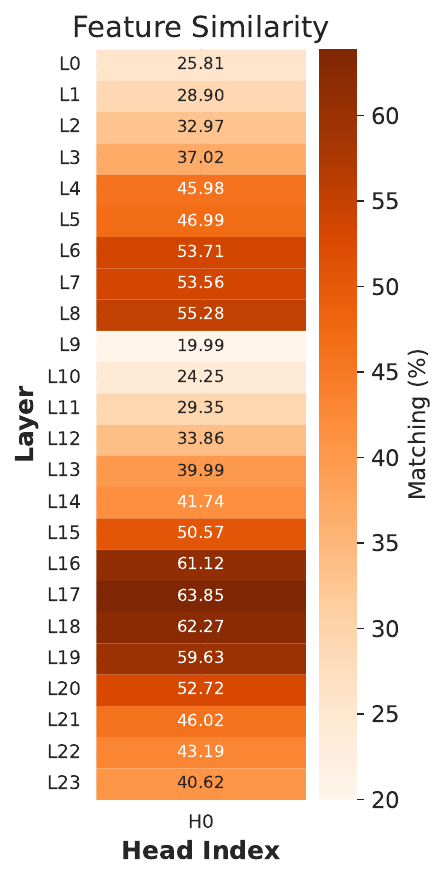} &
  \includegraphics[width=0.12\textwidth]{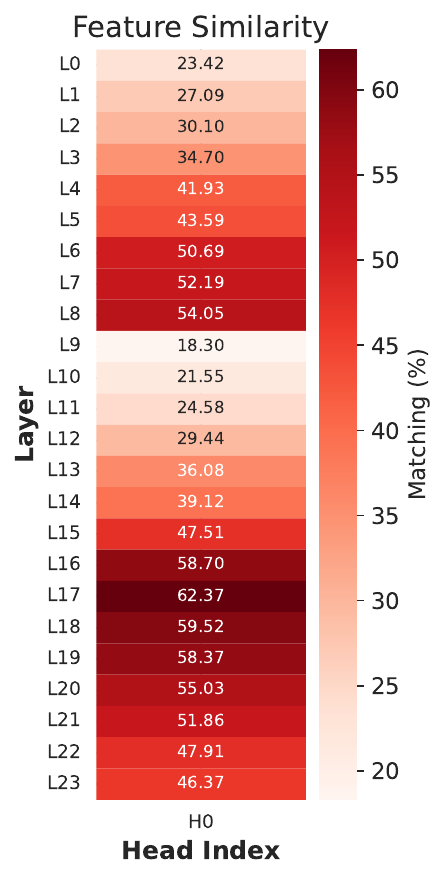} &
  \includegraphics[width=0.12\textwidth]{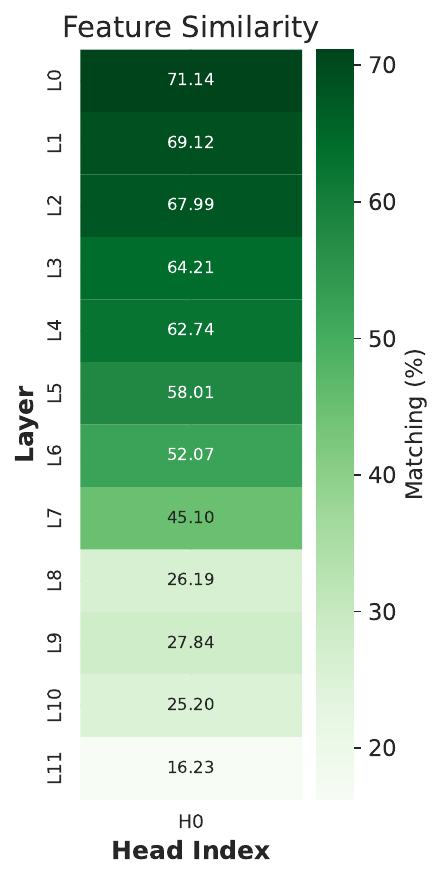} &
  \includegraphics[width=0.12\textwidth]{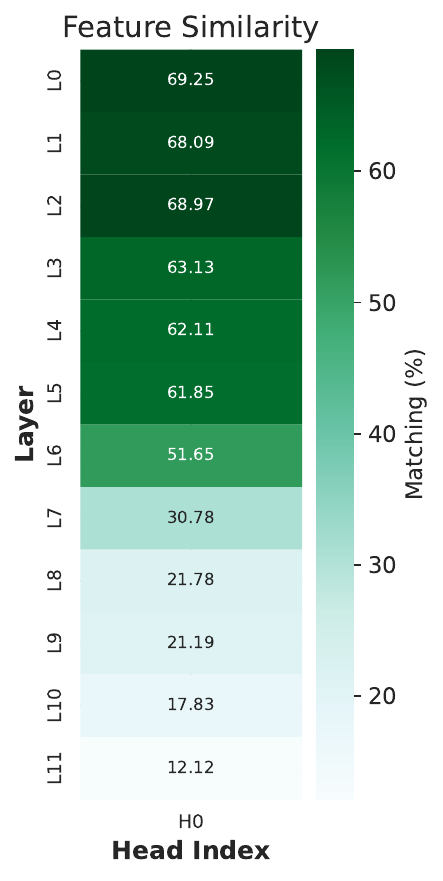} \\[2pt]
  \datasetlab{DTU} &
  \includegraphics[width=0.12\textwidth]{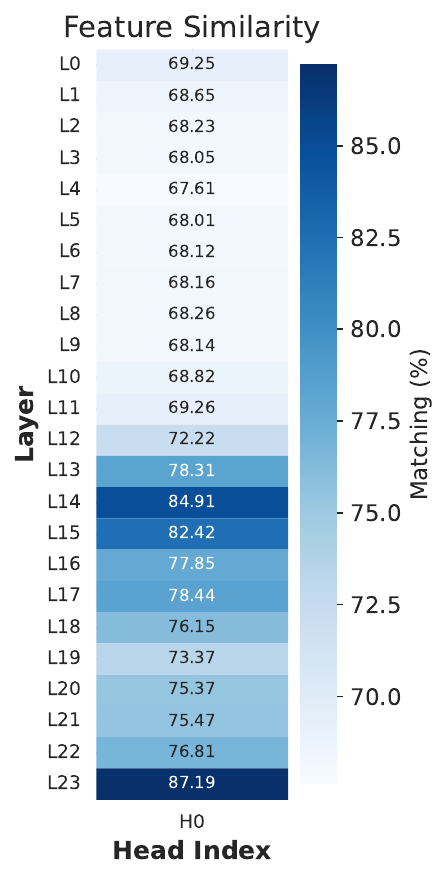} &
  \includegraphics[width=0.12\textwidth]{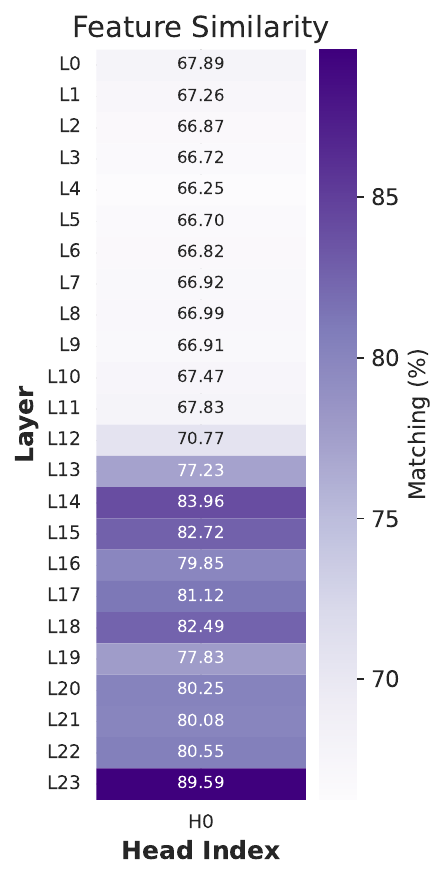} &
  \includegraphics[width=0.12\textwidth]{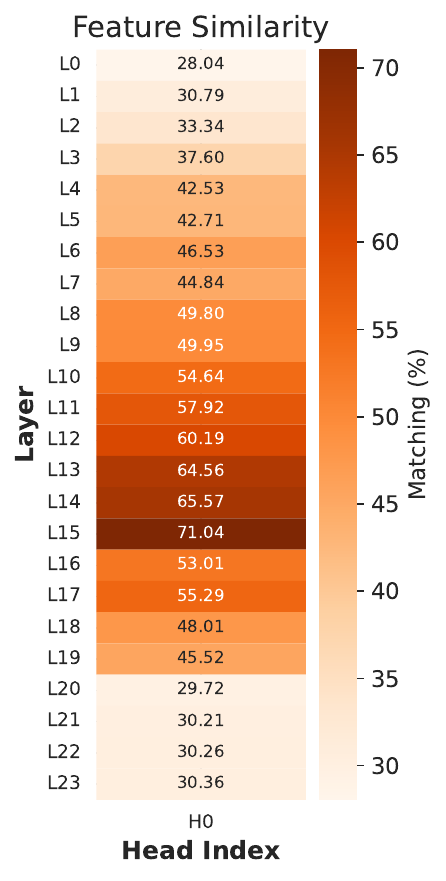} &
  \includegraphics[width=0.12\textwidth]{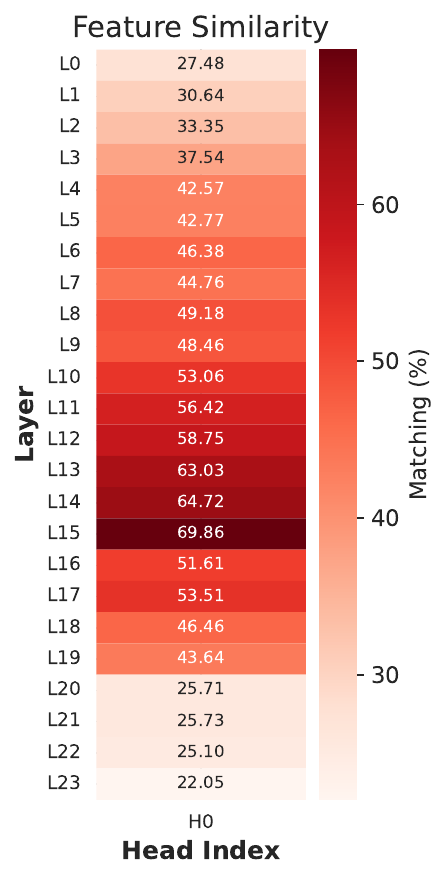} &
  \includegraphics[width=0.12\textwidth]{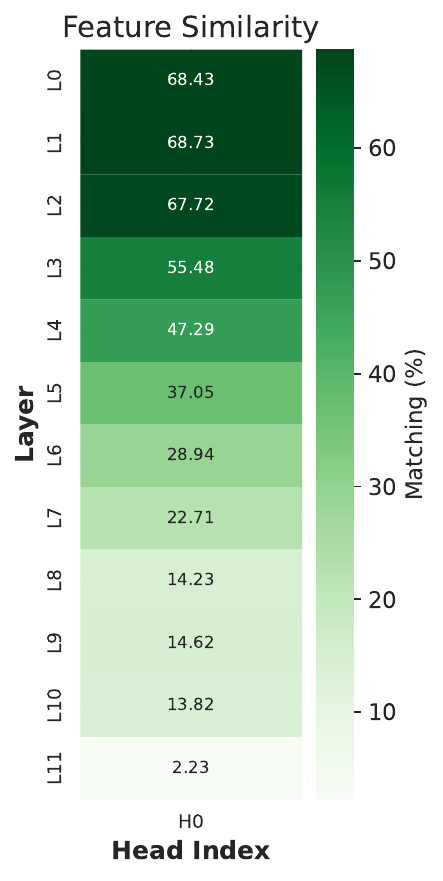} &
  \includegraphics[width=0.12\textwidth]{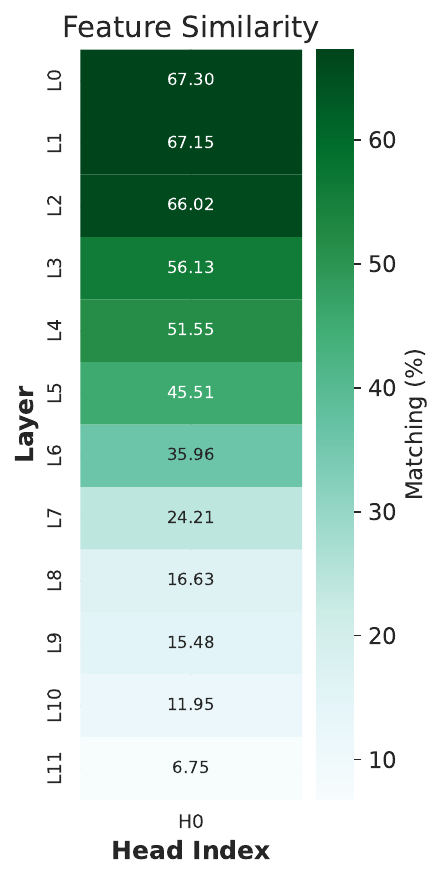} \\
\end{tabular}

\caption{\textbf{Correspondence matching through feature similarity.} We compare correspondence matching performance using the feature similarity for forward (left in pair) and reverse (right in pair) directions. We observe that models retain geometric correspondence similarity in their features, and this similarity becomes stronger after the matching ability in the QK attention space emerges.}
\label{fig:corr_matching_features}
\end{figure}

\subsection{Analysis through feature similarity} 

Since QK attention space can be misleading~\cite{Chefer2021CVPR}, we further analyze the correspondence matching performance in the feature space. We aim to confirm whether the attention patterns observed in the QK space are retained in the features. We calculate the cosine similarity of the features at the end of the layer and calculate the nearest neighbors, where we assume a correspondence is matched accurately if its nearest neighbor (patch with maximum feature cosine similarity) is maximally 3 patches away spatially. We report the results on the test splits. 

In~\cref{fig:corr_matching_features}, we present results on synthetic and real-world datasets, with each row corresponding to a different dataset (first row ShapeNet, second row ETH3D, third row DTU). We observe similar or the same trends.

\begin{figure}[b!]
\centering
\newcommand{\rowlabelwidth}{0.55cm}
\newcommand{\dslabwidth}{0.45cm}
\newcommand{\imwidth}{0.305\textwidth}
\setlength{\tabcolsep}{2pt}
\renewcommand{\arraystretch}{0}

\setlength{\tabcolsep}{3.5pt}
\begin{tabular}{@{}l l c c r r@{}}
  \multicolumn{2}{c}{\textbf{VGGT}} &
  \multicolumn{2}{c}{\textbf{Depth Anything 3}} &
  \multicolumn{2}{c}{\textbf{DUSt3R}} \\[2pt]
  \includegraphics[width=0.12\textwidth]{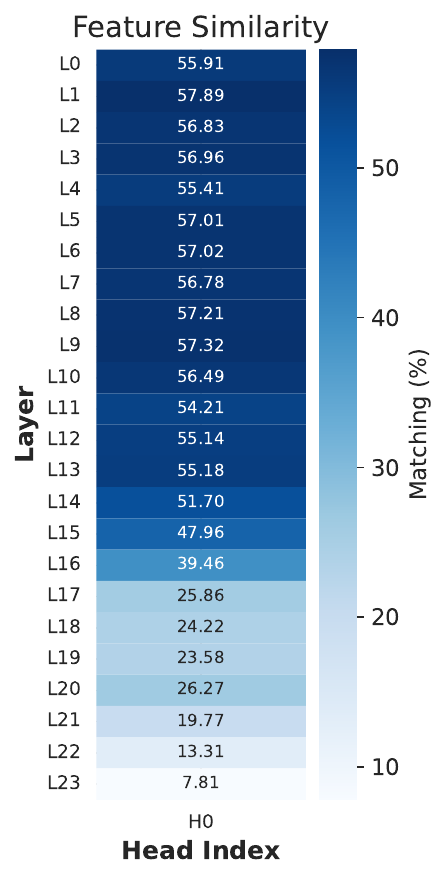} &
  \includegraphics[width=0.12\textwidth]{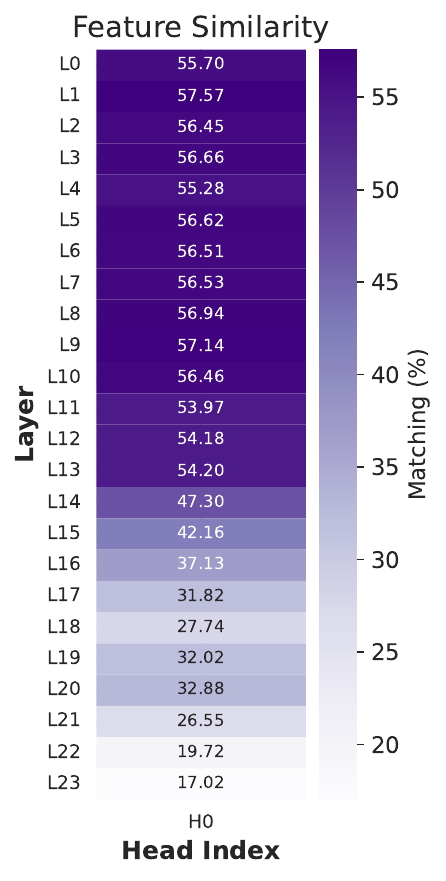} &
  \includegraphics[width=0.12\textwidth]{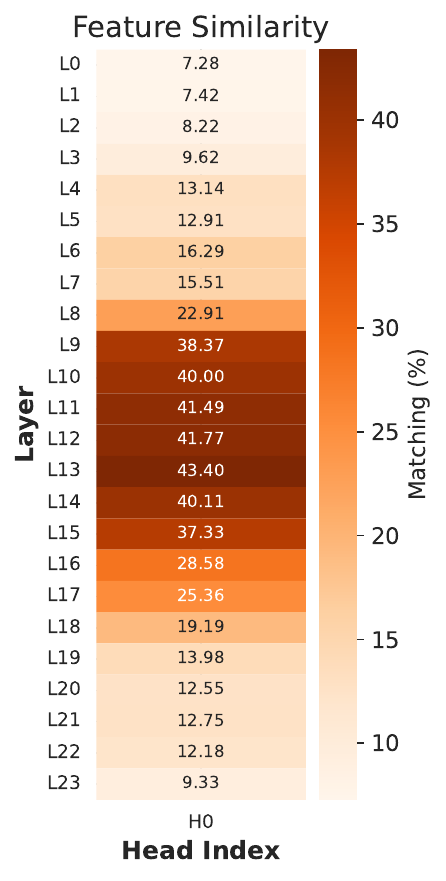} &
  \includegraphics[width=0.12\textwidth]{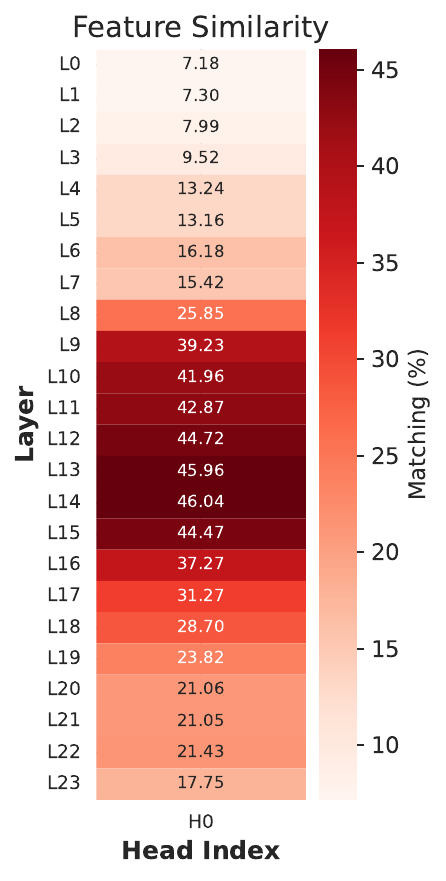} &
  \includegraphics[width=0.12\textwidth]{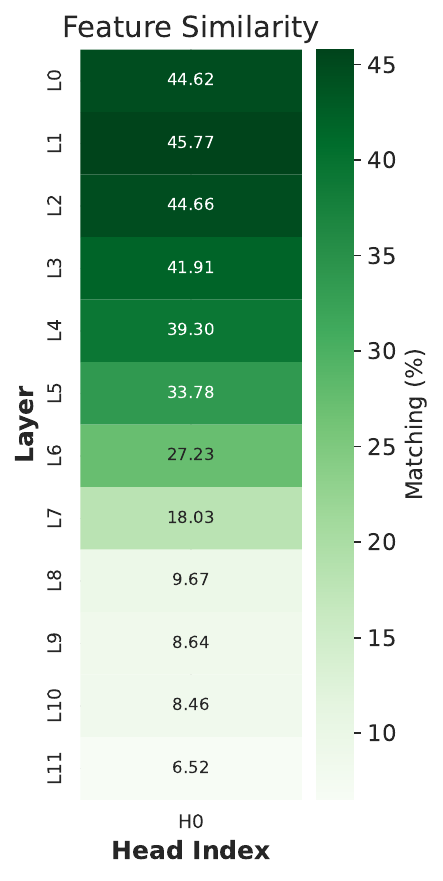} &
  \includegraphics[width=0.12\textwidth]{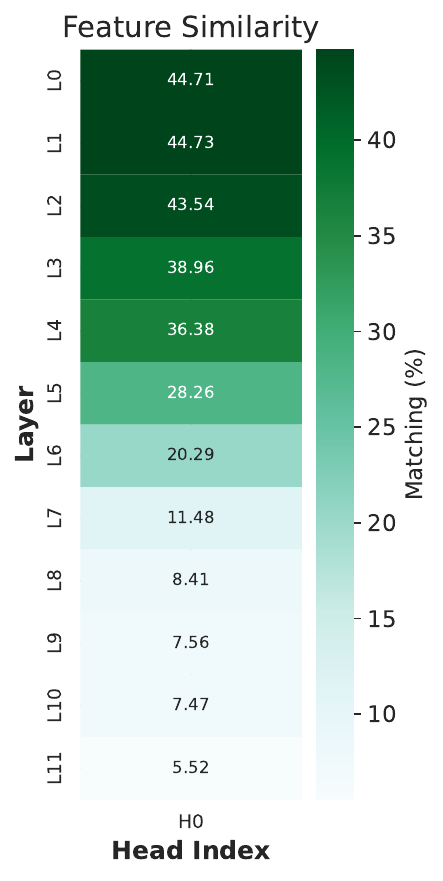} \\[2pt]

\end{tabular}

\caption{\textbf{Semantic correspondence matching through feature similarity.} We compare semantic correspondence matching performance using the feature similarity for forward (left in pair) and reverse (right in pair) directions from SPair71k dataset. We observe different trends for different model families. }
\label{fig:sem_matching}
\end{figure}

For \vggt, we observe that early-to-middle layers have lower geometric feature similarity than middle-to-late layers, and we see that information about geometric similarity is retained in the features, with maximum similarity towards the end of the network, since this layer is also passed to the downstream heads. 

For \da, we observe the highest geometric feature similarity in the middle layers. We believe this stems from the architectural design, in which multiple layers are provided to the downstream heads. 

Finally, for \duster, we observe that geometric feature similarity is already present in the early layers and slightly decreases towards the end of the asymmetrical decoder layers.

\subsection{Analysis of the semantic correspondences}

Our initial analysis with VGGT and ShapeNet data shows that the early layers of VGGT perform semantic correspondence matching, meaning the model matches the same semantic parts in the scene, but not necessarily to the same instance. To quantify this matching, we follow the standard procedure in semantic correspondence benchmarking and report feature similarity for the semantic correspondences using cosine similarity and nearest-neighbor distance. We report the accuracy of the distance within 3 patches of the SPair71k test set.

\begin{figure}[b!]
\centering
\newcommand{\rowlabelwidth}{0.45cm}
\newcommand{\imwidth}{0.28\textwidth}
\setlength{\tabcolsep}{2pt} 
\renewcommand{\arraystretch}{0} 

\newcommand{\rowlab}[1]{%
  \makebox[\rowlabelwidth][c]{\raisebox{2.5ex}{\rotatebox{90}{\textbf{#1}}}}%
}

\newcommand{\datasetlab}[1]{%
  \makebox[\rowlabelwidth][r]{\raisebox{5.5ex}{\rotatebox{90}{\textbf{#1}}}}%
}

\begin{tabular}{@{}c c c c c@{}}
  && \textbf{DA3 Small} & \textbf{DA3 Large} & \textbf{DA3 Giant} \\

  \multirow{2}{*}{\datasetlab{Synthetic}} &
  \rowlab{F (1$\to$2)} &
  \includegraphics[width=\imwidth]{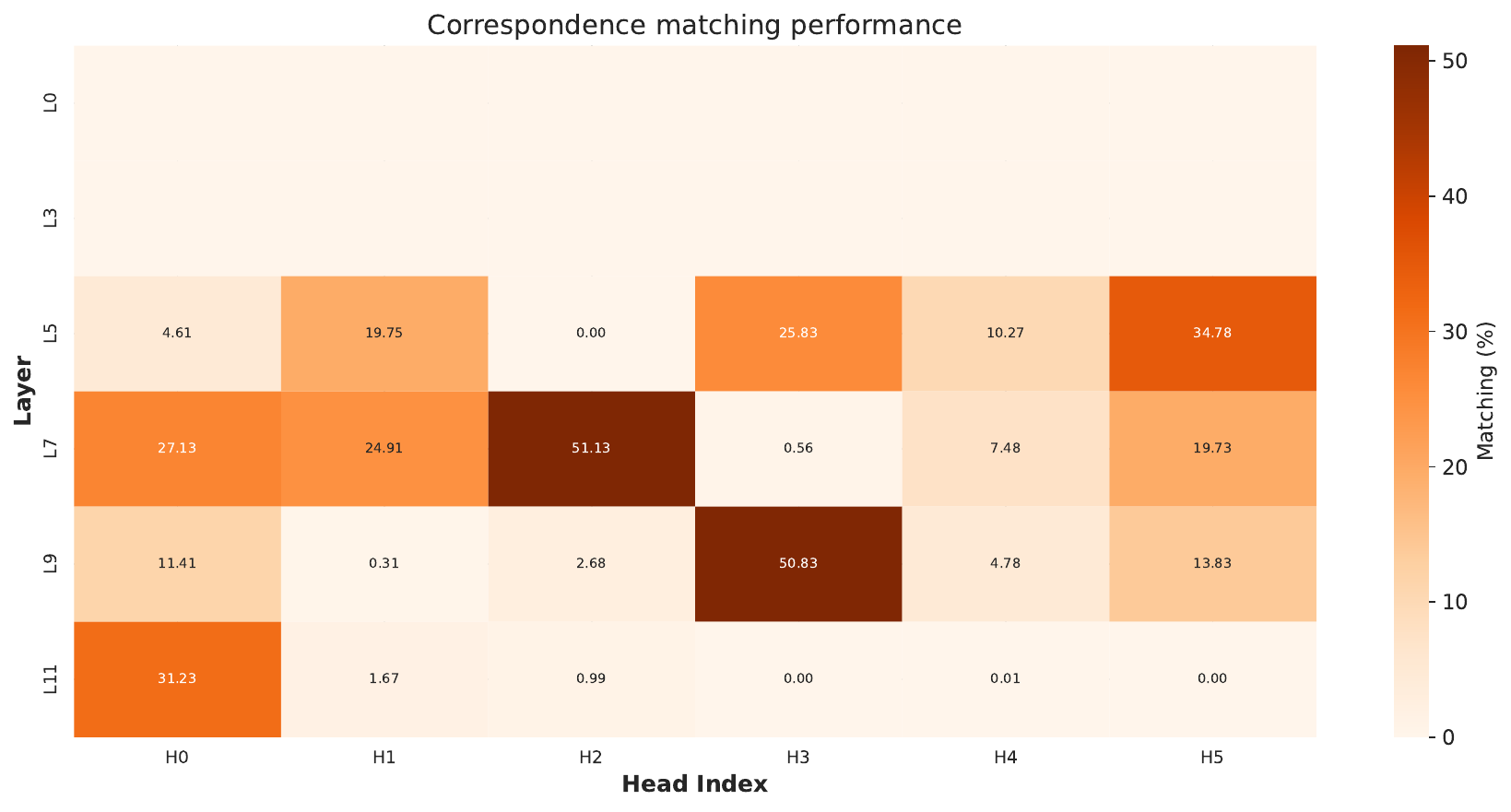} &
  \includegraphics[width=\imwidth]{figures/corr_matching/da3_large/forward_matching_heatmap.pdf} &
  \includegraphics[width=\imwidth]{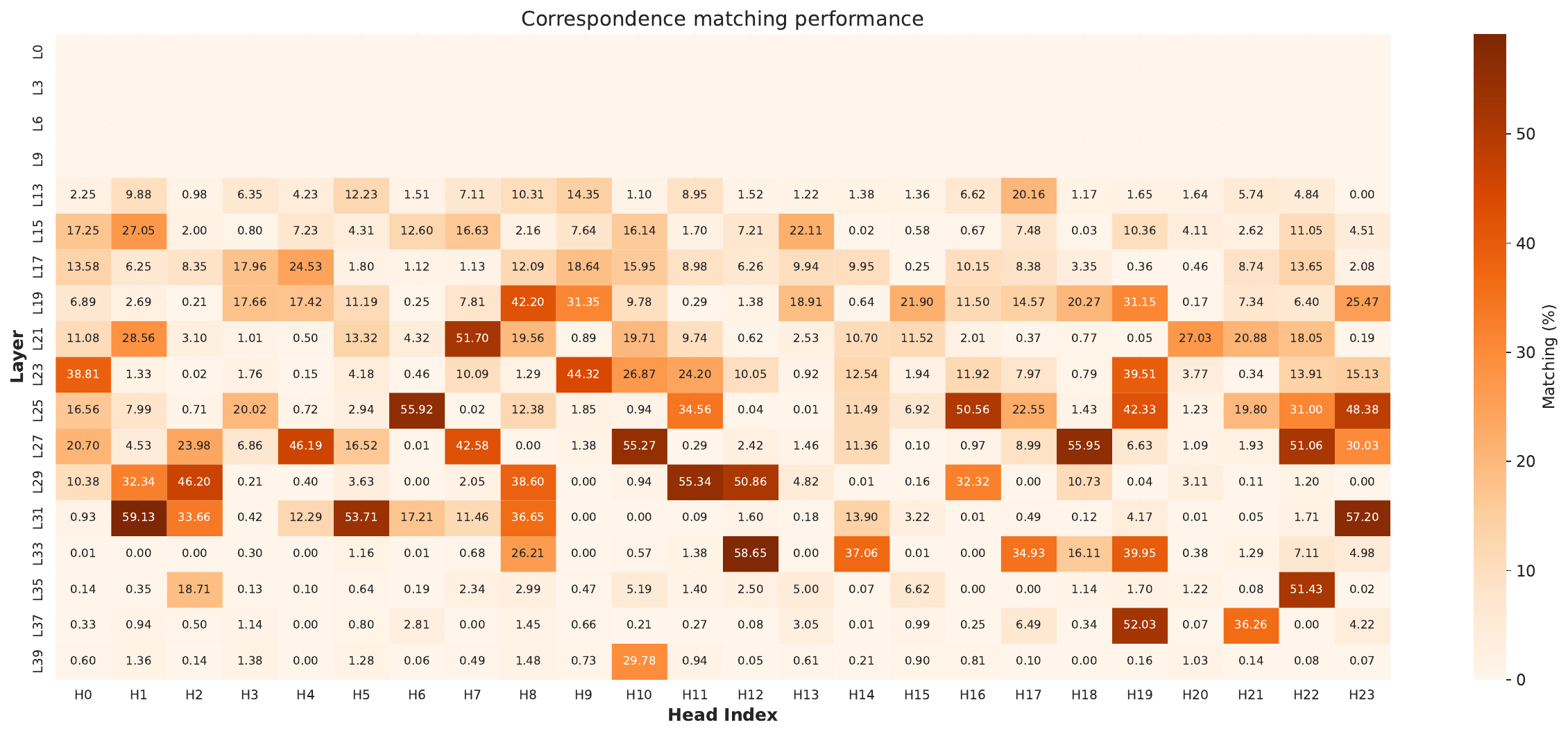} \\
  &
  \rowlab{R (2$\to$1)} &
  \includegraphics[width=\imwidth]{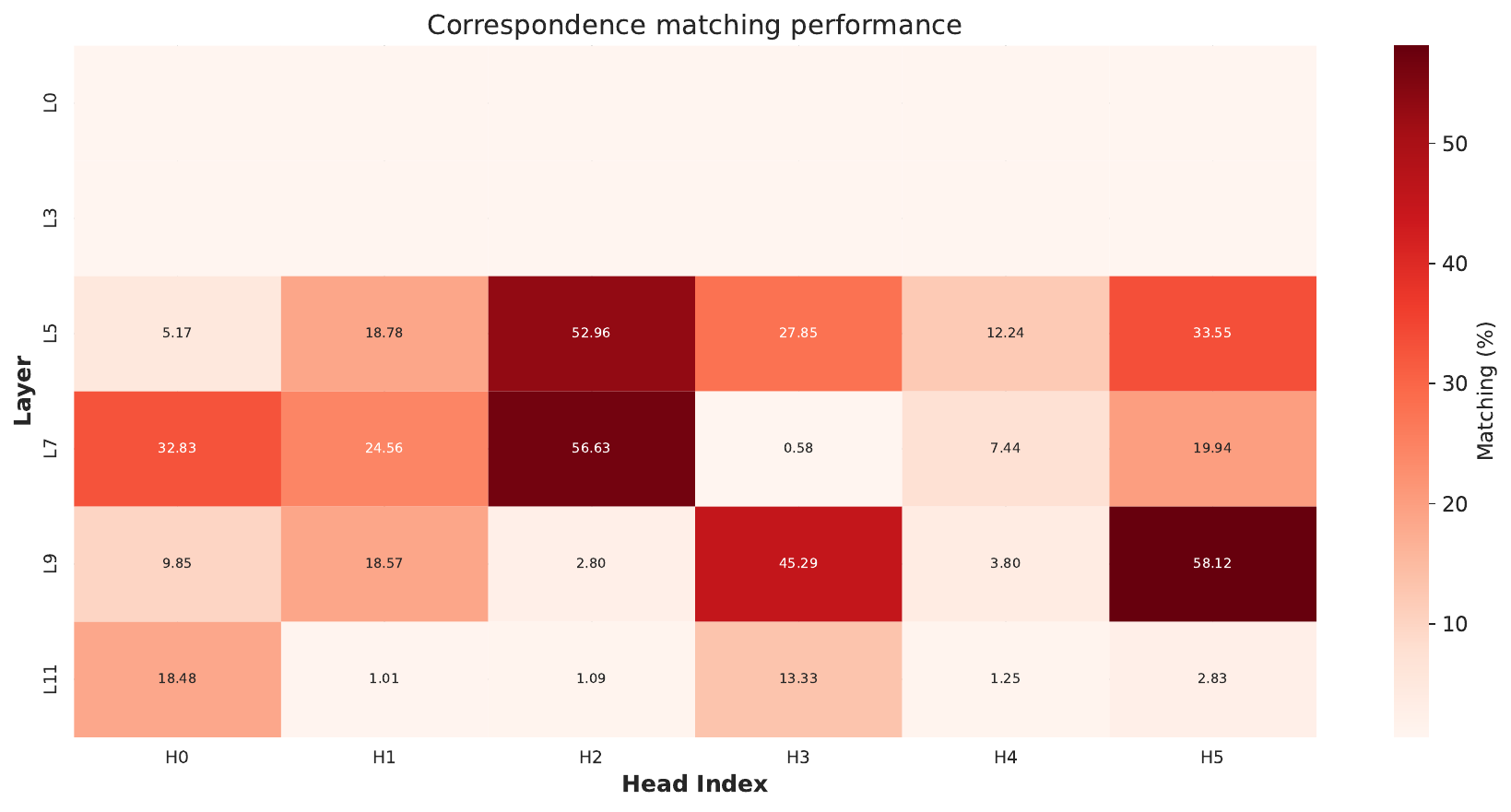} &
  \includegraphics[width=\imwidth]{figures/corr_matching/da3_large/reverse_matching_heatmap.pdf} &
  \includegraphics[width=\imwidth]{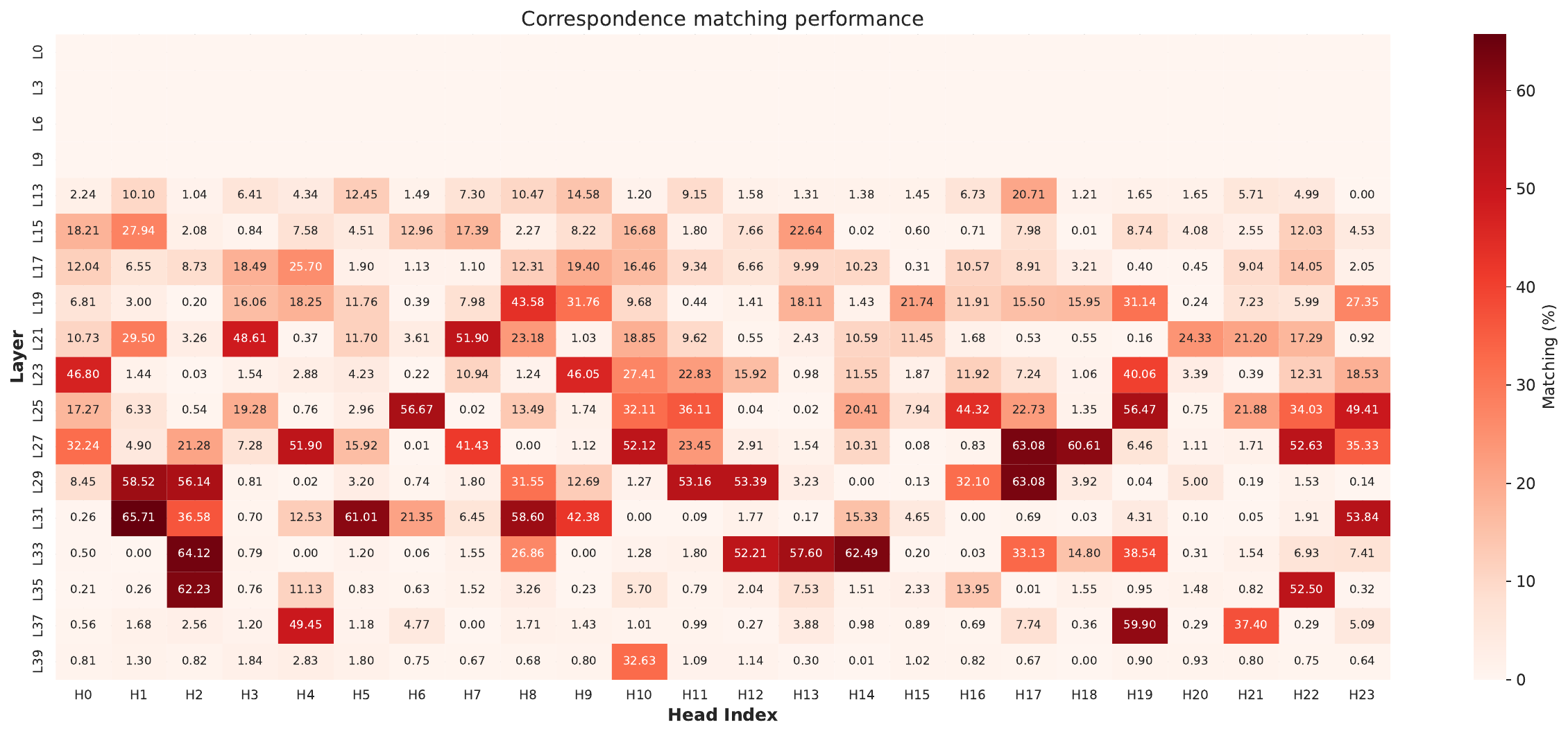} \\

  \multirow{2}{*}{\datasetlab{ETH3D}} &
  \rowlab{F (1$\to$2)} &
  \includegraphics[width=\imwidth]{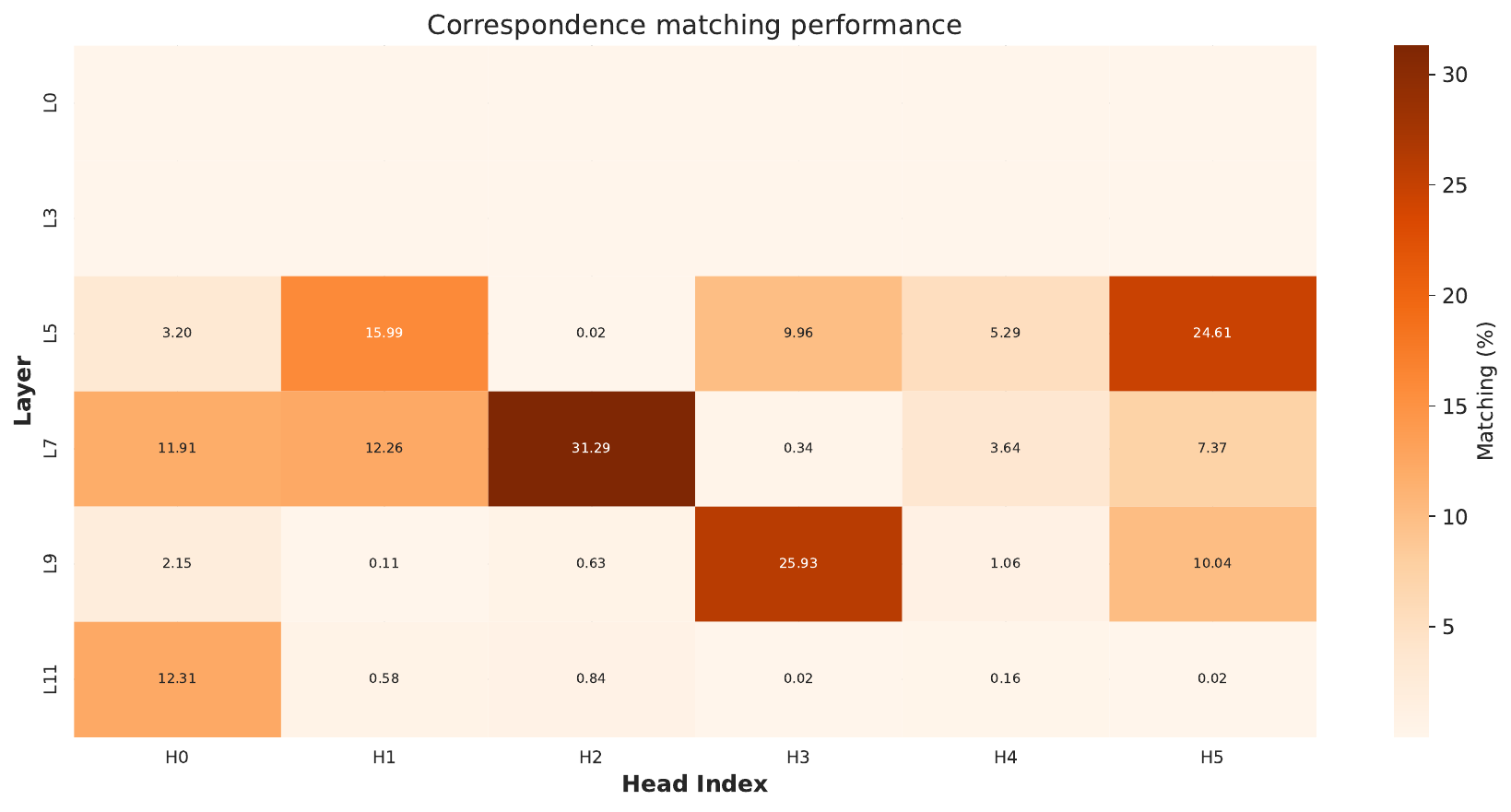} &
  \includegraphics[width=\imwidth]{figures/corr_matching/da3_large/eth_forward_matching_heatmap.pdf} &
  \includegraphics[width=\imwidth]{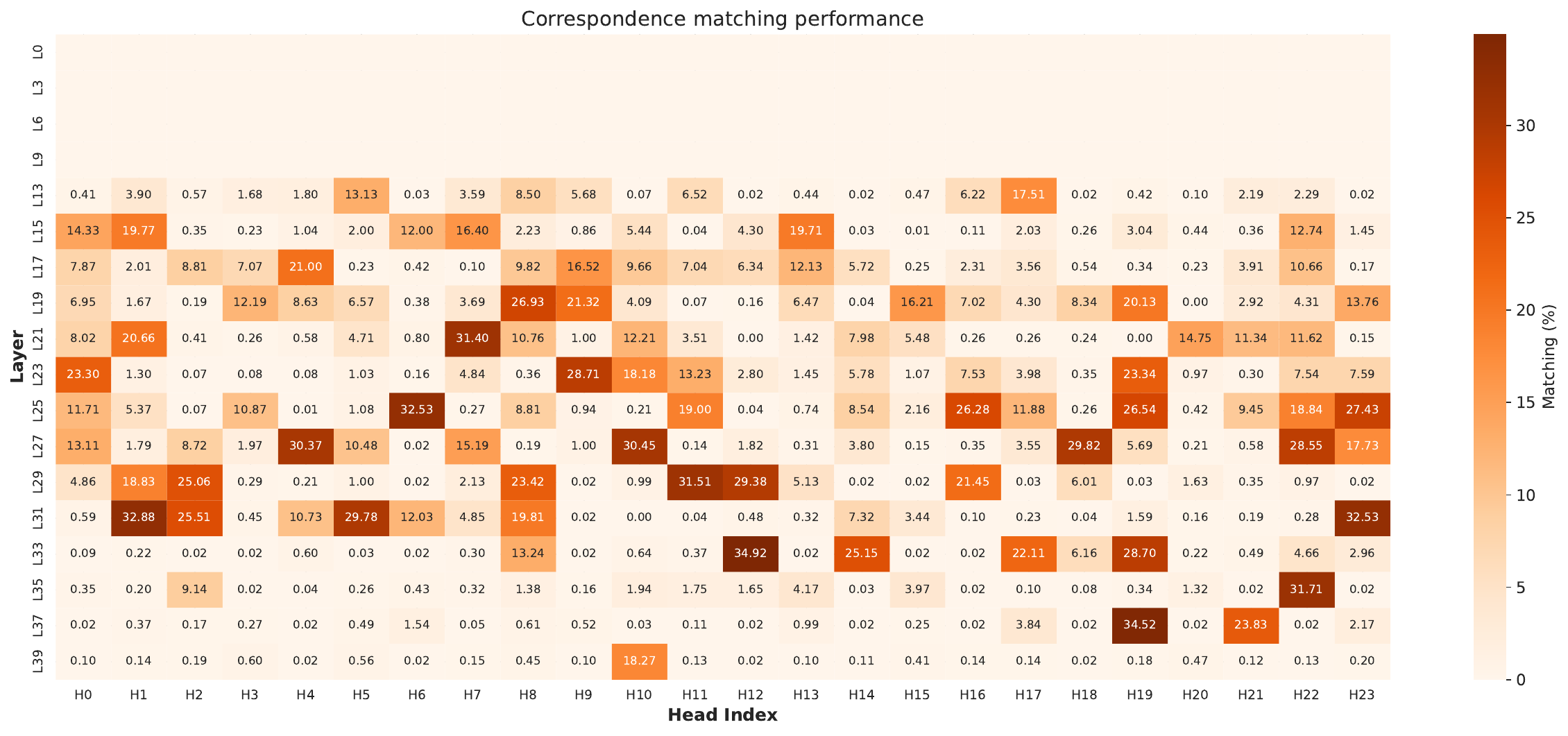} \\
  &
  \rowlab{R (2$\to$1)} &
  \includegraphics[width=\imwidth]{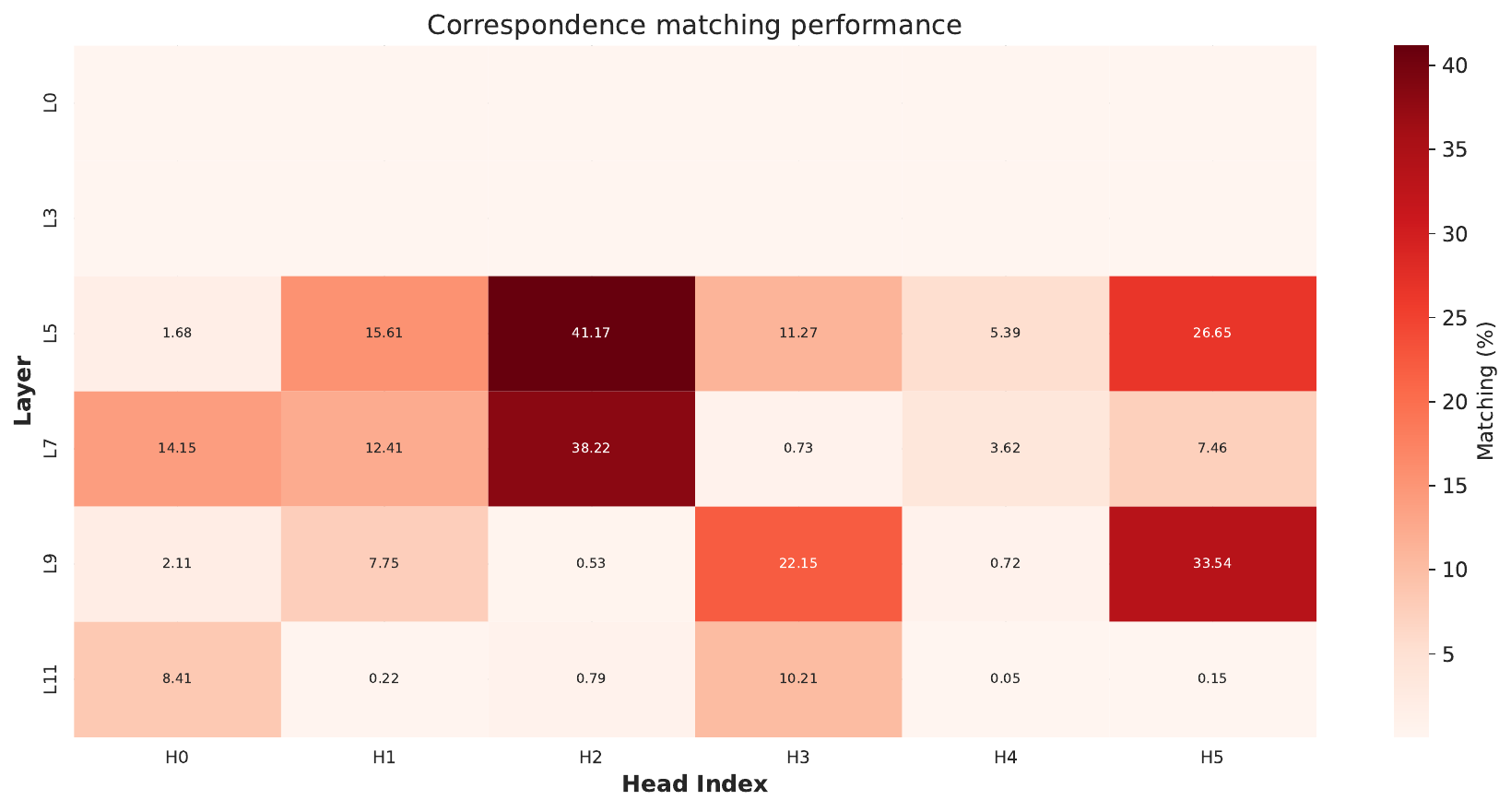} &
  \includegraphics[width=\imwidth]{figures/corr_matching/da3_large/eth_reverse_matching_heatmap.pdf} &
  \includegraphics[width=\imwidth]{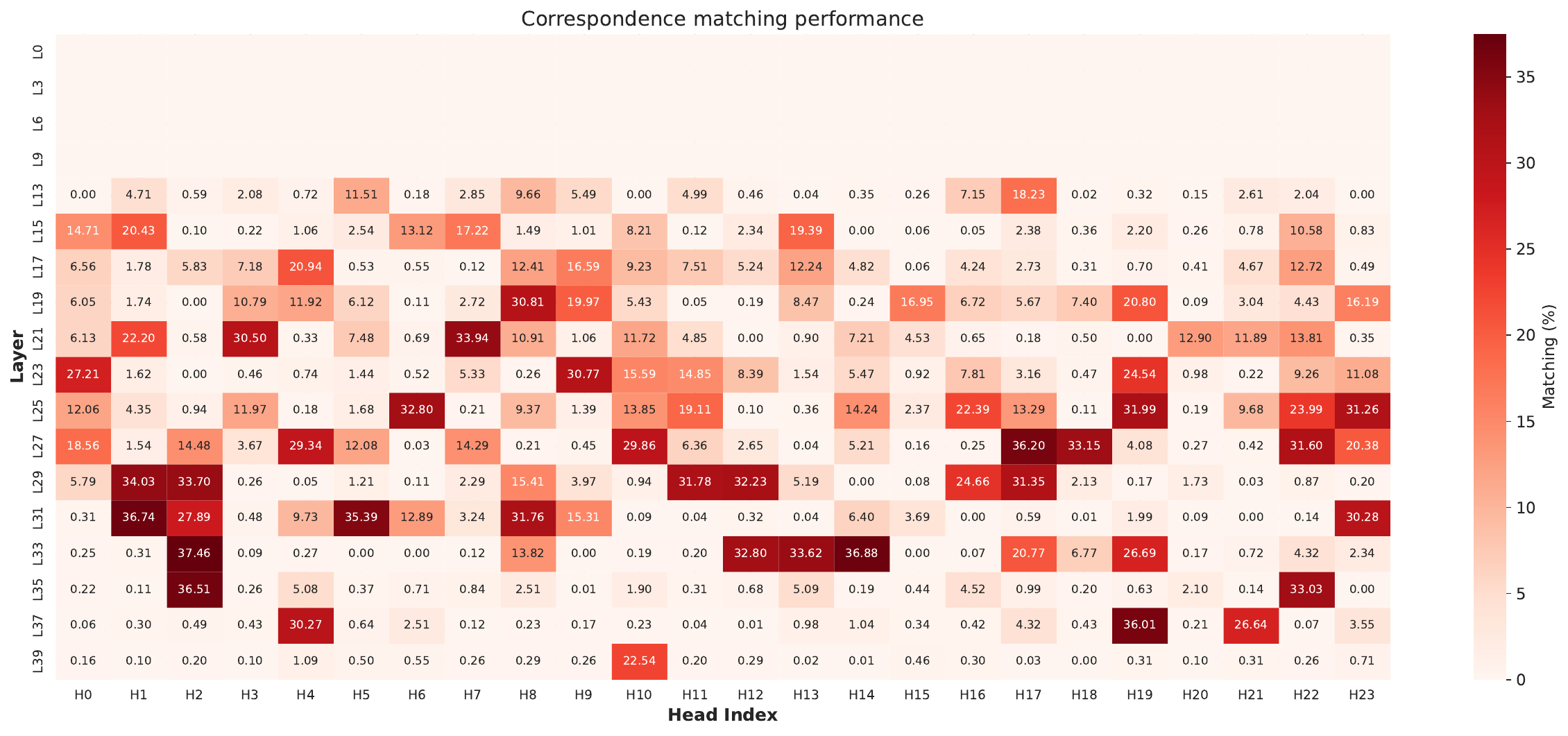} \\

  \multirow{2}{*}{\datasetlab{DTU}} &
  \rowlab{F (1$\to$2)} &
  \includegraphics[width=\imwidth]{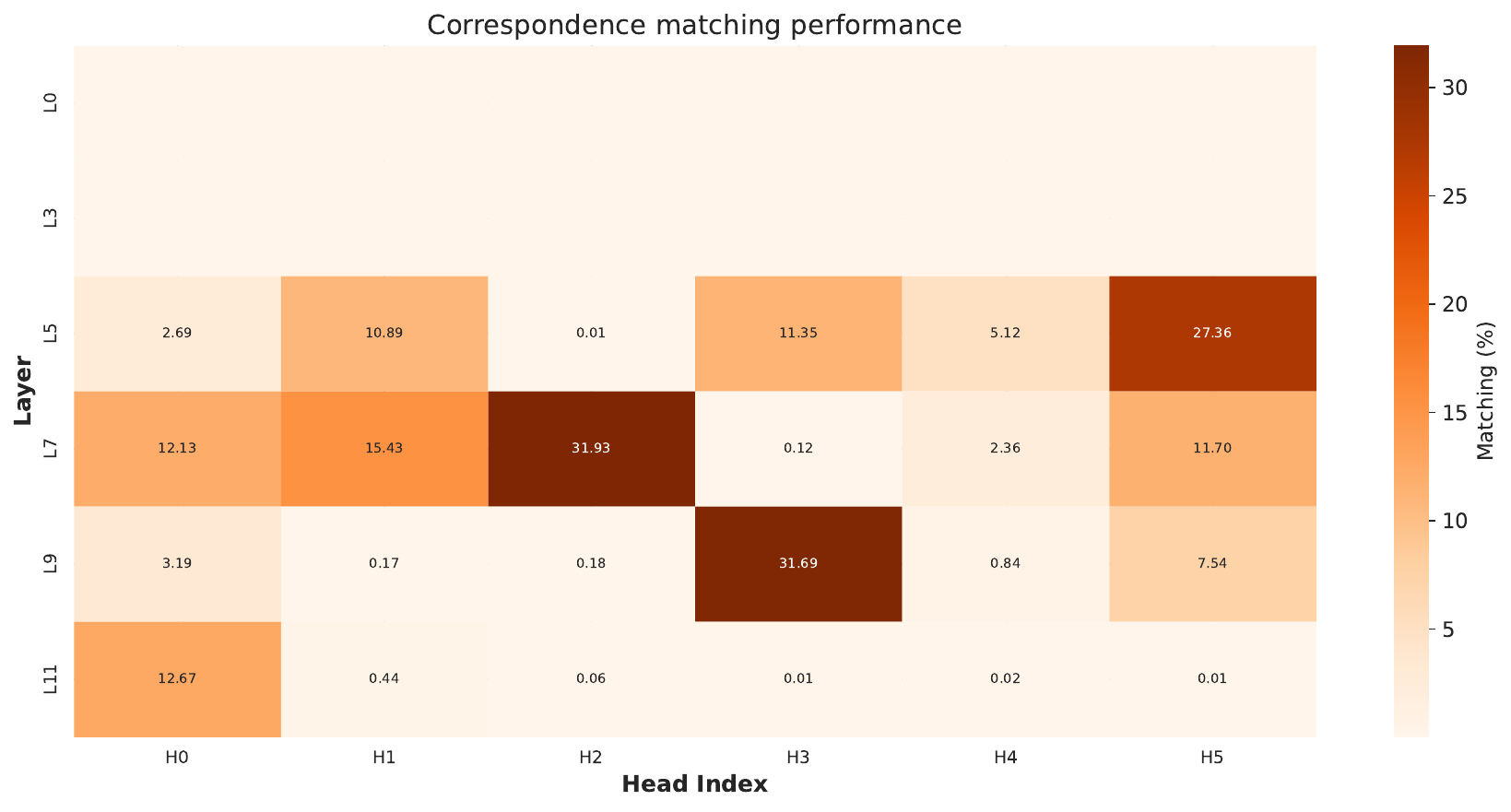} &
  \includegraphics[width=\imwidth]{figures/corr_matching/da3_large/dtu_forward_matching_heatmap.pdf} &
  \includegraphics[width=\imwidth]{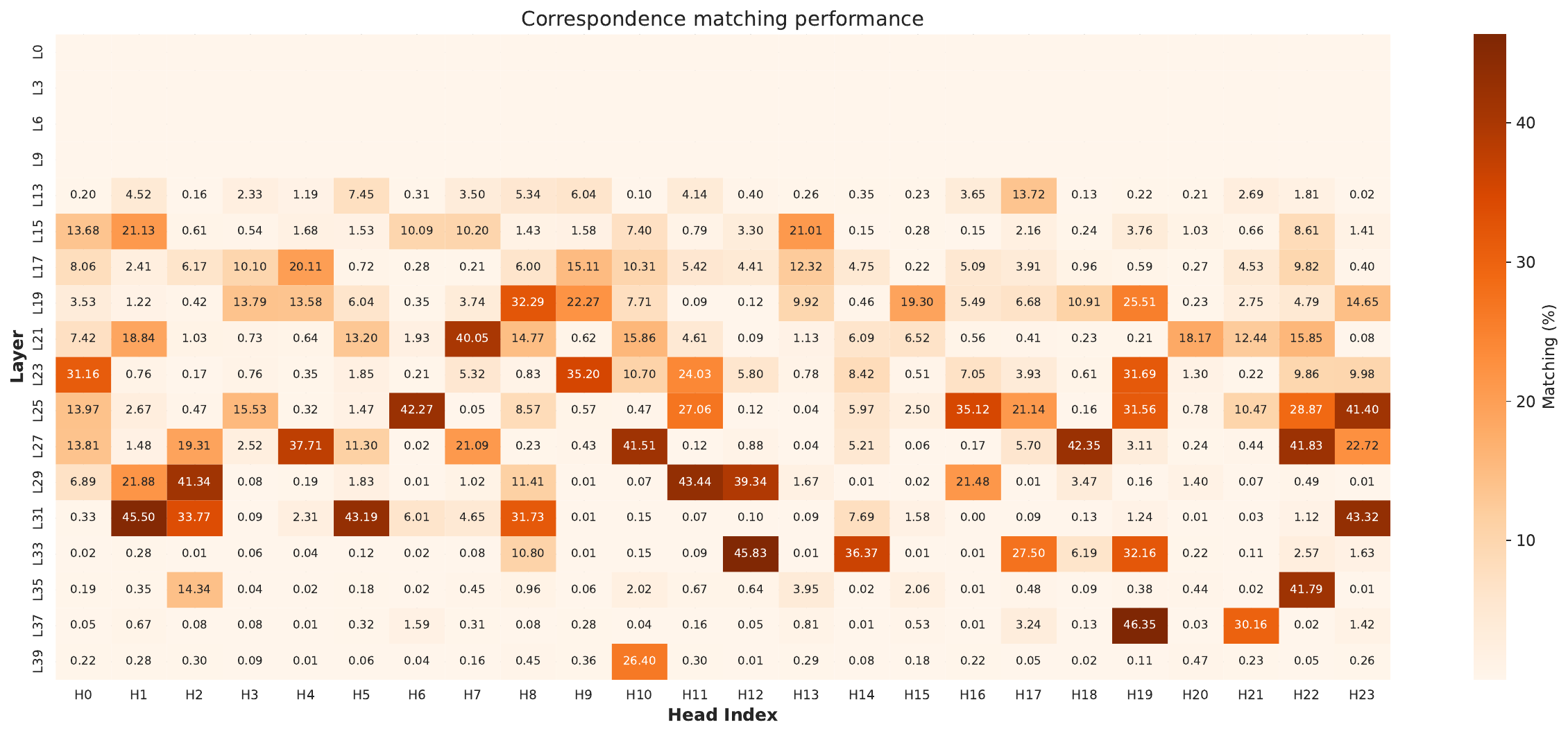} \\
  &
  \rowlab{R (2$\to$1)} &
  \includegraphics[width=\imwidth]{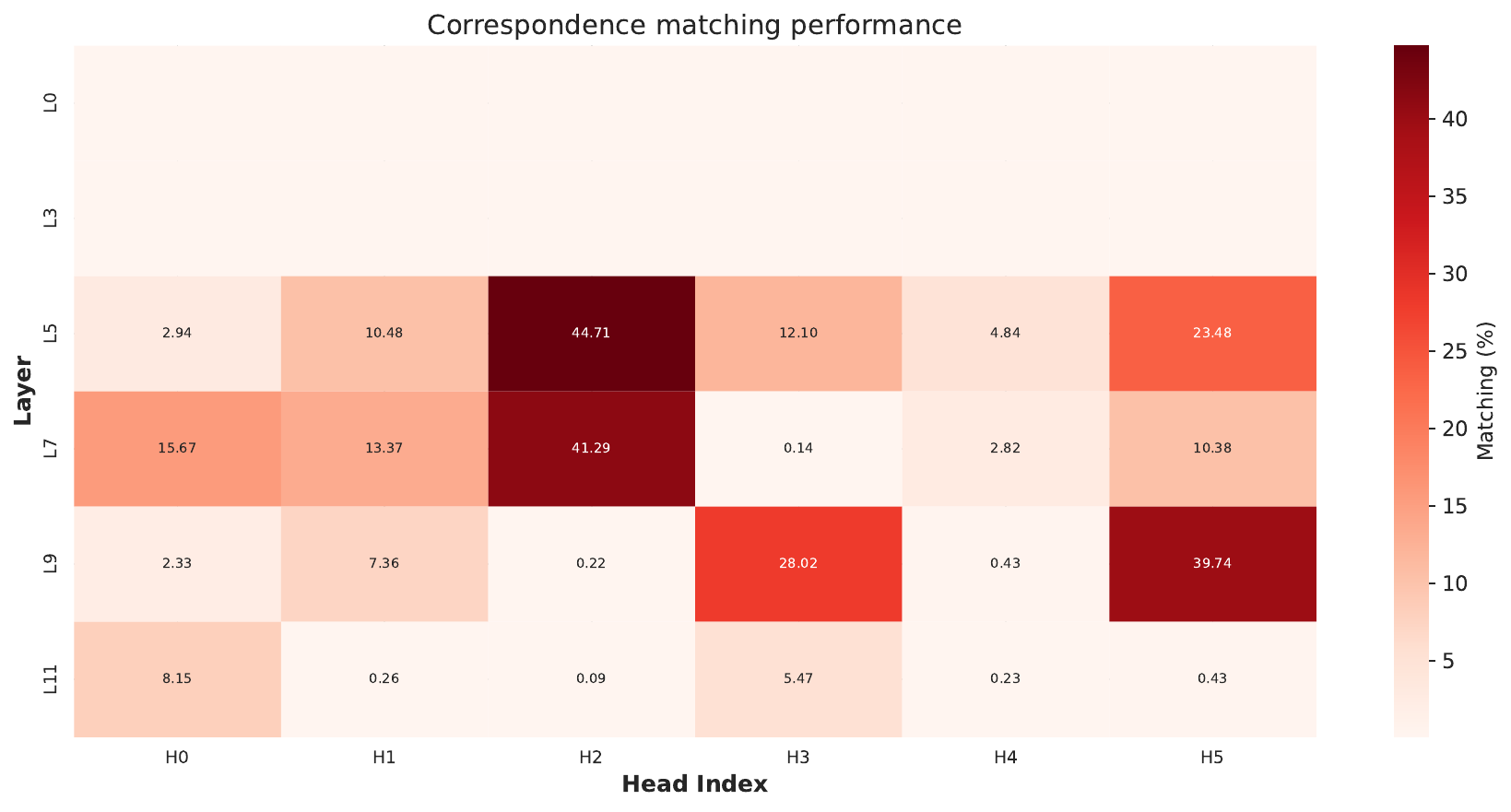} &
  \includegraphics[width=\imwidth]{figures/corr_matching/da3_large/dtu_reverse_matching_heatmap.pdf} &
  \includegraphics[width=\imwidth]{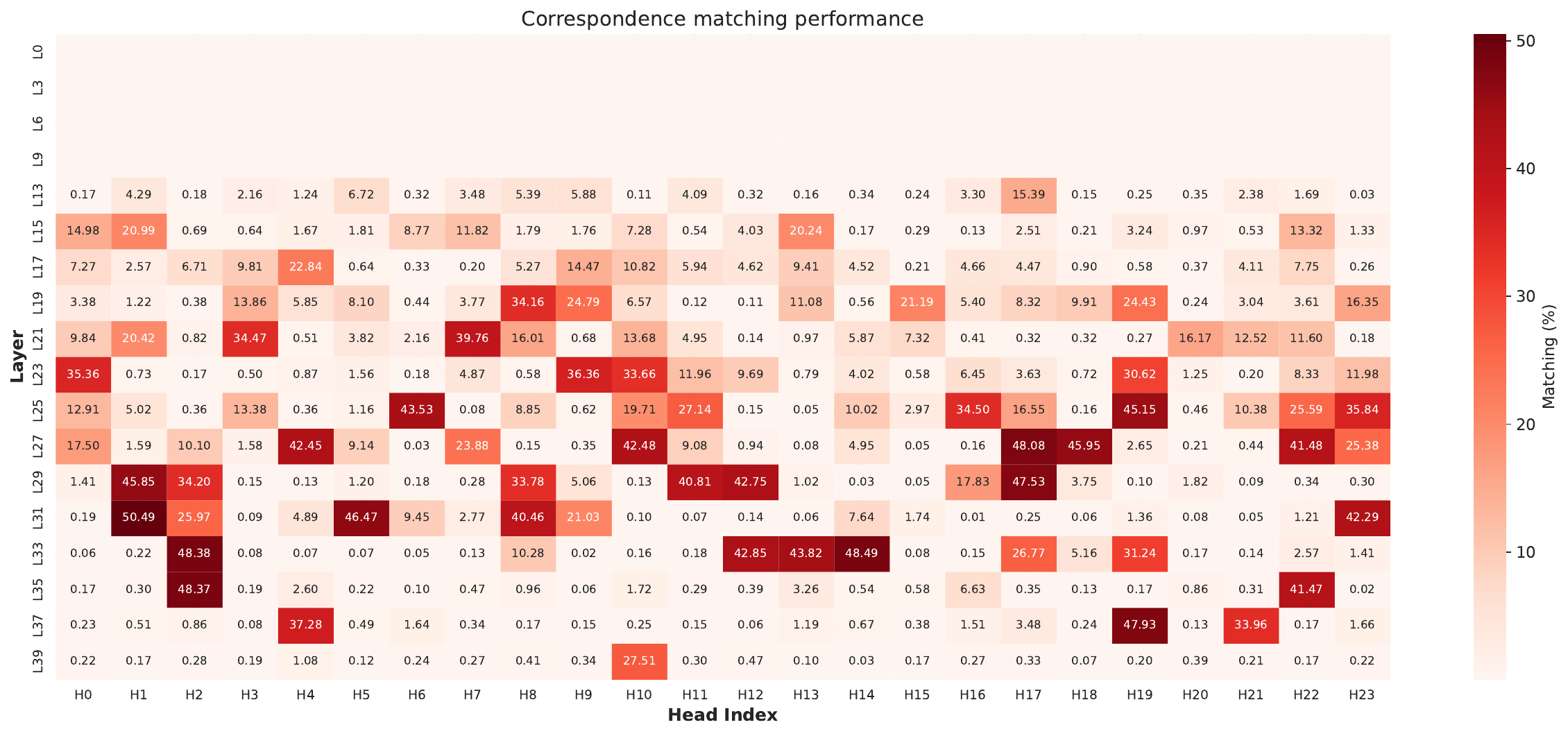} \\
  
\end{tabular}

\caption{\textbf{Correspondence matching for different \da's model sizes.} We compare correspondence matching performance using the QK attention space for forward (view 1$\xrightarrow{}$view 2) and reverse (view 2$\xrightarrow{}$view 1) directions for the giant, large, and small versions of the model. We observe that the middle layers exhibit strong correspondence-matching activity across all three sizes, with slightly decreased activity towards the end.}
\label{fig:da3_sizes}
\end{figure}

We show the results on~\cref{fig:sem_matching}, and report interesting patterns. In \vggt, there is a clear transition in the model between the semantic correspondences and geometric correspondences, where early layers have stronger semantic correspondence matching, and only after the geometric matching emerges in the QK attention space, we see decrease in the semantic correspondences, and, as shown in~\cref{fig:corr_matching_features}, increase in the geometric matching in the feature space. We believe this could stem from the model's large capacity and its use of a DINOv2 backbone, which exhibits emergent semantic matching in its representations~\cite{zhang2023tale}.  

For \da, we observe the strongest semantic correspondence activity in the middle layers; that activity is lower than for the \vggt. And finally, for \duster, we observe a similar trend as for geometric matching, where early layers have higher activity and it decreases towards the end of the network.

\subsection{Analysis of Depth Anything model sizes}

Since \da~comes in different model sizes, we analyzed the correspondence-matching performance across 3 sizes: small, large, and giant. In~\cref{fig:da3_sizes}, we show the differences in model size and correspondence-matching ability on our ShapeNet dataset (first two rows), ETH3D (middle two rows), and DTU (last two rows) datasets. We observe that the correspondence-matching ability emerges in the middle layers of the network, meaning that for larger model sizes, it occurs later than for smaller ones. We further observe a slight decrease in correspondence-matching ability in the final layers.

\begin{figure}[b!]
\centering
\newcommand{\rowlabelwidth}{0.55cm}
\newcommand{\imwidth}{0.305\textwidth}
\setlength{\tabcolsep}{2pt} 
\renewcommand{\arraystretch}{0} 

\newcommand{\rowlab}[1]{%
  \makebox[\rowlabelwidth][c]{\raisebox{9.5ex}{\rotatebox{90}{\textbf{#1}}}}%
}

\newcommand{\rowlabf}[1]{%
  \makebox[\rowlabelwidth][c]{\raisebox{4.5ex}{\rotatebox{90}{\textbf{#1}}}}%
}

\begin{tabular}{@{}c c c c@{}}
  & \textbf{VGGT} & \textbf{Depth Anything 3} & \textbf{DUSt3R*} \\

  \rowlabf{Whole heads} &
  \includegraphics[width=\imwidth]{figures/int/vggt_sampson_error_increase_median_barplots_poisoned_patches_square_zeroing_out.pdf} &
  \includegraphics[width=\imwidth]{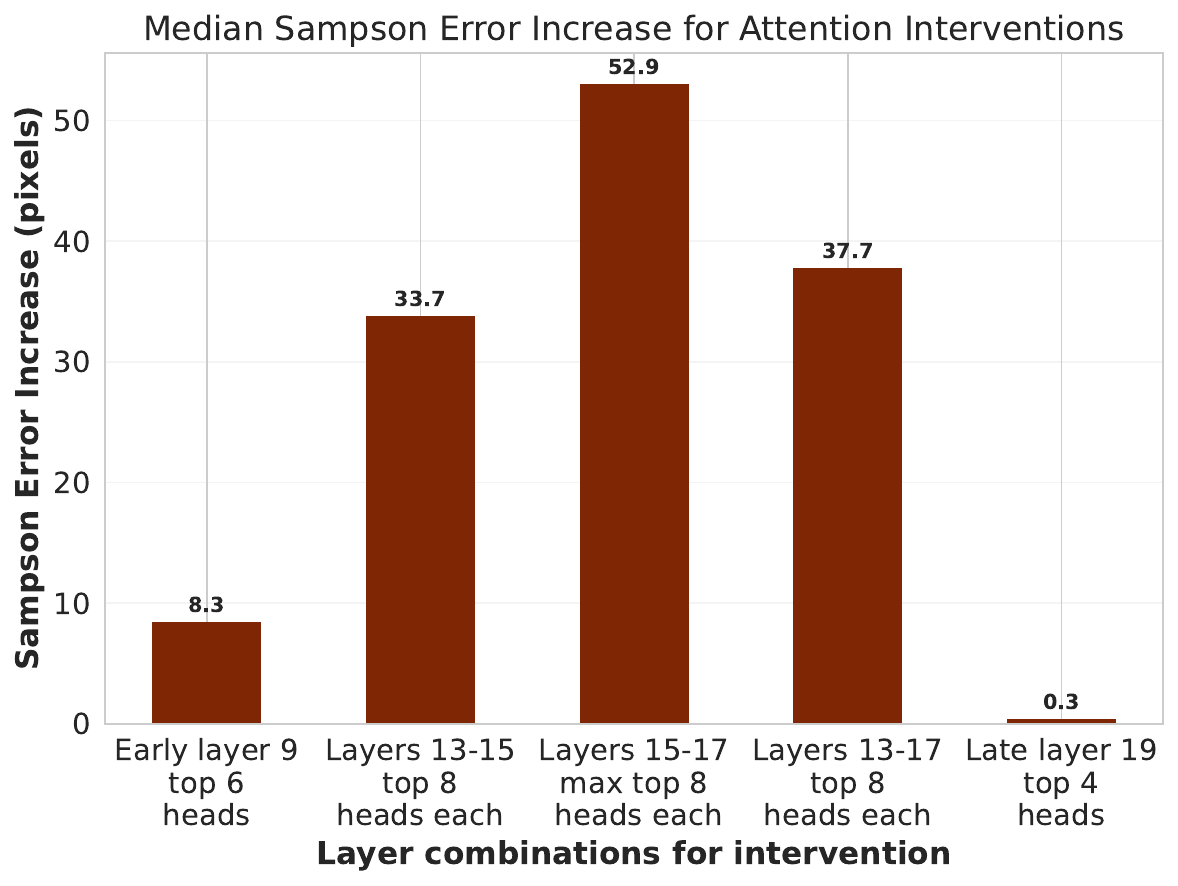} &
  \includegraphics[width=\imwidth]{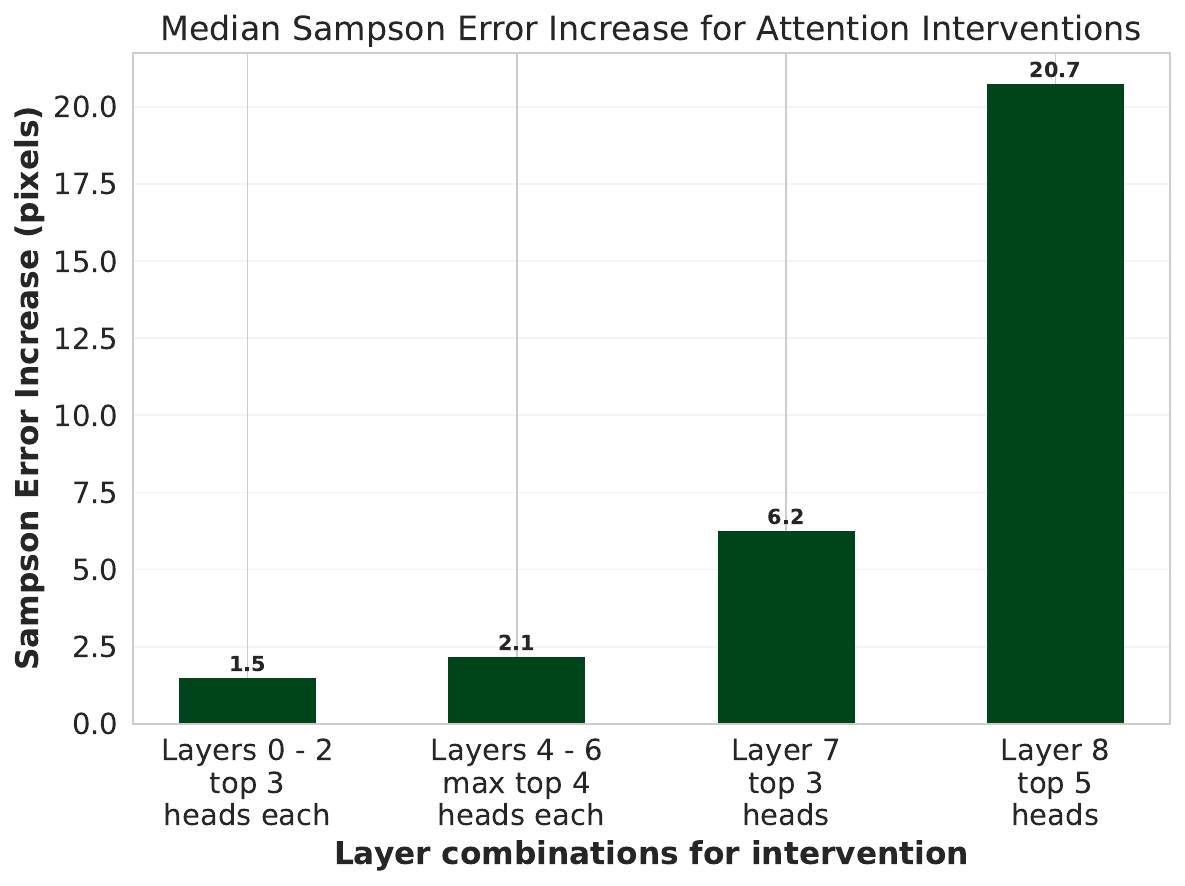} \\

  \rowlabf{Image map} &
  \includegraphics[width=\imwidth]{figures/int/vggt_sampson_error_increase_median_barplots_poisoned_patches_square_swap_zeros.pdf} &
  \includegraphics[width=\imwidth]{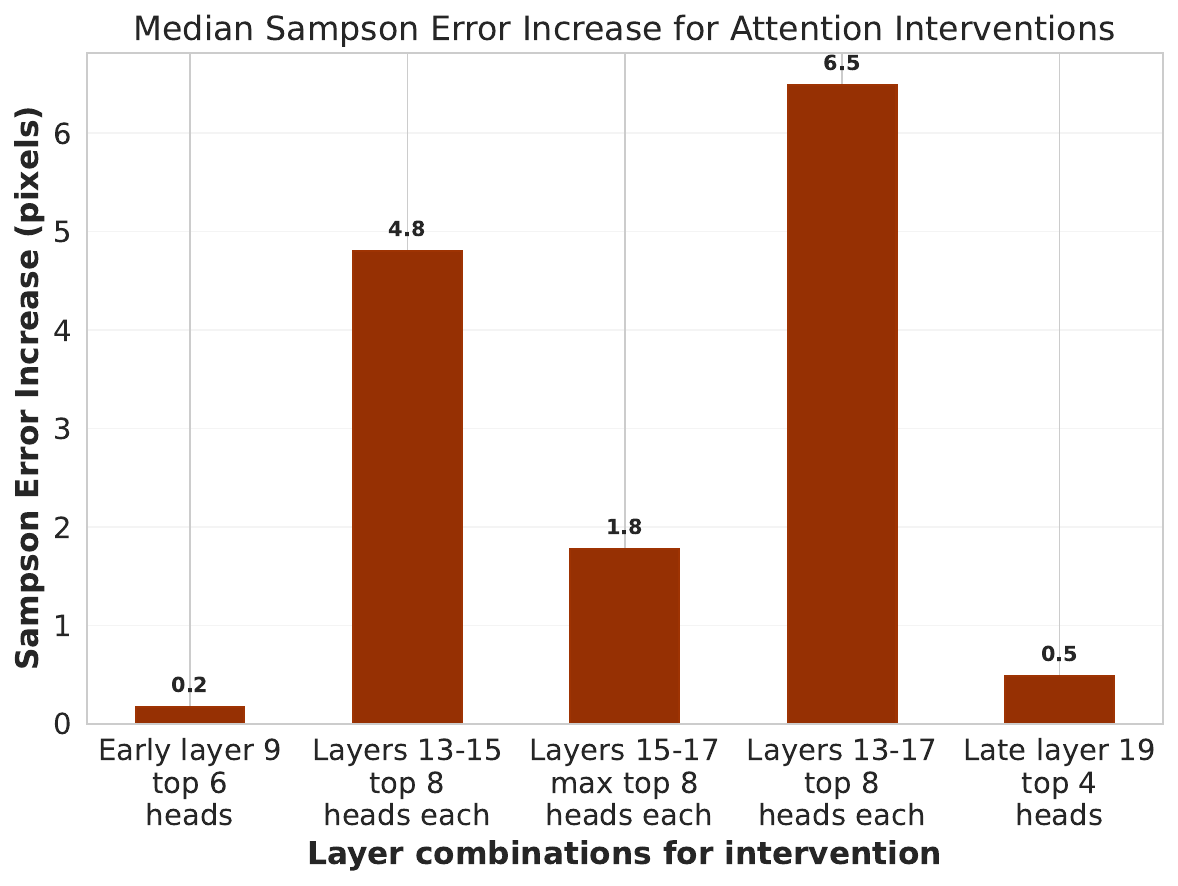} & 
  \includegraphics[width=\imwidth]{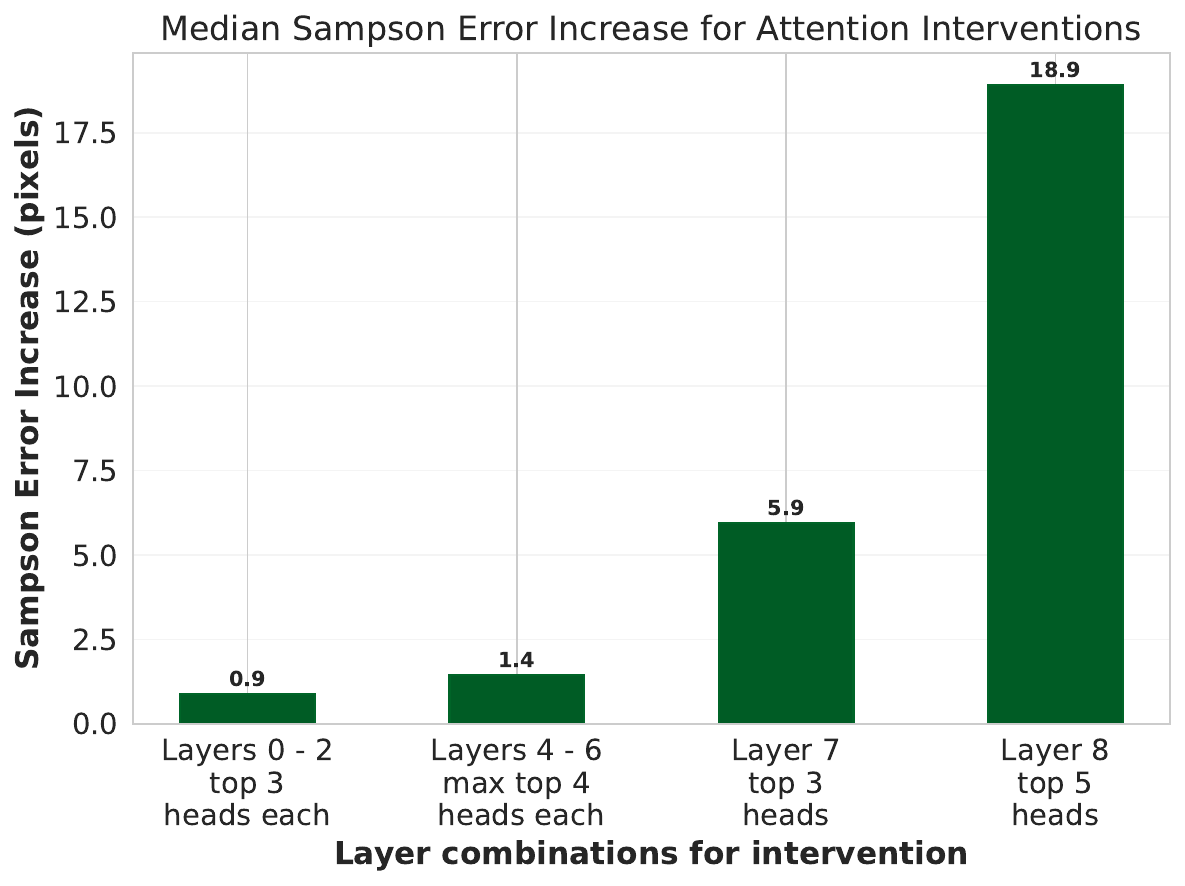}\\

  \rowlab{Block} &
  \includegraphics[width=\imwidth]{figures/int/vggt_sampson_error_increase_median_barplots_poisoned_patches_square_only_match_block.pdf} &
  \includegraphics[width=\imwidth]{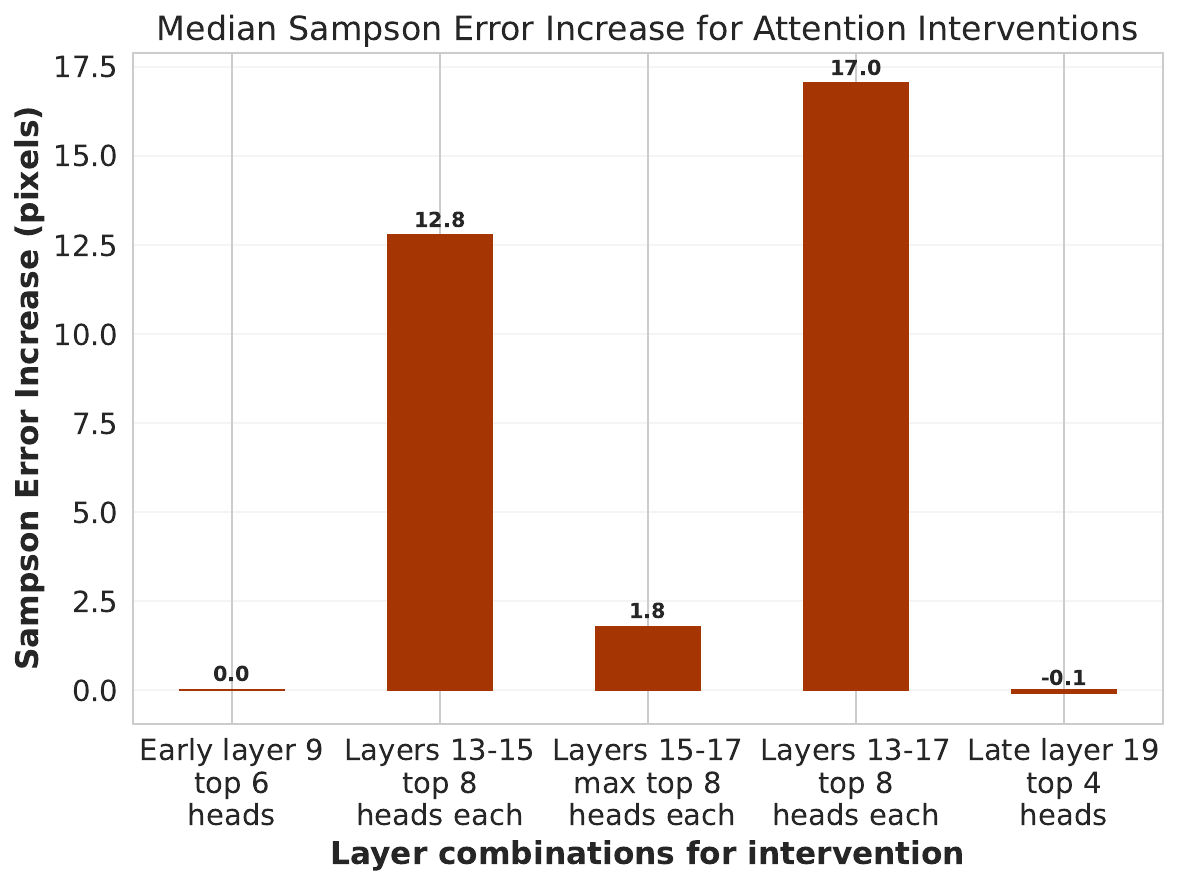} &
  \includegraphics[width=\imwidth]{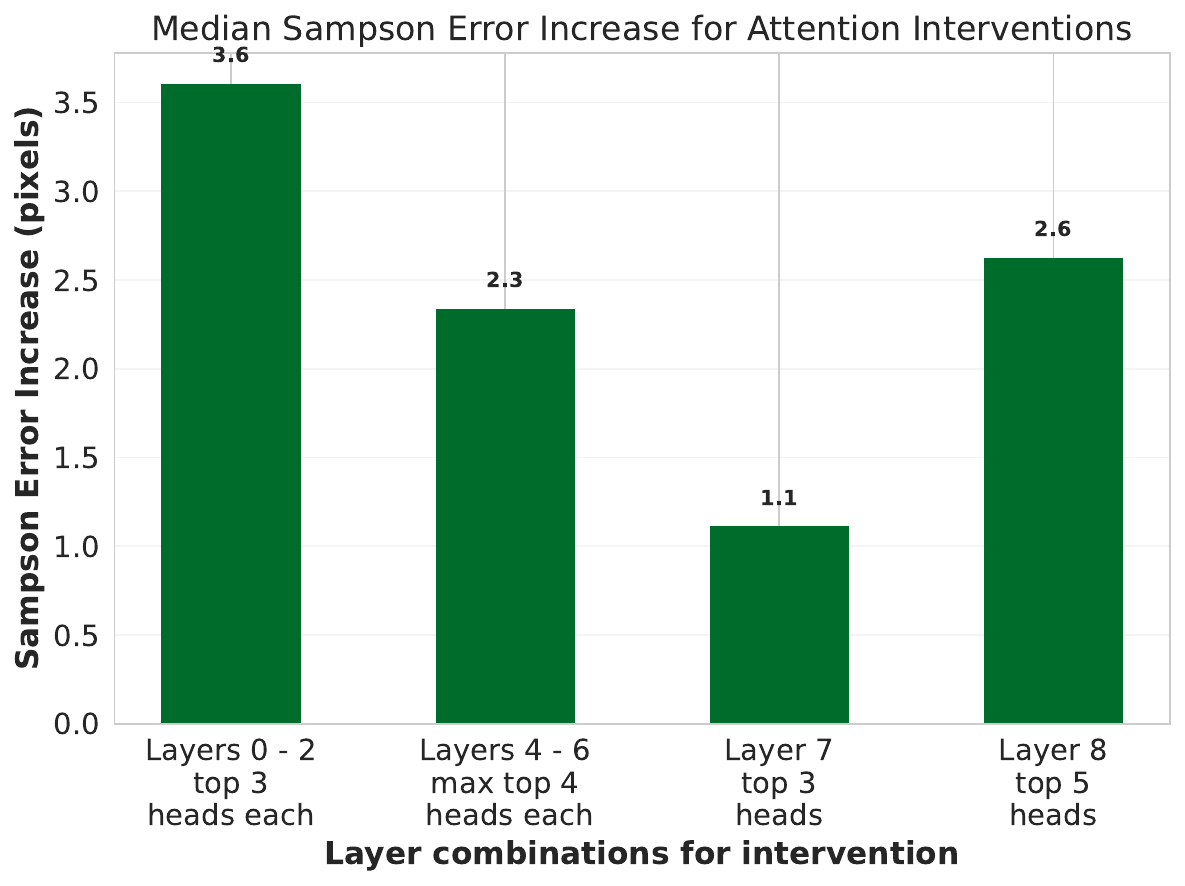} \\

  \rowlab{Patch} &
  \includegraphics[width=\imwidth]{figures/int/vggt_sampson_error_increase_median_barplots_poisoned_patches_square_only_match.pdf} &
  \includegraphics[width=\imwidth]{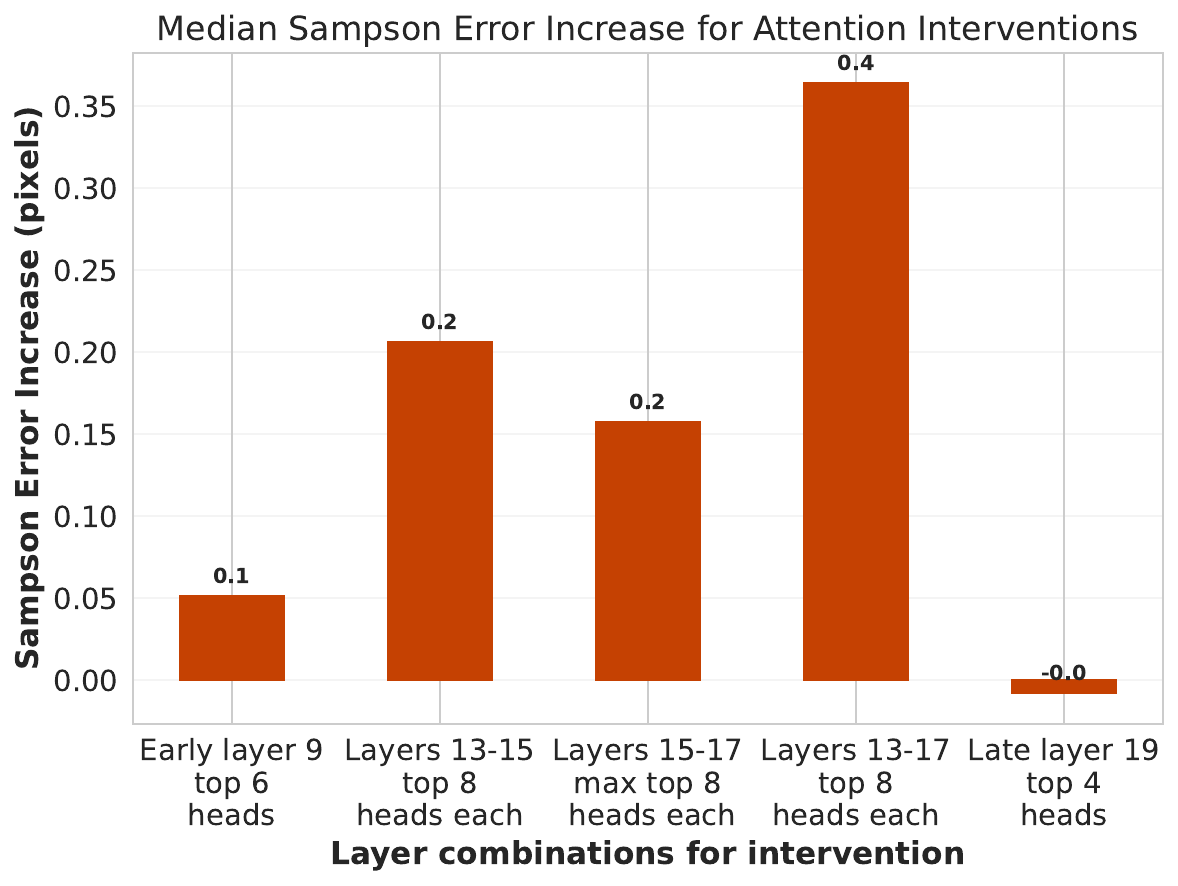} &
  \includegraphics[width=\imwidth]{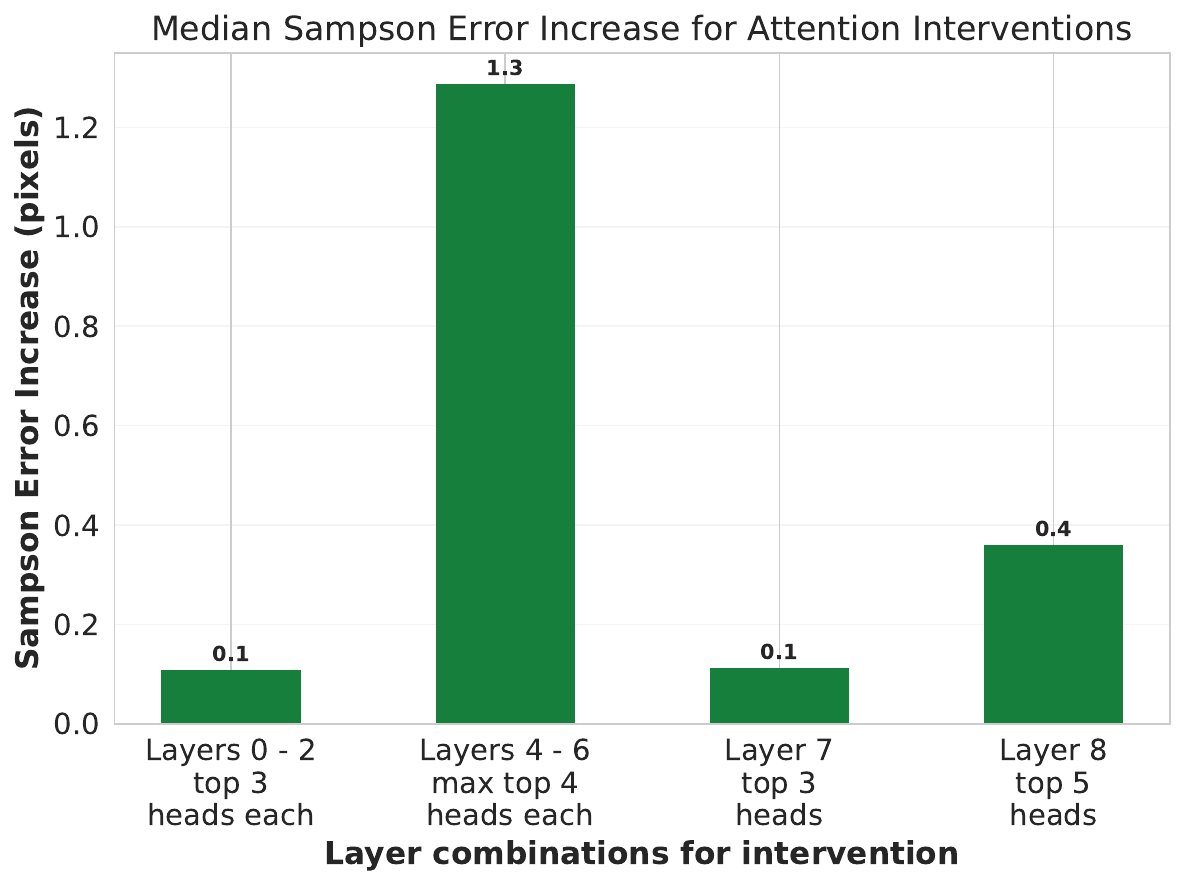} \\
\end{tabular}

\caption{\textbf{Performance degradation under attention interventions on 4 localities.} We show additional combinations of interventions. *For \duster, we report results only for valid cases for Image map and Whole heads.} 
\label{fig:attn_intervention_results_new}
\end{figure}

\section{Causal role of attention patterns}\label{suppl:attn_knockout}
To determine whether the previously identified attention patterns in the intermediate layers are truly the cause of the geometric understanding, we perform targeted interventions in the attention space. We perform the interventions using our ShapeNet data and 4 levels of locality. Here, we provide more information on our experiment design and the differences between the localities.

We compare 4 localities: Whole heads, Image map, Block, and Patch. For each intervention locality level, we apply the interventions to the specified layer-head combinations, which we chose based on the layer's matching performance, selecting the strongest matching heads to intervene on and using random early and later layers as a reference baseline.

The least local intervention type is \textbf{Whole heads}, in which we set the global and cross-attention space to 0 for a given layer and head. This results in the most severe degradation in the model's performance, as together with the geometry-related representation, we might be intervening on some other aspects, so a more local versions of the interventions are needed. 

Next, a more local intervention is applied on the \textbf{Image maps}, in which we set the tokens corresponding to the target image to 0 for a given layer-head combination and correspondence mapping. For example, for a patch x in image 1 (source), we set all patches in image 2 (target) to 0. In VGGT and DA3, this almost effectively enables global attention, as the global attention layers now also act as local ones for the corresponding pairs; in DUSt3R, it enables large portions of the cross-attention layers. With these interventions, we see large degradations in the model's performance, but less severe than in the previous case.

Further, in \textbf{Block} level, we apply the interventions directly following the correspondence mapping between the source and target views. For a given patch in image 1 (source), we set the corresponding patch in image 2 (target) and its 5 neighboring patches to zero (forming a 5x5 block). This allows us to have more targeted interventions and clearly assess the contribution of the layers with strong correspondence-matching ability. We observe again a degradation in the model's performance, which, although less severe than earlier, still shows the connection between the correspondence-matching ability and the final model's performance.

And finally, we perform the most local intervention on the \textbf{Patch} level, where we follow the same approach as for \textbf{Block}, but set to zero only the direct correspondence mappings between the source and target patches. With this intervention, for some layer-head combinations, we might not observe significant degradation, as the model could still leverage similarities and match the immediate neighboring areas. 

We show additional layer-head combinations on~\cref{fig:attn_intervention_results_new}.

\section{VGGT's robustness to scene modifications}\label{suppl:robustness}

Traditional multi-view geometry methods rely on locally unique correspondences and therefore degrade in noisy and difficult scenarios. Although VGGT is claimed to be robust to these challenges, this has not been systematically tested. Here, we evaluate VGGT on challenging cases that require reliable local matching and strong global spatial reasoning. We compare it against the classical SIFT + 8-point + RANSAC~\cite{lowe2004distinctive} pipeline as well as a learned matching method SuperGlue~\cite{sarlin20superglue}, and demonstrate that VGGT outperforms both methods.

\subsection{Lighting and color variations.} 
\begin{figure}[t!]
    \centering
    \begin{subfigure}[b]{0.15\textwidth}
        \includegraphics[width=\textwidth]{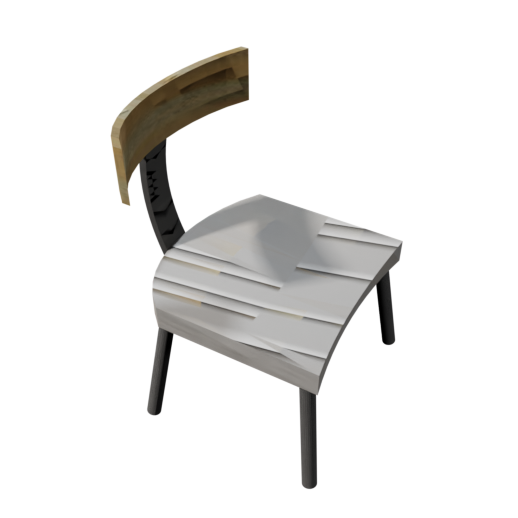}
    \end{subfigure}
    \begin{subfigure}[b]{0.15\textwidth}
        \includegraphics[width=\textwidth]{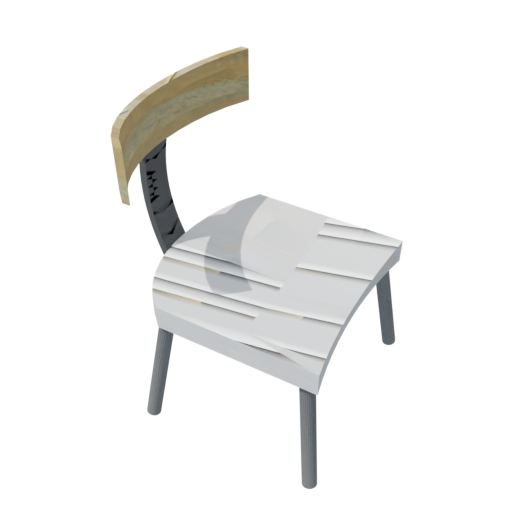}
    \end{subfigure}
    \begin{subfigure}[b]{0.15\textwidth}
        \includegraphics[width=\textwidth]{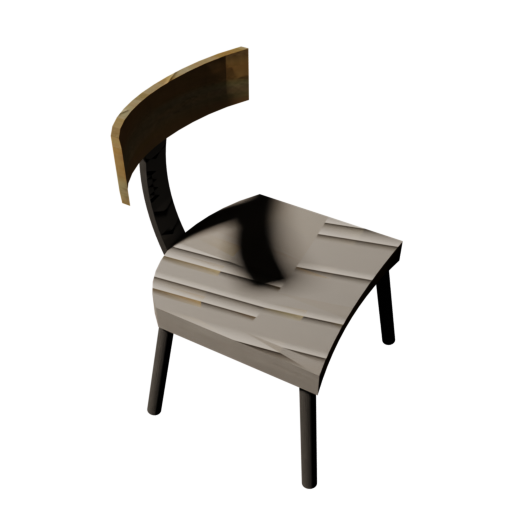}
    \end{subfigure}
    \begin{subfigure}[b]{0.15\textwidth}
        \includegraphics[width=\textwidth]{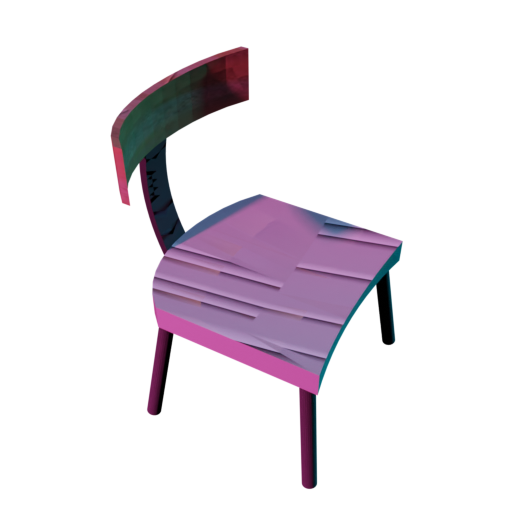}
    \end{subfigure}
    \begin{subfigure}[b]{0.15\textwidth}
        \includegraphics[width=\textwidth]{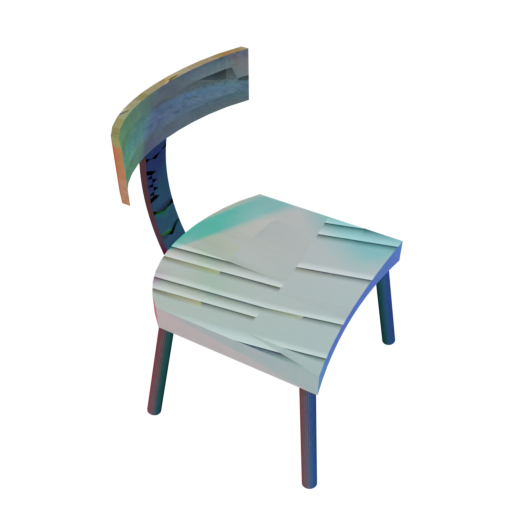}
    \end{subfigure}
    \begin{subfigure}[b]{0.15\textwidth}
        \includegraphics[width=\textwidth]{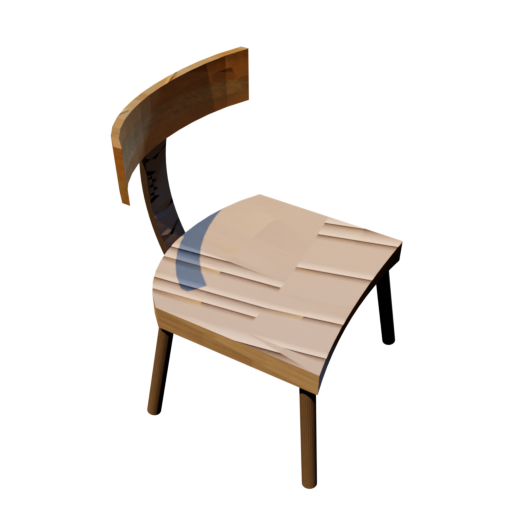}
    \end{subfigure}
    \begin{subfigure}[b]{0.15\textwidth}
        \includegraphics[width=\textwidth]{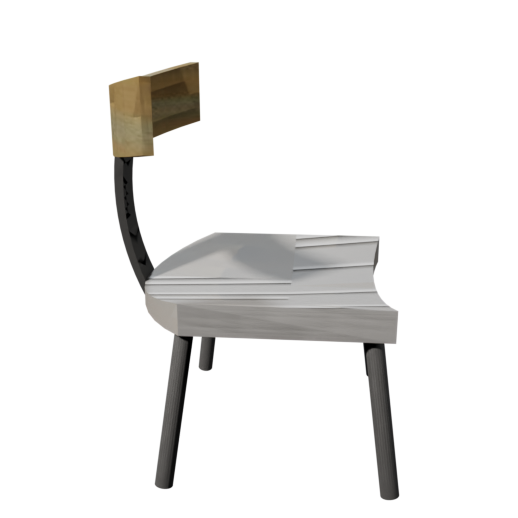}
        \caption{Normal\\ Normal}
    \end{subfigure}
    \begin{subfigure}[b]{0.15\textwidth}
        \includegraphics[width=\textwidth]{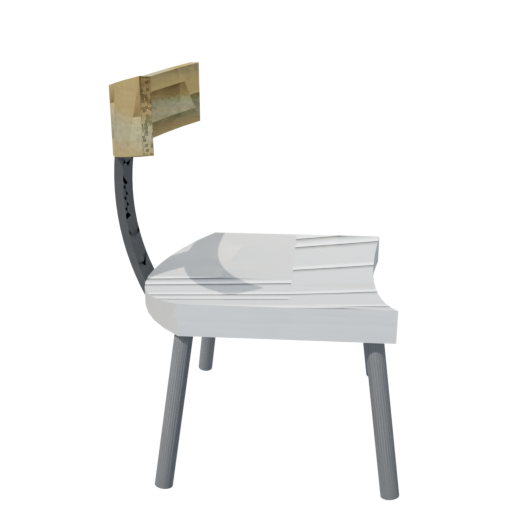}
        \caption{Brighter\\ Brighter}
    \end{subfigure}
    \begin{subfigure}[b]{0.15\textwidth}
        \includegraphics[width=\textwidth]{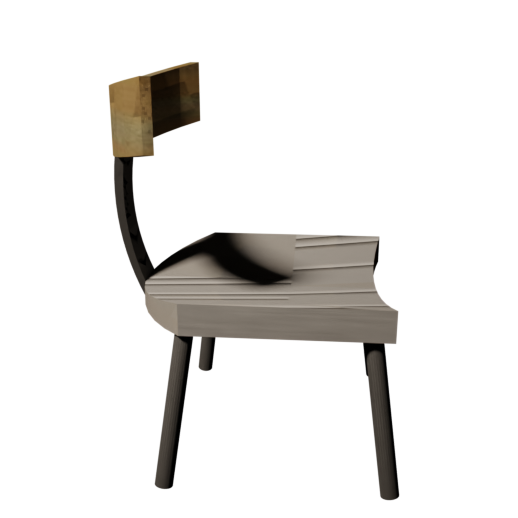}
        \caption{Darker\\ Darker}
    \end{subfigure}
    \begin{subfigure}[b]{0.15\textwidth}
        \includegraphics[width=\textwidth]{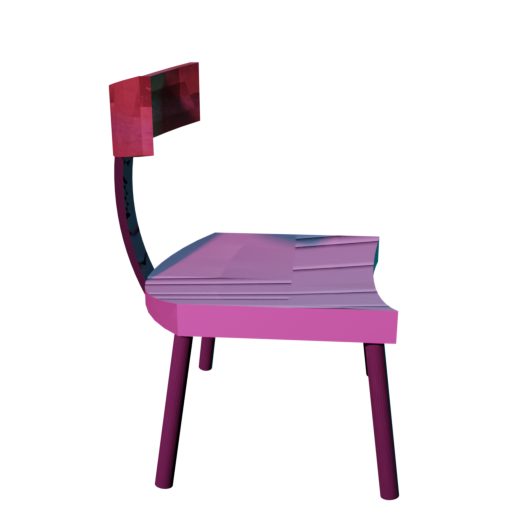}
        \caption{Neon\\ Neon}
    \end{subfigure}
    \begin{subfigure}[b]{0.15\textwidth}
        \includegraphics[width=\textwidth]{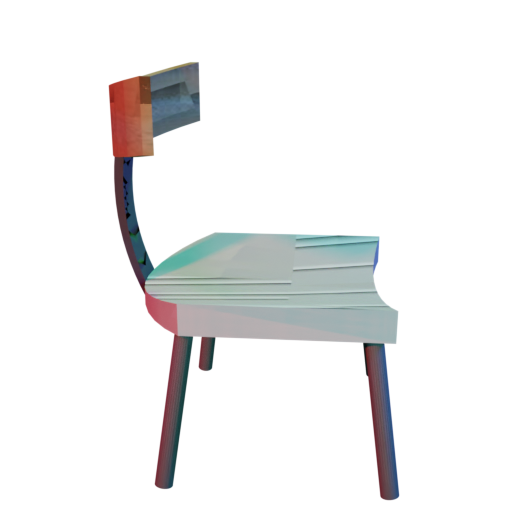}
        \caption{Rainbow\\ Rainbow}
    \end{subfigure}
    \begin{subfigure}[b]{0.15\textwidth}
        \includegraphics[width=\textwidth]{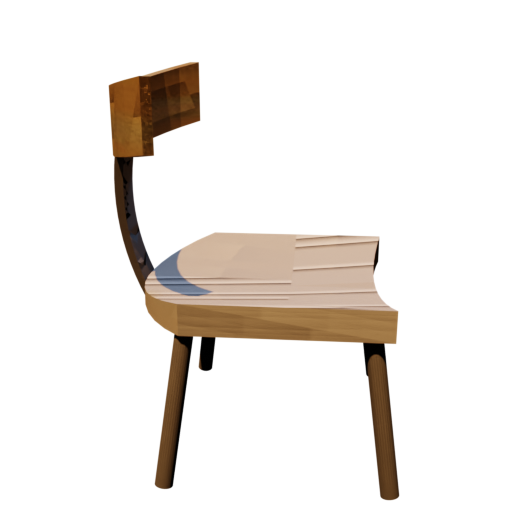}
        \caption{Warm\\ Warm}
    \end{subfigure}
    \caption{Examples of the light and color modifications used in~\cref{fig:ood_tests} over one scene.}
    \label{fig:suppl_ood}
\end{figure}

We design controlled experiments to test robustness under diverse lighting conditions by replacing the standard three-point illumination with darker, brighter, and top-lit configurations, as well as colored lighting that alters object appearance. We evaluate both asymmetric (one image modified) and symmetric (both images modified) conditions by comparing median Sampson errors in pixels on success cases and report failure rates for VGGT and the SuperGlue + RANSAC, where failure is an invalid model output or a Sampson error greater than 10 pixels. We show examples of used modifications on~\cref{fig:suppl_ood}.

\begin{figure*}[t]
    \centering
    \begin{minipage}[b]{0.65\textwidth}
        \centering
        \includegraphics[width=\textwidth]{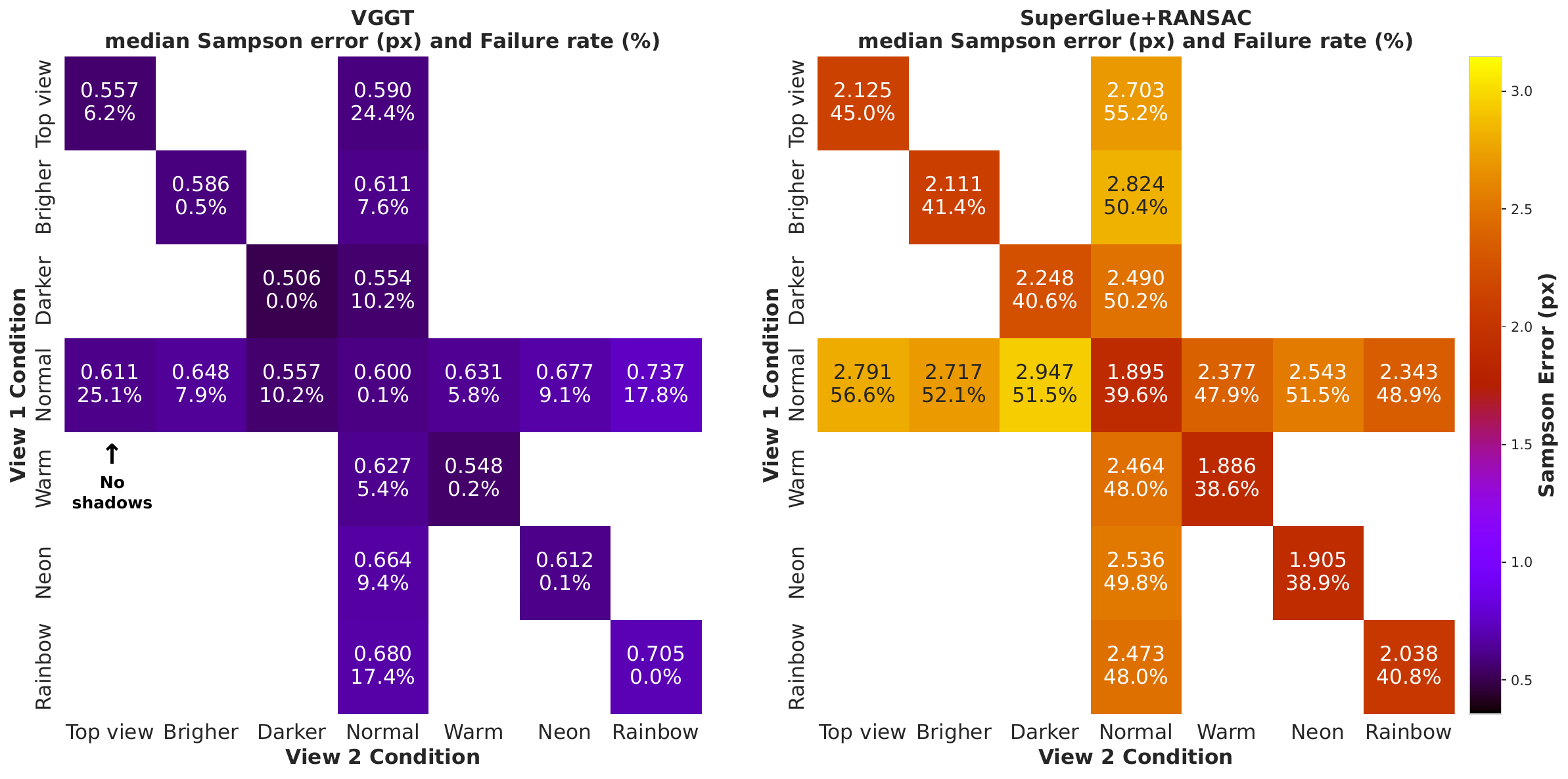} 
        \caption{\textbf{Light and color robustness.} We compare VGGT against SuperGlue + RANSAC on large viewpoint changes. VGGT shows only slight degradation under the changes, while SuperGlue shows higher sensitivity and a larger failure rate. However, VGGT's failure rate (Sampson error larger than 10 pixels) increases under extreme lighting changes and shadow variations. }
        \label{fig:ood_tests}
    \end{minipage}
    \hfill
    \begin{minipage}[b]{0.32\textwidth}
        \centering
        \includegraphics[width=\textwidth]{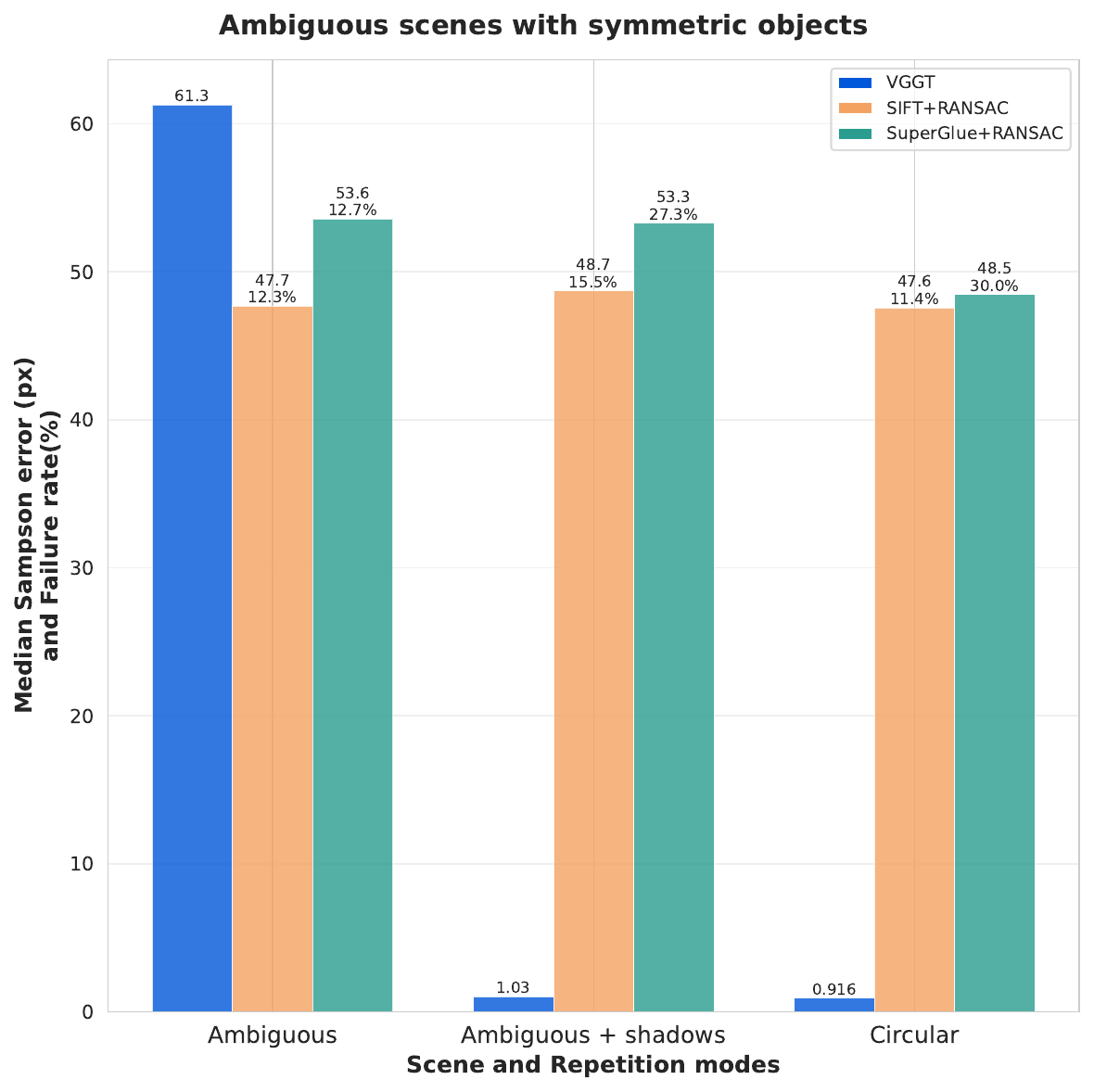}
        \caption{\textbf{Ambiguous scenes with repetitions.} VGGT outperforms baselines, even when only shadows disambiguate symmetric objects, but fails on completely ambiguous scenes without shadows.}
        \label{fig:amb_reps}
    \end{minipage}
\end{figure*}

~\Cref{fig:ood_tests} shows the performance on medium viewpoint differences for VGGT and SuperGlue. VGGT demonstrates superior robustness compared to classical methods, likely due to the strong color augmentation used in training, with its performance practically unchanged and with a low failure rate. SuperGlue and other baselines have a high failure rate, and their performance is affected by these changes. 
However, VGGT exhibits a higher failure rate under large lighting and color changes, particularly for top-view lighting, which removes shadows from the scene, and for neon and rainbow lighting, which introduces color inconsistencies. 
However, when the same interventions are applied to both views, VGGT performs well. It is specifically the case where only one view is intervened, it fails (Sampson error over 10 pixels) more often than for other modifications. 

\begin{figure*}[b!]
    \centering
    \begin{subfigure}[b]{0.16\textwidth}
        \includegraphics[width=\textwidth]{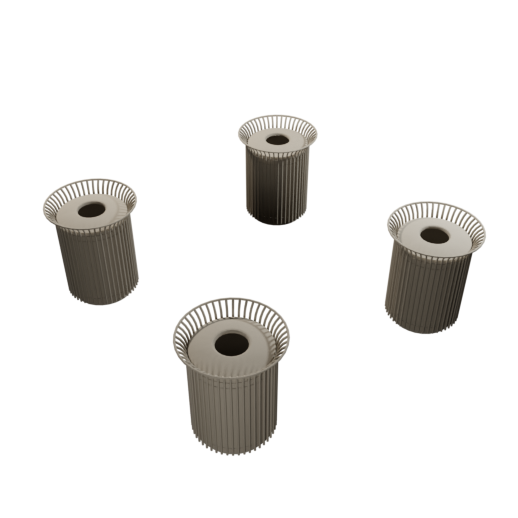}
    \end{subfigure}
    \begin{subfigure}[b]{0.16\textwidth}
        \includegraphics[width=\textwidth]{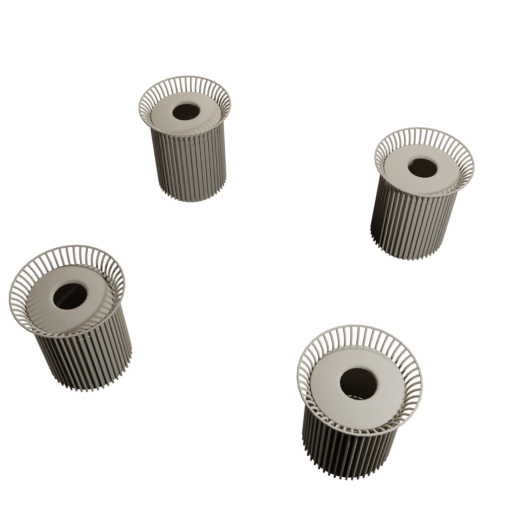}
    \end{subfigure}
    \begin{subfigure}[b]{0.16\textwidth}
        \includegraphics[width=\textwidth]{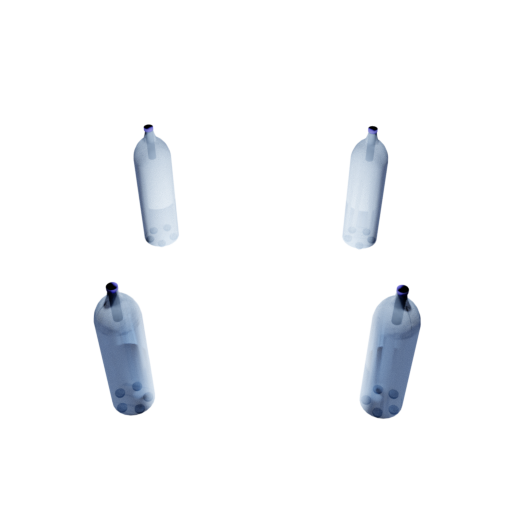}
    \end{subfigure}
    \begin{subfigure}[b]{0.155\textwidth}
        \includegraphics[width=\textwidth]{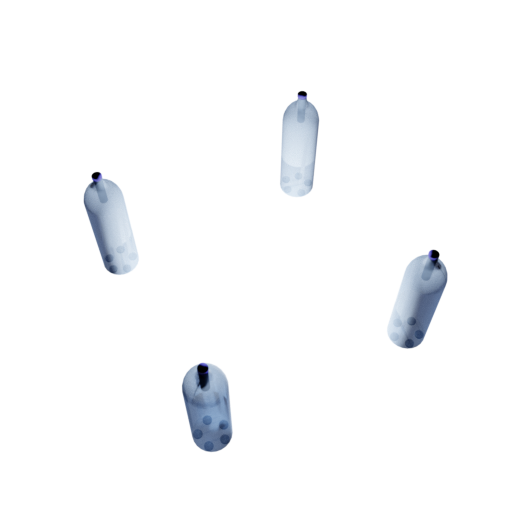}
    \end{subfigure}
    \caption{\textbf{Few simulated examples for repetition analysis.} Repetitive scene (left) corresponds to \textit{Ambiguous + shadow} setup from \cref{fig:amb_reps} with asymmetric lightning that creates shadows in the scene, ambiguous scene (right) corresponds to \textit{Ambiguous} setup from \cref{fig:amb_reps} with complete ambiguity in the scene.}
    \label{fig:ood_examples}
\end{figure*}

\subsection{Repetition and symmetry.}
We generate challenging scenes with multiple instances of the same object arranged in a circular pattern. We distinguish between unique (easier) and fully symmetric objects, such as bottles and vases (harder), where geometric ambiguity arises. We present our results for symmetric objects in \cref{fig:amb_reps}, where even with asymmetric lighting that creates shadows that break symmetry, classical methods fail to establish reliable correspondences, whereas VGGT effectively leverages these subtle cues. For completely ambiguous scenes with symmetric objects and symmetric lighting, all methods fail as expected, confirming the task is geometrically underconstrained.

\subsection{Focal length variations.}

\begin{figure*}[t!]
    \centering
    \begin{subfigure}[b]{.48\linewidth}
        \includegraphics[width=\linewidth]{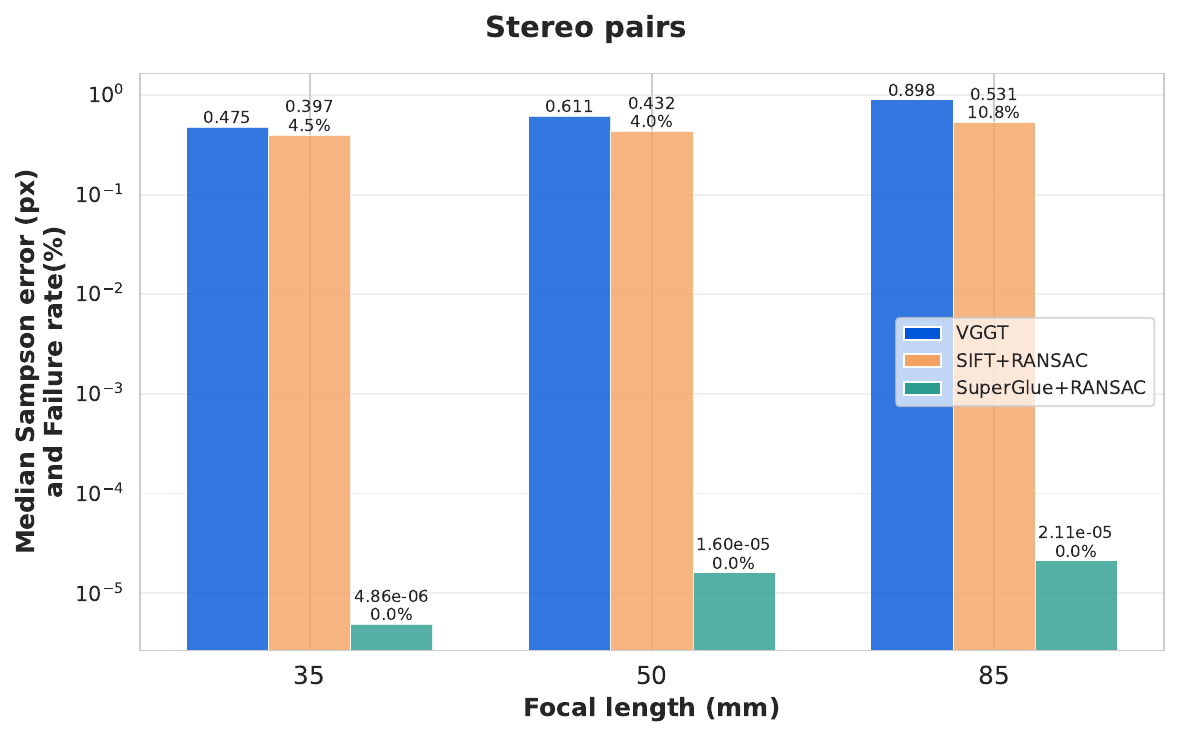}
    \end{subfigure}
    \begin{subfigure}[b]{.48\linewidth}
        \includegraphics[width=\linewidth]{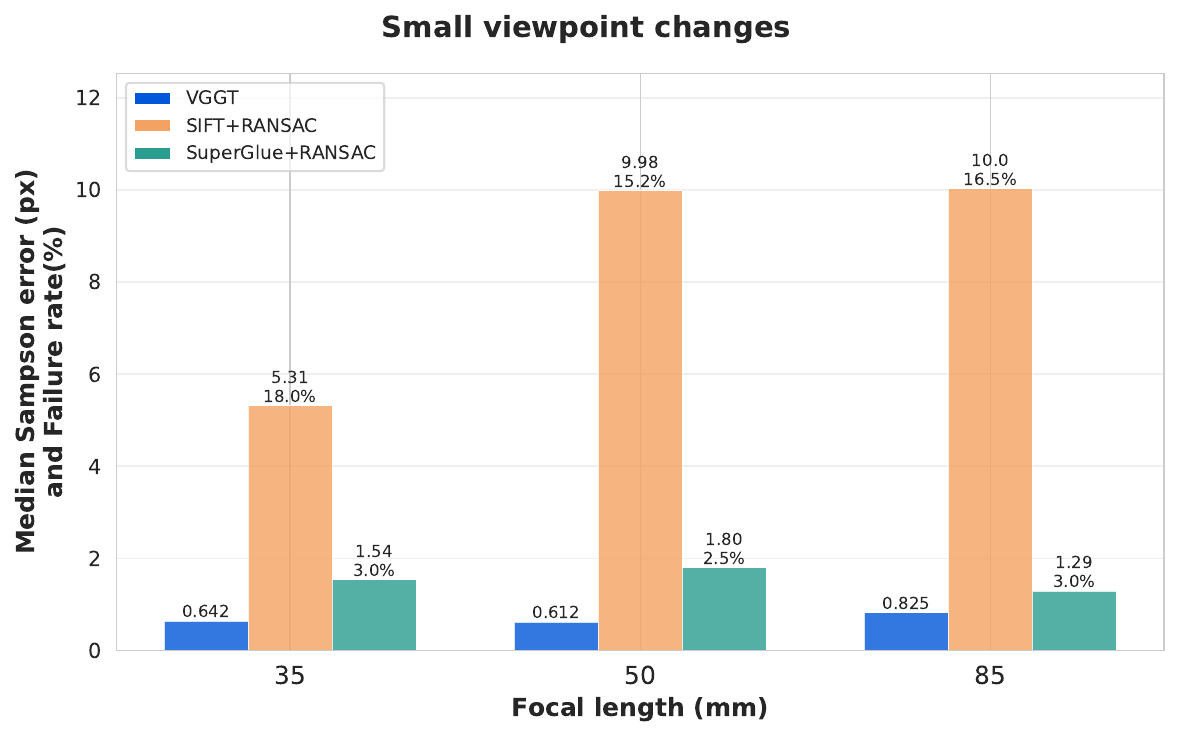}
    \end{subfigure}
    \begin{subfigure}[b]{.48\linewidth}
        \includegraphics[width=\linewidth]{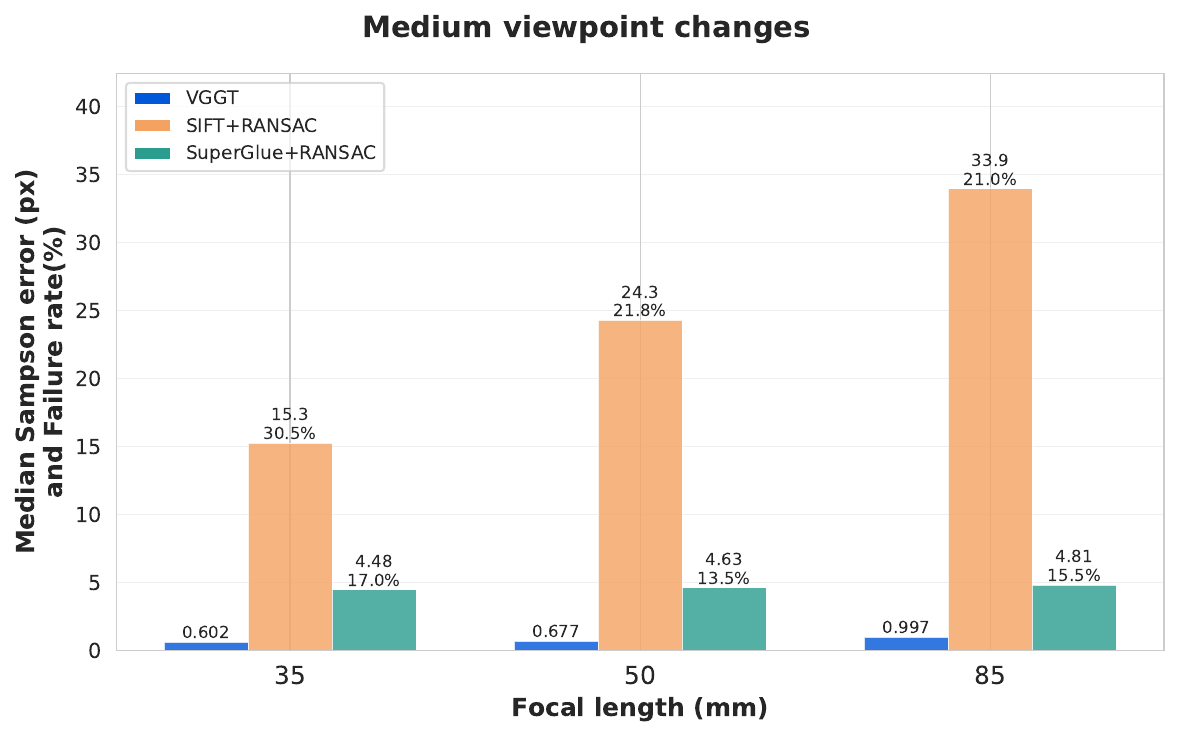}
    \end{subfigure}
    \begin{subfigure}[b]{.48\linewidth}
        \includegraphics[width=\linewidth]{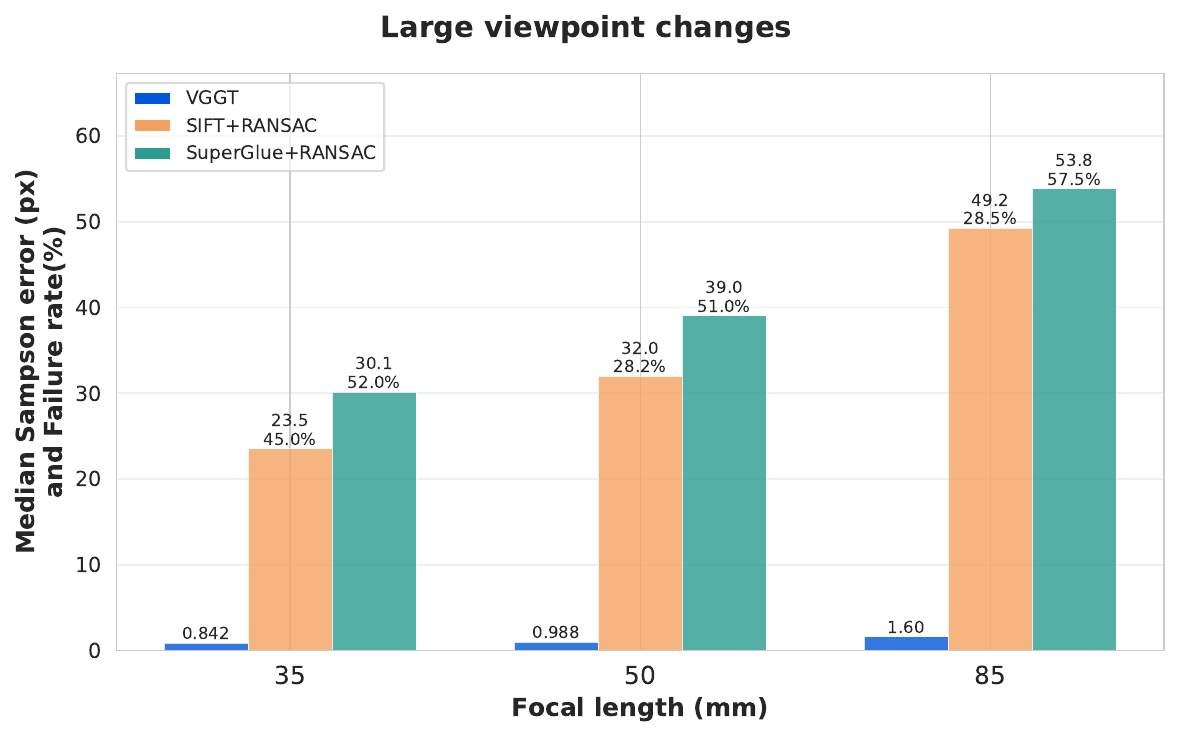}
    \end{subfigure}
    \caption{\textbf{Robustness to focal length changes for different viewpoint differences between the cameras.} VGGT outperforms the baselines for the case of large viewpoint differences, but performs on par with the baselines for stereo pair cameras.}
    \label{fig:focal_length_robustness_spherical}
\end{figure*}

Focal length changes effectively zoom in or out, altering feature scales and correspondence difficulty. We test focal lengths from 24 mm to 100 mm (36 mm sensor) across different viewpoint difference configurations. We compare focal lengths of 35 mm, 50 mm, and 85 mm for different viewpoint differences on ~\cref{fig:focal_length_robustness_spherical}, which shows that VGGT maintains consistent performance across all focal lengths, while classical methods exhibit high failure rates, particularly at extreme focal lengths and significant viewpoint differences. For close stereo pairs, classical methods achieve lower median error when successful, but VGGT demonstrates superior robustness by consistently producing valid predictions.

\bibliographystyle{splncs04}
\bibliography{main}

\end{document}